\documentclass[journal,compsoc,cspaper,final]{IEEEtran}
\usepackage{cite}
\usepackage{color}
\usepackage{xcolor}
\usepackage{graphicx}
\usepackage{colortbl}
\usepackage{amsmath,amssymb} % define this before the line numbering.
\usepackage{multirow}
\usepackage{makecell}
\usepackage{bbding}
\usepackage{tabularx}
\usepackage{subcaption}
\usepackage[pagebackref=true,breaklinks=true,colorlinks,bookmarks=false]{hyperref}
\usepackage{url}

\newcommand{\myeqref}[1]{\eqref{#1}}
\newcommand{\myfigref}[1]{Fig.~\ref{#1}}

\newcommand{\eg}[0]{\textit{e.g.}}
\newcommand{\ie}[0]{\textit{i.e.}}
\newcommand{\etal}[0]{\textit{et. al}}

\newcommand{\colored}[1]{\textcolor{black}{#1}}

\usepackage{tikz}

\usetikzlibrary{spy}

\usepackage{ragged2e}

\begin{document}
% \title{High-Quality Image Recovery from Event Cameras}
% \title{Recover High-Quality Images from \\Event Cameras}
% \title{Super-Resolving Blurry Images with Events}
\title{Learning to Super-Resolve Blurry \\ Images with Events}
% Denoising, Deblurring and Super-resolution}
% \title{Enhancing High-Speed Photography }

%\author{Michael~Shell,~\IEEEmembership{Member,~IEEE,}
%        John~Doe,~\IEEEmembership{Fellow,~OSA,}
%        and~Jane~Doe,~\IEEEmembership{Life~Fellow,~IEEE}% <-this % stops a space
%\thanks{M. Shell was with the Department
%of Electrical and Computer Engineering, Georgia Institute of Technology, Atlanta,
\author{Lei Yu, %~\IEEEmembership{Member,~IEEE,}
    \and Bishan Wang, 
    \and Xiang Zhang,
    \and Haijian Zhang,
	\and Wen Yang, %,~\IEEEmembership{Senior Member,~IEEE,}
	\and Jianzhuang Liu,
	\and Gui-Song Xia %~\IEEEmembership{Senior Member,~IEEE,}
 \IEEEcompsocitemizethanks{
\IEEEcompsocthanksitem L. Yu, B. Wang, X. Zhang, H. Zhang, and W. Yang are with the School of Electronic Information, Wuhan University, Wuhan 430072, China. E-mail: \{ly.wd, wangbs, xiangz, haijian.zhang, yangwen\}@whu.edu.cn.
\IEEEcompsocthanksitem G. S. Xia is with the School of Computer Science, Wuhan University, Wuhan 430072, China. E-mail: guisong.xia@whu.edu.cn.
\IEEEcompsocthanksitem J. Liu is with the Huawei Noah’s Ark Lab, Shenzhen 518000, China. E-mail: liu.jianzhuang@huawei.com.
\IEEEcompsocthanksitem The research was partially supported by the National Natural Science Foundation of China under Grants 62271354, 61871297, 61922065, 41820104006, 61871299, and the Natural Science Foundation of Hubei Province, China under Grant 2021CFB467.
\IEEEcompsocthanksitem Corresponding authors: L. Yu and G. S. Xia.
}
}


% \markboth{Journal of \LaTeX\ Class Files,~Vol.~14, No.~8, August~2015}%
% {Shell \MakeLowercase{\textit{et al.}}: Bare Demo of IEEEtran.cls for IEEE Journals}

\markboth{TPAMI Submission}%
{TPAMI Submission}



\IEEEtitleabstractindextext{%
\begin{abstract}
% \justifying
% Real world 

\justifying
{\it S}uper-{\it R}esolution from a single motion {\it B}lurred image (SRB) is a severely ill-posed problem due to the joint degradation of motion blurs and low spatial resolution. In this paper, we employ events to alleviate the burden of SRB and propose an {\it E}vent-enhanced SRB (E-SRB) algorithm, which can generate a sequence of sharp and clear images with {\it H}igh {\it R}esolution (HR) from a single blurry image with {\it L}ow {\it R}esolution (LR). To achieve this end, we formulate an event-enhanced degeneration model to consider the low spatial resolution, motion blurs, and event noises simultaneously. We then build an {\it e}vent-enhanced {\it S}parse {\it L}earning {\it Net}work ({\bf eSL-Net++}) upon a dual sparse learning scheme where both events and intensity frames are modeled with sparse representations. Furthermore, we propose an event shuffle-and-merge scheme to extend the single-frame SRB to the sequence-frame SRB without any additional training process. Experimental results on synthetic and real-world datasets show that the proposed eSL-Net++ outperforms state-of-the-art methods by a large margin. Datasets, codes, and more results 
are available at \url{https://github.com/ShinyWang33/eSL-Net-Plusplus}.

\end{abstract}

\begin{IEEEkeywords}
Event camera, intensity reconstruction, denoising, deblurring, super-resolution, sparse learning.
\end{IEEEkeywords}}

\maketitle
\IEEEdisplaynontitleabstractindextext

% \fi
\IEEEpeerreviewmaketitle

\IEEEraisesectionheading{\section{Introduction}}

\IEEEPARstart{S}{uper}-resolution (SR) is a fundamental low-level vision task that aims at recovering a {\em high-resolution} (HR) image from a {\em low-resolution} (LR) input~\cite{wang2020i}. 
It is known to be an ill-posed problem and often coupled with motion blurs under dynamic scenes with fast-moving objects~\cite{dong2015,bascle1996,nah2019ntire}.
% In general, it is an ill-posed problem due to the degradation caused by image down-samplings~\cite{srcnn2015}. Moreover, the SR problem is often coupled with motion blurs in real-world scenarios, especially under dynamic scenes with fast-moving objects~\cite{bascle1996,nah2019ntire}. 
The concurrence of multiple image degradations makes the SR problem more challenging. 
Even though both the problems
of image SR and motion deblurring have been investigated separately for decades and promising results have been achieved~\cite{aharonKSVDAlgorithmDesigning2006,wang2020i,nah2021,tian2020,nah2019ntire}, it is reported that simply superimposing a motion deblurring module on an image SR module may either amplify the unwanted artifacts or lose detailed information~\cite{singh2014,zhang2018g}.
%, thus usually leading to sub-optimal results 

% 合起来考虑：blind和non-blind，但是不能考虑复杂的运动；噪声的问题；
Instead of cascading approaches, it has been demonstrated that {\it S}uper-{\it R}resolution from a single motion {\it B}lurred LR image (SRB) can be better tackled by simultaneously resolving motion ambiguities~\cite{zhang2018g,liang2021,niu2021}, which is itself severely ill-posed~\cite{gu2019a,nah2021}. %Existing approaches are devoted to decoupling the motion ambiguities by assuming uniform motions~\cite{liang2021,pan2020b,zhang2018g}
Recently, promising results for SRB have been obtained by kernel-based approaches if uniform motions are valid~\cite{liang2021,pan2020b,zhang2018g,yamaguchi2010,park2017}. However, most scenes from the real-world scenario are with non-uniform motions, \eg, non-rigid or moving objects, violating the uniform motion assumption. To address this problem, various approaches have been proposed either by estimating motion flows from video sequences~\cite{park2017} or learning end-to-end deep neural networks~\cite{yu2018,xu2017,zhang2018e,zhang2018f}. These methods are either domain-specific only for face and text images~\cite{yu2018,xu2017} or heavily rely on the performance of the deblurring sub-module~\cite{zhang2018e,zhang2018f}, and thus are not reliable for general image SR tasks oriented to natural images with complex motions.

% on the other hand, 

\begin{figure}[t]	
	\centering	
	\includegraphics[width=0.45\textwidth]{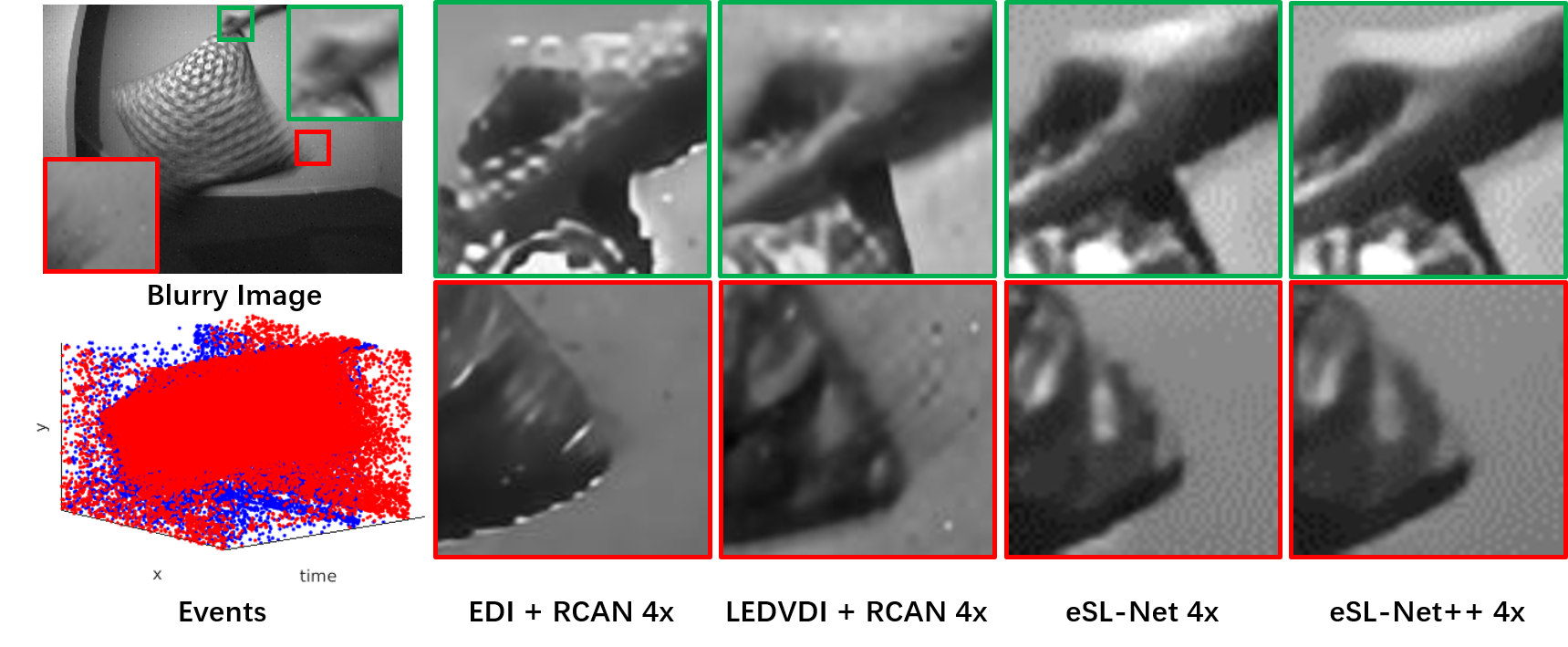}	
	\caption{Our eSL-Net++ reconstructs high-resolution, sharp, and clear intensity images for event cameras by {\it A}ctive {\it P}ixel {\it S}ensor (APS) frames and the corresponding event sequences. The eSL-Net++ performs much better than EDI~\cite{pan2019bringing} and LEDVDI~\cite{lin2020ledvdi} that are followed by an SR network RCAN~\cite{zhang2018image}.}	
	\label{highlight}
\end{figure}

% \cite{yamaguchi2010,zhang2018g,park2017,liang2021,zhang2020d,zhang2019b,zhang2018f}, while 
% \cite{bascle1996,yamaguchi2010,xu2017,zhang2018g,park2017,liang2021,zhang2020d,zhang2019b,zhang2018f}. 

% Motion ambiguity: non-blind / blind; uniform kernel based zhang2018f,... / of-flow based (using multiple frames)
% Unknown noise
% Only output one single frame

% 本文提出利用事件相机xxx，利用事件补偿时间上的ambiguity，同时空间上xxx的信息通过网络从数据中学习。event噪声问题很严重，单独考虑event和图像的噪声问题。

In this paper, we propose to utilize the event camera to enhance the performance of SRB.
Event cameras are bio-inspired sensors that can perceive dynamic scenes and emit asynchronous events, which are triggered to respond to the brightness change and generally represented as compositions of position, time stamp, and polarity~\cite{lichtsteiner2008128,gallego2019}.  With extremely low latency (in the order of $\mu$s), the triggered events inherently encode the intra-frame motions and textures with extremely high temporal resolution~\cite{benosman2013event,rebecq2019}. This property motivates us to super-resolve blurry images with events for more challenging SR problems, \eg, complex natural scenes with large and non-uniform motions. 
\begin{itemize}
    \item \colored{{\bf Motion Deblurring.} The embedded intra-frame information behind events compensates the erased motions and textures from blurry LR images~\cite{zhu2018ev,pan_high_2020}, which significantly relieves the burden of motion deblurring~\cite{pan_high_2020}. }
    \item \colored{{\bf Super-Resolution.} The extremely high temporal resolution of events preserves intra-frame temporal continuities of dynamic scenes when encountering motion blurs~\cite{gallego2019}. Therefore, similar to video SR~\cite{chan2022investigating}, the temporal correlation can be leveraged through events to boost the SR performance even with a single motion-blurred image.}
    % The extremely high temporal resolution of events enables the restoration of the missing temporal correlations~\cite{gallego2019}, which has been widely employed to compensate for the missing spatial information and further enhance the SR performance~\cite{chan2022investigating,jing2021turning}.}
    % The extremely high temporal resolution of events enables the potential to preserve temporal continuities of dynamic scenes even when encountering motion blurs~\cite{gallego2019}. Therefore, the temporal correlation can be adopted to boost the SR performance even with a single motion blurred image~\cite{chan2022investigating}. }
\end{itemize}
% severe image degradation due to the contamination of event noises and the violation of event statistics (as shown in Fig.~\ref{highlight}), which motivates us to design a comprehensive approach.

% Many algorithms have been proposed to deal with motion blurs with events and revealed t
% Even though the superiority of event-based motion deblurring has been validated~\cite{pan2019bringing,lin2020,jiang2020b,wang2020d,wang2019b}, the potential of event based super-resolution from LR blurry images are still vague. 

Besides motion ambiguities, noises and disturbances can lead to visually unpleasant results since these imperfections would be amplified if image SR algorithms are fed with noisy inputs~\cite{singh2014,zhang2018g}. %For event-based SRB, noises are from both LR blurry images and events. Specifically, 
Due to the special imaging mechanism, the event camera generally contains more noises and disturbances than the conventional frame-based camera~\cite{baldwin2020event}. Furthermore, noises of the event camera would be induced from both spatial and temporal domains, which raises the difficulty for event noise suppression~\cite{gallego2019}. It brings vagueness to the potential of event-based SRB even if the superiority of event-based motion deblurring has been validated~\cite{pan2019bringing,lin2020ledvdi,jiang2020b,wang2020d,wang2019b}. Existing algorithms of event denoising (\eg,~\cite{wang2020joint,baldwin2020event}) can be applied separately, but it might suffer from over-suppression of events, leading to deteriorated deblurring performance.  %, which raises the difficulties for event noise suppression.

Thus it is necessary to cope with motion ambiguities and event noises simultaneously for {\it E}vent-enhanced SRB (E-SRB). This paper is devoted to achieving this end by adopting the \textit{sparse learning} framework. Particularly, the degeneration process from HR sharp images to LR blurry images is revisited by adopting events, where motion blurs and noises are both considered. Based on this {\it E}vent-enhanced {\it D}egeneration {\it M}odel (EDM), the solution of E-SRB is then feasible by imposing sparsity on HR sharp images under specific dictionaries. And finally, an {\it e}vent-enhanced {\it S}parse {\it L}earning {\it Net}work (eSL-Net++) is proposed by unfolding the iterative algorithm for the $\ell_1$-norm penalized optimization problem and training it on a synthetic dataset.

The contributions of this work are three-fold: 
\begin{itemize}
    \item We propose to adopt events to enhance the performance of SRB, where an EDM is presented by taking into account both event noises, motion blurs, and the low spatial resolution. 
    \item We propose an eSL-Net++ to tackle the challenge of E-SRB based on a dual sparse learning scheme, where event noise suppression, motion deblurring, and image SR are simultaneously addressed. 
    \item We propose a rigorous event shuffle-and-merge scheme to extend the eSL-Net++ to high frame-rate HR video sequence recovery from a single blurry LR image without any additional training process. Both synthetic and real-world datasets are built for training and testing.
\end{itemize}

%A preliminary version of this work has been published in~\cite{wang2020event}, \ie, eSL-Net~\cite{wang2020event},
This paper is an extended version of our preliminary work, \ie, eSL-Net~\cite{wang2020event}, with several significant improvements: (i) the event statistics are further considered in the EDM, based on which a {\it D}ual {\it S}parse {\it L}earning scheme (DSL) is proposed to suppress noises from events and blurry images by mutual compensations, and the resultant network is called eSL-Net++; (ii) a rigorous {\it e}vent {\it s}huffle-and-{\it m}erge scheme (ESM) is proposed to extend eSL-Net++ for video sequence recovery which achieves better performance than its previous version, \ie, eSL-Net; (iii) extended evaluations on both synthetic and real-world datasets are implemented. Fig.~\ref{highlight} shows an example by eSL-Net and eSL-Net++.

% The proposed eSL-Net and its extension to eSL-Net++ can both achieve superior performance %这里根据第一张图描述一下我们方法的优越性。

\section{Related Works}
% Both tasks of image SR and motion deblur are highly ill-posed

\subsection{SR from Blurry LR Images}
% 超分辨率问题是什么？早期方法主要利用模型和hand crafted prior去提升性能；自从Dong将DL方法用于SR，取得了primising的结果；DL逐渐成为了SR的dominant的方法。
% 图像模糊丢失了时间上的信息，引入了运动ambiguities，降低了图像质量。非常challenge，而且在realistic的场景中很常见。Recent works mainly focus on DL method。
% The image SR and the motion deblurring are both highly ill-posed problems and many algorithms have been proposed to address them.
% The image degeneration from HR to LR would erase detailed information and bring spatial ambiguities.
Image SR is to resolve the spatial ambiguities and recover missing detailed information from the LR images~\cite{wang2020i,liang2021,zhang2020d,niu2021}. Early attempts address this problem by optimizing an under-determined problem regularized by various handcrafted priors, which often suffer from undesired artifacts, \eg, over-smoothness~\cite{yang2008,yang2010image,shi2015}. Recent state-of-the-art methods turn to adopt deep priors which often surpass the traditional methods~\cite{dong2015,mao2016,kim2016,zhang2018residual}. However, when dealing with blurry LR inputs, the task of image SR becomes more challenging due to motion ambiguities and texture erasures~\cite{jin2018learning,jin2019learning,purohit2019bringing}. Thus, it is essential to decouple motion deblurring from image SR. 
% For uniform motions, blurs can be modelled as the convolution of a blur kernel with a sharp image, leading to the kernel-based motion deblurring approaches by means of regularized deconvolution~\cite{krishnan2011blind,sun2013edge}. For motion blurs caused by complex motion behaviors, pixel/patch-wise motion flows are often required to flexibly depict non-uniform motions. And various approaches have been devoted to achieve this end~\cite{gong2017motion,zhang2015intra,hyun2015generalized,jin2018learning,pan2016blind,xu2013unnatural,fergus2006removing,liu2020self,zhou2019spatio}.

Even though image SR and motion deblurring have been separately investigated for decades~\cite{aharonKSVDAlgorithmDesigning2006,wang2020i,nah2021,tian2020,nah2019ntire}, simply cascading existing SR and deblurring methods may amplify unwanted artifacts and suffer from sub-optimal results~\cite{singh2014,zhang2018g}. It is essential to resolve SR and the deblurring in a joint manner, where decoupling of motion ambiguities plays an important role. The uniform motions are commonly assumed to alleviate the difficulty where blurs can be considered within a unified degeneration model parameterized by a blurring kernel~\cite{zhang2018g,zhang2019b,yamaguchi2010,park2017}. \colored{Jointly optimizing the blurring kernel and image SR can effectively boost the overall performance in either non-blind~\cite{zhang2018g,zhang2019b} or blind~\cite{yamaguchi2010,park2017,liang2021,wang2021unsupervised} approaches. Gu \etal~\cite{gu2019b} propose an iterative kernel correction (IKC) to refine the blur kernel iteratively based on the previous SR results. Flow-based kernel prior (FKP) \cite{liang2021}, kernel space representation~\cite{cornillere2019}, and mutual affine network (MANet)~\cite{liang2021mutual} take into account the spatially variant degradations. To bridge the real-to-synthesis gap, Wang \etal~\cite{wang2021unsupervised} propose a degradation-aware SR network (DASR) by learning degradation representations through contrastive learning. The kernel estimation accuracy plays an important role in blind SR since the degradation error can further be magnified by the SR process~\cite{gu2019b}, which poses challenges to real-world scenarios, especially when encountering large and non-uniform motion blurs. }
% Gu \etal~\cite{gu2019b} propose an iterative kernel correction (IKC) to refine the blur kernel iteratively based on the previous SR results. Flow-based kernel prior (FKP) \cite{liang2021flow}, kernel space representation \cite{cornillere2019}, and mutual affine network (MANet) \cite{liang2021mutual} are proposed to take into account the spatially variant degradations. To bridge the real-to-synthesis gap, Wang \etal \cite{wang2021unsupervised} propose a degradation-aware SR network (DASR) by learning degradation representations through the contrastive learning strategy. The kernel estimation accuracy plays an important role for blind SR since the degradation error can further be magnified by the SR process~\cite{gu2019b}, which poses challenges to real-world scenarios, especially when encountering large and non-uniform motion blurs.}
% However, inaccurate blur kernel estimation will inevitably produce incorrect deblurring results 
% % However, those kernel-based approaches would result in inaccurate motion estimation in the real-world scenario 
% especially when encountering with non-uniform motions and thus deteriorate the overall image SR performance. }

% \textcolor{blue}{Another 
% }

\colored{Image SR from LR images degraded by non-uniform motion blur is more challenging~\cite{zhang2018f,zhang2021plug}. To tackle this problem, Park {\etal}~\cite{park2017} propose to predict accurate motion flows from the camera pose and depth, estimated by stereo matching between inter-frame video sequences, which however is only valid for static scenes. On the other hand, several recent works jointly resolve image SR and motion deblurring using end-to-end trained deep networks~\cite{zhang2018e,zhang2018f,yu2018,xu2017}. Specifically, some domain-specific priors are adopted to alleviate the ill-posedness but they are only valid for blurry face and text images~\cite{yu2018,xu2017}. Deblur-then-SR approaches heavily rely on the performance of the deblurring sub-modules~\cite{zhang2018e,zhang2020d,zhang2018f}, thus not reliable for general image SR tasks oriented to natural images with complex motions.
% Several blind SR approaches have been proposed by predicting blur kernels to consider multiple image degeneration in a supervised~\cite{liang2021,liang2021flow} or unsupervised~\cite{wang2021unsupervised} manner, thus inaccurate blur kernel estimation will inevitably produce incorrect results especially for general image SR tasks oriented to natural images and complex motions~\cite{liang2021}.
% However, existing approaches are either domain-specific only for face and text images~\cite{yu2018,xu2017} or heavily relying on the performance of the deblurring sub-module~\cite{zhang2018e,zhang2020d,zhang2018f}, and thus not reliable for general image SR tasks oriented to natural images and complex motions.
}

% To address this problem, some recent work turns to using end-to-end deep networks~\cite{zhang2018e,zhang2018f,yu2018,xu2017}.

% Decoupling the motion ambiguities plays an important role for this joint SR problem and inaccurate motion estimation would deteriorate the overall image SR performance~\cite{zhang2018g,liang2021}. 

% Blind and non-blind

\subsection{Event-Based Low-Level Image Enhancement}
% Early attempts of reconstructing intensity from pure events are commonly based on the assumption of brightness constancy, i.e. static scenes~\cite{kim2014}. The intensity reconstruction is then addressed by simultaneously estimating the camera movement, optical flow and intensity gradient~\cite{kim2016}. In~\cite{cook2011}, Cook et al. propose a bio-inspired and interconnected network to simultaneously reconstruct intensity frames, optical flow and angular velocity for small rotation movements. Later on, Bardow \etal~\cite{bardow2016} formulate the intensity change and optical flow in a unified variational energy minimization framework. By optimization, one can simultaneously reconstruct the video frames together with the optical flow. On the other hand, another research line on intensity reconstruction is the direct event integration method~\cite{scheerlinck2018,munda2018}, which does not rely on any assumption about the scene structure or motion dynamics. Since events can only depicts the dynamic scenes, reconstruction of intensity from pure events are not adaptive to the static scenes.

%\subsubsection{Fusion with APS Frames}
\textbf{Event-Based Motion Deblurring.} Event cameras can ``continuously'' emit events asynchronously with extremely low latency (in the order of $\mu$s)~\cite{lichtsteiner2008128,gallego2019}, inherently embedding motions and textures~\cite{benosman2013event}. Thus the task of motion deblurring~\cite{pan2019bringing,pan_high_2020}  can be essentially alleviated by compensating blurry images with events~\cite{scheerlinck2019a,jiang2020b,wang2020event}.
%  Even though the calibration of events to image frames is further required for fusion based methods, it can be easily tackled either by the DAVIS~\cite{brandli2014b} sensors or with a dual camera set connecting the event camera and the RGB camera with a beam splitter~\cite{wang2020joint}. Generally, the fusion based methods can typically achieve better performance than pure event based methods~\cite{pan_high_2020,jiang2020b,lin2020}. In~\cite{scheerlinck2019a}, events are approximated as the time differential of intensity frames. Based on this, a
In~\cite{scheerlinck2019a}, events are approximated as the time differential of intensity frames, a complementary filter is proposed as a fusion engine and nearly continuous-time intensity frames can be generated. Pan \etal~\cite{pan2019bringing} propose an event-based deblurring approach by relating blurry {\it A}ctive {\it P}ixel {\it S}ensor (APS) frames and events with an {\it E}vent-based {\it D}ouble {\it I}ntegration (EDI) model. Afterward, a multiple-frame EDI model is proposed for high-rate video reconstruction by further considering inter-frame relations~\cite{pan_high_2020}. \colored{Recent works turn to convolutional neural networks (CNNs) for event-based motion deblurring by supervised learning from synthesized datasets composed of paired events, blurry inputs, and sharp images~\cite{wang2019,lin2020ledvdi,sun2022event,song2022cir}. And then the semi-supervised \cite{xu2021motion} and self-supervised \cite{zhang2022unifying} learning frameworks are respectively proposed to bridge the real-to-synthesis gap for event-based motion deblurring. 
}
Existing approaches solely focus on event-based motion deblurring, but rarely exploit events to resolve HR clear images from blurred inputs, \ie, E-SRB. %And two wo challenges s, \ie, {\it event noises} and {\it low spatial resolution}.

\noindent \textbf{Event SR.} Even though event cameras have an extremely high temporal resolution, their spatial (pixel) resolution is relatively low and yet not easy to be resolved physically~\cite{gallego2019}. Directly resolving the resolution purely from events is very challenging due to the heavily interfered events ~\cite{wang2020joint}. To achieve this end, Duan and Wang \etal \cite{wang2020joint, duan2021eventzoom} propose spatial guidance for events by leveraging gradients and motions provided by images with a high spatial resolution from traditional cameras. If further providing camera poses, intensity images with a high resolution can be directly reconstructed purely from events by leveraging the Poisson equation~\cite{kim2014}. On the other hand, recent works turn to exploit deep neural networks for event-based intensity recovery~\cite{rebecq2019,wang2019,lin2020ledvdi} and correspondingly, E2SRI~\cite{mostafavi2020e2sri} and EventSR~\cite{wang2020d} are proposed for event-based image recovery and super-resolution where classical techniques of generative adversary network (GAN), recurrent neural network (RNN) and U-Net are respectively exploited. Both E2SRI and EventSR are devoted to recovering images purely from events. Benefiting from the high temporal resolution, one can adopt events to enhance video frame interpolation (VFI)~\cite{tulyakov2021time} and video super-resolution (VSR)~\cite{jing2021turning}. However, they only accept sharp and clear inputs and require two consecutive image frames. Instead, only a single blurry LR image is available in this paper, making the problem more challenging.

\noindent\textbf{Event Denoising.} The collected events from event cameras often suffer from noises and disturbances due to the thermal effects and the environmental brightness fluctuations~\cite{baldwin2020event}. And it becomes the main obstacle to follow-up applications, especially for low-level imaging tasks~\cite{brandli2014,gallego2019,hu2021}. 
% Even though the background activity noise has been reported to follow the Poisson distribution~\cite{Khodamoradi2017}, event statistics have not been fully unveiled yet~\cite{gallego2019}, which raises the difficulty of event denoising. 
% Thus event denoising is necessary even though event statistics have not been fully unveiled yet~\cite{gallego2019}. 
Temporal and spatial consistencies are commonly employed to remove noisy events~\cite{czech2016,wu2020a}, but real events may violate such consistencies, especially with complex textures. To address this problem, the event time surface~\cite{lagorce2016hots} provides a smooth manifold~\cite{munda2018} while intensity images generate an event occurrence probability~\cite{baldwin2020event,wang2020joint}, with which one can largely improve the robustness to event noises~\cite{munda2018,baldwin2020event,wang2020joint}. %These methods only focus on event denoising and are often exploited as the prepossessing step.
On the other hand, various methods have been proposed to address the problem of event noises jointly within the framework of event-based motion deblurring. Barua \etal~\cite{barua2016} propose a learning-based approach to smooth the image gradients by imposing sparsity regularization and then exploit Poisson integration to recover the intensity image from gradients.  Instead of sparsity, Munda \etal~\cite{munda2018} introduce the manifold regularization over the event time surface~\cite{lagorce2016hots} and propose a real-time intensity reconstruction algorithm. %Since most of existing methods rely on the hand-crafted regularization instead of data-driven priors, noises can be largely alleviated but at the cost of producing artifacts (e.g. blurry edges). 

Even though the problem of SRB can be essentially alleviated by events, motion deblurring and event denoising are two challenges. Straightforwardly, concatenating modules of event denoising~\cite{baldwin2020event} and event-based motion deblurring~\cite{xu2021motion} can tackle the task of event-based SRB. However, errors or artifacts might be aggregated, leading to sub-optimal solutions. It motivates us to resolve the event-based SRB and jointly tackle the problems of motion deblurring and super-resolution in the presence of tremendous event noises.

% Both event noises and low spatial resolution degenerate the  obstacle While for image recovery based on fusion of APS intensity images and events, the image super-resolution is still not able to be addressed by the state-of-the-art algorithms.

% Comparing to~\cite{mostafavi2020e2sri}, our proposed approach differs in the following aspects: (1) we proposed a unified framework to simultaneously resolve the tasks including denoising, deblurring and super-resolution, while E2SRI~\cite{mostafavi2020e2sri} cannot directly deal with blurring or noisy inputs; (2) the proposed network is completely interpretable with meaningful intermediate processes; (3) our framework reconstructs the intensity frame by fusing events and APS frames, while E2SRI is proposed for reconstruction from pure events.event-enhanced sparse learning (eSL) module,
\begin{figure*}[!htp]
    \centering
    \includegraphics[width=.9\textwidth]{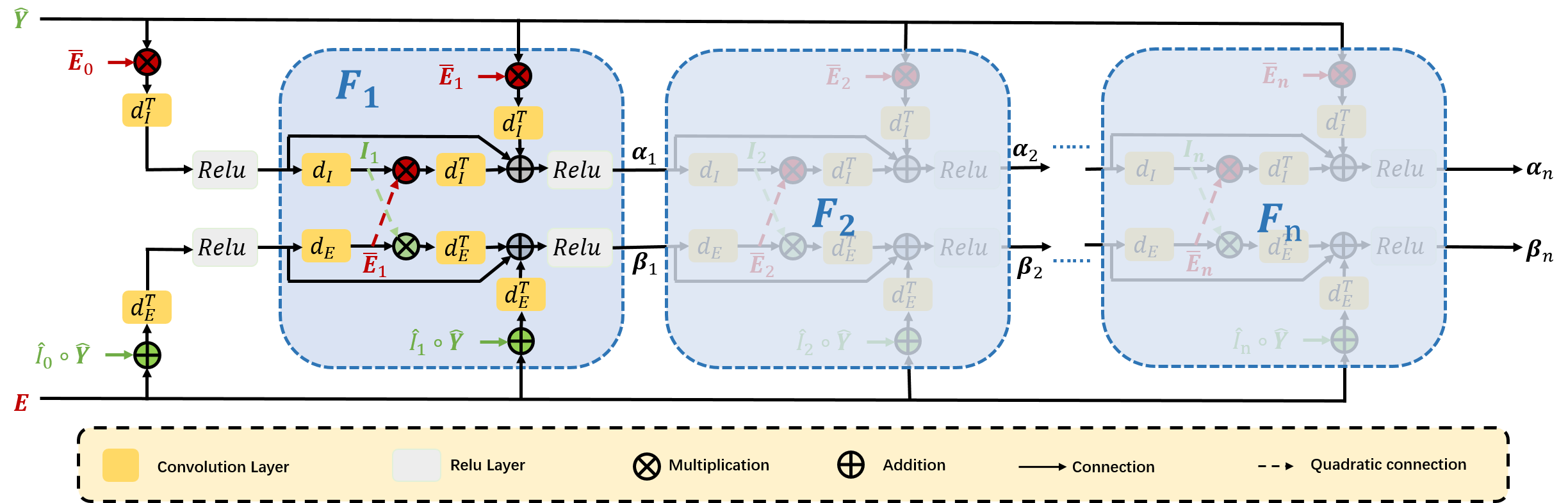}
    \caption{The {\it D}ual {\it S}parse {\it L}earning (DSL) scheme, where \myeqref{ev-lasso} is optimized through the deep neural network by unfolding \myeqref{ev-ista}. Note that quadratic connection means that the input is squared and then fed into the block to which the arrow connects.}
    \label{fig:fusionframework}
\end{figure*}
\section{Problem Statement}
% Super-resolution 

% Event cameras such as DAVIS~\cite{brandli2014b} can produce both events and gray-scale intensity APS frames.and correspondingly, $\mathcal{E}_{\mathcal{T}} \triangleq\left\{\boldsymbol{e}(t)\right\}_{t \in \mathcal{T}}$ the set of event streams $\boldsymbol{e}(t)$ collected within $\mathcal{T}$. 
Let $ \boldsymbol{Y} $ be the image frame captured during the exposure time interval $\mathcal{T}\triangleq [0,T)$. Due to the imperfection of image sensors, the observed image $\boldsymbol{Y}$ may suffer from non-negligible quality degeneration due to noises, motion blurs, and low spatial resolution. And the degraded observation $ \boldsymbol{Y}$ can be related to the high-quality (sharp, clear, and high-resolution) latent images $\boldsymbol{X}(t)$ as follows,
\begin{equation}\label{eq:overall_noevents}
\begin{split}
    \boldsymbol{Y} &= \frac{1}{T} \int_{\mathcal{T}} \boldsymbol{I}(t)dt + \boldsymbol\varepsilon_Y, \\ \boldsymbol{I}(t) &= \boldsymbol{PX}(t),
\end{split}
\end{equation}
where $\boldsymbol{I}(t)$ denotes the down-sampled version of $\boldsymbol{X}(t)$ via the operator $\boldsymbol{P}$ at time $t$, the integration over $\mathcal{T}$ represents the motion blur process~\cite{Nah2017Deep} and $\boldsymbol{\varepsilon}_Y\sim \mathcal{N}(0,\sigma_Y^2)$ is the Gaussian noise with standard deviation $\sigma_Y$. Thus the super-resolution from an LR blurred image (SRB) can be represented as,
\begin{equation}\label{eq:srb}
    \boldsymbol{X}(t) = \mbox{SRB}(\boldsymbol{Y}).
\end{equation}
Obviously, finding a single HR image $\boldsymbol{X}(t), t\in \mathcal{T}$ or a sequence of HR images $\{\boldsymbol{X}(t)\}_{t\in \mathcal{T}}$ from a single blurry image $ \boldsymbol{Y}$ is severely ill-posed. %And to the best of our knowledge, it has never been tackled in existing literature.

To tackle this problem, we propose to alleviate the SRB problem \eqref{eq:srb} via the event camera, which
% , \ie,
outputs events whenever the logarithm of the intensity changes over a pre-setting threshold $c>0$, 
\begin{equation}\label{con:event_trigger}
\log(\boldsymbol{I}(t))-\log(\boldsymbol{I}(\tau)) = p \cdot c,
\end{equation}
where $\boldsymbol{I}(t)$ and $\boldsymbol{I}(\tau)$ denote the latent instantaneous intensities at time $t$ and $\tau$, respectively, and $p \in \left\lbrace +1,-1 \right\rbrace $ is the polarity representing the direction (increase or decrease) of the intensity change. 
% Thus an event triggered at $(x,y)$ on time $\tau$ can be expressed as a spike function $p \delta(t-\tau)$ and a collection of events during time interval $[f,t]$ can be written as a spike train:
% \begin{equation}\label{event_with_dirac}
% e_{x}(t) \triangleq  p \sum_{\tau\in \mathcal{T}^{x,y}_{f\to t}}\delta(t-\tau)
% \end{equation}
% where $\mathcal{T}^{x,y}_{f\to t}$ is a set of time stamps of triggered events at $(x,y)$ in $[f,t]$ and $\delta(\cdot)$ is the Dirac function~\cite{dirac1981principles}. 
Correspondingly, we have the following relationship,
\begin{equation}\label{eq:e2latent}
    \log\left(\boldsymbol{I}(t)\right)=\log\left(\boldsymbol{I}(f)\right)+c \int_{f}^{t} e(s) d s,
\end{equation}
where $\boldsymbol{I}(f)$ is the latent image of time $f$, and $e(t)\triangleq  \sum_{\tau\in \mathcal{T}_{f\to t}}p(\tau)\cdot \delta(t-\tau)$ denotes the event stream (represented as the spike train) triggered at position $x$ during $\mathcal{T}_{f\to t} \triangleq [f,t)$. Finally, one can derive the following relation from \eqref{eq:e2latent},
% \begin{equation}\label{eq:e2latent_simple}
%     \log\left(\boldsymbol{I}(t)\right)=\log\left(\boldsymbol{I}(f)\right)+c \int_{f}^{t} e(s) d s
% \end{equation}
% or equivalently,
\begin{equation}\label{eq:e2latent_simple}
    \boldsymbol{I}(t)=\boldsymbol{I}(f) \circ \exp \left(c \int_{f}^{t} e(s) d s \right),
\end{equation}
where $\circ$ represents the Hadamard product.
% and $\boldsymbol{\Lambda}(f,t)$ is a matrix with element at $x,y$, \ie, $\boldsymbol{\Lambda}_{x}(f,t)  \triangleq \int_{f}^{t} e_{x}(s) d s$ representing the accumulation of triggered events during the time interval $[f,t]$.
% And the followin, the intensity at $(x,y)$ % As a result, a sequence of discrete events is turned into a continuous time signal.
%\subsection{General Model for Events and Intensity Images}
%\begin{equation}\label{event_image}
% $\log\left(\boldsymbol{I}_{x}(t)\right)=\log\left(\boldsymbol{I}_{x}(f)\right)+c \int_{f}^{t} e_{x}(s) d s$, then
%\end{equation}
%Consequently, plug   \eqref{event_image} into   \eqref{blur_model} to get:
%  \eqref{mathmodel} indicates the relationship among events, the observed intensity image and the latent image ideally. 
% Exploiting events can largely enhance the temporal resolution of the collected low rate frames, i.e. $\boldsymbol{Y} $. 
Then according to \eqref{eq:overall_noevents} and \eqref{eq:e2latent_simple}, it has
\begin{equation}\label{event_image_clear}
\begin{split}
\boldsymbol{Y} &=\boldsymbol{E}(f)\circ \boldsymbol{I}(f) + \boldsymbol{\varepsilon}_Y, \\ \boldsymbol{I}(f) &= \boldsymbol{PX}(f),
\end{split}
\end{equation}
with  
\begin{equation}\label{eq:etr}
\boldsymbol{E}(f)\triangleq \frac{1}{T} \int_{\mathcal{T}} \exp \left(c \int_{f}^{t} e(s) d s \right) d t,
\end{equation}
which is called the EDI in~\cite{pan2019bringing} representing the average of accumulated events. Benefiting from the high temporal resolution, equation \myeqref{eq:etr} can provide missing intra-frame information caused by the blurring process and thus largely alleviate the difficulty of SRB. Moreover, equation \eqref{event_image_clear} implies that one can even super-resolve $\boldsymbol{X}$ at any specific time $f\in \mathcal{T}$, which provides a convenient approach to recover a sequence of HR images $\boldsymbol{X}(f)$ from one single blurry LR image $\boldsymbol{Y}$.

% \subsection{Robust Event-based SRB } \label{degeneration-model}
% Benefiting from the high temporal resolution, events provide the missing intra-frame information caused by blurring process and largely alleviate the difficulty of SRB.
However, the thermal effects or current leakage largely disturb the triggered events~\cite{baldwin2020event}, resulting in violation of \myeqref{event_image_clear}. Define the event accumulation $\boldsymbol{\Lambda}(f,t)\triangleq \int_{f}^t e(s)ds$. Then it can be assumed as a random variable drawn from the Poisson distribution~\cite{Khodamoradi2017}, 
\begin{equation}
    \boldsymbol{\Lambda}(f,t) \sim \textnormal{Poisson}(\lambda|t-f|),
\end{equation}
with $\lambda$ representing the event firing rate, \ie, the number of events triggered in a unit time interval. Applying the Taylor expansion, one can obtain 
$
 \exp(c\boldsymbol{\Lambda}(f,t)) \approx 1+c\boldsymbol{\Lambda}(f,t) %\\ &\sim \mathcal{N}(1+c\lambda_{x}(t-f),c^2+c^3\lambda_{x}(t-f))
$ and then \eqref{eq:etr} becomes,
\begin{equation}\label{eq:etr2}
    \boldsymbol{E}(f) \approx 1 + \frac{c}{T}\int_{\mathcal{T}} \boldsymbol{\Lambda}(f,t) dt.
\end{equation}
Since $\boldsymbol{\Lambda}(f,t)$ is a random variable, the integration on the right side of \eqref{eq:etr2} is still a random variable and obeys the Poisson distribution,
\[
\begin{split}
   \int_{\mathcal{T}} \boldsymbol{\Lambda}(f,t) dt &\sim \textnormal{Poisson}\left(\lambda\rho \right) ,
  \end{split}
\]
with $\rho = f^2-fT+\frac{T^2}{2}$. And when $\lambda$ is large enough, Poisson distribution can be approximated by a Gaussian distribution with mean and variance equal to $\lambda \rho$. According to \eqref{eq:etr2}, we can finally derive that $\boldsymbol{E}(f)$ is approximately drawn from a Gaussian distribution
\begin{equation}\label{eq:event_noise}
    \boldsymbol{E}(f) \sim \mathcal{N}(\mu,\sigma^2),
\end{equation}
where $\mu=1+c\lambda\rho/T$ and $\sigma= \frac{c\sqrt{\lambda\rho}}{T}$.

% And when $\lambda_{x}$ is large enough, Poisson distribution can be approximated by a Gaussian distribution, thus
% \[
%     \boldsymbol{\Lambda}_{x}(f,t) \sim \mathcal{N}(\lambda_{x}(t-f),\lambda_{x}(t-f))
% \]

% and its integral over $[t_f,t_f+T]$ will be degenerated to summation of $N$ split intervals, thus
% \[
%     \begin{split}
%       \boldsymbol{E}(f) &\approx \frac{1}{N} \sum_{n=0}^{N-1} \exp(c\boldsymbol{\Lambda}(f+n/T,f+(n+1)/T)) \\
%       &\sim \textnormal{Poisson}(1+c\lambda_{x}/T)
%     \end{split}
% \]
%Apparently, $\lambda_{x}$ depends on the change of intensity $\boldsymbol{I}_{x}(t)$ between $t$ and $f$, and event noises are 
Thus we can assume an additive Gaussian noise $\boldsymbol{ \varepsilon}_E\sim \mathcal{N}(0,\sigma^2)$ for $\boldsymbol{E}(f)$
% \begin{equation}
%     \boldsymbol{E} = \boldsymbol{E}_c +\boldsymbol{ \varepsilon}_E
% \end{equation}
% where $\boldsymbol{E}_c(f)=\mu$ represents the estimator and $\boldsymbol{ \varepsilon}_E\sim \mathcal{N}(0,\sigma^2)$ is the Gaussian white noise.
and finally, obtain the event-enhanced degeneration model,
\begin{equation}\label{general_model}
\begin{split}
\boldsymbol{Y}  &= \bar{\boldsymbol{E}}(f) \circ \boldsymbol{ I}(f) +\boldsymbol{\varepsilon}_Y, \\
\boldsymbol{E}(f) &= \bar{\boldsymbol{E}}(f) +\boldsymbol{ \varepsilon}_E, \\
 \boldsymbol{ {I}}(f) &= \boldsymbol{P}\boldsymbol{X}(f),
\end{split}
\end{equation}
where $\bar{\boldsymbol{E}}(f)\triangleq \mu$ represents the estimator of $\boldsymbol{E}(f)$. And our goal is to solve \eqref{general_model} for $\boldsymbol{X}(f)$. Specifically, given the observed image $\boldsymbol{Y} $ and the corresponding triggered events $\mathcal{E}_{\mathcal{T}} \triangleq\left\{{e}(t),t \in \mathcal{T}\right\}$, our goal is to reconstruct a high-quality intensity image $\boldsymbol{X}$ of the specified time $f\in \mathcal{T}$, \ie, %Obviously, it is an ill-posed problem with multiple tasks including denoising, deblurring and super-resolution. 
% Thus, the problem of  super-resolving $\boldsymbol{X}(f)$ from blurry LR image $\boldsymbol{Y}$ by adopting events can be formulated as follows,
\begin{equation}\label{eq:esrb}
        \boldsymbol{X}(f) = \mbox{E-SRB}(\boldsymbol{Y},\mathcal{E}_{\mathcal{T}},f),
\end{equation}
where the events $\mathcal{E}_{\mathcal{T}}$ and the blurry image $\boldsymbol{Y}$ are calibrated in the spatial and temporal domains~\cite{wang2020joint,tulyakov2021time}.

% In addition to events, gray-scale intensity images can be simultaneously collected and calibrated both in time and space~\cite{brandli2014b,wang2020joint}, but at a relatively lower frame rate.

% Since each pixel can be treated separately, subscripts $x$, $y$ are often omitted henceforth. Finally, considering the whole pixels, we can get a simple model connecting events, the observed intensity image and the latent intensity image:
% \begin{equation}\label{mathmodel}
% \boldsymbol{Y} = \boldsymbol{E}(f) \circ \boldsymbol{I}(f)
% \end{equation}
% with $\boldsymbol{E}(f) =\frac{1}{T} \int_{t_f}^{t_f+T} \exp (c \int_{f}^{t} e(s) d s) d t$ being double integral of events at time $f$~\cite{pan2019bringing} and $\circ$ denoting the Hadamard product. 

\begin{figure*}[!tb]
    \centering
    \includegraphics[width=\linewidth]{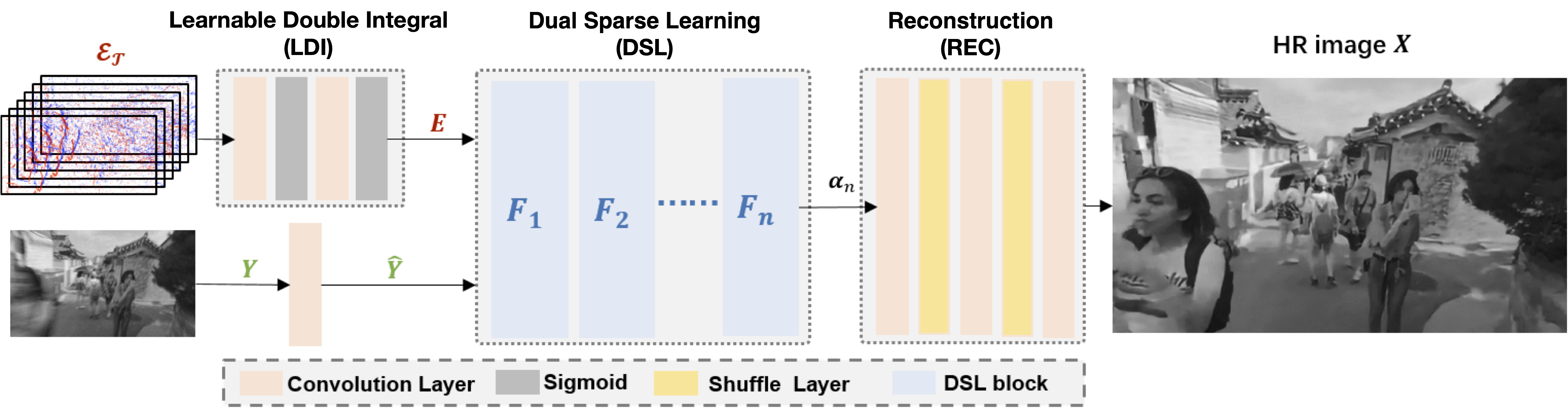}
    \caption{Event-enhanced Sparse Learning Network for super-resolving blurry images with events, \ie, eSL-Net++. Coefficients $\bf \beta$ and event reconstruction $\bar{\boldsymbol{E}}$ are omitted here.}
    \label{fig:highframework}
\end{figure*}

%\section{Event Enhanced High-Quality Image Recovery}\label{singleframe}
\section{Event-based Sparse Learning for SRB}\label{singleframe}
In this section, we first formulate the E-SRB problem using a DSL scheme by imposing sparsity on both noiseless events and latent clear intensities. Then, we build the eSL-Net++ network accordingly by unfolding the iteration stages of the sparse recovery algorithm and training eSL-Net++ with synthetic datasets. Finally, we present a rigorous approach to extend eSL-Net++ from the single to sequence E-SRB by event shuffles without further training.
\subsection{Dual Sparse Learning with Events} \label{sparse-learning}
Exploiting sparsity techniques, we assume that the LR sharp clear image $ {\boldsymbol{I}} $, the HR sharp clear image $ \boldsymbol{X} $ and the accumulated events $ \bar{\boldsymbol{E}} $ can be sparsely represented by convolutional  kernels or dictionaries $ \boldsymbol{d}_I, \boldsymbol{d}_X,\boldsymbol{d}_E$, \textit{i.e.}, 
\begin{equation}\label{convkernel}
\begin{split}
    {\boldsymbol{ I}} &=  \boldsymbol{d}_I * \boldsymbol{\alpha}_{ I},\\
    {\boldsymbol{ X}} &=  \boldsymbol{d}_X * \boldsymbol{\alpha}_{ X},\\
    \bar{\boldsymbol{ E}} &=  \boldsymbol{d}_E * \boldsymbol{\beta},
\end{split}
\end{equation}
% $ {\boldsymbol{ I}} = \boldsymbol{D}_{{I}} \boldsymbol{\alpha}_{ I} $, $ \boldsymbol{X} = \boldsymbol{D}_{X} \boldsymbol{\alpha}_{X} $ and $ \bar{\boldsymbol{E}} = \boldsymbol{D}_{E} \boldsymbol{\beta}$, 
where \colored{$*$ denotes the convolutional operator and} $ \boldsymbol{\alpha}_{I} $, $ \boldsymbol{\alpha}_{X} $, and $ \boldsymbol{\beta} $ are correspondingly the sparse representations. Assume that the LR sharp clear image $ {\boldsymbol{I}} $ and the HR sharp clear image $ \boldsymbol{X} $ share the same sparse representation, \ie, $ \boldsymbol{\alpha} = \boldsymbol{\alpha}_{ I} = \boldsymbol{\alpha}_{ X}$, if $ \boldsymbol{d}_I $ and $ \boldsymbol{d}_X $ are defined properly. Therefore, given an observed blurry image $ \boldsymbol{Y} $ and the corresponding events $\mathcal{E}_{\mathcal{T}}$, we can find the sparse coefficients $\boldsymbol{\alpha}$ and $\boldsymbol{\beta}$ over the convolutional dictionaries $ \boldsymbol{d}_I $ and $ \boldsymbol{d}_E$ by solving the following problem:
\begin{equation} \label{ev-lasso}
\min _{{\boldsymbol{\alpha}}, {\boldsymbol{\beta}}} \frac{1}{2}\|{\boldsymbol{Y}}-\boldsymbol{I}\circ\bar{\boldsymbol{E}}\|_{2}^{2}+\frac{\lambda_1}{2}\|\boldsymbol{E}-\bar{\boldsymbol{E}}\|_{2}^{2}+\lambda_2\|{\boldsymbol{\alpha}}\|_{1}+\lambda_3\|{\boldsymbol{\beta}}\|_{1},
\end{equation}
where \colored{$\circ$ denotes the Hadamard product}. The first two terms represent the data fidelity and the last two terms are the sparse regularization with $ \|\cdot\|_p $ denoting the $\ell_p$-norm. The coefficients $\lambda_1$, $\lambda_2$, $\lambda_3$ balance the data fidelity and the sparse regularization. \colored{Actually, we can consider \myeqref{ev-lasso} as two sub-problems with respect to $\boldsymbol{I}$ and $\boldsymbol{E}$, and each is a typical LASSO problem~\cite{tibshirani1996regression,candes2008enhancing}.} Thus, the classical solver for \myeqref{ev-lasso} is to use the iterative shrinkage thresholding algorithm (ISTA)~\cite{daubechies2004an} and the solutions can be found via the following iterative updates:
\begin{equation}\label{ev-ista}
\begin{split}
\boldsymbol{\alpha}^+=\Gamma_{\eta\lambda_2}&\left(\boldsymbol{\alpha}-\eta\boldsymbol{d}_{I}^T*\bar{\boldsymbol{E}}\circ\left(\bar{\boldsymbol{E}}\circ(\boldsymbol{d}_I *\boldsymbol{\alpha})-\boldsymbol{Y}\right)\right),\\
% \boldsymbol{\beta}^+&=\Gamma_{\eta\lambda_3}\left(\boldsymbol{\beta}-\eta\boldsymbol{d}_{E}^T*\left( ({\boldsymbol{I}} \circ {\boldsymbol{I}}+\lambda_1\boldsymbol{A})\circ \boldsymbol{d}_E*\boldsymbol{\beta}- ({\boldsymbol{I}}+\lambda_1\boldsymbol{E})\right)\right)\\
\boldsymbol{\beta}^+=\Gamma_{\eta\lambda_3}&\left(\boldsymbol{\beta}-\eta\boldsymbol{d}_{E}^T*{\boldsymbol{I}} \circ \left( \boldsymbol{I}\circ (\boldsymbol{d}_E*\boldsymbol{\beta})- \boldsymbol{Y} \right) - \right.\\ &\left.\eta\lambda_1\boldsymbol{d}_{E}^T*(\boldsymbol{d}_{E} * \boldsymbol{\beta} - \boldsymbol{E})\right),
% ({\boldsymbol{I}}+\lambda_1\boldsymbol{E})\right)\right)
\end{split}
\end{equation}
with $\bar{\boldsymbol{E}}=\boldsymbol{d}_E*\boldsymbol{\beta}$,  $\boldsymbol{I}=\boldsymbol{d}_I*\boldsymbol{\alpha}$, $\eta$ denoting the step size, and $ \Gamma_{\theta}(\boldsymbol{\iota}) = \text{sign}(\boldsymbol{\iota})\text{max}(|\boldsymbol{\iota}|-\theta,0) $ denoting the element-wise soft-thresholding function. After obtaining the optimum solution of the sparse coefficients $\boldsymbol{\alpha}$, we can finally recover the LR sharp clear image $\boldsymbol{I}$ and the HR sharp clear image $ \boldsymbol{X} $ according to \myeqref{convkernel}.

% the  are optimized in a manner of turbo-like iterations. 

The iterative updates \myeqref{ev-ista} provide a DSL scheme for the task of E-SRB, which implicitly unifies noise suppression and motion deblurring. In the DSL scheme, the sparse coefficients $\boldsymbol{\alpha}$ of the intensity images and $\boldsymbol{\beta}$ of the events can mutually compensate each other by leveraging the smoothness of frames and high temporal resolution of events. On the other hand, when eliminating the updates of $\boldsymbol{\beta}$, equation \myeqref{ev-ista} degenerates to the preliminary version, \ie, eSL-Net~\cite{wang2020event}, where $\boldsymbol{E}$ is directly fed without the noise suppression and mutual enhancement.

\subsection{Network Modules} \label{esl-network}

% Inspired by~\cite{gregor2010learning}, we can solve the sparse coding problem efficiently by integrating it into the CNN architecture.~\cite{gregor2010learning}

Instead of directly optimizing \eqref{ev-lasso}, we unfold the iterations \eqref{ev-ista} and build a CNN architecture for sparse learning, which consists of a fixed number of phases corresponding to the iterations of \myeqref{ev-ista}. 

%Therefore eSL-Net++ is an interpretable deep network.
% Therefore we propose an Event-enhanced Sparse Learning Network (\textbf{eSL-Net++}) to solve problem of noises, motion blurs and low spatial resolution in a unified framework.

% Thus we divide the collected events $\boldsymbol{\mathcal{E}}_\mathcal{T}$ into $K$ equal time intervals, \ie, $\boldsymbol{\mathcal{E}}_1,\boldsymbol{\mathcal{E}}_2,...,\boldsymbol{\mathcal{E}}_K$, as shown in Fig.~\ref{input event frame}a. 

\noindent\textbf{Dual Sparse Learning Module.}
The DSL scheme is shown in Fig.~\ref{fig:fusionframework}, where $\{\bar{\boldsymbol{E}}_i\}_{i=0,...,n}$ and $\{\boldsymbol{I}_i\}_{i=0,...,n}$ are respectively the reconstructions of the accumulated events $\bar{\boldsymbol{E}}$ and the restored sharp image $\boldsymbol{I}$ at the $i$-th iteration via updated sparse coefficients $\boldsymbol{\alpha}_i$ and $\boldsymbol{\beta}_i$. Obviously, DSL jointly optimizes both $\boldsymbol{\alpha}$ and $\boldsymbol{\beta}$ in a manner of mutual compensation that efficiently leverages information from image frames and events. Specifically, the DSL of \myeqref{ev-ista} degenerates to its preliminary version~\cite{wang2020event} where noisy events $\boldsymbol{E}$ are directly used to update $\boldsymbol{\alpha}$ and thus may produce noisy reconstructions. 

Instead of directly inputting image frame $\boldsymbol{Y}$ to the DSL module, we linearly transform it into the feature domain, \ie, $\hat{\boldsymbol{Y}}$. Then both $\hat{\boldsymbol{Y}}$ and $\boldsymbol{E}$ are fed into the DSL block to compute the sparse coefficients. The DSL block is implemented in a recursive manner that is composed of $n$ recursive blocks $\boldsymbol{F}_1, ..., \boldsymbol{F}_n$ and the unfolded details are depicted in Fig.~\ref{fig:fusionframework} where the ReLU layer is employed to implement $\Gamma_\theta$. In summary, the DSL module accepts $\boldsymbol{E}$ and $\hat{\boldsymbol{Y}}$ and outputs sparse coefficients $\boldsymbol{\alpha,\beta}$, \ie,
\[
\begin{split}
\boldsymbol{\alpha,\beta} &= \mbox{DSL}(\boldsymbol{E},\hat{\boldsymbol{Y}}).
\end{split}
\]

\noindent\textbf{Learnable Double Integral Module.}
According to \eqref{eq:etr}, calculating $\boldsymbol{E}(f)$ from events $\boldsymbol{\mathcal{E}}_\mathcal{T}$ requires two integral operations. To approximate such integration, a {\it l}earnable {\it d}ouble {\it i}ntegral (LDI) network is proposed where two convolution layers and two sigmoid layers are adopted as shown in Fig.~\ref{fig:highframework}. And the input and output relation of LDI can be written as
\[
    \boldsymbol{E} = \mbox{LDI}(\boldsymbol{\mathcal{E}}_\mathcal{T}).
\]

\noindent\textbf{Reconstruction Module.}
Then, the outputs of the last recursion $\boldsymbol{F_n}$ are optimized sparse codes $\boldsymbol{\alpha}$ and $\boldsymbol{\beta}$. 
% According to \myeqref{convkernel}, we can easily acquire LR/SR sharp clear image reconstruction. 
Then we use convolutional layers followed by the shuffle layer~\cite{Ahn_2018_ECCV} as the HR conventional kernel $\boldsymbol{d}_{X}$ to reconstruct the final sharp clear HR image frame $\boldsymbol{X}$. Besides, the denoised events $\bar{\boldsymbol{E}}$ can also be easily obtained by $\boldsymbol{d}_{E}*\boldsymbol{\beta}$. Finally, the input and output relation of the {\it REC}onstruction (REC) module can be written as
\[
    \boldsymbol{X},\bar{\boldsymbol{E}} = \mbox{REC}(\boldsymbol{\alpha},\boldsymbol{\beta}).
\]
\subsection{Overall eSL-Net++ and Its Training Strategy}
The overall architecture of the proposed network for E-SRB is shown in \myfigref{fig:highframework}. The eSL-Net++ is fed with a pair of inputs, \ie, the LR blurry image frame $\boldsymbol{Y} $ and the corresponding events during the exposure time interval $\mathcal{T}$, and outputs the HR sharp clear intensity image $\boldsymbol{X}$, \ie,
\begin{equation}\label{eq:eslnet}
\begin{split}
\boldsymbol{X},\bar{\boldsymbol{E}} &= \mbox{eSL-Net++}\left(\boldsymbol{Y} ,\boldsymbol{\mathcal{E}}_{\mathcal{T}}\right)
\\
&\triangleq \mbox{REC}\left(\mbox{DSL}\left(\mbox{LDI}\left(\boldsymbol{\mathcal{E}}_\mathcal{T}\right),\hat{\boldsymbol{Y}}\right)\right).
\end{split}
\end{equation}
% by supervision of the ground-truth $\boldsymbol{X}^{gt}$.% and the noise-free events $\boldsymbol{\mathcal{E}}^{gt}_{\mathcal{T}}$.

The proposed eSL-Net++ is trained over the synthetic dataset, where noise free events $\boldsymbol{\mathcal{E}}_{\mathcal{T}}^{gt}$ and the ground truth HR sharp image $\boldsymbol{X}^{gt}$ are both available. Correspondingly, the total training loss is composed of two aspects, \ie, event denoising error and image SR error:
\begin{equation}\label{eq:loss}
    \mathcal{L} = \zeta_1 \mathcal{L}_E + \zeta_2 \mathcal{L}_{X},
\end{equation}
where the event denoising error is defined as
$
    \mathcal{L}_E = \|\boldsymbol{E}^{gt} - \bar{\boldsymbol{E}}\|_1
$
with $\boldsymbol{E}^{gt} = \mbox{LDI}(\boldsymbol{\mathcal{E}}_{\mathcal{T}}^{gt})$, the image SR error is defined as
$
    \mathcal{L}_X = \|\boldsymbol{X}^{gt} - {\boldsymbol{X}}\|_1
$, and $\zeta_1,\zeta_2$ are balancing parameters. \colored{In our experiments, we set $[\zeta_1,\zeta_2]=[1,1]$.}
% where $\boldsymbol{E}^{gt} = \mbox{LDI}(\boldsymbol{\mathcal{E}}_{\mathcal{T}}^{gt})$.

\subsection{From Single to Sequence E-SRB by Event Shuffles}

% \textbf{Single-Frame E-SRB.} 
Training eSL-Net++ with the ground truth HR sharp image $\boldsymbol{X}^{gt}(f)$ at the specific time $f\in \mathcal{T}$, one can obtain a solver for the task of the single-frame E-SRB \eqref{eq:srb} at time $f$. Thus, we name the resultant network to resolve the HR image at time $f$ as \textit{eSL-Net++$_f$} and the corresponding LDI as \textit{LDI$_f$}, while the modules of DSL and REC are independent of $f$.

% \textbf{Sequence E-SRB.} 
Even though one can train eSL-Net++ for different time $f$ to get the sequence solver of E-SRB, it is time consuming for the training stage and cannot solve E-SRB for arbitrary time stamps. Thus we propose an easy method to extend the single-frame solver to the sequence-frame solver without any additional training procedure.

\begin{figure}[!htb]
	\centering
	% \subfigure[]{
		\begin{minipage}[t]{0.49\linewidth}
			\centering
			\includegraphics[height=5.5cm]{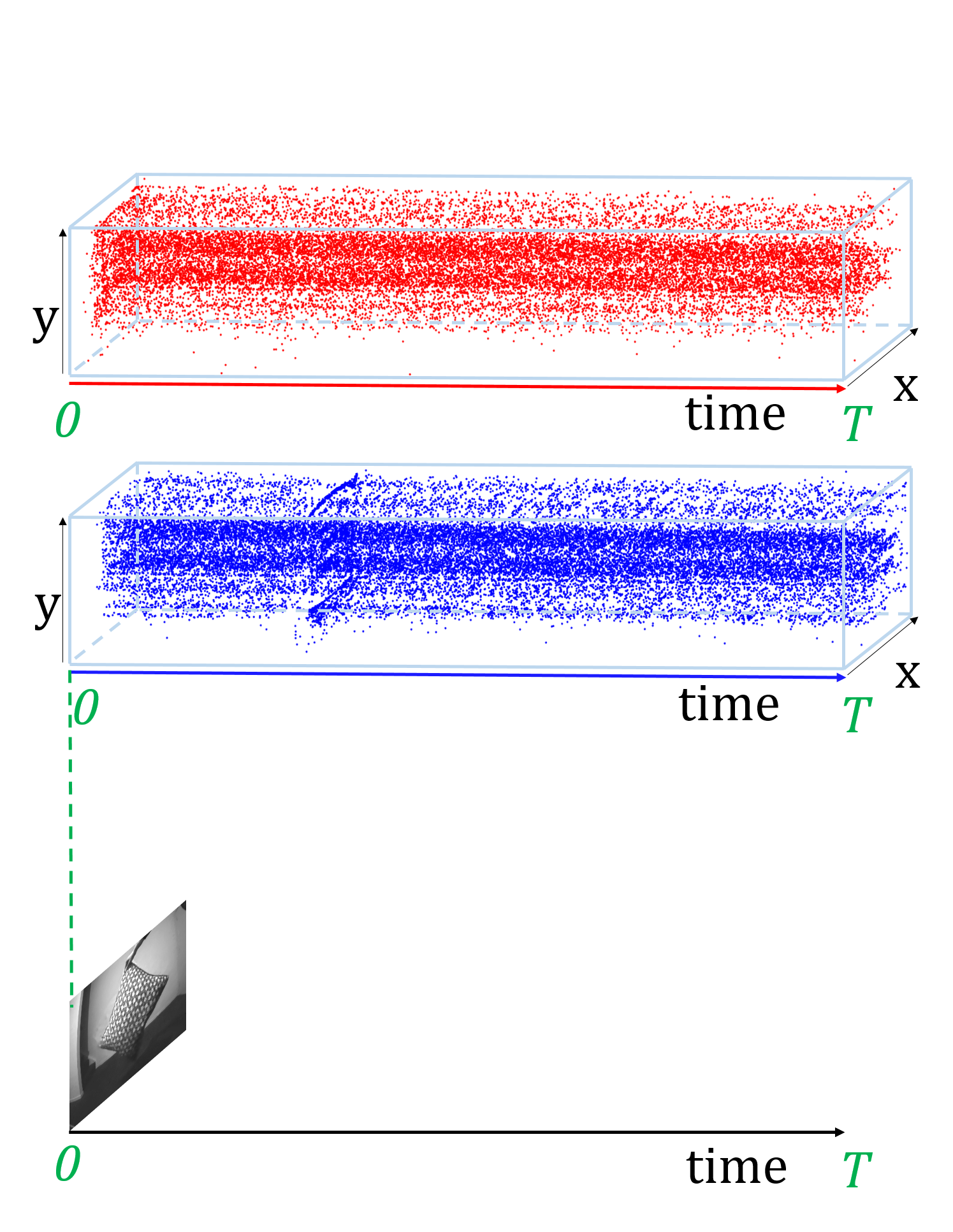}
			%\caption{fig1}
		\end{minipage}%
	% }
 \hspace*{-1.5mm}
	% \subfigure[]{
		\begin{minipage}[t]{0.49\linewidth}
			\centering
			\includegraphics[height=5.5cm]{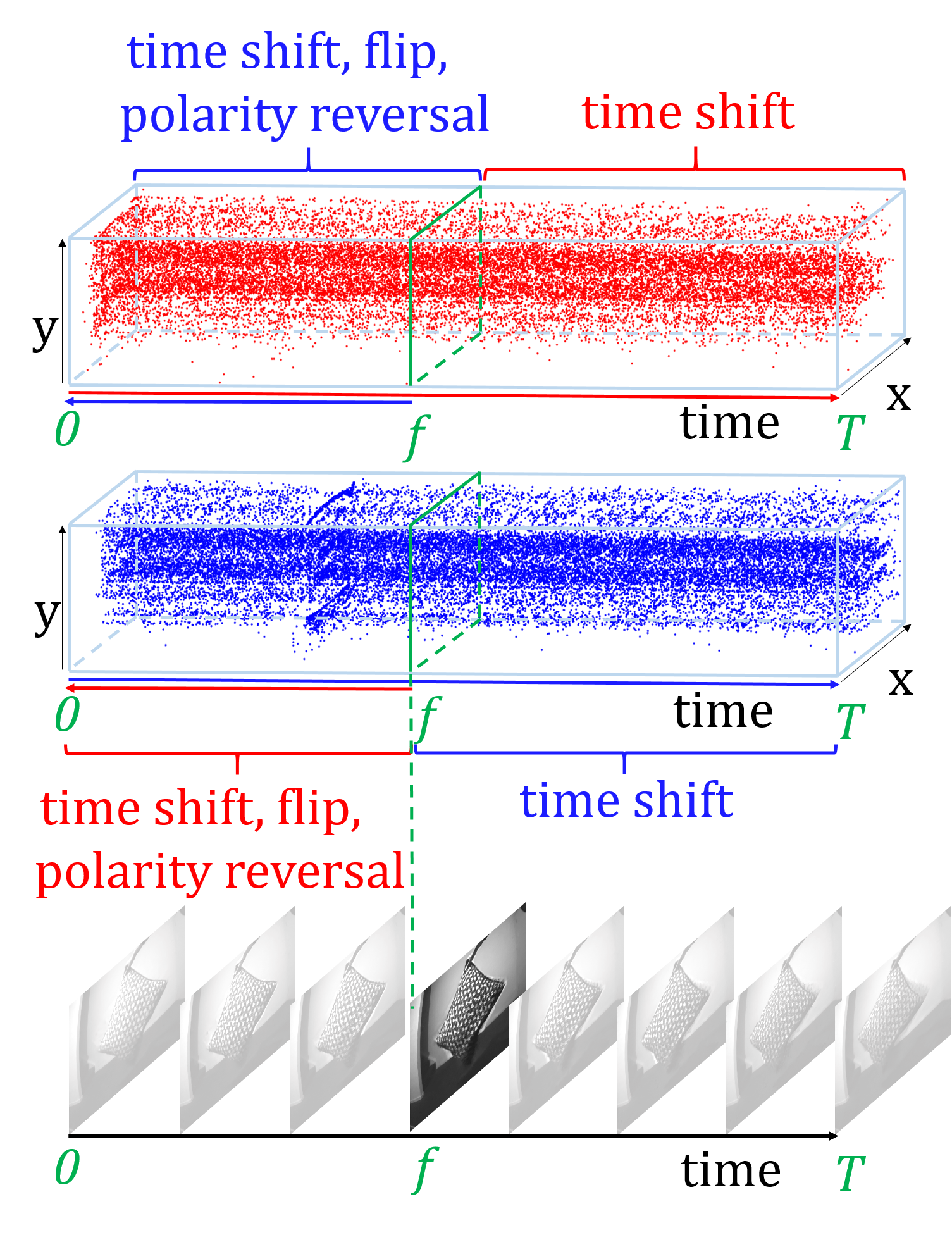}
			%\caption{fig1}
		\end{minipage}%
	% }
% 	这个图要修改一下，图中的符号都要整理清楚。
	\caption{The {\it R}igorous {\it E}vent {\it S}huffle-and-{\it M}erge (RESM) scheme: (a) the input event tensors when reconstructing the image of time $f=0$ (from top to bottom are  positive event stream, negative event stream, and recovered latent image of time $f=0$); (b) the shuffle of events to input tensors when reconstructing the image of time $f > 0$.}	
	\label{input event frame}
\end{figure}

Suppose that eSL-Net++$_0$ has been trained, LDI$_0$ can be considered as the approximation of $\boldsymbol{E}(0) \approx \mbox{LDI$_0$}(\boldsymbol{\mathcal{E}}_\mathcal{T})$, \ie,
\begin{equation}\label{eq:ldinet0}
    \mbox{LDI$_0$}(\boldsymbol{\mathcal{E}}_\mathcal{T}) \approx \frac{1}{T}\int_0^T\exp\left(c\int_0^t e(s) ds\right)dt.
\end{equation}
% According to \myeqref{general_model}, E-SRB can reconstruct clear image frame at any time stamp $f\in \mathcal{T}$ which is dependent to the double integral $\boldsymbol{E}(f)$. 
% \[
% \begin{split}
% \mbox{LDI-$0$}\left(\boldsymbol{\mathcal{E}}_{\mathcal{T}_{[0,f)}}\right) &\approx \frac{1}{f}\int_0^f\exp\left(c\int_0^t e(s) ds\right)dt \\
% \mbox{LDI-$0$}\left(\boldsymbol{\mathcal{E}}_{\mathcal{T}_{[0,T-f)}}\right) &\approx \frac{1}{f}\int_0^f\exp\left(c\int_0^t e(s) ds\right)dt 
% \end{split}
% \]
From \myeqref{eq:etr}, we have
% \begin{equation}\label{eq:ef_new}
% \small
\begin{align}\label{eq:ef_new}
        \boldsymbol{E}(f) 
        % =&  \frac{1}{T}\int_{0}^T \exp\left(c\int_{f}^{t}e(s)ds\right) dt \\
        =& \frac{1}{T}\!\left(\!\int_{0}^{f} \!\exp\!\left(\!c\!\int_{f}^{t}\! e(s)ds\!\right)\! dt \! + \! \int_{f}^{T}\! \exp\!\left(\!c\!\int_{f}^{t}\!e(s)ds\!\right)\! dt \!\right) \notag\\
        % =&  \int_{0}^{f} \exp\left(c\int_{0}^{f-t}-e(-s-f)ds\right) dt +  \\
        % &\int_{f}^{T} \exp\left(c\int_{0}^{t-f}e(s-f)ds\right) dt \\
        =& \frac{1}{T}\int_{0}^{f} \exp\left(c\int_{0}^{t'}-e(-s+f)ds\right) dt'  \notag \\
        & +\frac{1}{T}\int_{0}^{T-f} \exp\left(c\int_{0}^{t'}e(s+f)ds\right) dt' ,
        % &\approx \frac{1}{T}\left(\mbox{LDI-0}(\boldsymbol{\mathcal{E}}_{[0,f)}) + \mbox{LDI-0}(\boldsymbol{\mathcal{E}}_{[f,T)})\right)
    \end{align}
% \end{equation}
where the first term is related to the events triggered in $[0,f)$ and the second term is related to the events in $[f,T)$. Correspondingly, we can divide events $\boldsymbol{\mathcal{E}}_\mathcal{T}$ into two subsets, \ie, $\boldsymbol{\mathcal{E}}_{{[0,f)}}$ and $\boldsymbol{\mathcal{E}}_{{[f,T)}}$. Then one can approximate $\boldsymbol{E}(f)$ via LDI$_0$ according to \eqref{eq:ldinet0} and \eqref{eq:ef_new}, which corresponds to a {\it rigorous event shuffle-and-merge} (RESM) scheme, \ie,
% \begin{equation}\label{eq:ldief}
% \small
\begin{align}\label{eq:ldief}
    \boldsymbol{E}(f) \triangleq & \mbox{RESM}(\boldsymbol{\mathcal{E}}_{\mathcal{T}},f) \\
    \approx & \frac{f}{T}\cdot \mbox{LDI$_0$}\left(\mathcal{R}\left(\boldsymbol{\mathcal{E}}_{{[0,f)}}\right)\right)  \notag \\ & + \left(1-\frac{f}{T}\right)\cdot \mbox{LDI$_0$}\left(\mathcal{S}\left(\boldsymbol{\mathcal{E}}_{{[f,T)}}\right)\right),
\end{align}
% \end{equation}
where $\mathcal{R}$ and $\mathcal{S}$ are event shuffle operators, \ie,
% \[
% \begin{split}
$\mathcal{R}\left(\boldsymbol{\mathcal{E}}_{{[0,f)}}\right) \triangleq \{-e(-t+f),t\in [0,f)\} $ represents the event operator consisting of time shift, flip, and polarity reversal, and $\mathcal{S}\left(\boldsymbol{\mathcal{E}}_{{[f,T)}}\right) \triangleq \{e(t+f),t\in [0,T-f)\} $ represents the event operator of time shift, as shown in Fig.~\ref{input event frame}(b). 
    % \end{split}
% \]

It implies by \eqref{eq:ldief} that the double integral $\boldsymbol{E}(f)$ can be approximately calculated with LDI$_0$ by simply shuffling collected events $\boldsymbol{\mathcal{E}}_\mathcal{T}$. Thus for arbitrary time $f$, we can get the corresponding HR sharp reconstruction $\boldsymbol{X}(f)$,
\[
    \boldsymbol{X}(f),\bar{\boldsymbol{E}}(f) = \mbox{REC}\left(\mbox{DSL}\left(\boldsymbol{E}(f),\hat{\boldsymbol{Y}}\right)\right), %\mbox{eSL-Net-0}(\boldsymbol{Y},\boldsymbol{E}(f))
\]
with $\bar{\boldsymbol{E}}(f)$ the denoised version of $\boldsymbol{E}(f)$. Theoretically, we can generate a video with frame-rate as high as the DVS's ({\it D}ynamic {\it V}ision {\it S}ensor) eps ({\it e}vents {\it p}er {\it s}econd).

\def\imwidth{0.2}

\def\ssxxs{(1.5,0.7)} % 图案
\def\ssxxsr{(-1.4,-0.1)} % 门

\def\ssxxsz{(1.55,0.72)} % 图案中间帧

\def\ssyys{(1.8,-2.0)}
\def\ssyysr{(-0.05,-2.0)}

\def\hqfred{(0.8,-0.5)}
\def\hqfgreen{(0.2,-0.8)}

\def\hqfredz{(0.725,-0.507)}%中间帧
\def\hqfgreenz{(0.095,-0.82)}%中间帧

\def\hqfredpos{(1.8,-2.35)}
\def\hqfgreenpos{(-0.05,-2.35)}

\def\ssizz{1.77cm}
\def\sswidth{0.245\textwidth}
\def\ssmag{5}
\def\scc{(2.12,1.4)}
\begin{figure*}[!ht]
\footnotesize
	\centering
    % \subfigure{
    	\begin{minipage}[t]{\imwidth\linewidth}
    		\centering
    		%gopro blurry image and gt
			\begin{tikzpicture}[spy using outlines={green,magnification=\ssmag,size=\ssizz},inner sep=0]
				\node {\includegraphics[width=\linewidth]{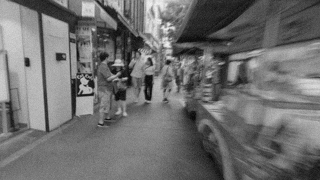}};
				\spy on \ssxxsr in node [left] at \ssyysr;
				\spy [red] on \ssxxs in node [left,red] at \ssyys;
				\end{tikzpicture}
            Blurry Image\\
            (PSNR, SSIM)\vspace{0.5em}
            \begin{tikzpicture}[spy using outlines={green,magnification=\ssmag,size=\ssizz},inner sep=0]
				\node {\includegraphics[width=\linewidth]{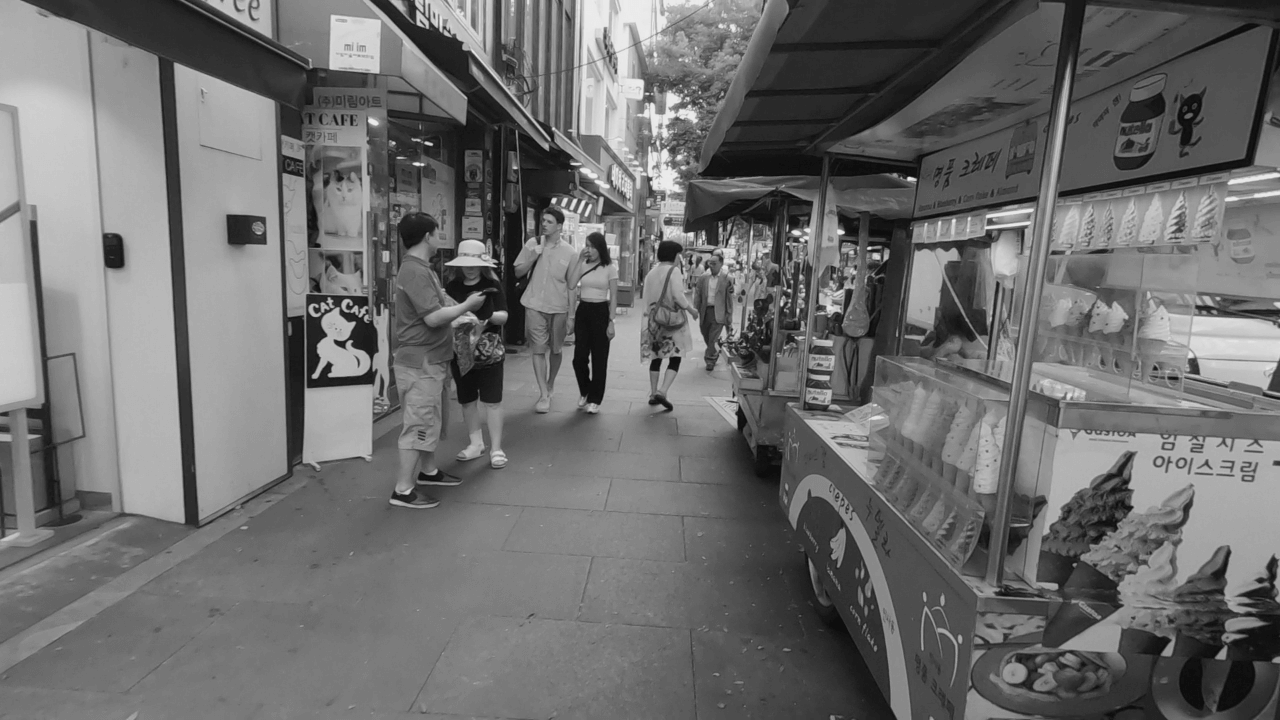}};
				\spy on \ssxxsr in node [left] at \ssyysr;
				\spy [red] on \ssxxs in node [left,red] at \ssyys;
				\end{tikzpicture}
            
            Ground Truth\\
            (PSNR, SSIM)\vspace{0.5em}
            
            %HQF blurry image and gt
            \begin{tikzpicture}[spy using outlines={green,magnification=\ssmag,size=\ssizz},inner sep=0]
				\node {\includegraphics[width=\linewidth]{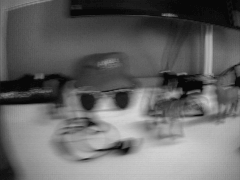}};
				\spy on \hqfgreen in node [left] at \hqfgreenpos;
				\spy [red] on \hqfred in node [left,red] at \hqfredpos;
				\end{tikzpicture}
            Blurry Image\\
            (PSNR, SSIM)\vspace{0.5em}
            \begin{tikzpicture}[spy using outlines={green,magnification=\ssmag,size=\ssizz},inner sep=0]
				\node {\includegraphics[width=\linewidth]{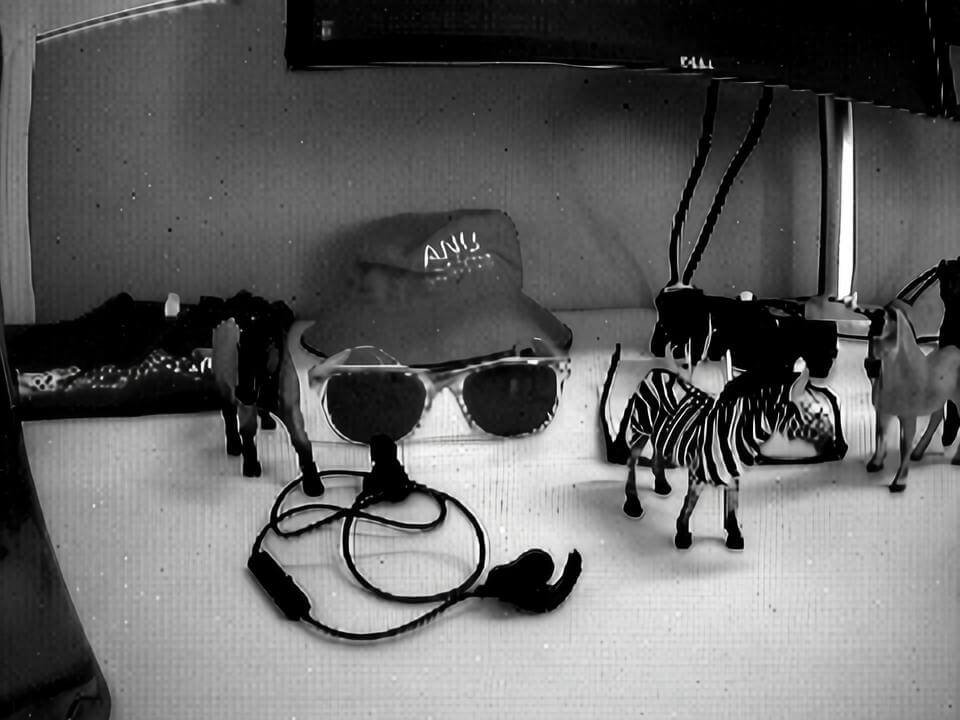}};
				\spy on \hqfgreen in node [left] at \hqfgreenpos;
				\spy [red] on \hqfred in node [left,red] at \hqfredpos;
				\end{tikzpicture}
            
            Ground Truth\\
            (PSNR, SSIM)\vspace{0.5em}
            
    	\end{minipage}%
    % }
    \hspace*{0mm}
    % \subfigure{
    	\begin{minipage}[t]{\imwidth\linewidth}
    		\centering
    		%gopro GFN and LEDVDI+RCAN
    		\begin{tikzpicture}[spy using outlines={green,magnification=\ssmag,size=\ssizz},inner sep=0]
				\node {\includegraphics[width=\linewidth]{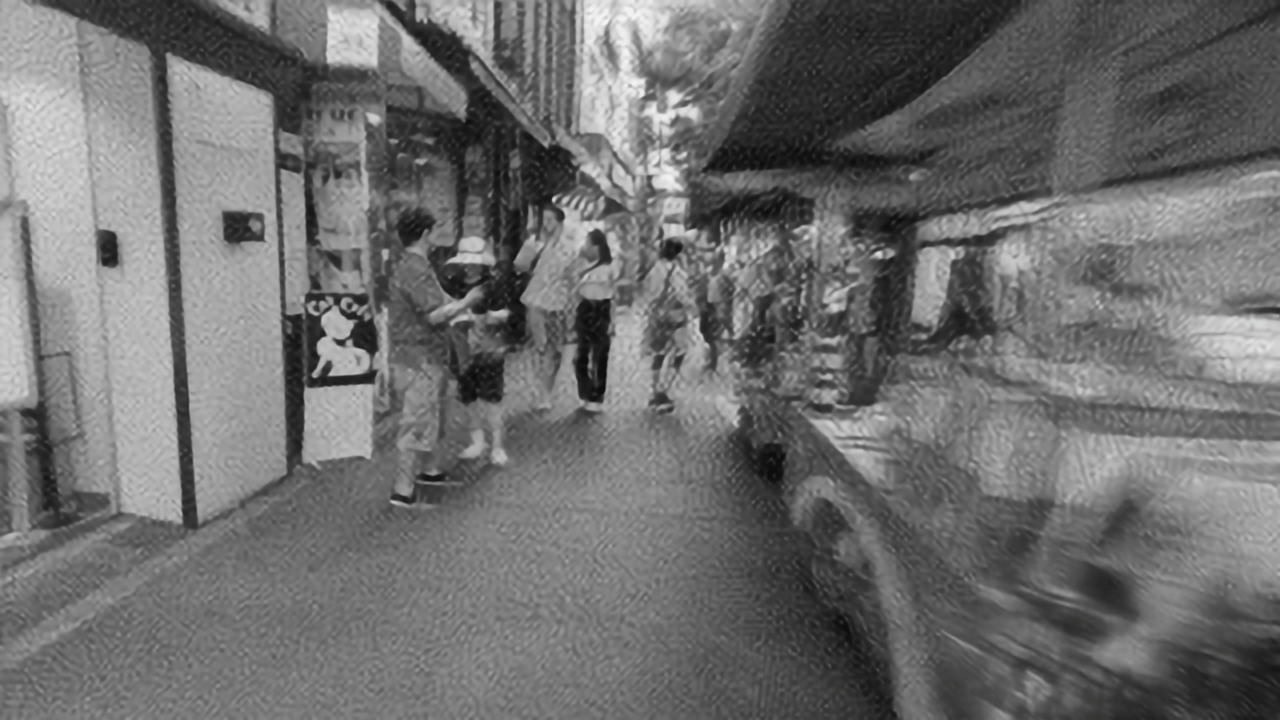}};
				\spy on \ssxxsr in node [left] at \ssyysr;
				\spy [red] on \ssxxsz in node [left,red] at \ssyys;
				\end{tikzpicture}
			GFN\\
			(22.83, 0.6142)\vspace{0.5em}
			
			\begin{tikzpicture}[spy using outlines={green,magnification=\ssmag,size=\ssizz},inner sep=0]
				\node {\includegraphics[width=\linewidth]{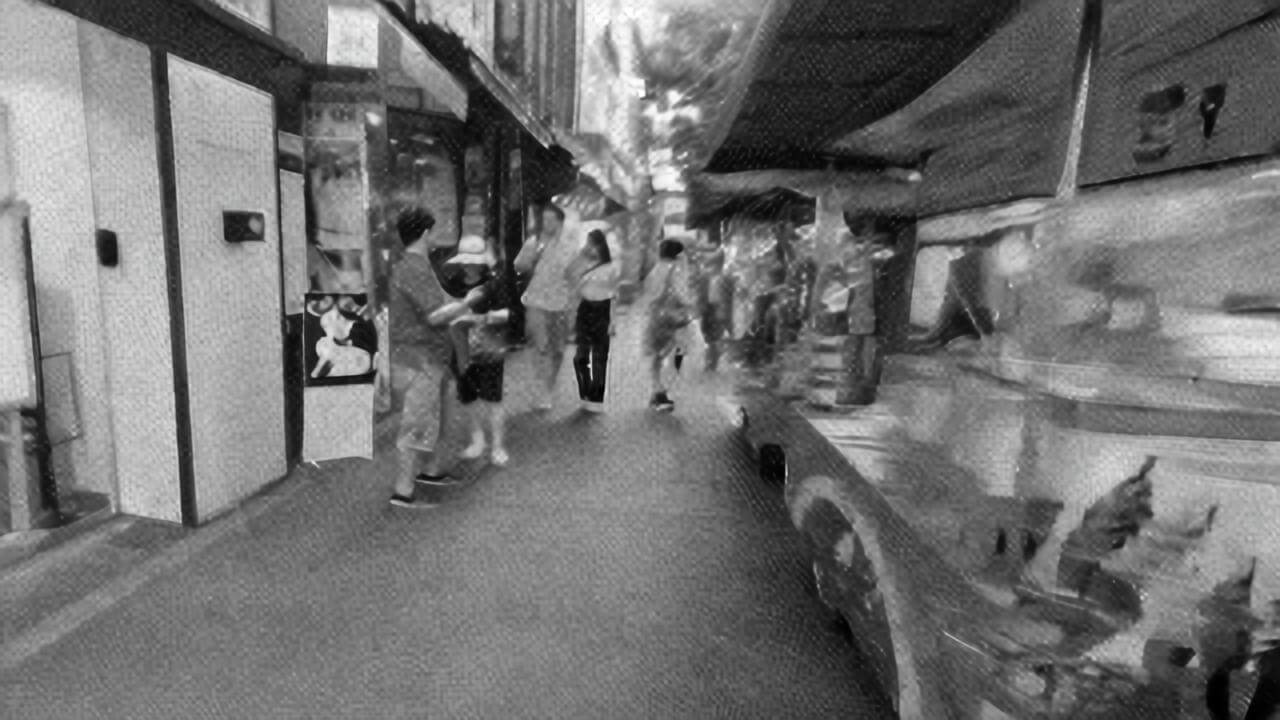}};
				\spy on \ssxxsr in node [left] at \ssyysr;
				\spy [red] on \ssxxsz in node [left,red] at \ssyys;
				\end{tikzpicture}
			LEDVDI+RCAN\\
		    (24.09, 0.6336)\vspace{0.5em}
			
			%HQF gopro GFN and LEDVDI+RCAN
			\begin{tikzpicture}[spy using outlines={green,magnification=\ssmag,size=\ssizz},inner sep=0]
				\node {\includegraphics[width=\linewidth]{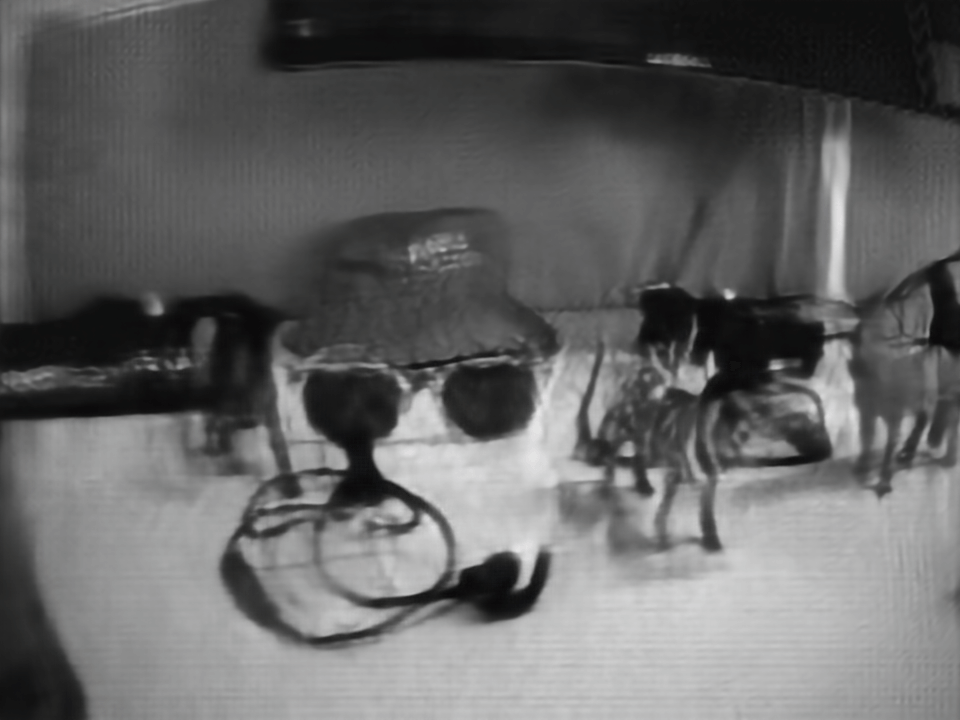}};
				\spy on \hqfgreenz in node [left] at \hqfgreenpos;
				\spy [red] on \hqfredz in node [left,red] at \hqfredpos;
				\end{tikzpicture}
			GFN\\
			(21.73, 0.7612)\vspace{0.5em}
			
			\begin{tikzpicture}[spy using outlines={green,magnification=\ssmag,size=\ssizz},inner sep=0]
				\node {\includegraphics[width=\linewidth]{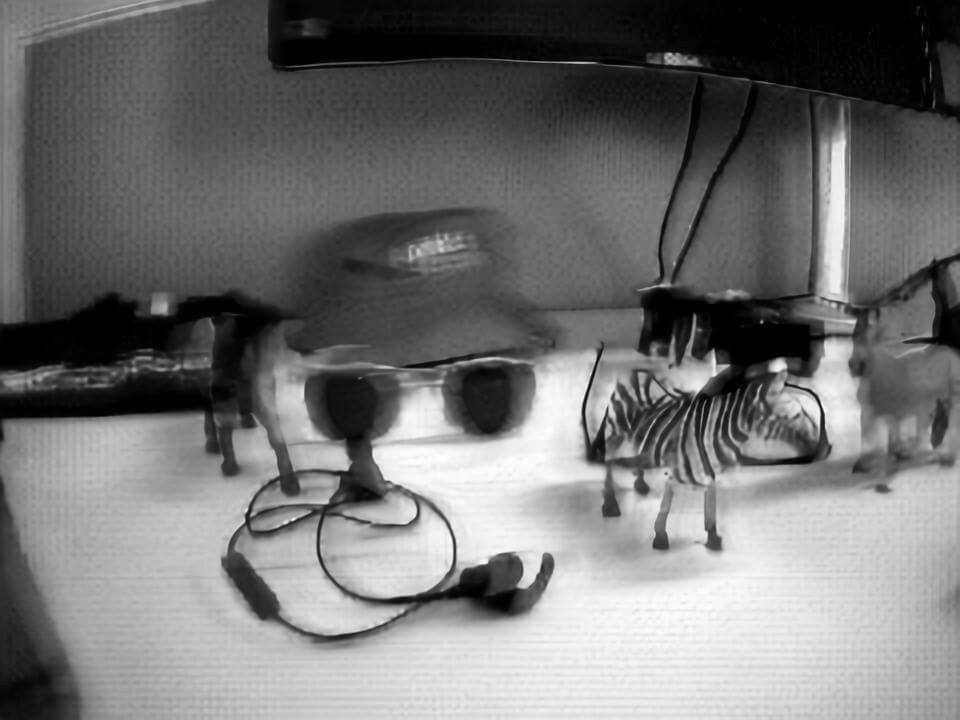}};
				\spy on \hqfgreenz in node [left] at \hqfgreenpos;
				\spy [red] on \hqfredz in node [left,red] at \hqfredpos;
				\end{tikzpicture}
			LEDVDI+RCAN\\
		    (20.36, 0.7790)\vspace{0.5em}
    	\end{minipage}%
    % }
    \hspace*{0mm}
    % \subfigure{
    	\begin{minipage}[t]{\imwidth\linewidth}
    		\centering
            \begin{tikzpicture}[spy using outlines={green,magnification=\ssmag,size=\ssizz},inner sep=0]
				\node {\includegraphics[width=\linewidth]{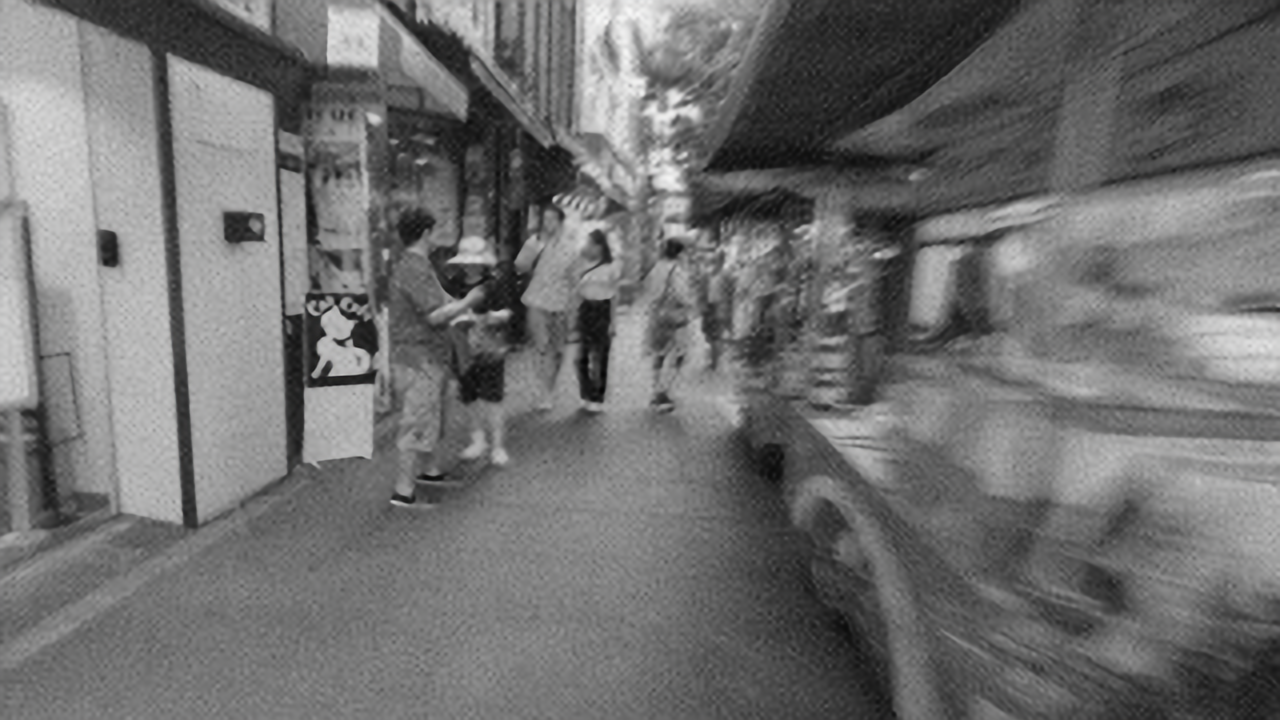}};
				\spy on \ssxxsr in node [left] at \ssyysr;
				\spy [red] on \ssxxsz in node [left,red] at \ssyys;
				\end{tikzpicture}
			DASR\\
			(20.73, 0.5710)\vspace{0.5em}
    		%gopro SRN+RCAN and RED+RCAN
   %  		\begin{tikzpicture}[spy using outlines={green,magnification=\ssmag,size=\ssizz},inner sep=0]
			% 	\node {\includegraphics[width=\linewidth]{pic/gopro/002_00000034_srn.jpg}};
			% 	\spy on \ssxxsr in node [left] at \ssyysr;
			% 	\spy [red] on \ssxxsz in node [left,red] at \ssyys;
			% 	\end{tikzpicture}
			% SRN+RCAN\\
			% (23.37, 0.6210)\vspace{0.5em}
			
			\begin{tikzpicture}[spy using outlines={green,magnification=\ssmag,size=\ssizz},inner sep=0]
				\node {\includegraphics[width=\linewidth]{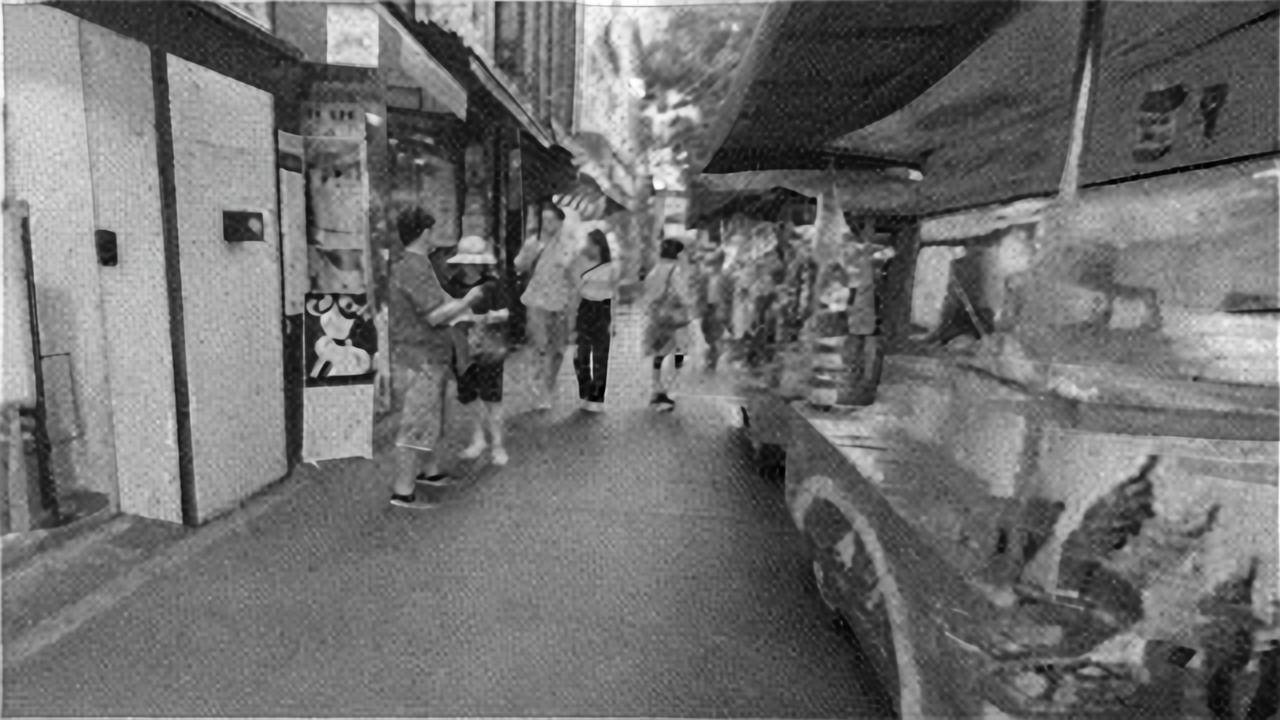}};
				\spy on \ssxxsr in node [left] at \ssyysr;
				\spy [red] on \ssxxsz in node [left,red] at \ssyys;
				\end{tikzpicture}
				
			RED-Net+RCAN\\
			(24.02, 0.6270)\vspace{0.5em}
			%(13.32, 0.4847)\vspace{0.5em} E2SRI
			%(12.22, 0.3836)\vspace{0.5em} e2vid

            \begin{tikzpicture}[spy using outlines={green,magnification=\ssmag,size=\ssizz},inner sep=0]
				\node {\includegraphics[width=\linewidth]{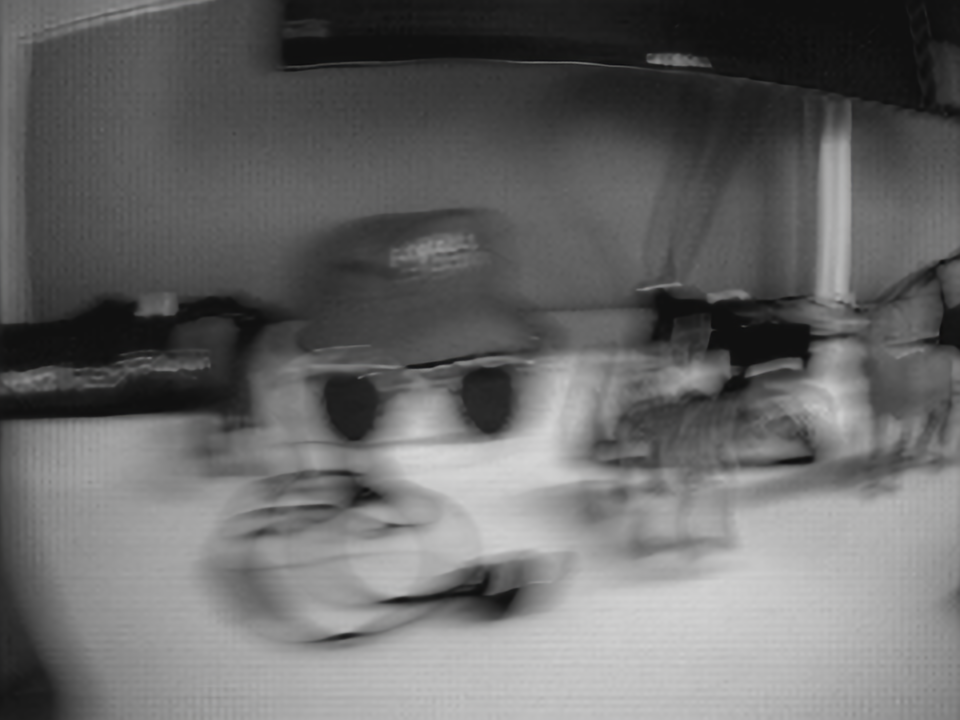}};
				\spy on \hqfgreenz in node [left] at \hqfgreenpos;
				\spy [red] on \hqfredz in node [left,red] at \hqfredpos;
				\end{tikzpicture}
			{DASR}\\
			{(16.21, 0.6249)}\vspace{0.5em}
		    %HQF SRN+RCAN and LEDVDI+RCAN
		 %    \begin{tikzpicture}[spy using outlines={green,magnification=\ssmag,size=\ssizz},inner sep=0]
			% 	\node {\includegraphics[width=\linewidth]{pic/HQF/000315_srn.jpg}};
			% 	\spy on \hqfgreenz in node [left] at \hqfgreenpos;
			% 	\spy [red] on \hqfredz in node [left,red] at \hqfredpos;
			% 	\end{tikzpicture}
			% SRN+RCAN\\
			% (20.68, 0.7509)\vspace{0.5em}
			
			\begin{tikzpicture}[spy using outlines={green,magnification=\ssmag,size=\ssizz},inner sep=0]
				\node {\includegraphics[width=\linewidth]{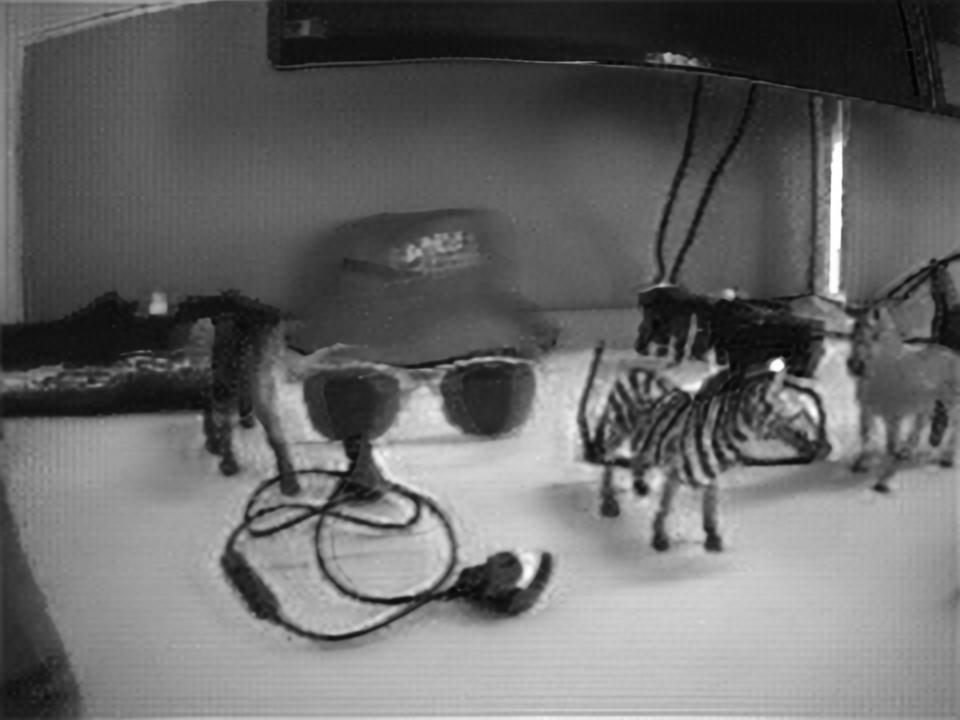}};
				\spy on \hqfgreenz in node [left] at \hqfgreenpos;
				\spy [red] on \hqfredz in node [left,red] at \hqfredpos;
				\end{tikzpicture}
				
			RED-Net+RCAN\\
		    (22.03, 0.7615)\vspace{0.5em}
			%(10.11, 0.5709)\vspace{0.5em}
    	\end{minipage}%
    % }
    \hspace*{0mm}
    % \subfigure{
    	\begin{minipage}[t]{\imwidth\linewidth}
    		\centering
    		%gopro CDVD+RCAN and eSL-Net
    		\begin{tikzpicture}[spy using outlines={green,magnification=\ssmag,size=\ssizz},inner sep=0]
				\node {\includegraphics[width=\linewidth]{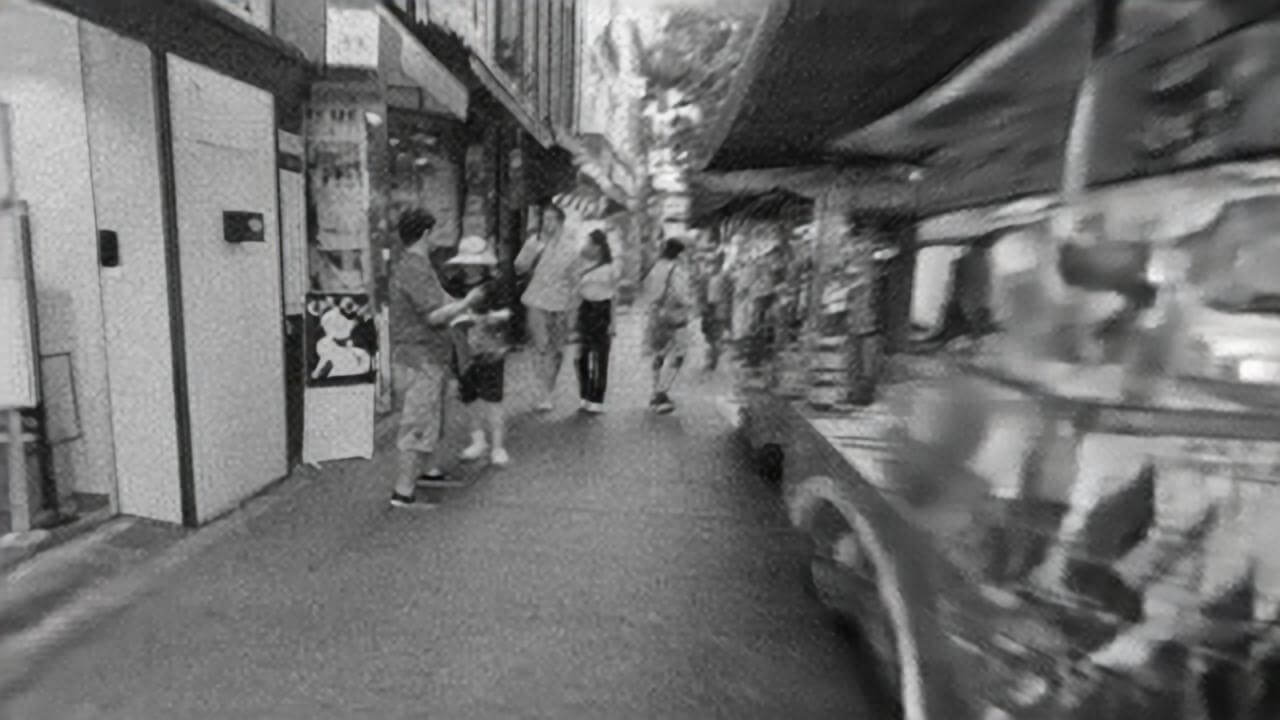}};
				\spy on \ssxxsr in node [left] at \ssyysr;
				\spy [red] on \ssxxsz in node [left,red] at \ssyys;
				\end{tikzpicture}
			CDVD+RCAN\\
			(22.89, 0.6667)\vspace{0.5em}
			
			\begin{tikzpicture}[spy using outlines={green,magnification=\ssmag,size=\ssizz},inner sep=0]
				\node {\includegraphics[width=\linewidth]{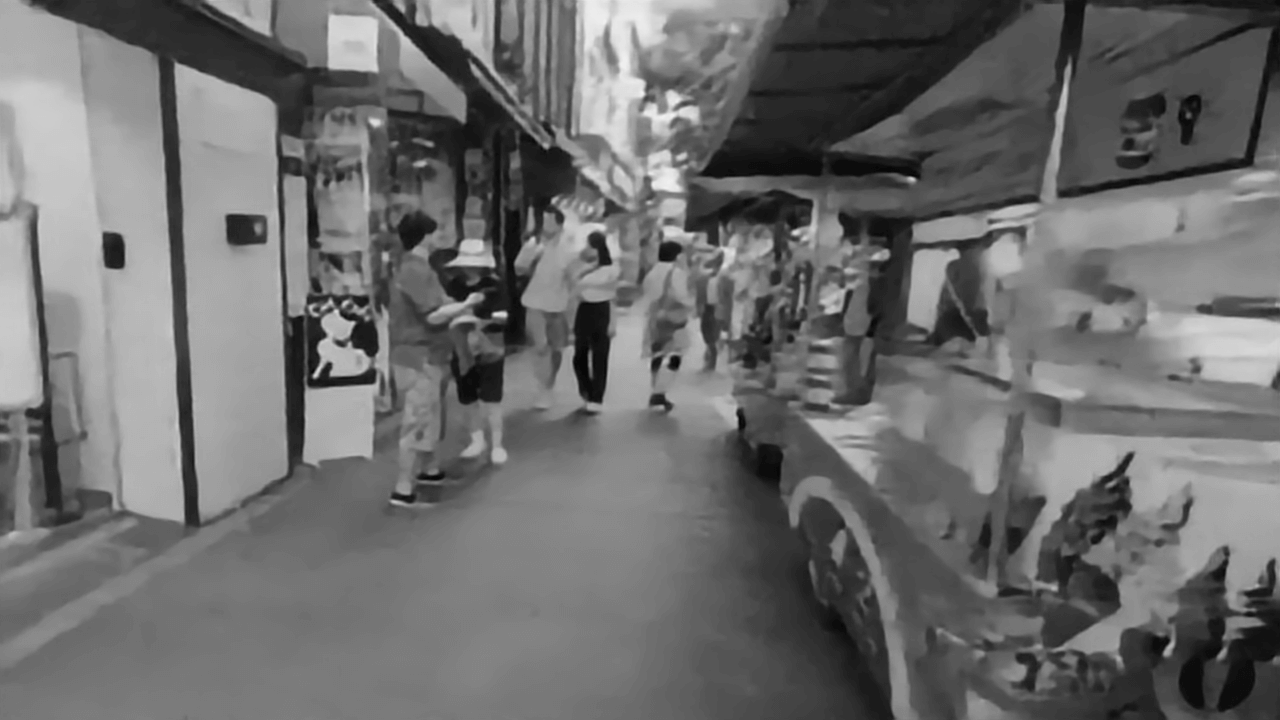}};
				\spy on \ssxxsr in node [left] at \ssyysr;
				\spy [red] on \ssxxs in node [left,red] at \ssyys;
				\end{tikzpicture}
				
			eSL-Net\\
			(25.50, 0.7600)\vspace{.5em}
			
			%HQF CDVD+RCAN and eSL-Net
			\begin{tikzpicture}[spy using outlines={green,magnification=\ssmag,size=\ssizz},inner sep=0]
				\node {\includegraphics[width=\linewidth]{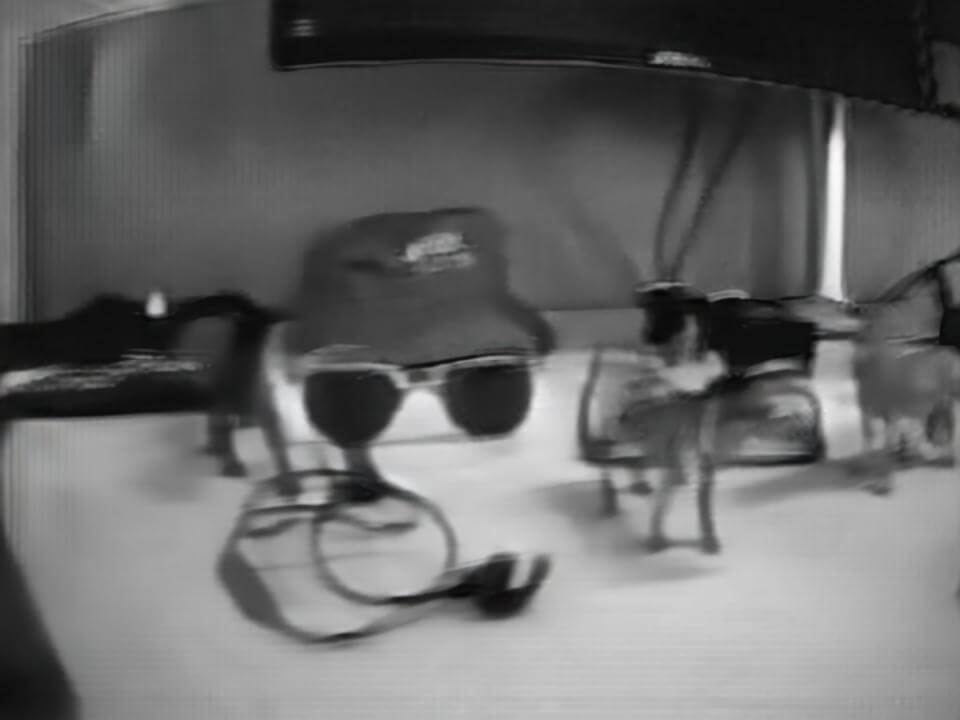}};
				\spy on \hqfgreenz in node [left] at \hqfgreenpos;
				\spy [red] on \hqfredz in node [left,red] at \hqfredpos;
				\end{tikzpicture}
			CDVD+RCAN\\
			(20.86, 0.7484)\vspace{0.5em}
			
			\begin{tikzpicture}[spy using outlines={green,magnification=\ssmag,size=\ssizz},inner sep=0]
				\node {\includegraphics[width=\linewidth]{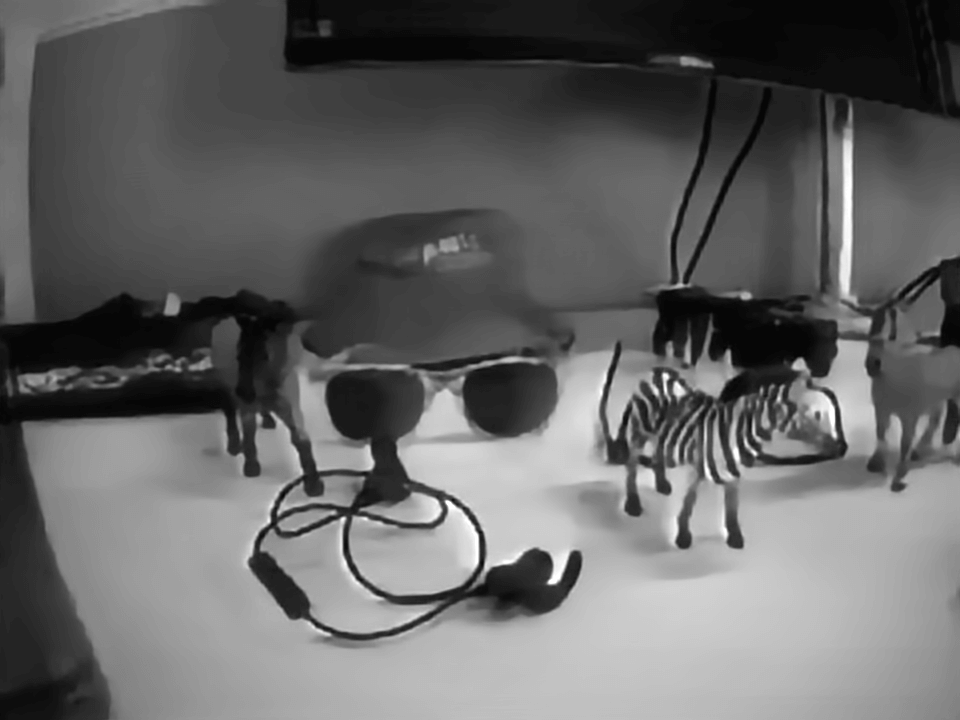}};
				\spy on \hqfgreen in node [left] at \hqfgreenpos;
				\spy [red] on \hqfred in node [left,red] at \hqfredpos;
				\end{tikzpicture}
				
			eSL-Net\\
			(23.25, 0.8016)\vspace{.5em}
    	\end{minipage}%
    % }
    \hspace*{0mm}
    % \subfigure{
    	\begin{minipage}[t]{\imwidth\linewidth}
    		\centering
    		%gopro EDI+RCAN and eSL-Net++
    		\begin{tikzpicture}[spy using outlines={green,magnification=\ssmag,size=\ssizz},inner sep=0]
				\node {\includegraphics[width=\linewidth]{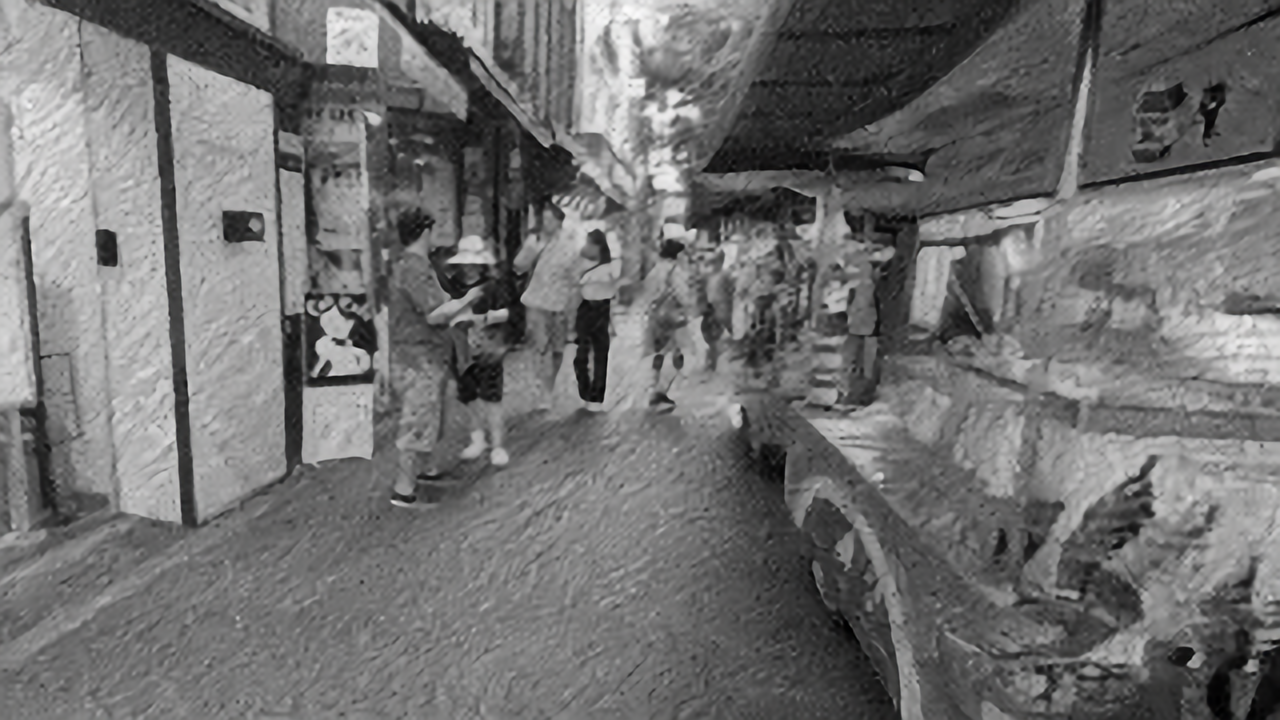}};
                \spy on \ssxxsr in node [left] at \ssyysr;
				\spy [red] on \ssxxsz in node [left,red] at \ssyys;
				% \spy on \ssxxsr in node [left] at \ssyysr;
				% \spy [red] on \ssxxsz in node [left,red] at \ssyys;
				\end{tikzpicture}
			EFNet+DASR\\
			(23.81, 0.6305)\vspace{0.5em}
			%(20.84, 0.6542)\vspace{0.5em}
			
			\begin{tikzpicture}[spy using outlines={green,magnification=\ssmag,size=\ssizz},inner sep=0]
				\node {\includegraphics[width=\linewidth]{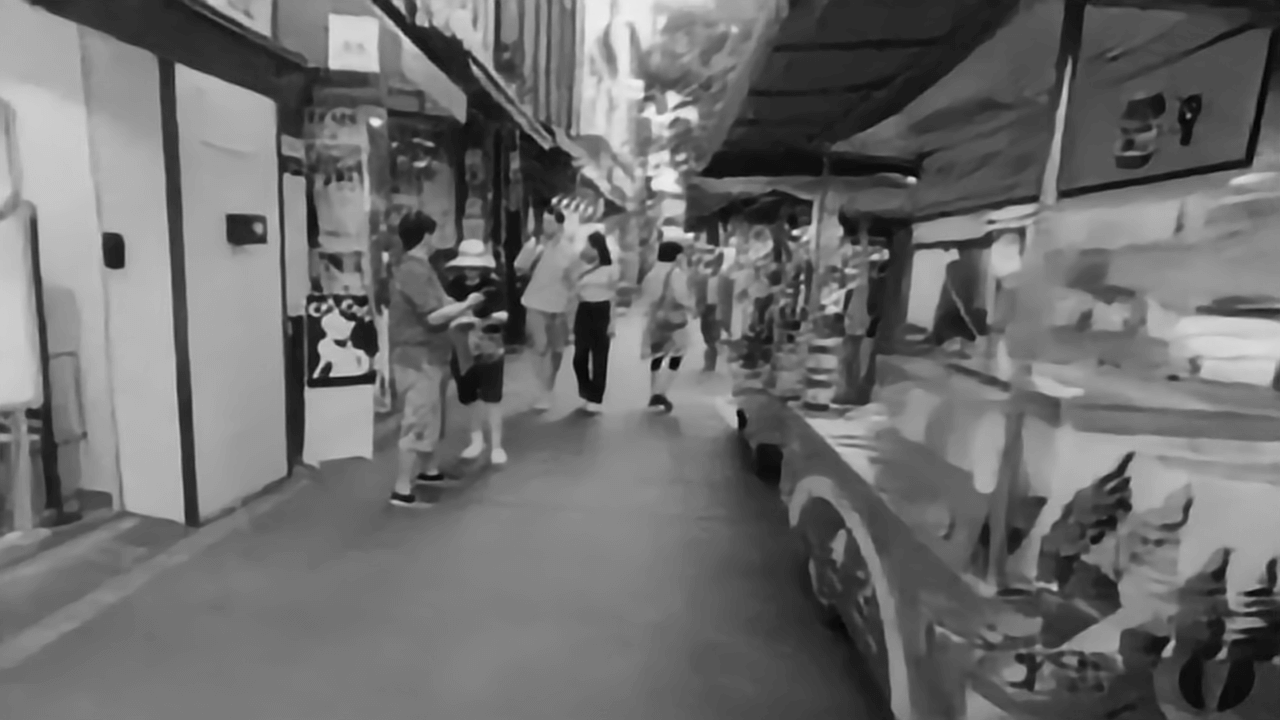}};
				\spy on \ssxxsr in node [left] at \ssyysr;
				\spy [red] on \ssxxs in node [left,red] at \ssyys;
				\end{tikzpicture}
				
    		{\bf eSL-Net++\\
    		(25.94, 0.7724)} \vspace{.5em}
    		
    		%HQF EDI+RCAN and eSL-Net++
    		\begin{tikzpicture}[spy using outlines={green,magnification=\ssmag,size=\ssizz},inner sep=0]
				\node {\includegraphics[width=\linewidth]{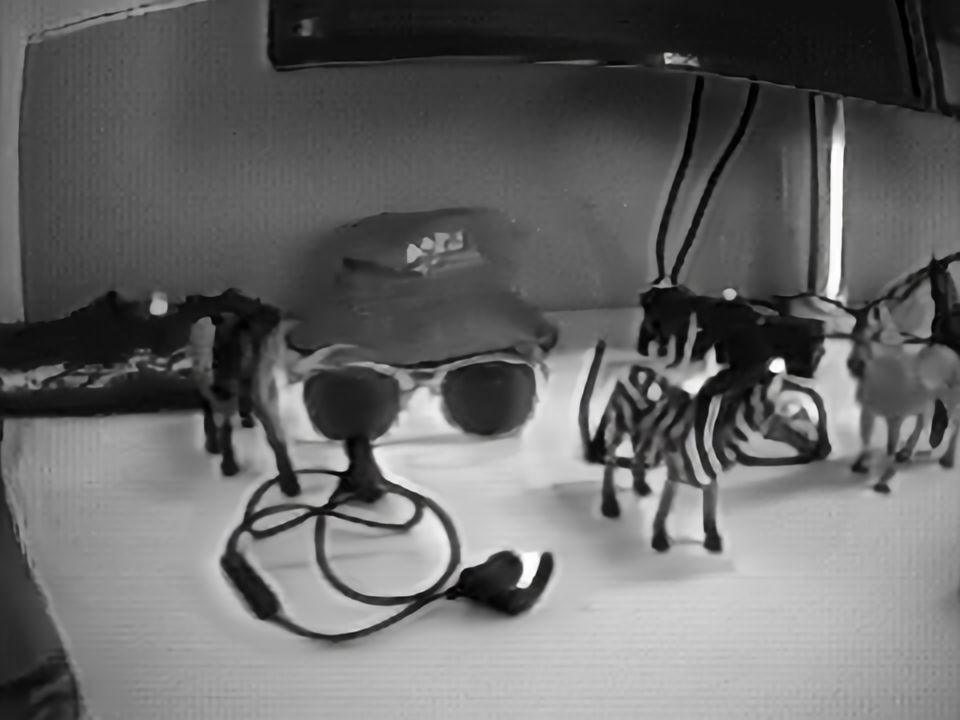}};
                \spy on \hqfgreenz in node [left] at \hqfgreenpos;
				\spy [red] on \hqfredz in node [left,red] at \hqfredpos;
				% \spy on \hqfgreenz in node [left] at \hqfgreenpos;
				% \spy [red] on \hqfredz in node [left,red] at \hqfredpos;
				\end{tikzpicture}
			EFNet+DASR\\
			(22.35, 0.7893)\vspace{0.5em}
			%(17.47, 0.6965)\vspace{0.5em}
			
			\begin{tikzpicture}[spy using outlines={green,magnification=\ssmag,size=\ssizz},inner sep=0]
				\node {\includegraphics[width=\linewidth]{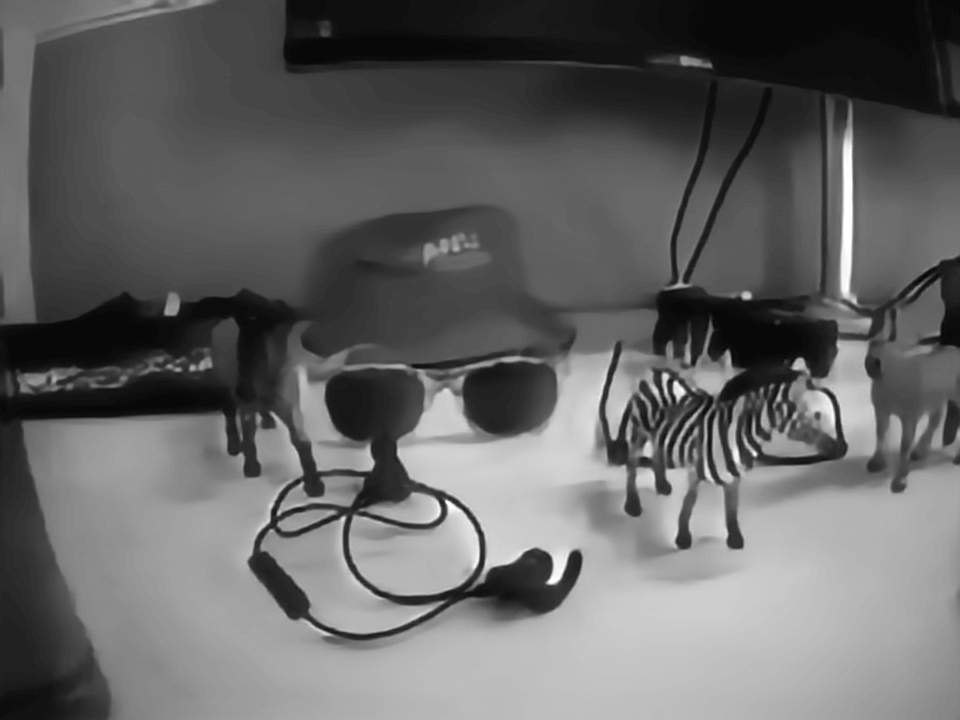}};
				\spy on \hqfgreen in node [left] at \hqfgreenpos;
				\spy [red] on \hqfred in node [left,red] at \hqfredpos;
				\end{tikzpicture}
				
    		{\bf eSL-Net++\\
    		(23.59, 0.8040)} \vspace{.5em}
    	\end{minipage}%
    % }
    
    \centering
	\caption{\colored{Quantitative and qualitative results on the GoPro dataset (top two rows) and the HQF dataset (bottom two rows), where our proposed eSL-Net and eSL-Net++ are compared to GFN, DASR, CDVD+RCAN, RED-Net+RCAN, LEDVDI+RCAN, and EFNet+DASR. }}
	\label{syn_sr}
\end{figure*}

\def\imwidth{0.2}

\def\ssxxs{(0.9,0.4)} % 跑道
\def\ssxxsr{(-0.3,0.9)} % 跑道 三角柱
\def\sxxx{(-0.4,-0.8)}%窗户 线
\def\sxxxr{(0.2,0.5)}%窗户 栏

\def\sxxxx{(-0.5,-0.6)}%she 手 绿
\def\sxxxxr{(1,0.25)}%she 右 红

\def\ssxxsl{(0.85,0.43)} % 跑道 ledvdi
\def\ssxxsrl{(-0.37,0.93)} % 跑道 三角柱ledvdi
\def\sxxxl{(-0.4,-0.75)}%窗户 线
\def\sxxxrl{(0.05,0.6)}%窗户 栏

\def\sxxxxl{(-0.5,-0.6)}%she 手 绿ledvdi
\def\sxxxxrl{(0.92,0.25)}%she 右 红ledvdi

\def\ssxxss{(0.68,0)} % 杯子 勺
\def\ssxxssr{(-0.55,-0.55)} % 杯子 垫
\def\ssxxssl{(0.64,0.12)} % 杯子 勺 ledvdi
\def\ssxxssrl{(-0.62,-0.46)} % 杯子 垫ledvdi

\def\ssyys{(1.8,-2.3)}
\def\ssyysr{(-0.05,-2.3)}

\def\ssyy{(-0.8,-0.85)}
\def\ssizz{1.77cm}
\def\sswidth{0.245\textwidth}
\def\ssmag{5}
\def\scc{(2.12,1.4)}
\begin{figure*}[!ht]
\footnotesize
	\centering

    % \subfigure{
    	\begin{minipage}[t]{\imwidth\linewidth}
    		\centering
    		%input
			\begin{tikzpicture}[spy using outlines={green,magnification=\ssmag,size=\ssizz},inner sep=0]
				\node {\includegraphics[width=\linewidth]{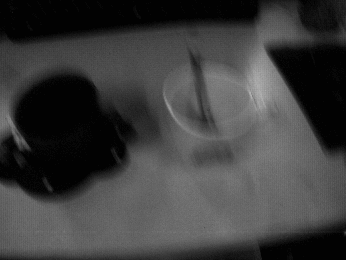}};
				\spy on \ssxxssr in node [left] at \ssyysr;
				\spy[red] on \ssxxss in node [left,red] at \ssyys;
				\end{tikzpicture}
			\vspace{-0.5em}
			
			%SHE	
			\begin{tikzpicture}[spy using outlines={green,magnification=\ssmag,size=\ssizz},inner sep=0]
				\node {\includegraphics[width=\linewidth]{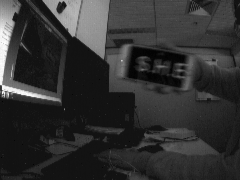}};
				\spy on \sxxxx in node [left] at \ssyysr;
				\spy[red] on \sxxxxr in node [left,red] at \ssyys;
				\end{tikzpicture}
			\vspace{-0.5em}

			% windows	
			\begin{tikzpicture}[spy using outlines={green,magnification=\ssmag,size=\ssizz},inner sep=0]
				\node {\includegraphics[width=\linewidth]{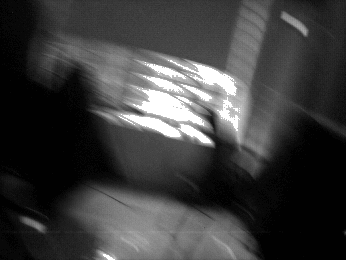}};
				\spy on \sxxx in node [left] at \ssyysr;
				\spy[red] on \sxxxr in node [left,red] at \ssyys;
				\end{tikzpicture}
			\vspace{-0.5em}
			
			%playground
			\begin{tikzpicture}[spy using outlines={green,magnification=\ssmag,size=\ssizz},inner sep=0]
				\node {\includegraphics[width=\linewidth]{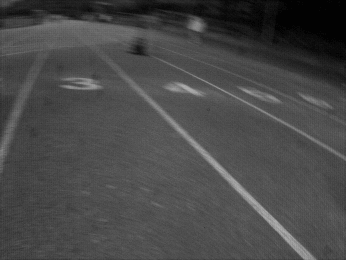}};
				\spy on \ssxxsr in node [left] at \ssyysr;
				\spy[red] on \ssxxs in node [left,red] at \ssyys;
				\end{tikzpicture}
			\vspace{-0.5em}
            Blurry Image \vspace{0.3em}
    	\end{minipage}%
    % }
    \hspace*{0mm}
    % \subfigure{
    	\begin{minipage}[t]{\imwidth\linewidth}
    		\centering
    		\begin{tikzpicture}[spy using outlines={green,magnification=\ssmag,size=\ssizz},inner sep=0]
				\node {\includegraphics[width=\linewidth]{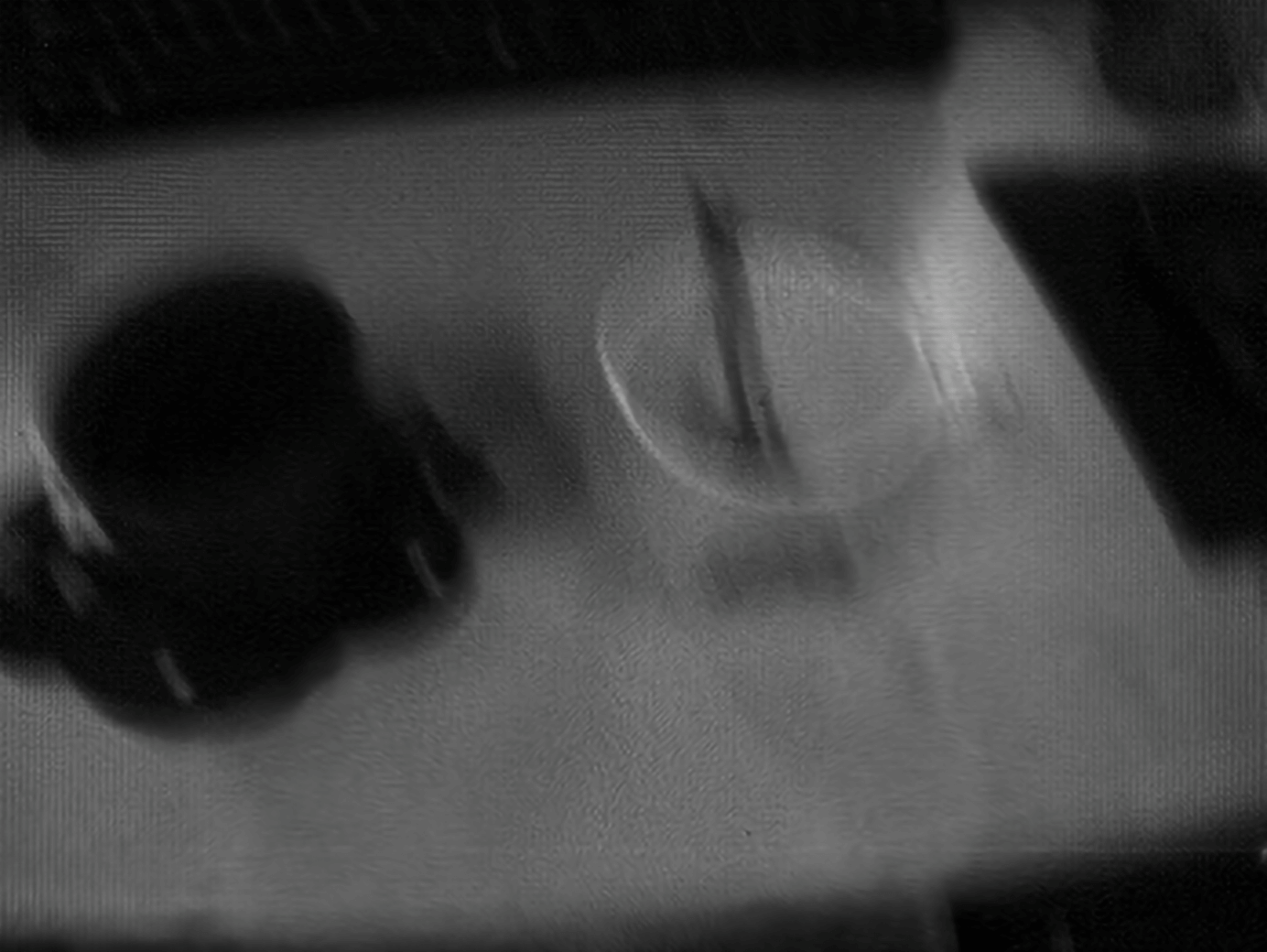}};
				\spy on \ssxxssrl in node [left] at \ssyysr;
				\spy[red] on \ssxxssl in node [left,red] at \ssyys;
				\end{tikzpicture}
				\vspace{-0.5em}
    		
			\begin{tikzpicture}[spy using outlines={green,magnification=\ssmag,size=\ssizz},inner sep=0]
				\node {\includegraphics[width=\linewidth]{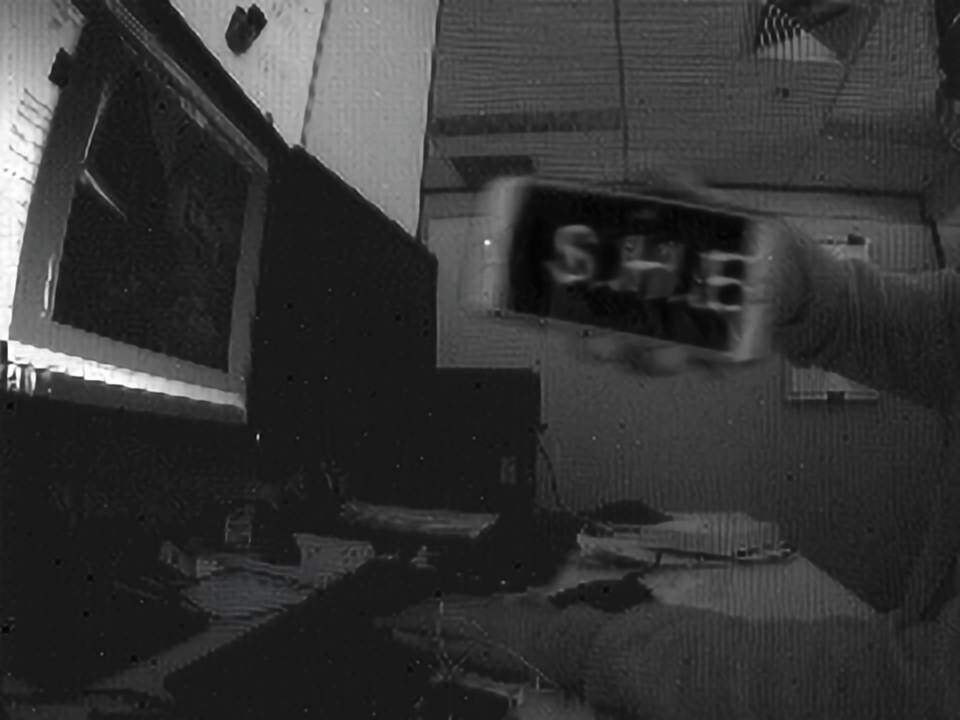}};
				\spy on \sxxxxl in node [left] at \ssyysr;
				\spy[red] on \sxxxxrl in node [left,red] at \ssyys;
				\end{tikzpicture}
			\vspace{-0.5em}
			
			\begin{tikzpicture}[spy using outlines={green,magnification=\ssmag,size=\ssizz},inner sep=0]
				\node {\includegraphics[width=\linewidth]{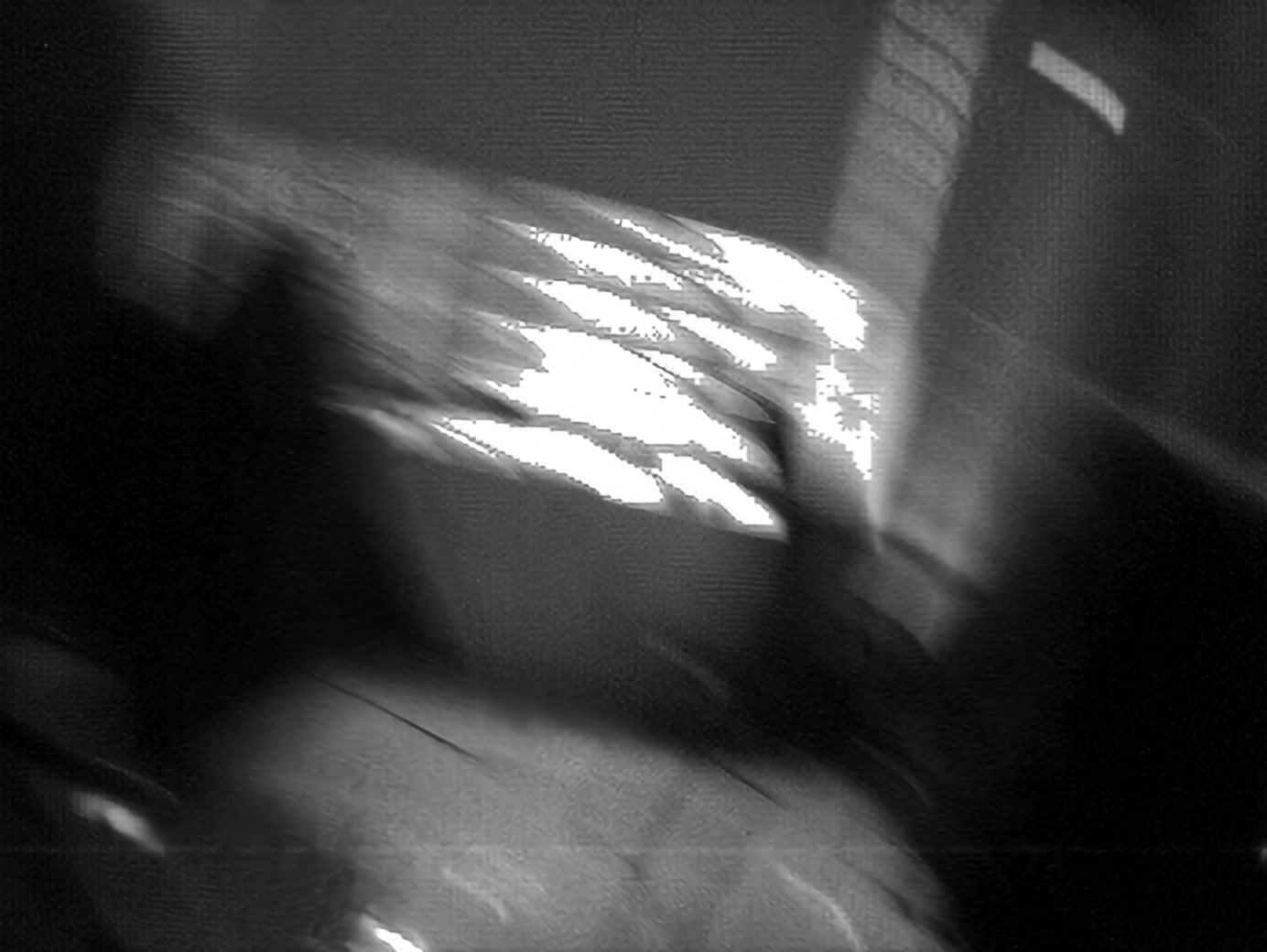}};
				\spy on \sxxxl in node [left] at \ssyysr;
				\spy[red] on \sxxxrl in node [left,red] at \ssyys;
				\end{tikzpicture}
			\vspace{-0.5em}
			
    		\begin{tikzpicture}[spy using outlines={green,magnification=\ssmag,size=\ssizz},inner sep=0]
				\node {\includegraphics[width=\linewidth]{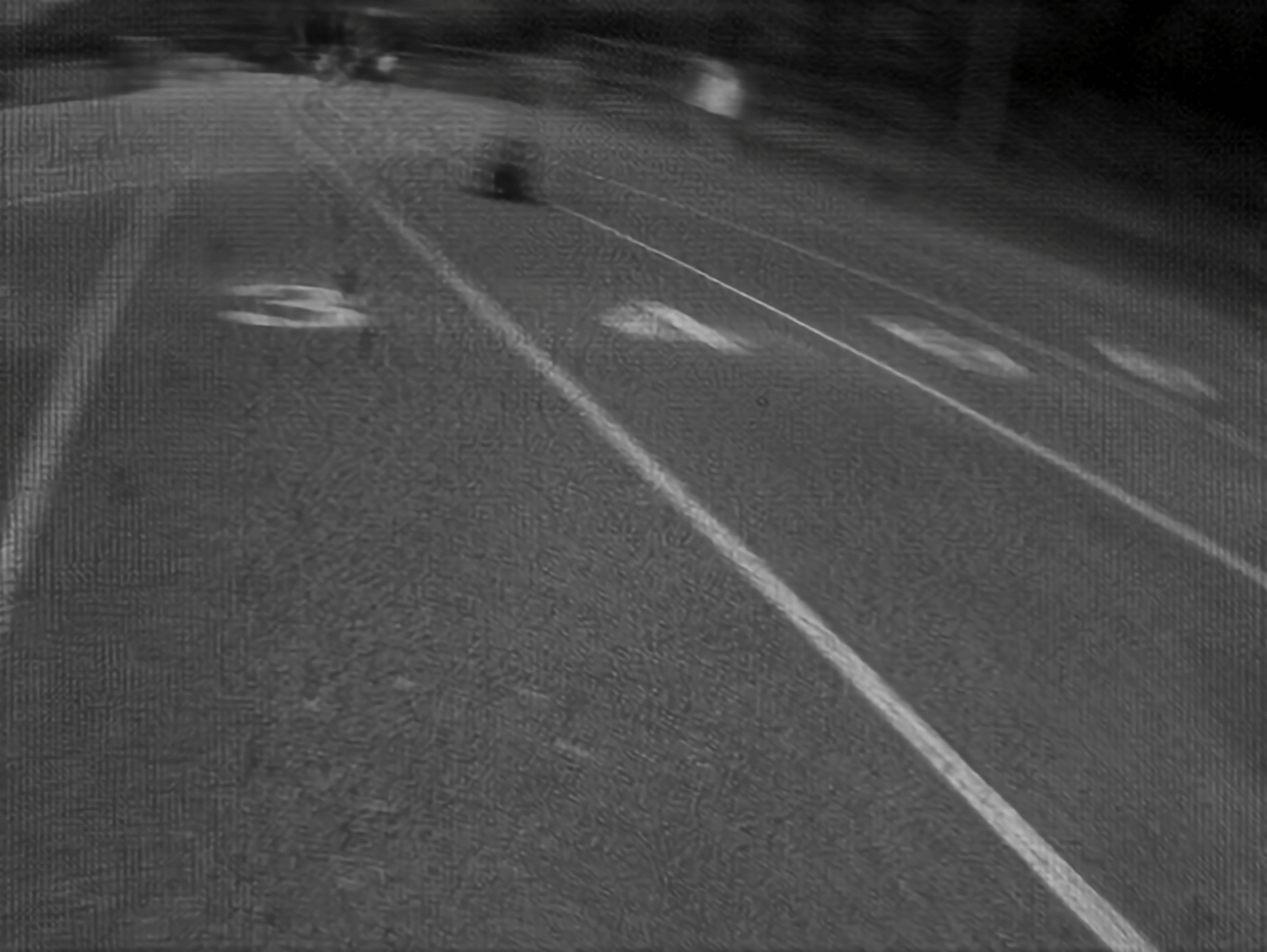}};
				\spy on \ssxxsrl in node [left] at \ssyysr;
				\spy[red] on \ssxxsl in node [left,red] at \ssyys;
				\end{tikzpicture}
			\vspace{-0.5em}			
			GFN \vspace{0.3em}
    	\end{minipage}%
    % }
    \hspace*{0mm}
    % \subfigure{
    	\begin{minipage}[t]{\imwidth\linewidth}
    		\centering
    		\begin{tikzpicture}[spy using outlines={green,magnification=\ssmag,size=\ssizz},inner sep=0]
				\node {\includegraphics[width=\linewidth]{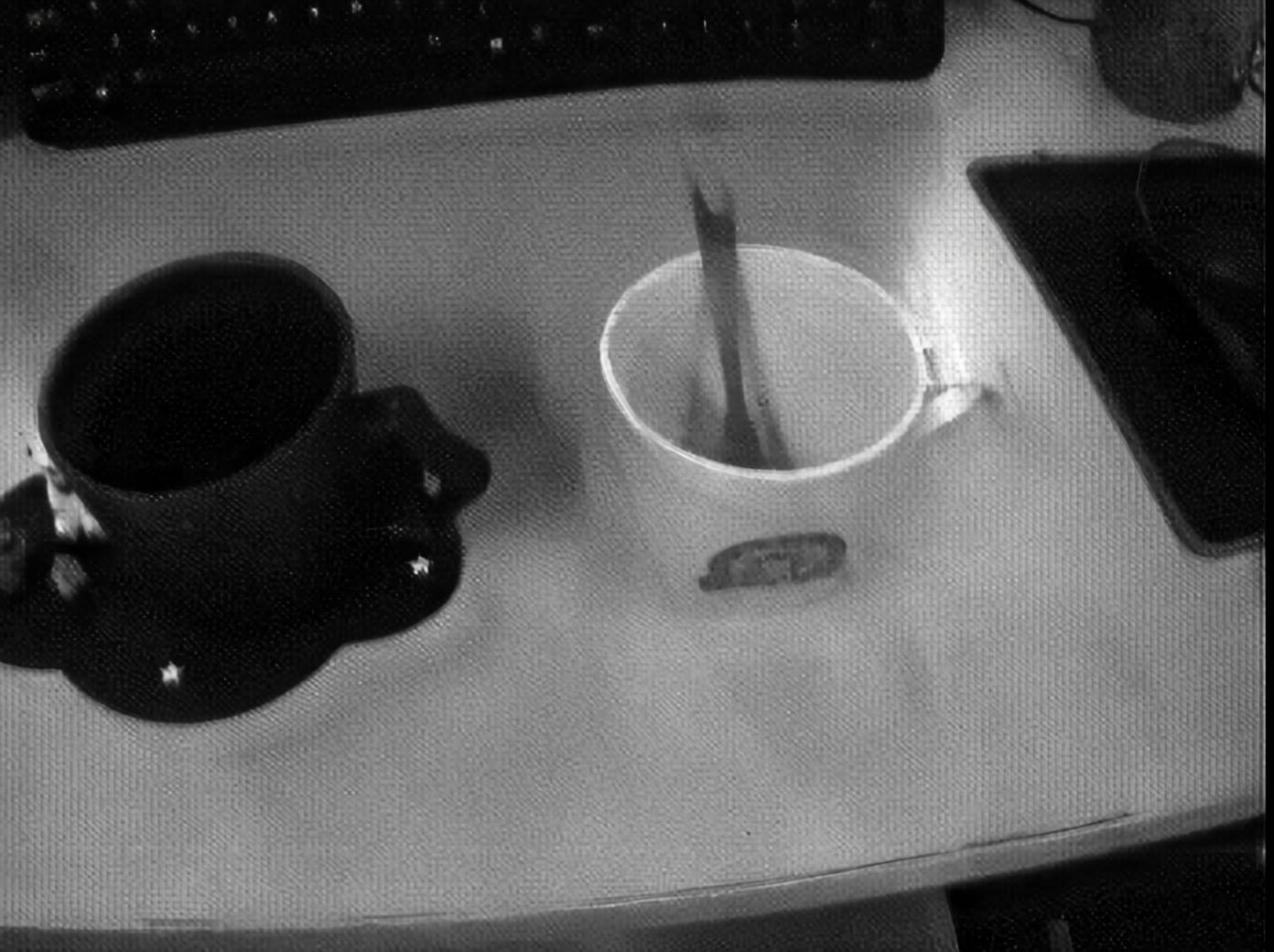}};
				\spy on \ssxxssrl in node [left] at \ssyysr;
				\spy[red] on \ssxxssl in node [left,red] at \ssyys;
				\end{tikzpicture}
				\vspace{-0.5em}
    		
            \begin{tikzpicture}[spy using outlines={green,magnification=\ssmag,size=\ssizz},inner sep=0]
				\node {\includegraphics[width=\linewidth]{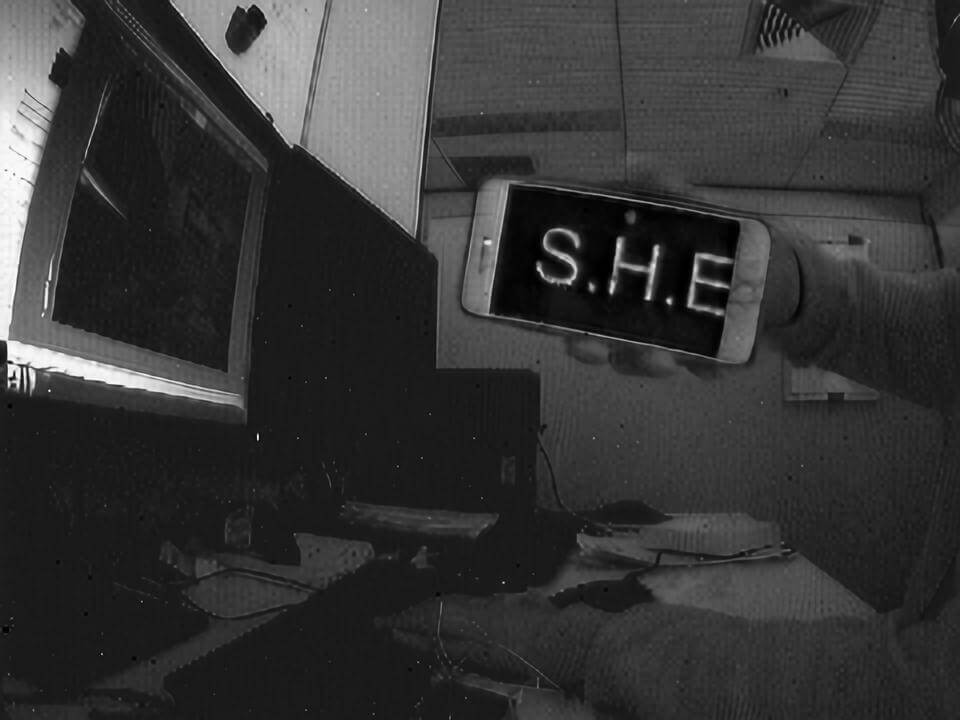}};
				\spy on \sxxxxl in node [left] at \ssyysr;
				\spy[red] on \sxxxxrl in node [left,red] at \ssyys;
				\end{tikzpicture}
			\vspace{-0.5em}
			
			\begin{tikzpicture}[spy using outlines={green,magnification=\ssmag,size=\ssizz},inner sep=0]
				\node {\includegraphics[width=\linewidth]{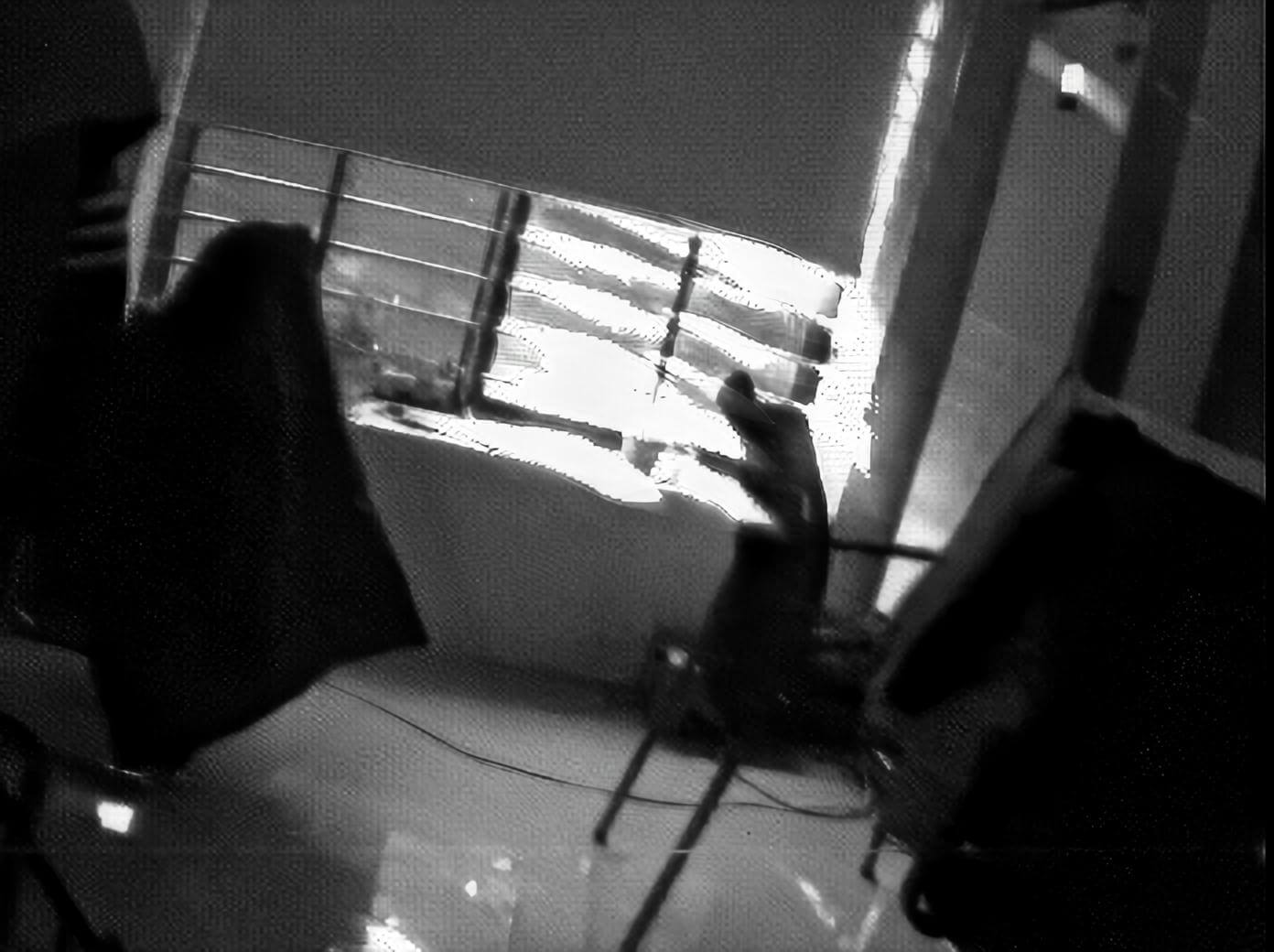}};
				\spy on \sxxxl in node [left] at \ssyysr;
				\spy[red] on \sxxxrl in node [left,red] at \ssyys;
				\end{tikzpicture}
			\vspace{-0.5em}
			
			\begin{tikzpicture}[spy using outlines={green,magnification=\ssmag,size=\ssizz},inner sep=0]
				\node {\includegraphics[width=\linewidth]{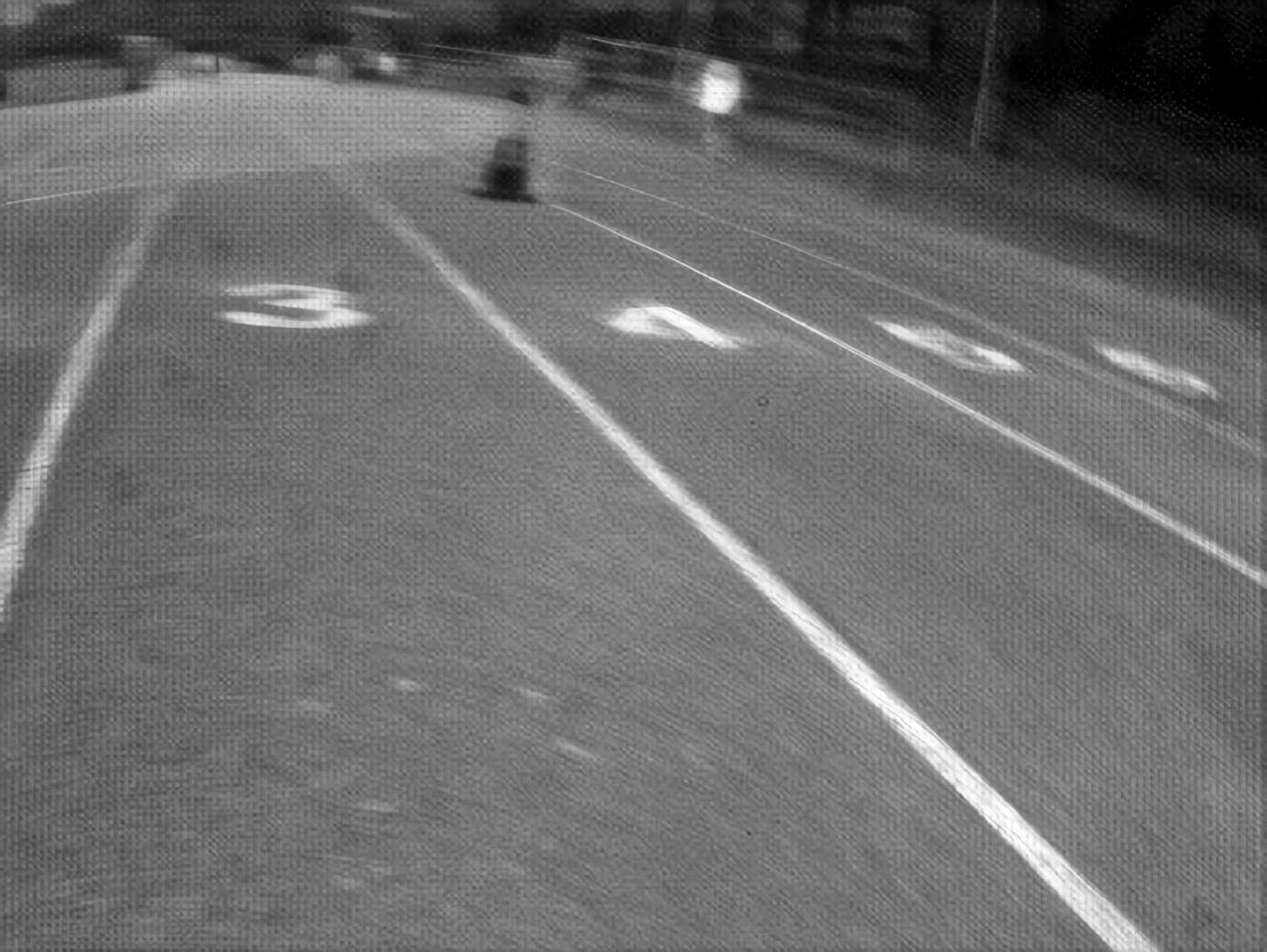}};
				\spy on \ssxxsrl in node [left] at \ssyysr;
				\spy[red] on \ssxxsl in node [left,red] at \ssyys;
				\end{tikzpicture}
			\vspace{-0.5em}
			LEDVDI+RCAN \vspace{0.3em}
    	\end{minipage}%
    % }
    \hspace*{0mm}
	% \subfigure{
		\begin{minipage}[t]{\imwidth\linewidth}
			\centering
			\begin{tikzpicture}[spy using outlines={green,magnification=\ssmag,size=\ssizz},inner sep=0]
				\node {\includegraphics[width=\linewidth]{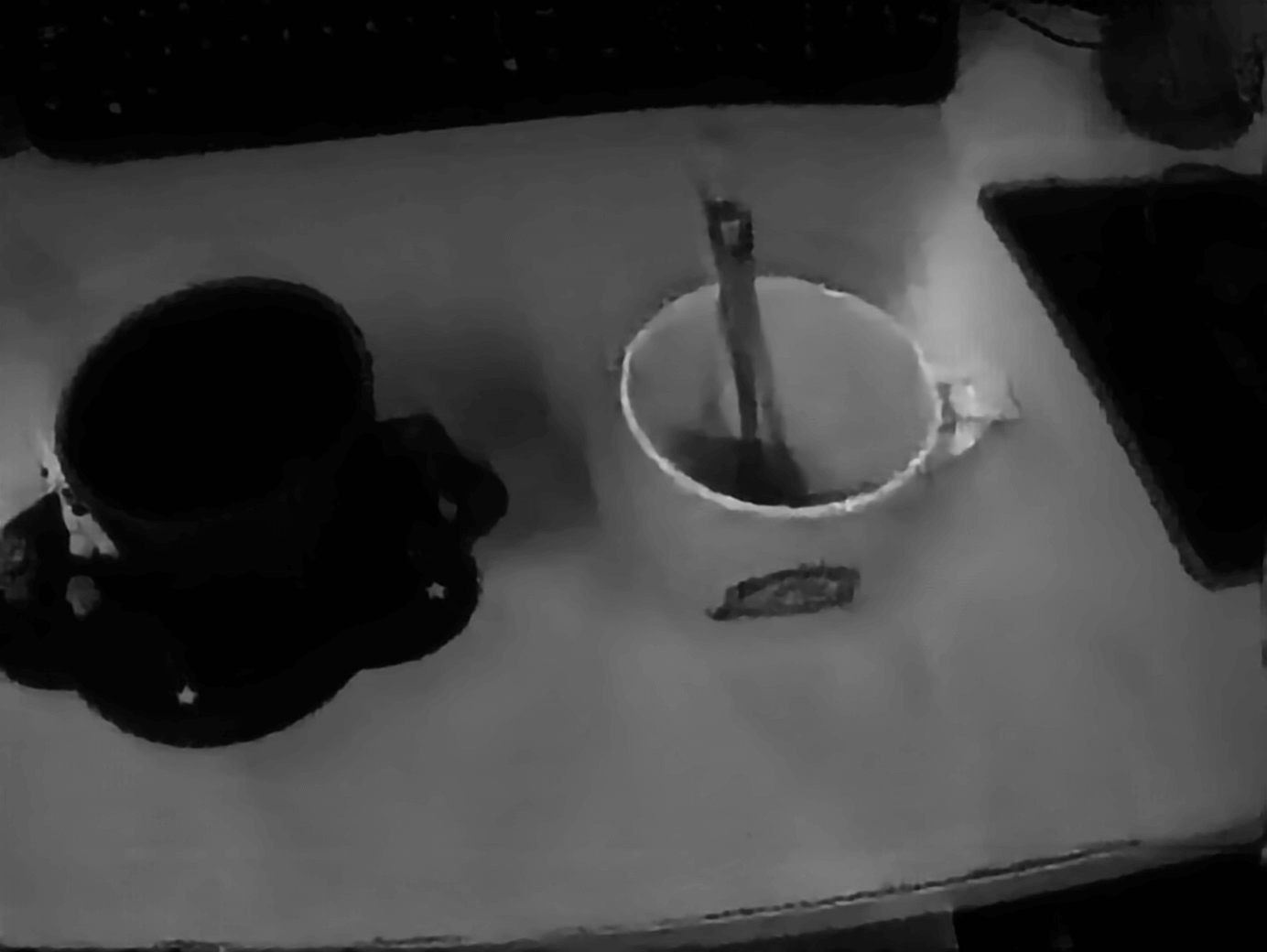}};
				\spy on \ssxxssr in node [left] at \ssyysr;
				\spy[red] on \ssxxss in node [left,red] at \ssyys;
				\end{tikzpicture}
				\vspace{-0.5em}
			
            \begin{tikzpicture}[spy using outlines={green,magnification=\ssmag,size=\ssizz},inner sep=0]
				\node {\includegraphics[width=\linewidth]{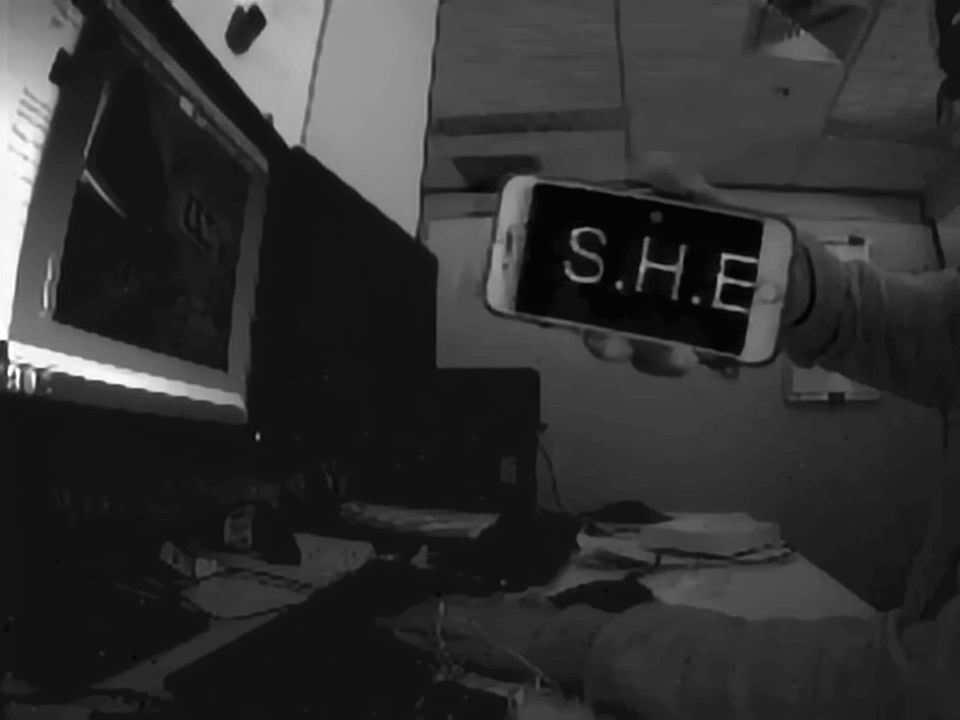}};
				\spy on \sxxxx in node [left] at \ssyysr;
				\spy[red] on \sxxxxr in node [left,red] at \ssyys;
				\end{tikzpicture}
			\vspace{-0.5em}
			
			\begin{tikzpicture}[spy using outlines={green,magnification=\ssmag,size=\ssizz},inner sep=0]
				\node {\includegraphics[width=\linewidth]{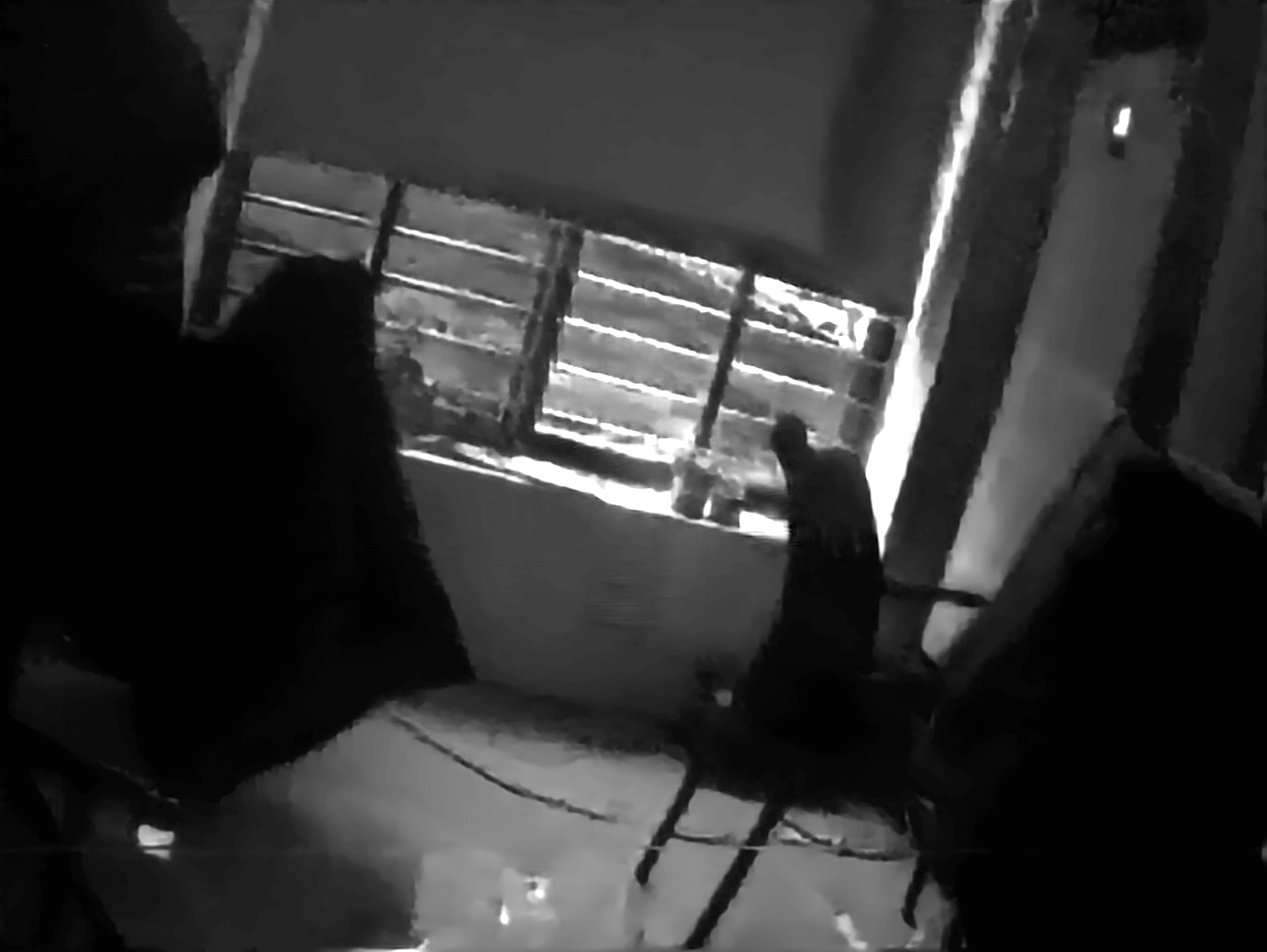}};
				\spy on \sxxx in node [left] at \ssyysr;
				\spy[red] on \sxxxr in node [left,red] at \ssyys;
				\end{tikzpicture}
			\vspace{-0.5em}
			
			\begin{tikzpicture}[spy using outlines={green,magnification=\ssmag,size=\ssizz},inner sep=0]
				\node {\includegraphics[width=\linewidth]{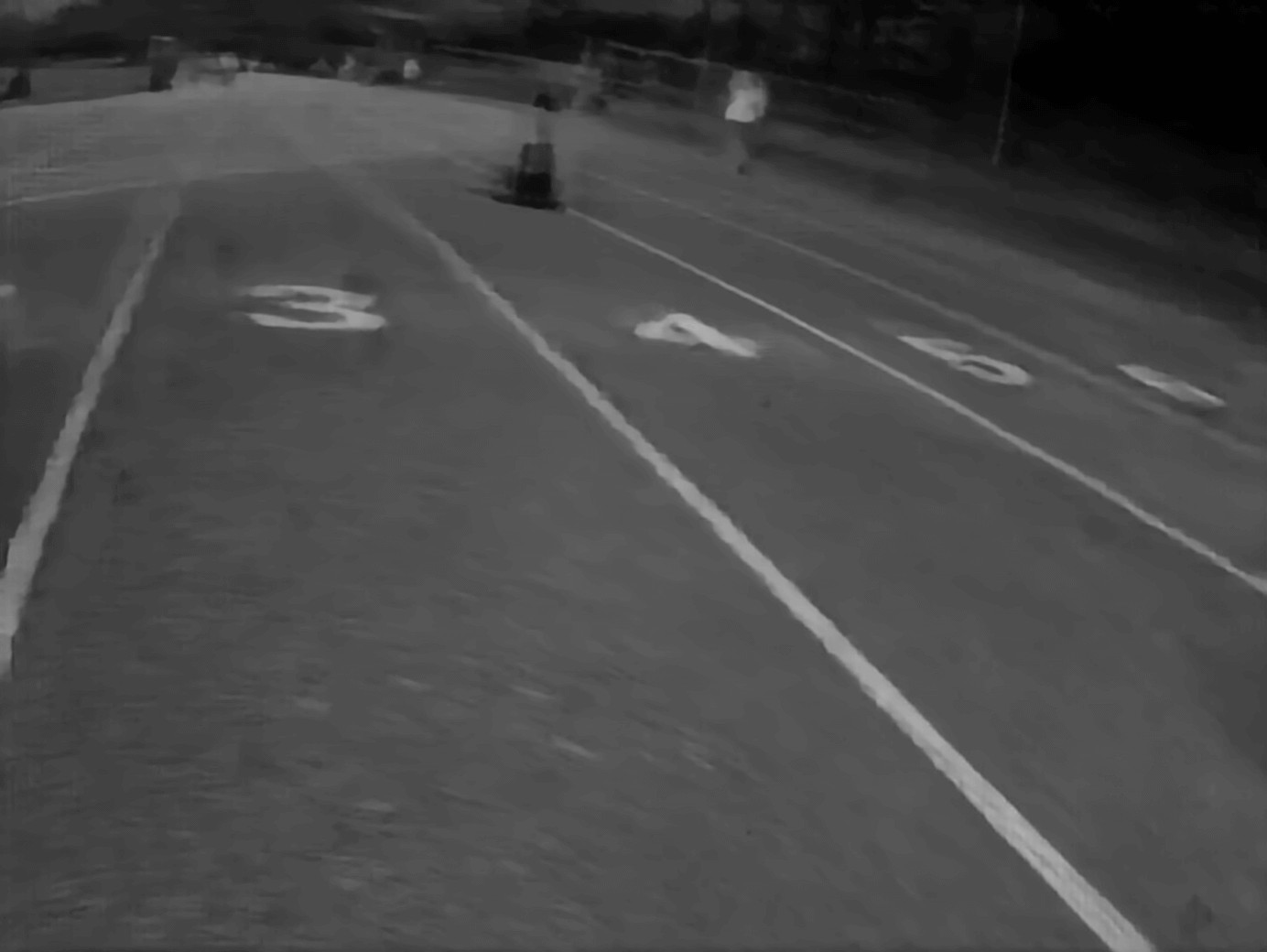}};
				\spy on \ssxxsr in node [left] at \ssyysr;
				\spy[red] on \ssxxs in node [left,red] at \ssyys;
				\end{tikzpicture}
			\vspace{-0.5em}
			eSL-Net \vspace{.3em}
			
		\end{minipage}%
	% }
 \hspace*{0mm}
    % \subfigure{
    	\begin{minipage}[t]{\imwidth\linewidth}
    		\centering
    		\begin{tikzpicture}[spy using outlines={green,magnification=\ssmag,size=\ssizz},inner sep=0]
				\node {\includegraphics[width=\linewidth]{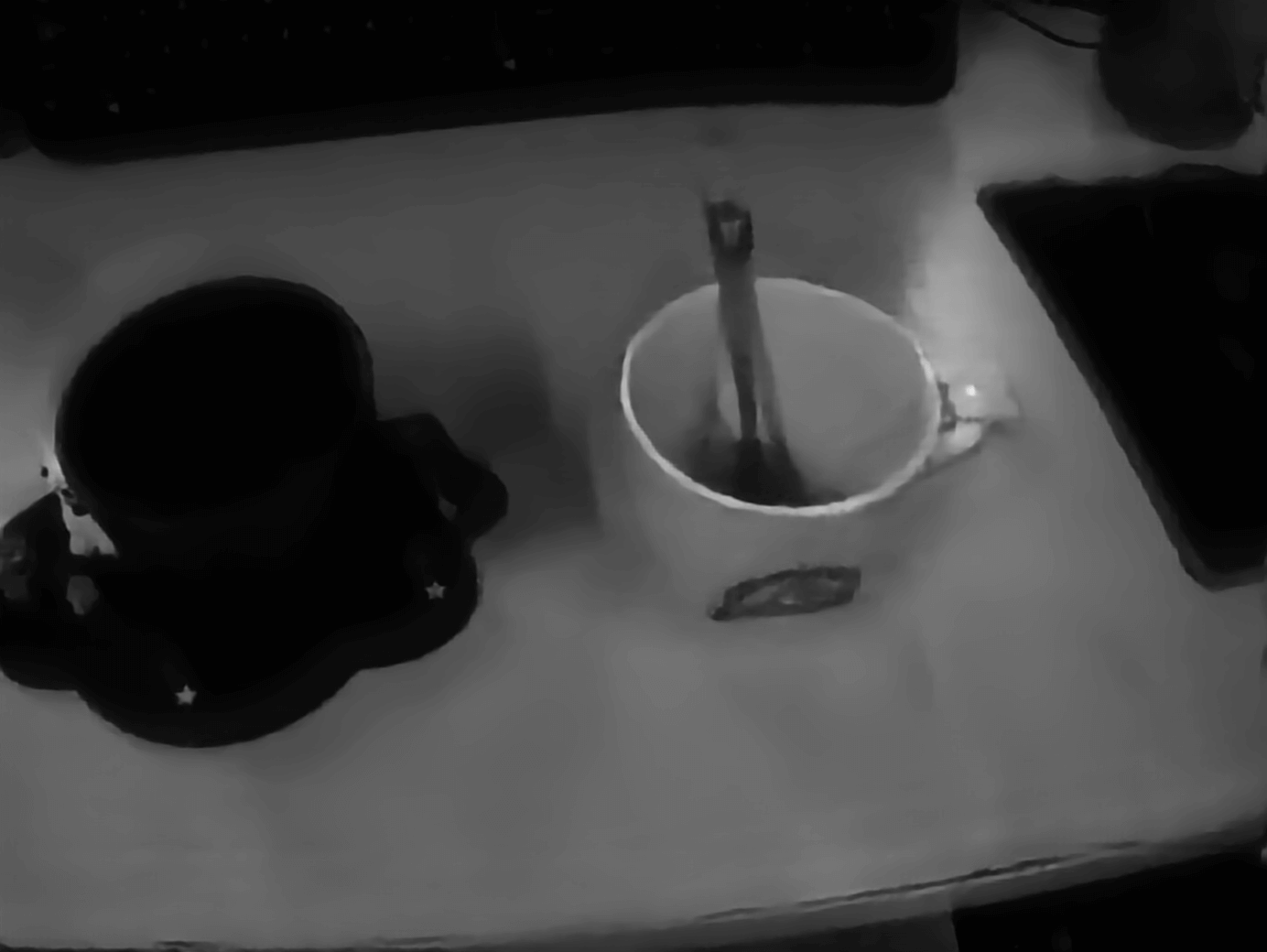}};
				\spy on \ssxxssr in node [left] at \ssyysr;
				\spy[red] on \ssxxss in node [left,red] at \ssyys;
				\end{tikzpicture}
				\vspace{-0.5em}
				
			\begin{tikzpicture}[spy using outlines={green,magnification=\ssmag,size=\ssizz},inner sep=0]
				\node {\includegraphics[width=\linewidth]{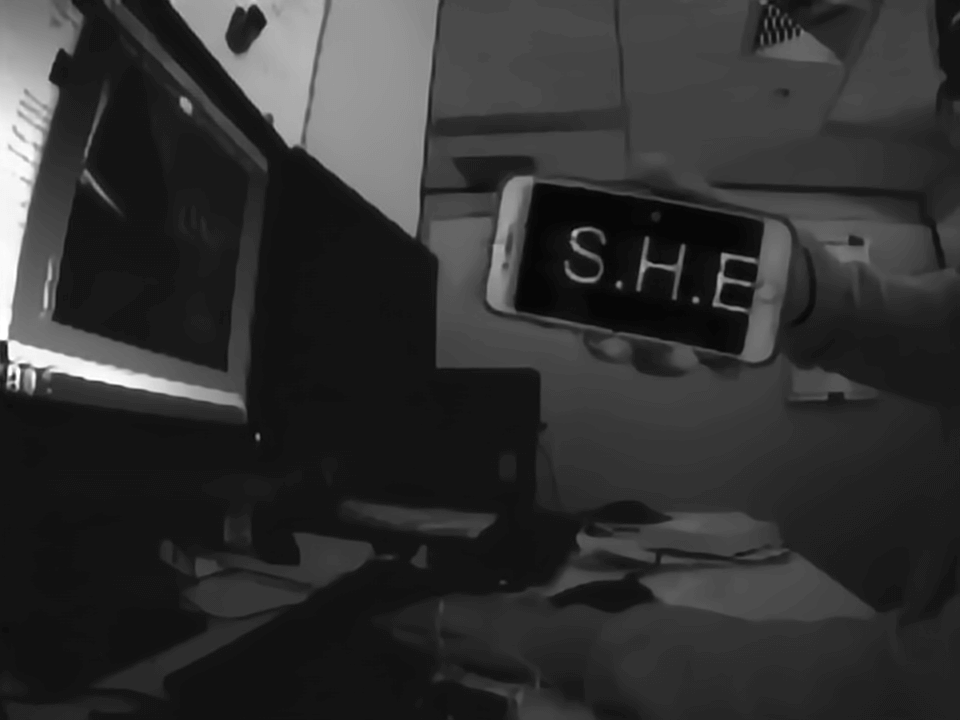}};
				\spy on \sxxxx in node [left] at \ssyysr;
				\spy[red] on \sxxxxr in node [left,red] at \ssyys;
				\end{tikzpicture}
			\vspace{-0.5em}
			
			\begin{tikzpicture}[spy using outlines={green,magnification=\ssmag,size=\ssizz},inner sep=0]
				\node {\includegraphics[width=\linewidth]{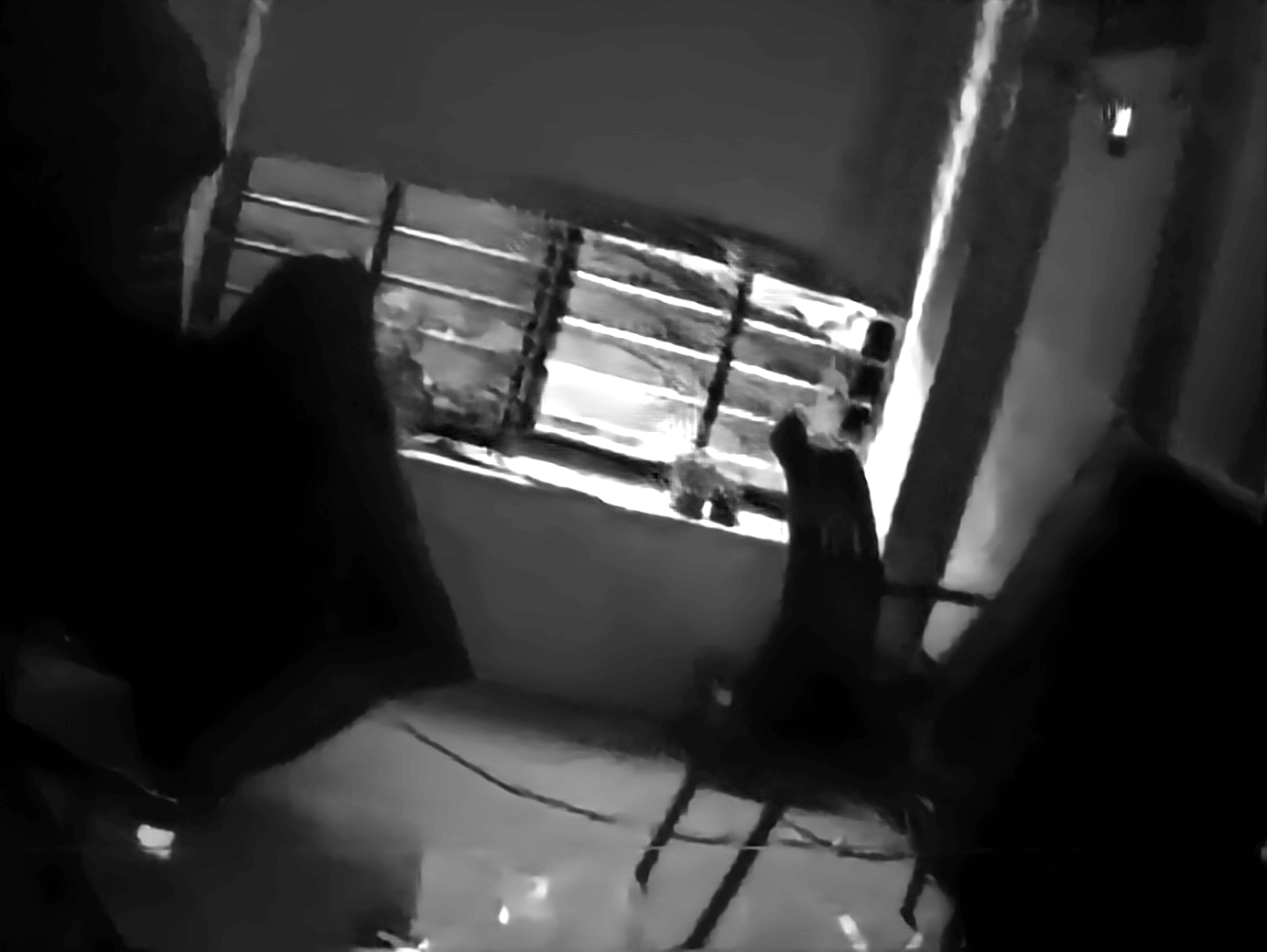}};
				\spy on \sxxx in node [left] at \ssyysr;
				\spy[red] on \sxxxr in node [left,red] at \ssyys;
				\end{tikzpicture}
			\vspace{-0.5em}
			
			\begin{tikzpicture}[spy using outlines={green,magnification=\ssmag,size=\ssizz},inner sep=0]
				\node {\includegraphics[width=\linewidth]{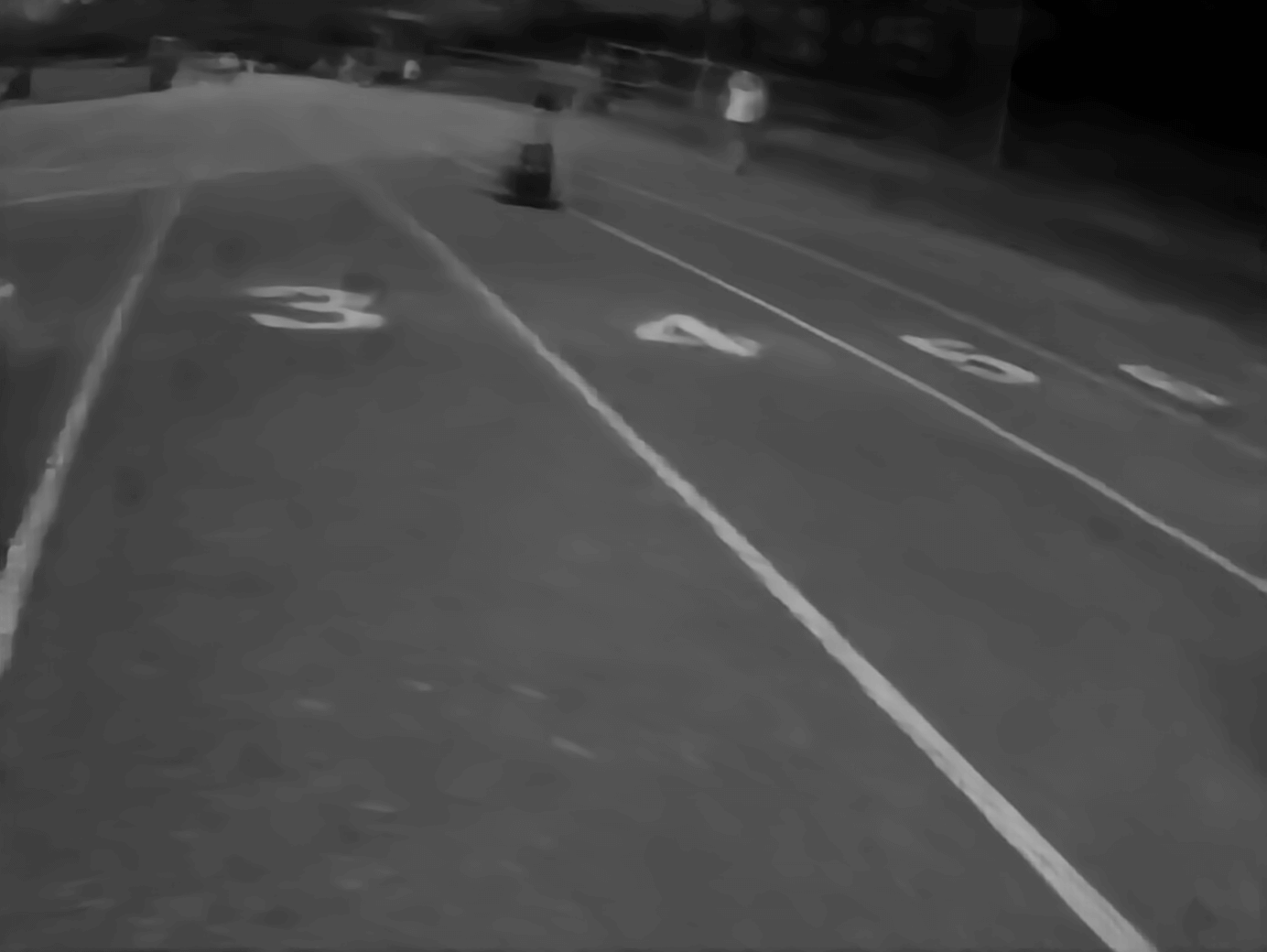}};
				\spy on \ssxxsr in node [left] at \ssyysr;
				\spy[red] on \ssxxs in node [left,red] at \ssyys;
				\end{tikzpicture}
			\vspace{-0.5em}
    		{\bf eSL-Net++} \vspace{.3em}
    	\end{minipage}%
    % }
    \centering
	\caption{Qualitative results on the RWS dataset, where our proposed eSL-Net and eSL-Net++ are compared with GFN and LEDVDI+RCAN.}
	\label{real_sr_more}
\end{figure*}

\def\imwidth{0.16}
\def\cimwid{0.072}

\def\zuoxia{(-0.8,-0.4)}
\def\youshang{(-0.05,0.8)}

\def\ssyy{(-0.8,-0.85)}
\def\ssizz{0.5cm}
\def\sswidth{0.245\textwidth}
\def\ssmag{3}
\def\scc{(2.12,1.4)}

\begin{figure*}[htb]
\footnotesize
	\centering

    % \subfigure{
    	\begin{minipage}[t]{\imwidth\linewidth}
    		\centering
    		%ground truth
			\begin{tikzpicture}[spy using outlines={rectangle,green,magnification=\ssmag,size=\ssizz},inner sep=0]
				\node {\includegraphics[width=\linewidth]{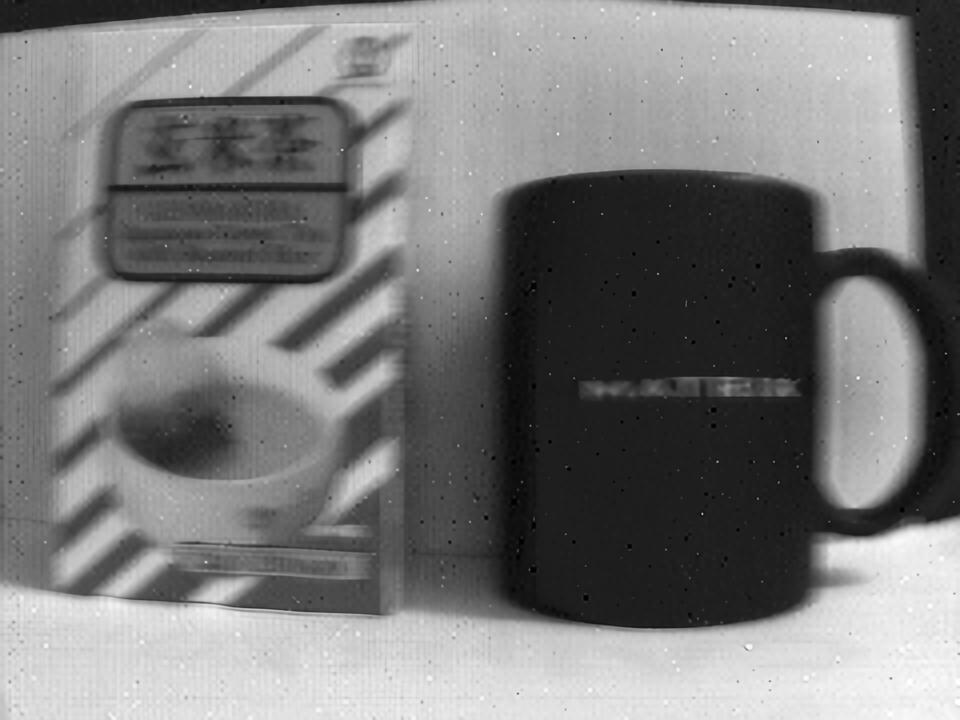}};
				% \spy on \ssxxs in node [left] at \ssyys;
				\draw[green] \zuoxia rectangle \youshang;
				\end{tikzpicture}
            Blurry Image\vspace{0.3em}
    	\end{minipage}%
    % }
    \hspace*{0mm}
    % \subfigure{
    	\begin{minipage}[t]{\imwidth\linewidth}
    		\centering
    		\begin{tikzpicture}[spy using outlines={rectangle,green,magnification=\ssmag,size=\ssizz},inner sep=0]
				\node {\includegraphics[width=\linewidth]{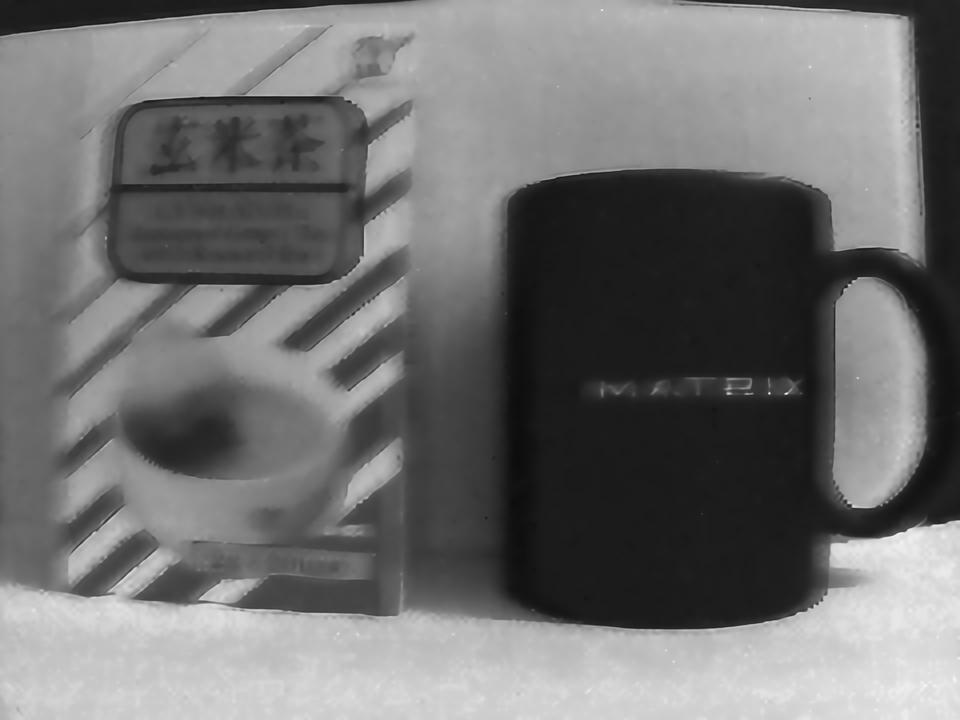}};
				% \spy on \ssxxs in node [left] at \ssyys;
				\draw[green] \zuoxia rectangle \youshang;
				\end{tikzpicture}
			
			EDI+RCAN\vspace{0.5em}
    	\end{minipage}%
    % }
    \hspace*{0mm}
    % \subfigure{
    	\begin{minipage}[t]{\imwidth\linewidth}
    		\centering
    	    \begin{tikzpicture}[spy using outlines={green,magnification=\ssmag,size=\ssizz},inner sep=0]
				\node {\includegraphics[width=\linewidth]{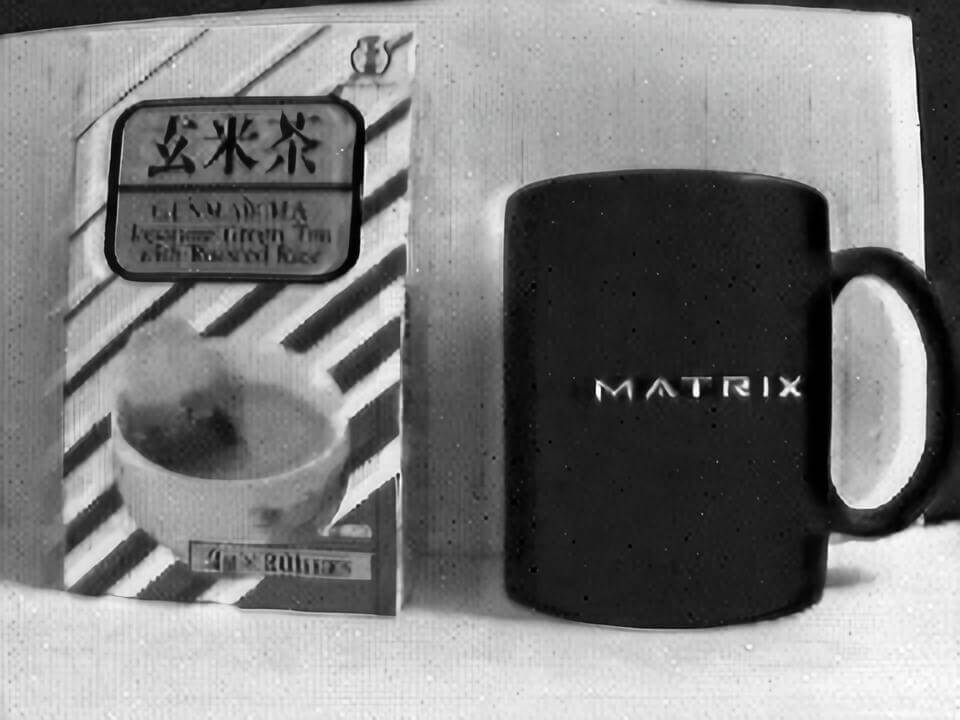}};
				% \spy on \ssxxs in node [left] at \ssyys;
				\draw[green] \zuoxia rectangle \youshang;
				\end{tikzpicture}
			LEDVDI+RCAN\vspace{0.5em}
    	\end{minipage}%
    % }
    \hspace*{0mm}
    % \subfigure{
    	\begin{minipage}[t]{\imwidth\linewidth}
    		\centering
    	    \begin{tikzpicture}[spy using outlines={green,magnification=\ssmag,size=\ssizz},inner sep=0]
				\node {\includegraphics[width=\linewidth]{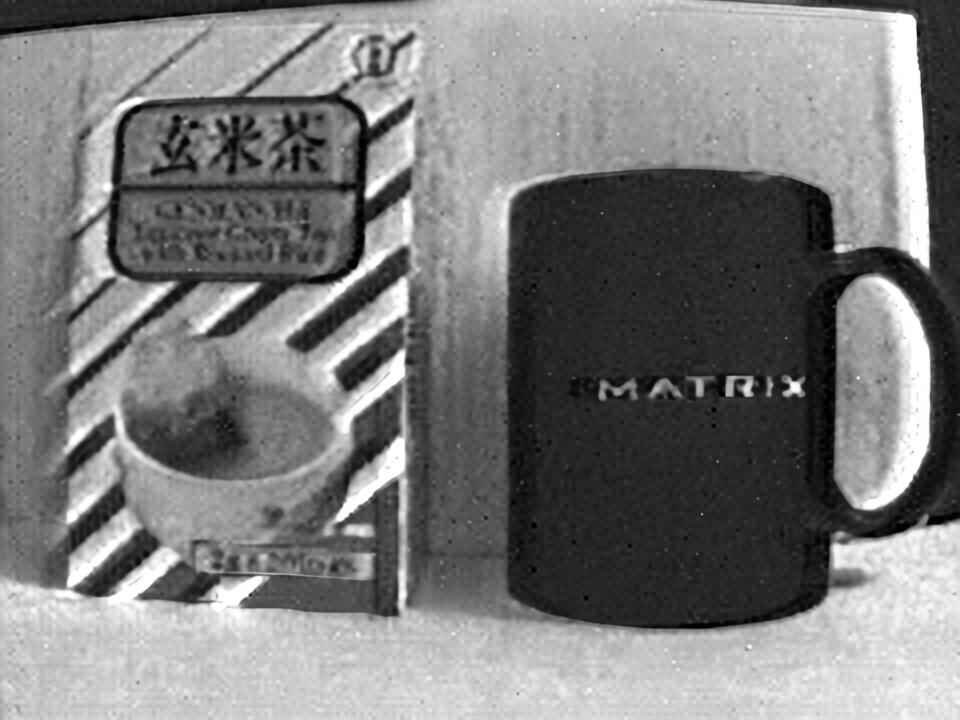}};
				% \spy on \ssxxs in node [left] at \ssyys;
				\draw[green] \zuoxia rectangle \youshang;
				\end{tikzpicture}
			RED-Net+RCAN\vspace{0.5em}
    	\end{minipage}%
    % }
    \hspace*{0mm}
	% \subfigure{
		\begin{minipage}[t]{\imwidth\linewidth}
			\centering
			\begin{tikzpicture}[spy using outlines={green,magnification=\ssmag,size=\ssizz},inner sep=0]
				\node {\includegraphics[width=\linewidth]{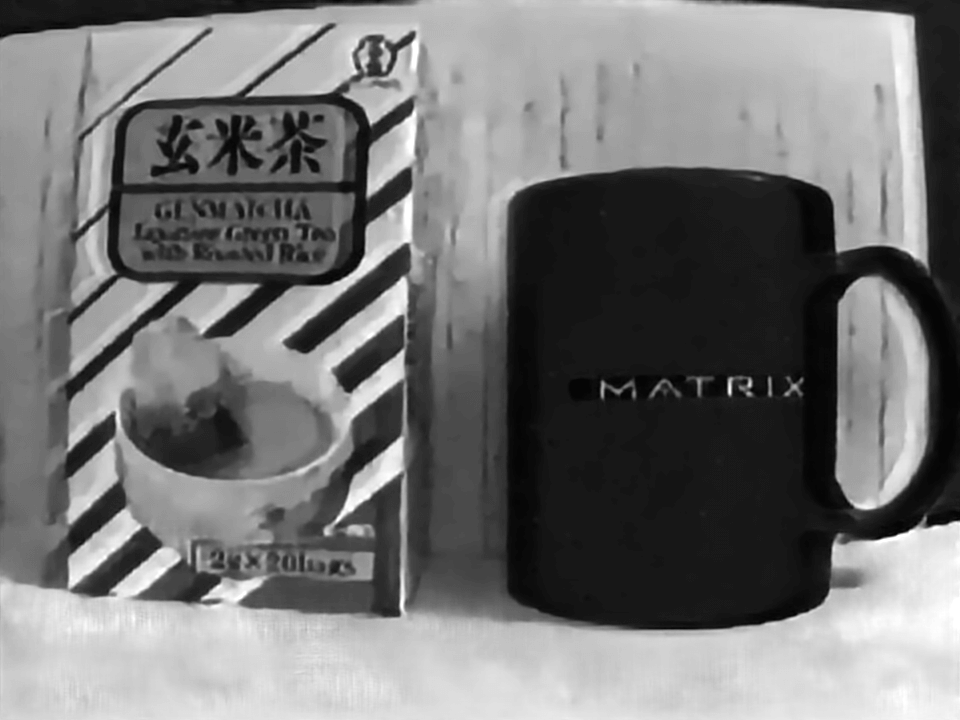}};
				% \spy on \ssxxs in node [left] at \ssyys;
				\draw[green] \zuoxia rectangle \youshang;
				\end{tikzpicture}
			eSL-Net\vspace{.5em}
			
% 			(27.60/26.49, 0.8318/0.7907)\\
			
		\end{minipage}%
	% }
 \hspace*{0mm}
    % \subfigure{
    	\begin{minipage}[t]{\imwidth\linewidth}
    		\centering
    		\begin{tikzpicture}[spy using outlines={green,magnification=\ssmag,size=\ssizz},inner sep=0]
				\node {\includegraphics[width=\linewidth]{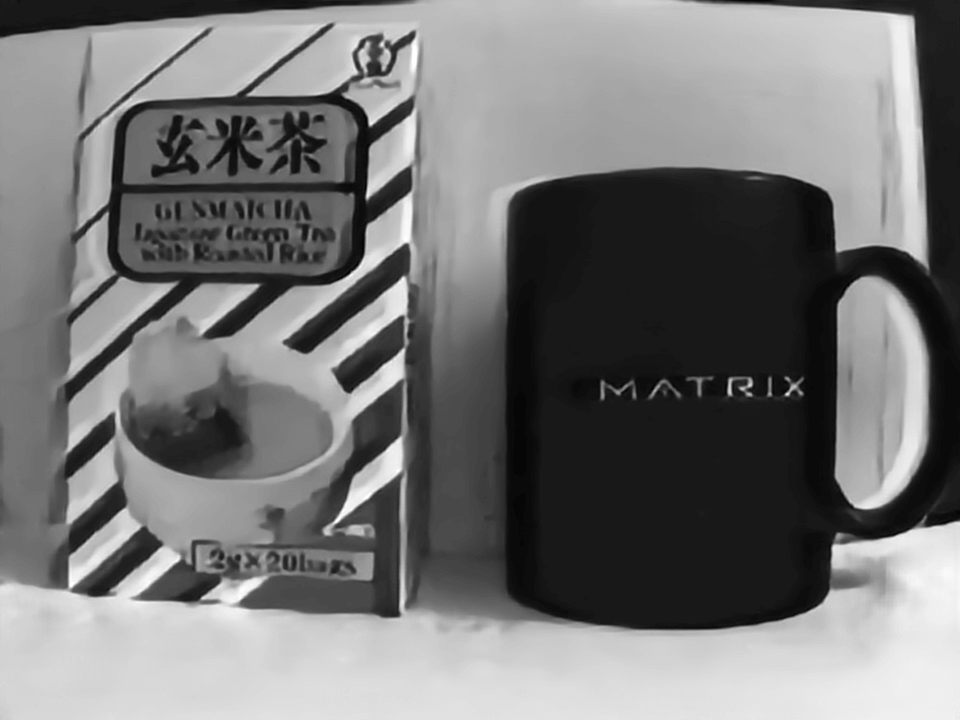}};
				% \spy on \ssxxs in node [left] at \ssyys;
				\draw[green] \zuoxia rectangle \youshang;
				\end{tikzpicture}
    		{\bf eSL-Net++} \vspace{.5em}
    % 		(28.50/26.98, 0.8548/0.8060)\\
    	\end{minipage}%
    % }
%     	\begin{tikzpicture}[overlay, inner sep=0]
% 	%\node (Adaptive) [noborder] {Adaptive};
% 	\node [label={[label distance=0.5cm,text depth=-1ex,rotate=270]right: {\tiny \textbf{Extremely Dense}}}] at (0,8.7) {};
% 	\node [label={[label distance=0.5cm,text depth=-1ex,rotate=270]right: {\tiny \textbf{Denser}}}] at (0,6.3) {};
% 	\node [label={[label distance=0.5cm,text depth=-1ex,rotate=270]right: {\tiny \textbf{Sparser}}}] at (0,2.3) {};
% 	\draw[-latex, line width = 1.5pt] (0,4.6) -- (0,2); 
% 		\end{tikzpicture} 

    \def\removelag{-1.3mm}
    % \subfigure{%left, bottom, right and top
            \begin{tikzpicture}[inner sep=0]
            \hspace*{-0.5mm}
            \node [label={[label distance=0.4cm,text depth=-1ex,rotate=90]right: {\scriptsize \text{EDI+RCAN}}}] at (0,8.7) {};
            \end{tikzpicture}
            \hspace*{-1.mm}
			\includegraphics[width=\cimwid\linewidth,trim={220px 230px 500px 100px},,clip=true]{pic/mulframe/edi/0019_0.jpg}\hspace*{-0.5mm}
			\includegraphics[width=\cimwid\linewidth,trim={220px 230px 500px 100px},clip]{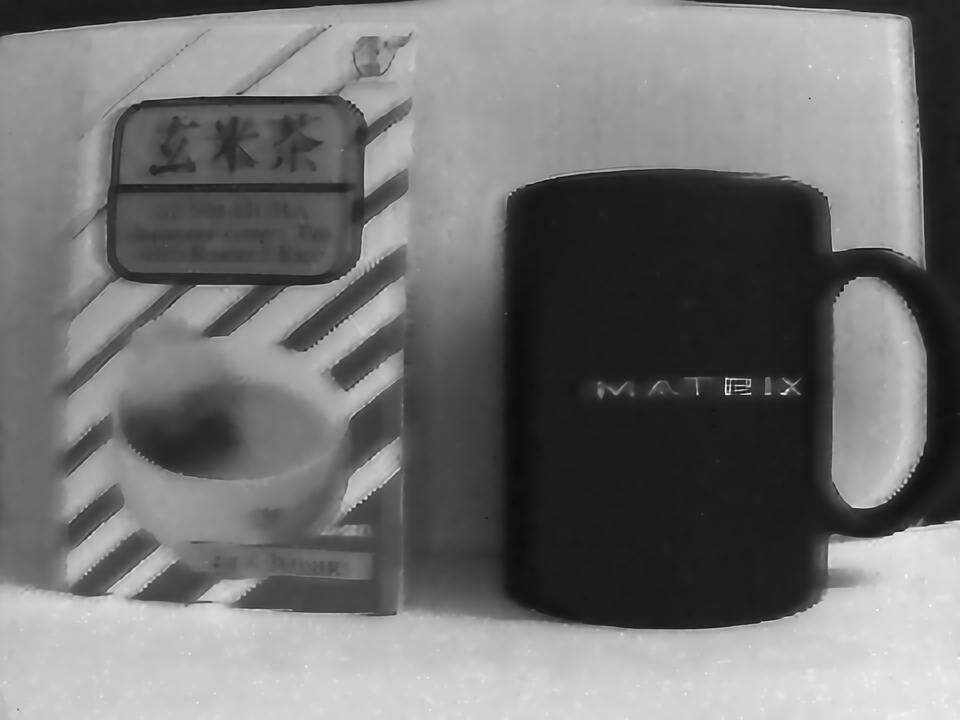}\hspace*{-0.5mm}
			\includegraphics[width=\cimwid\linewidth,trim={220px 230px 500px 100px},clip]{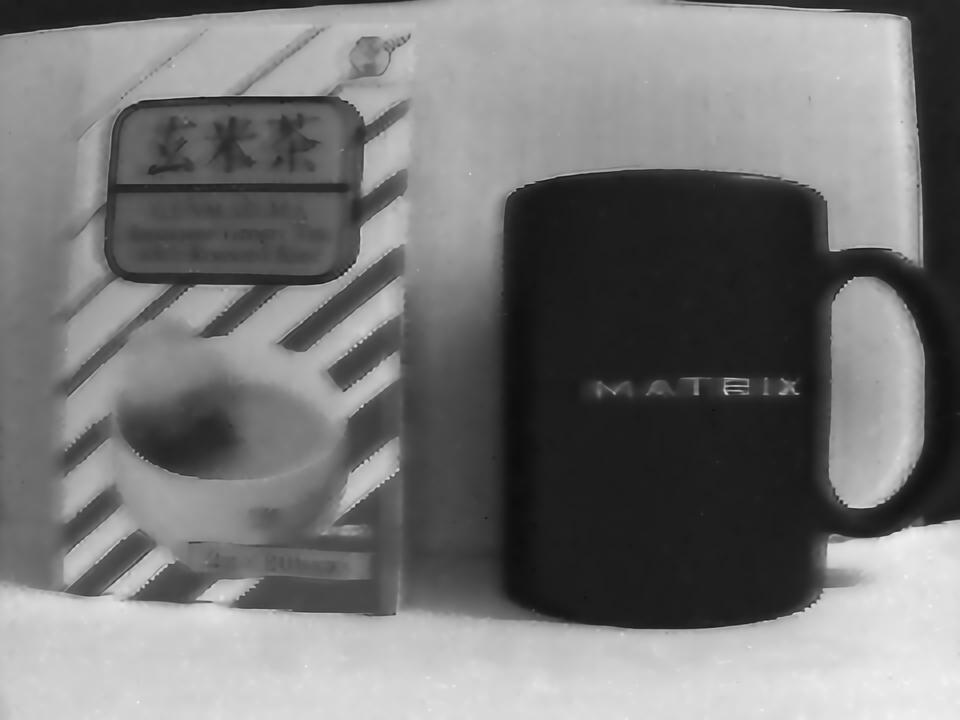}\hspace*{-0.5mm}
			\includegraphics[width=\cimwid\linewidth,trim={220px 230px 500px 100px},clip]{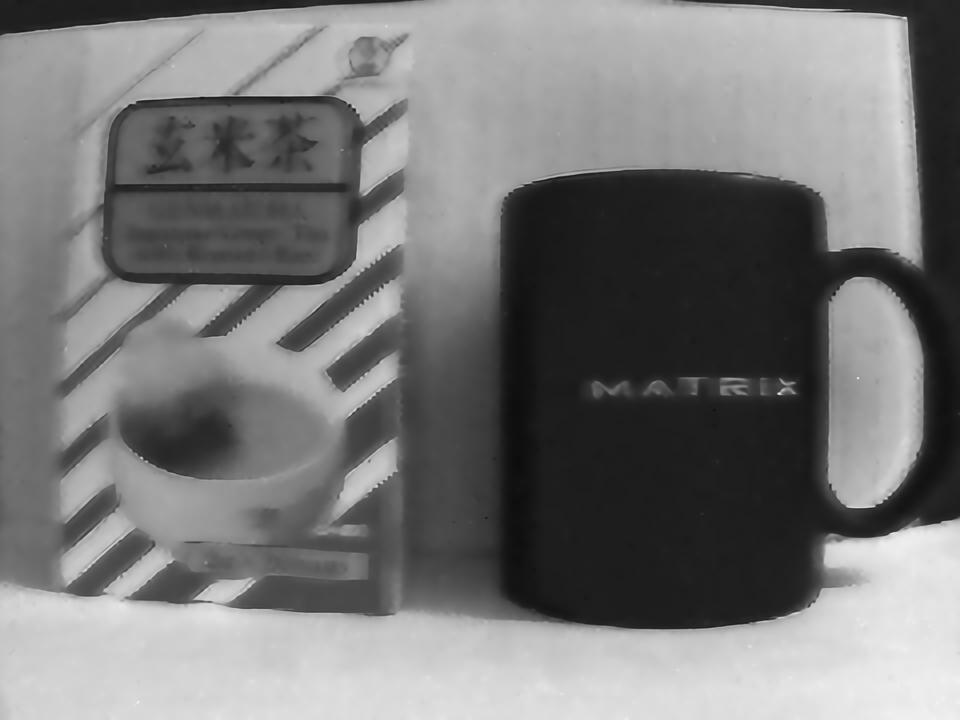}\hspace*{-0.5mm}
			\includegraphics[width=\cimwid\linewidth,trim={220px 230px 500px 100px},clip]{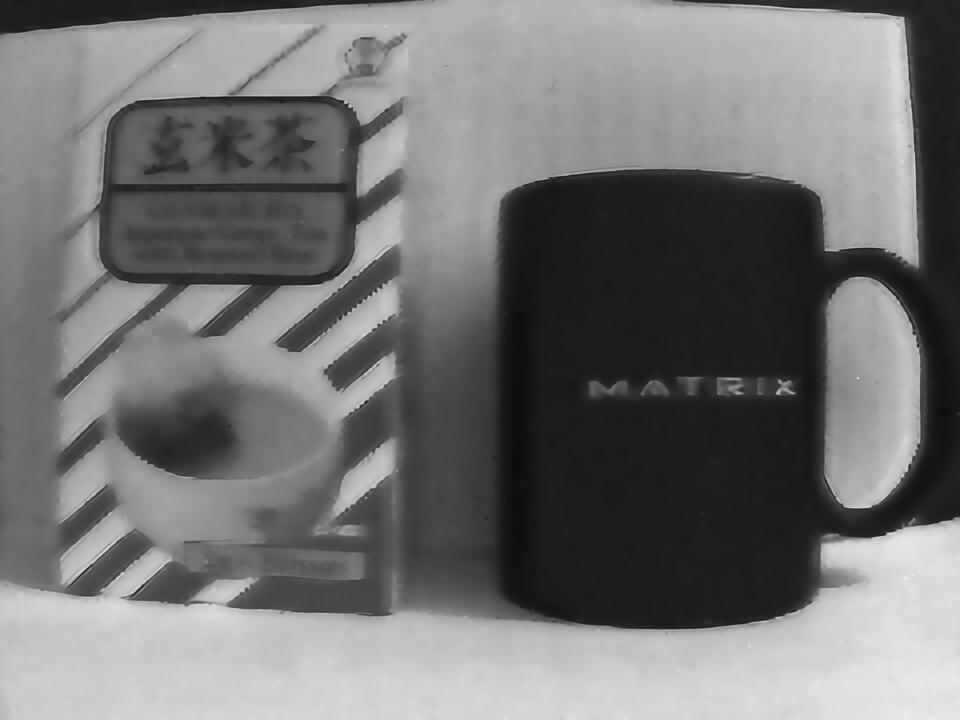}\hspace*{-0.5mm}
			\includegraphics[width=\cimwid\linewidth,trim={220px 230px 500px 100px},clip]{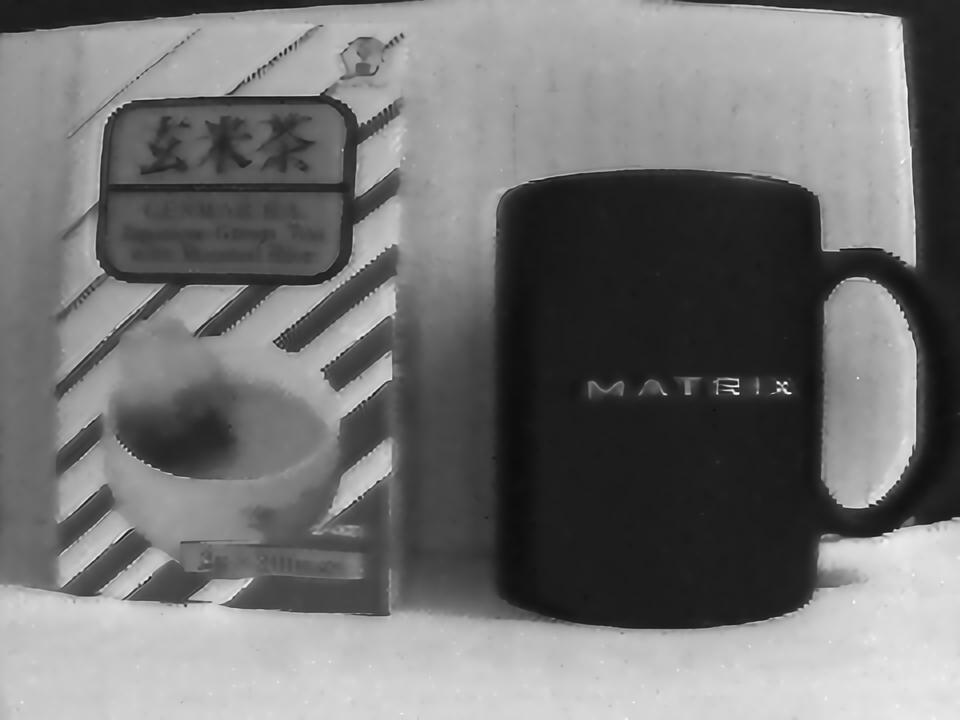}\hspace*{-0.5mm}
			\includegraphics[width=\cimwid\linewidth,trim={220px 230px 500px 100px},clip]{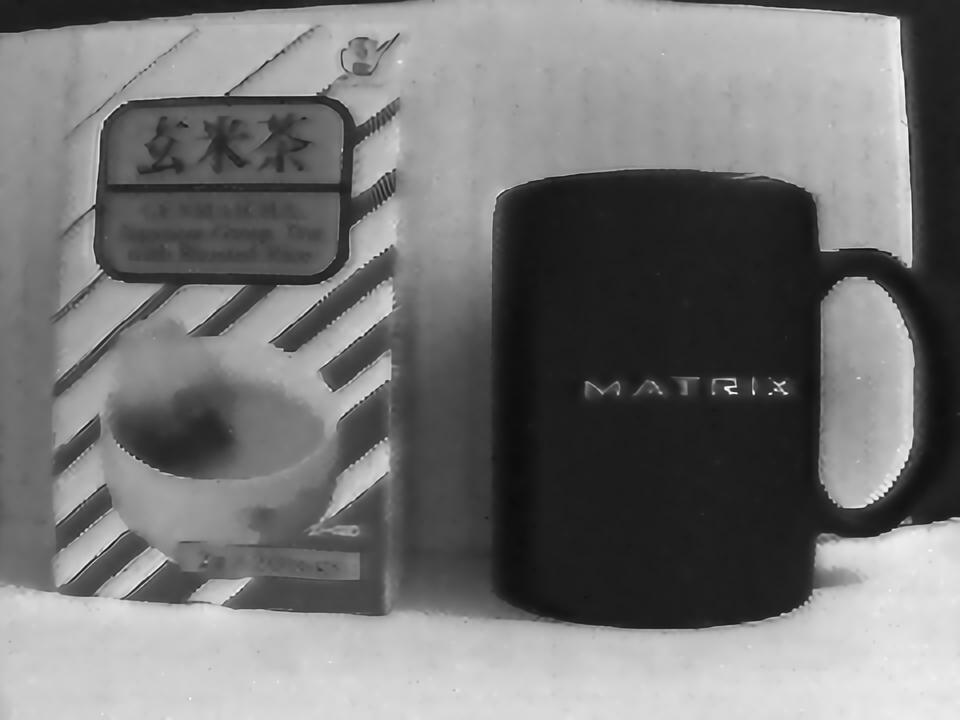}\hspace*{-0.5mm}
			\includegraphics[width=\cimwid\linewidth,trim={220px 230px 500px 100px},clip]{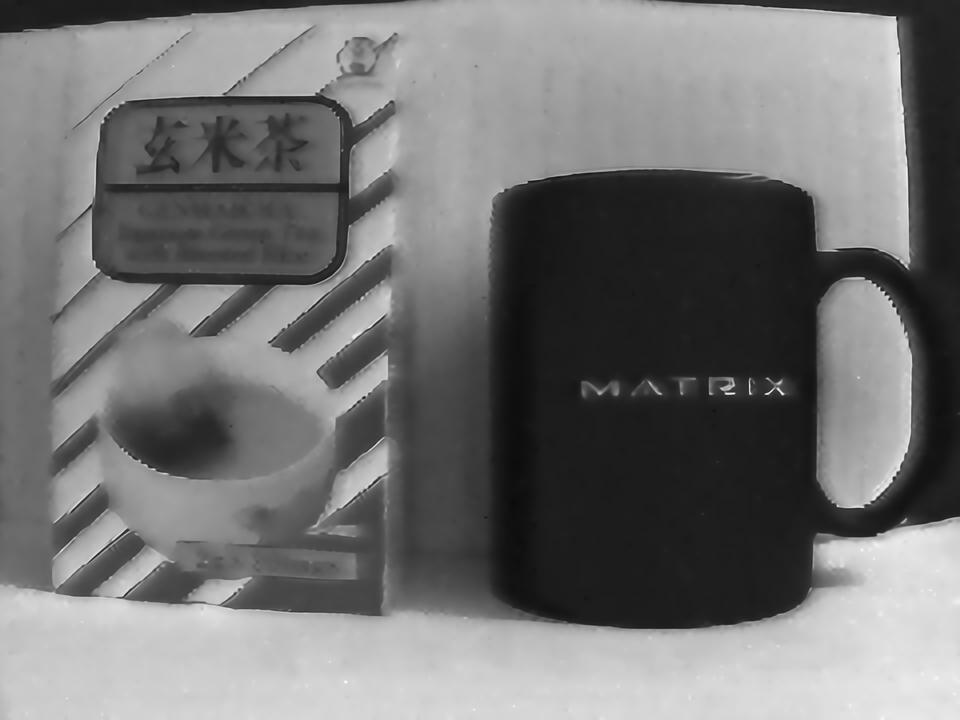}\hspace*{-0.5mm}
			\includegraphics[width=\cimwid\linewidth,trim={220px 230px 500px 100px},clip]{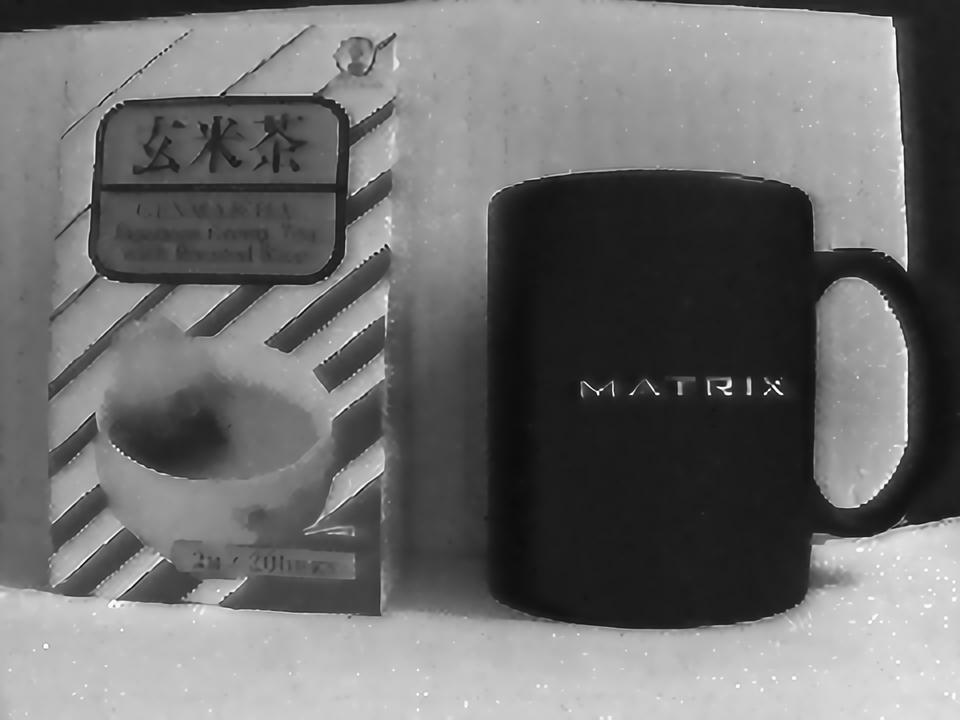}\hspace*{-0.5mm}
			\includegraphics[width=\cimwid\linewidth,trim={220px 230px 500px 100px},clip]{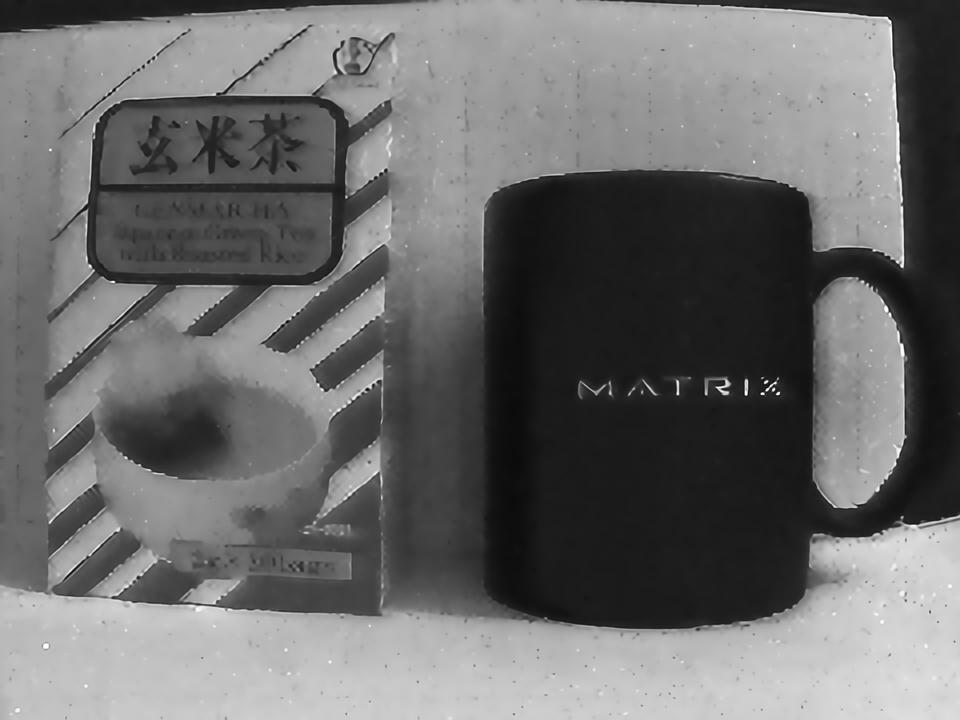}\hspace*{-0.5mm}
			\includegraphics[width=\cimwid\linewidth,trim={220px 230px 500px 100px},clip]{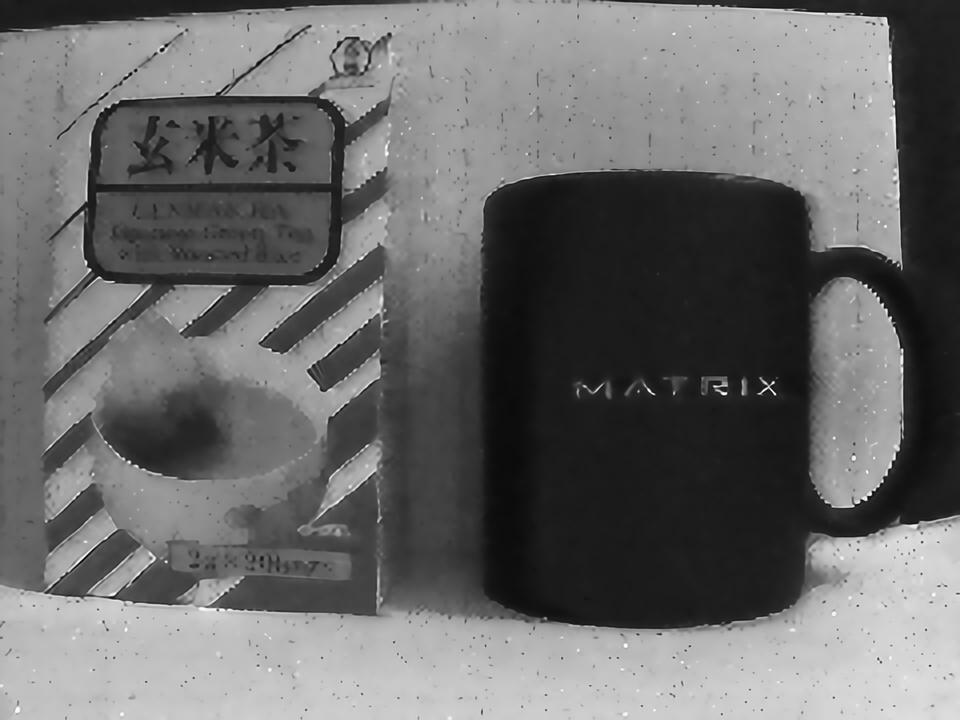}\hspace*{-0.5mm}
			\includegraphics[width=\cimwid\linewidth,trim={220px 230px 500px 100px},clip]{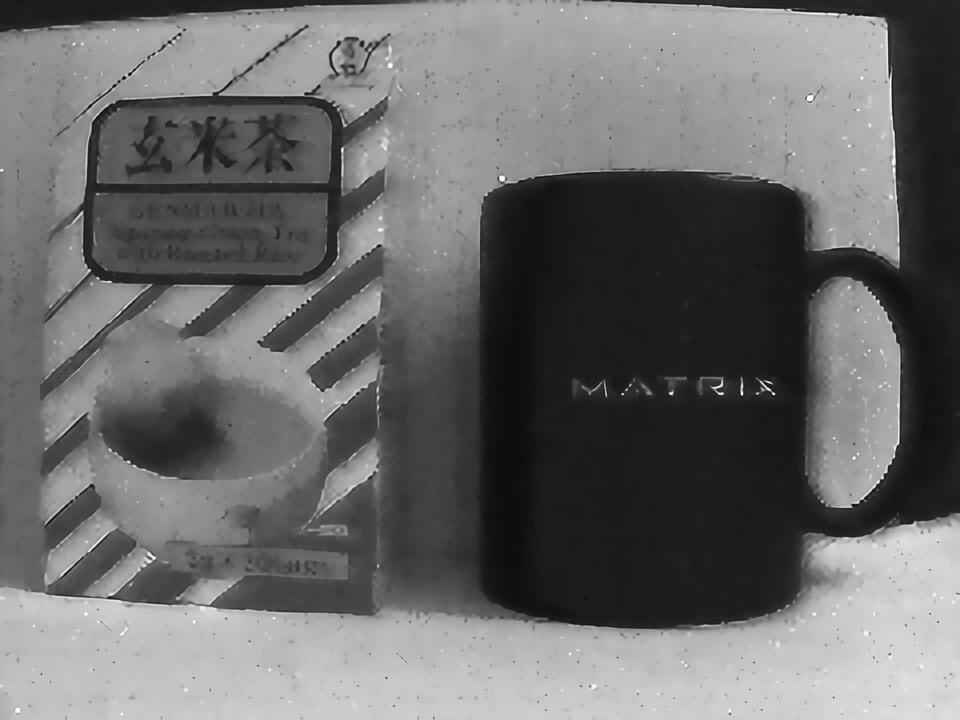}\hspace*{-0.5mm}
			\includegraphics[width=\cimwid\linewidth,trim={220px 230px 500px 100px},clip]{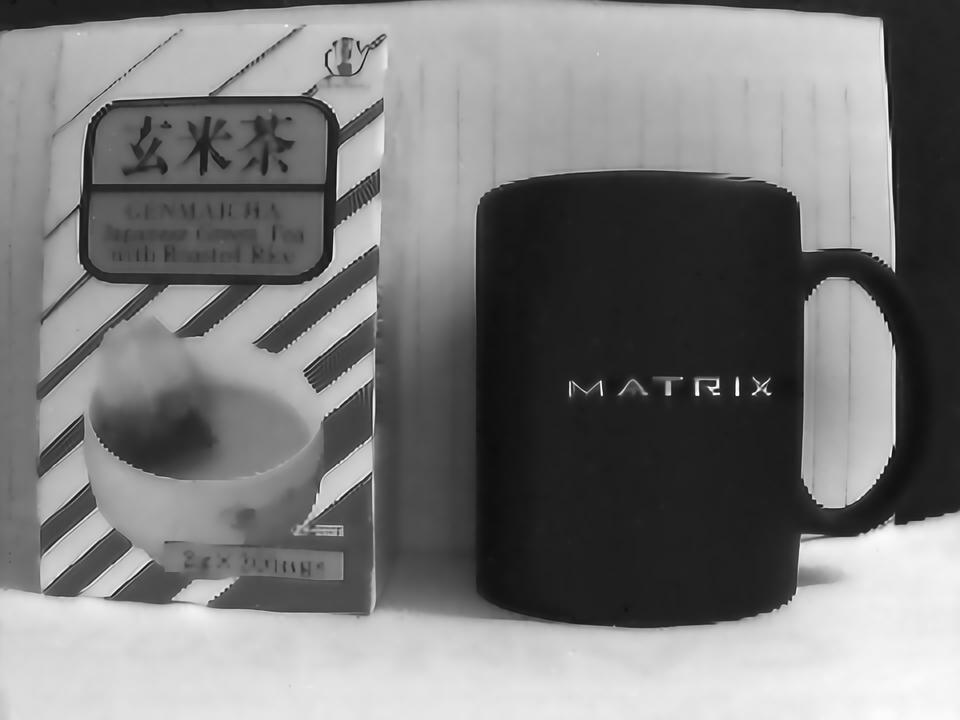}\hspace*{-0.5mm}
	% }
 \vspace{.1em}
	\\
	\def\removelag{-1.3mm}
	% \subfigure{%left, bottom, right and top
	        \begin{tikzpicture}[inner sep=0]
            \node [label={[label distance=0.1cm,text depth=-1ex,rotate=90]right: {\scriptsize \text{LEDVDI+RCAN}}}] at (0,8.7) {};
            \end{tikzpicture}
	        \includegraphics[width=\cimwid\linewidth,trim={220px 230px 500px 100px},,clip=true]{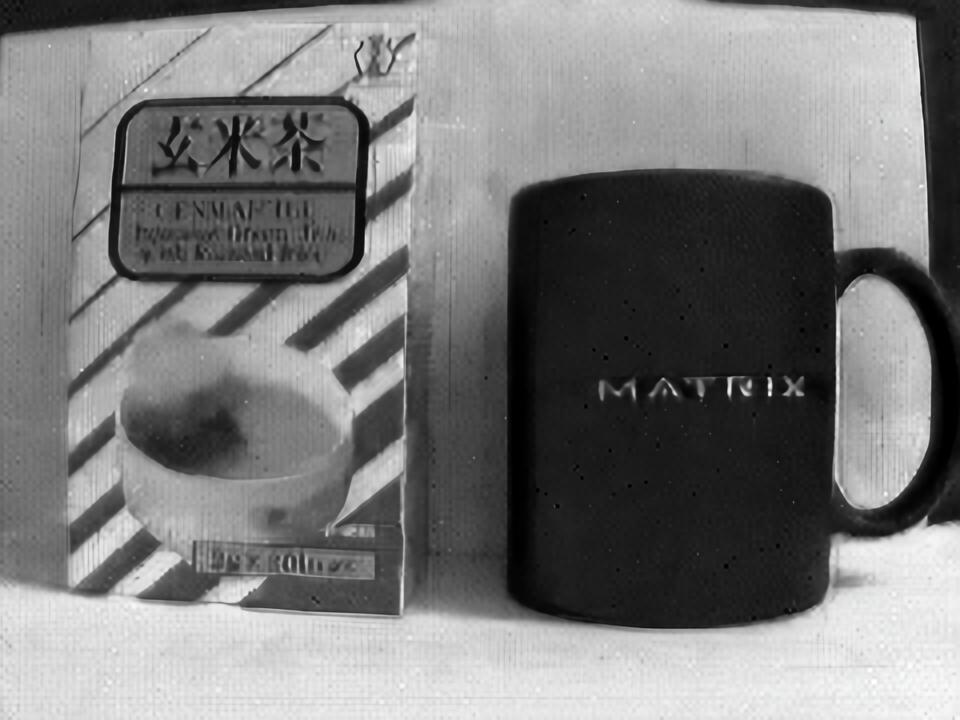}\hspace*{-0.5mm}
			\hspace*{\cimwid\linewidth}
			\hspace*{\removelag}
			\includegraphics[width=\cimwid\linewidth,trim={220px 230px 500px 100px},,clip=true]{pic/mulframe/ledvdi/102_reconstruction.jpg}\hspace*{-0.5mm}
            \hspace*{\cimwid\linewidth}
			\hspace*{\removelag}
			\includegraphics[width=\cimwid\linewidth,trim={220px 230px 500px 100px},clip]{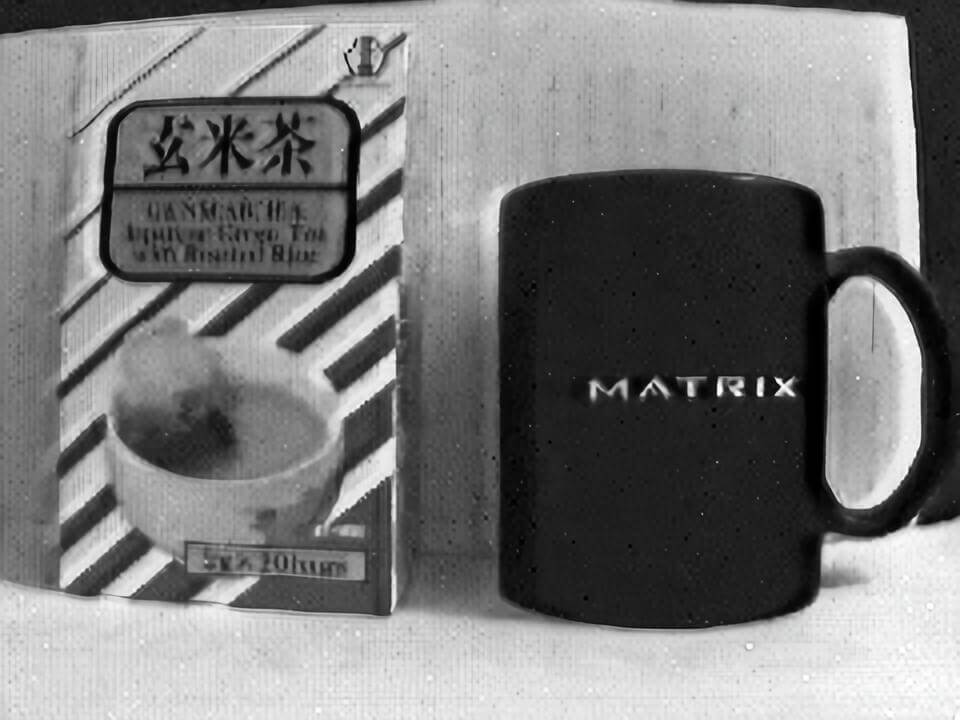}\hspace*{-0.5mm}
            \hspace*{\cimwid\linewidth}
			\hspace*{\removelag}
			\includegraphics[width=\cimwid\linewidth,trim={220px 230px 500px 100px},clip]{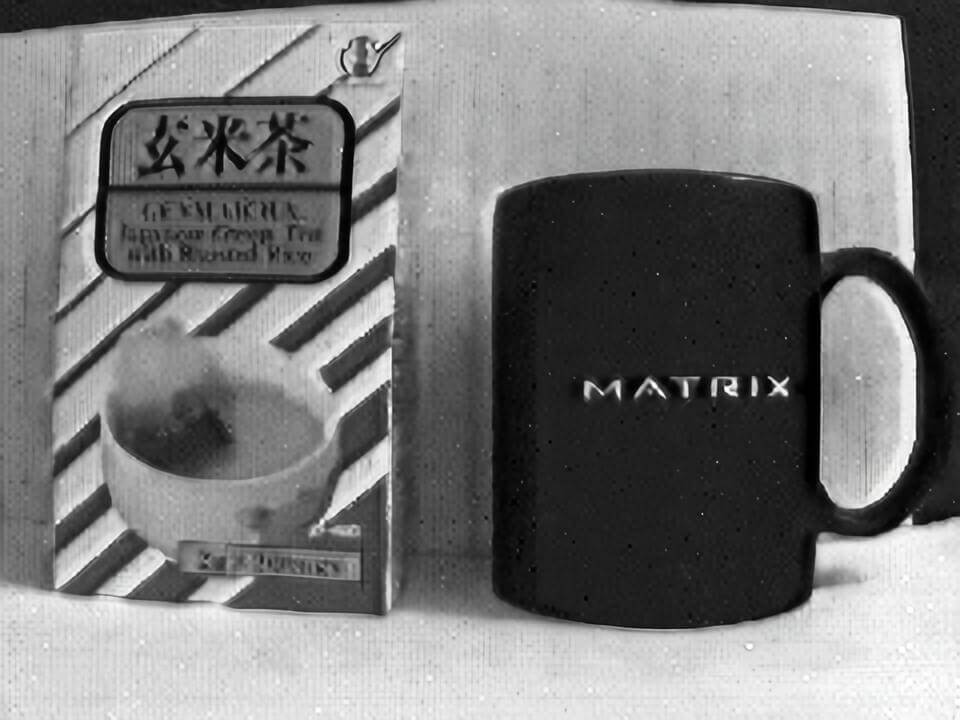}\hspace*{-0.5mm}
           \hspace*{\cimwid\linewidth}
			\hspace*{\removelag}
			\includegraphics[width=\cimwid\linewidth,trim={220px 230px 500px 100px},clip]{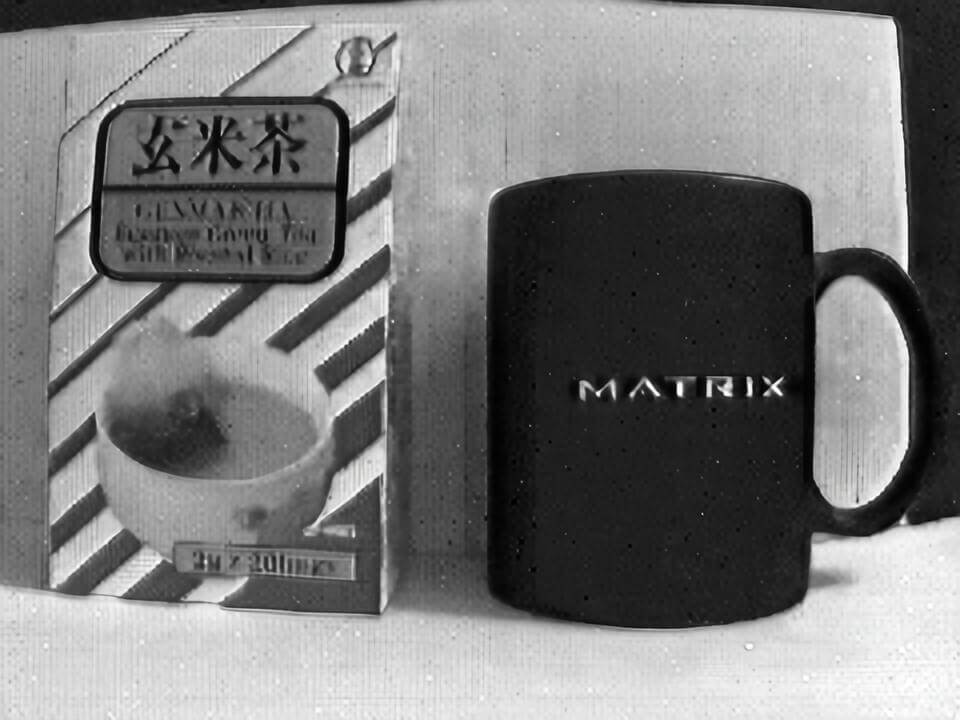}\hspace*{-0.5mm}
            \hspace*{\cimwid\linewidth}
			\hspace*{\removelag}
			\includegraphics[width=\cimwid\linewidth,trim={220px 230px 500px 100px},clip]{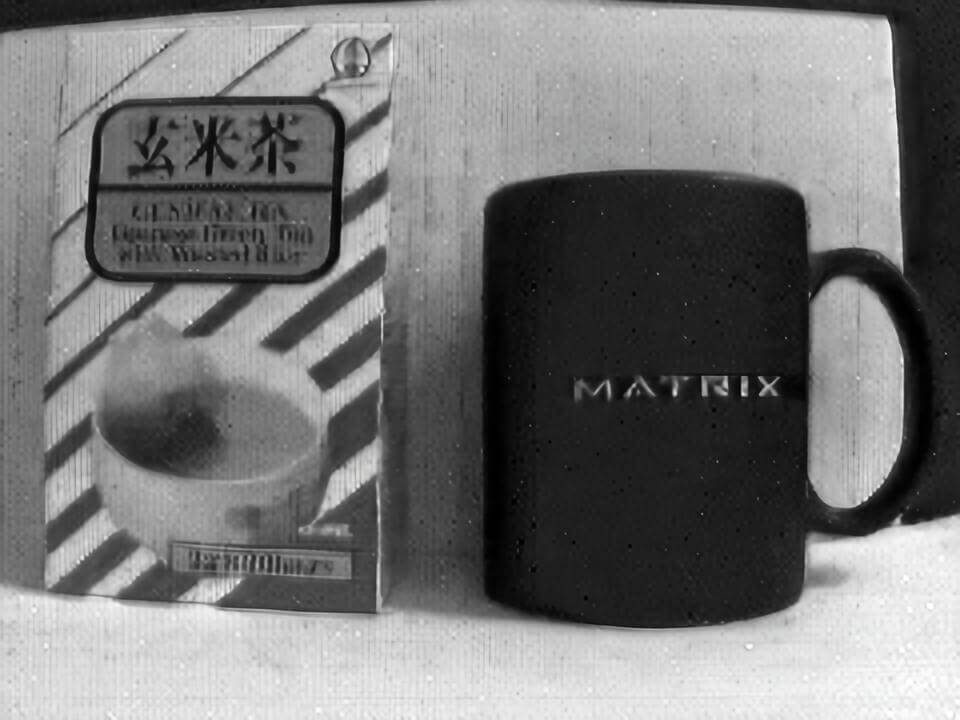}\hspace*{-0.5mm}
            \hspace*{\cimwid\linewidth}
			\hspace*{\removelag}
			\includegraphics[width=\cimwid\linewidth,trim={220px 230px 500px 100px},clip]{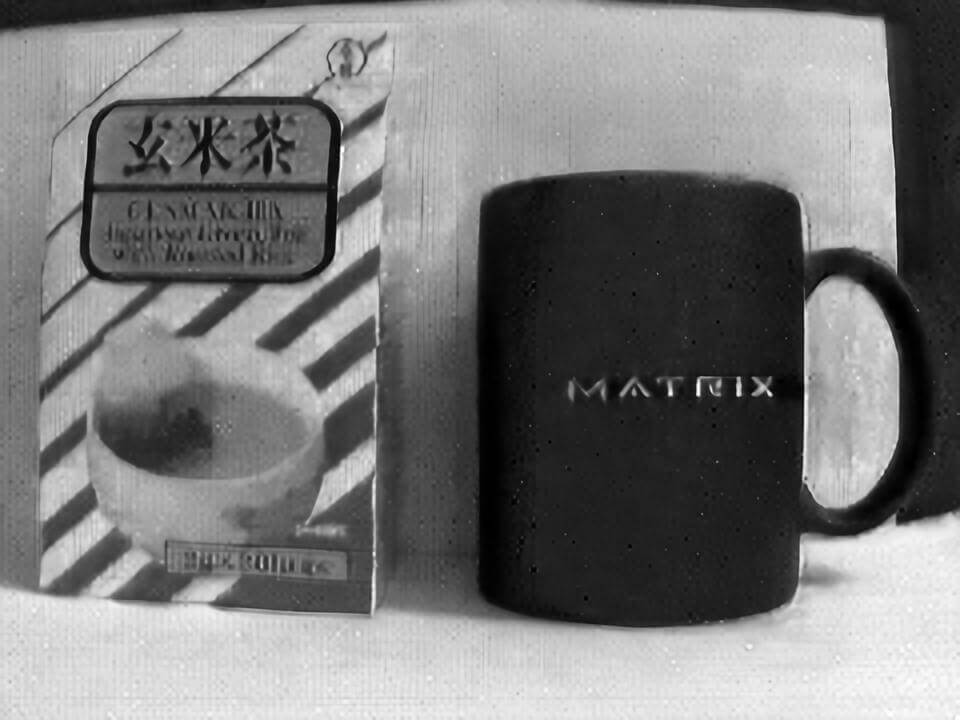}
% 			\hspace*{-0.5mm}
% 			\includegraphics[width=\cimwid\linewidth,trim={220px 230px 500px 100px},clip]{pic/mulframe/ledvdi/0019_11.jpg}\hspace*{-0.5mm}
% 			\includegraphics[width=\cimwid\linewidth,trim={220px 230px 500px 100px},clip]{pic/mulframe/ledvdi/0019_12.jpg}
	% }
\vspace{.1em}
	\\
		% \subfigure{%left, bottom, right and top
		    \begin{tikzpicture}[inner sep=0]
            \node [label={[label distance=0.1cm,text depth=-1ex,rotate=90]right: {\scriptsize \text{RED-Net+RCAN}}}] at (0,8.7) {};
            \end{tikzpicture}
	        \includegraphics[width=\cimwid\linewidth,trim={220px 230px 500px 100px},,clip=true]{pic/mulframe/red/0019_00.jpg}\hspace*{-0.5mm}
			\hspace*{\cimwid\linewidth}
			\hspace*{\removelag}
			\includegraphics[width=\cimwid\linewidth,trim={220px 230px 500px 100px},,clip=true]{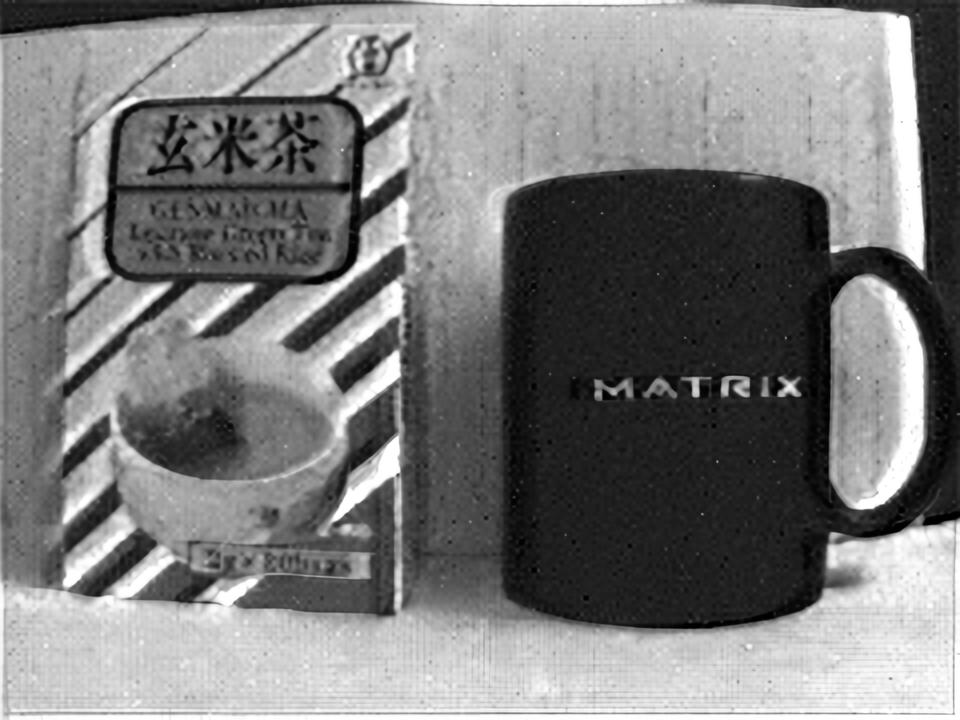}\hspace*{-0.5mm}
            \hspace*{\cimwid\linewidth}
			\hspace*{\removelag}
			\includegraphics[width=\cimwid\linewidth,trim={220px 230px 500px 100px},clip]{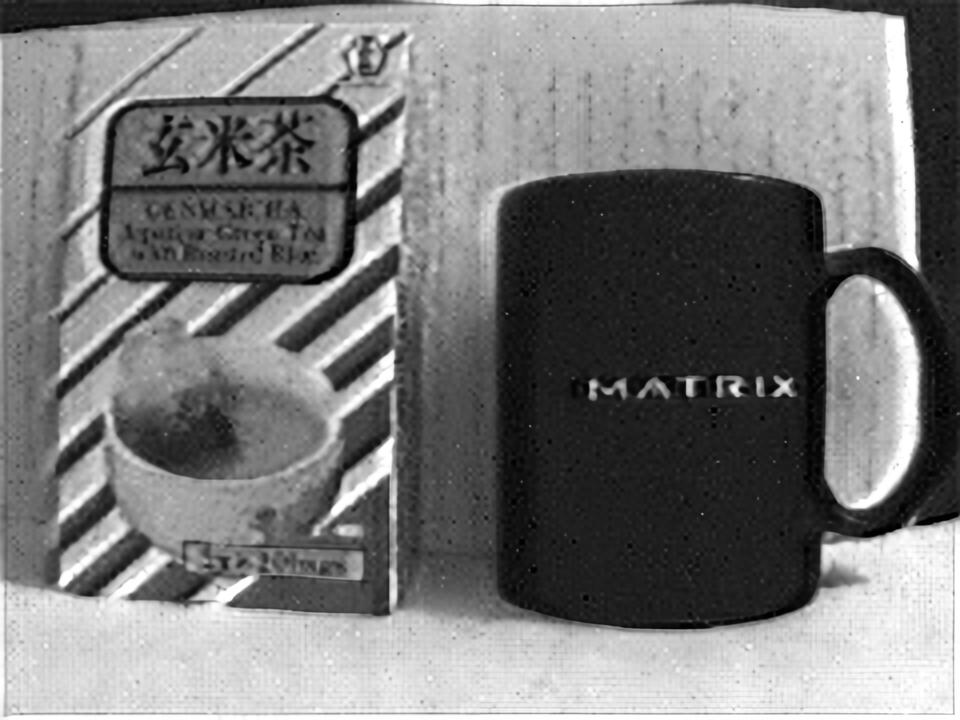}\hspace*{-0.5mm}
            \hspace*{\cimwid\linewidth}
			\hspace*{\removelag}
			\includegraphics[width=\cimwid\linewidth,trim={220px 230px 500px 100px},clip]{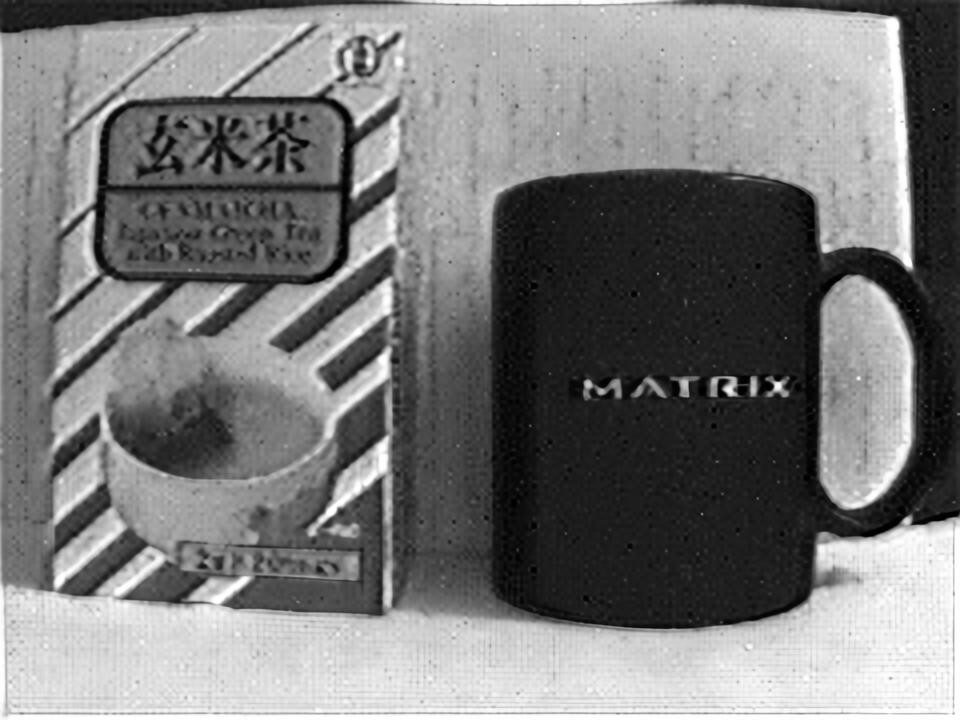}\hspace*{-0.5mm}
           \hspace*{\cimwid\linewidth}
			\hspace*{\removelag}
			\includegraphics[width=\cimwid\linewidth,trim={220px 230px 500px 100px},clip]{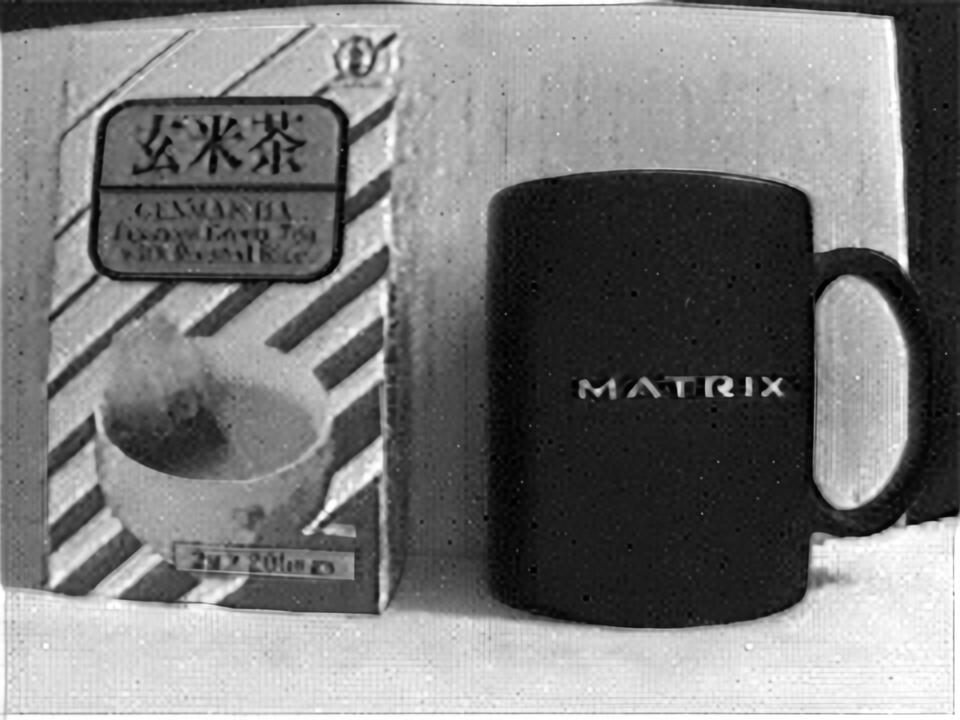}\hspace*{-0.5mm}
            \hspace*{\cimwid\linewidth}
			\hspace*{\removelag}
			\includegraphics[width=\cimwid\linewidth,trim={220px 230px 500px 100px},clip]{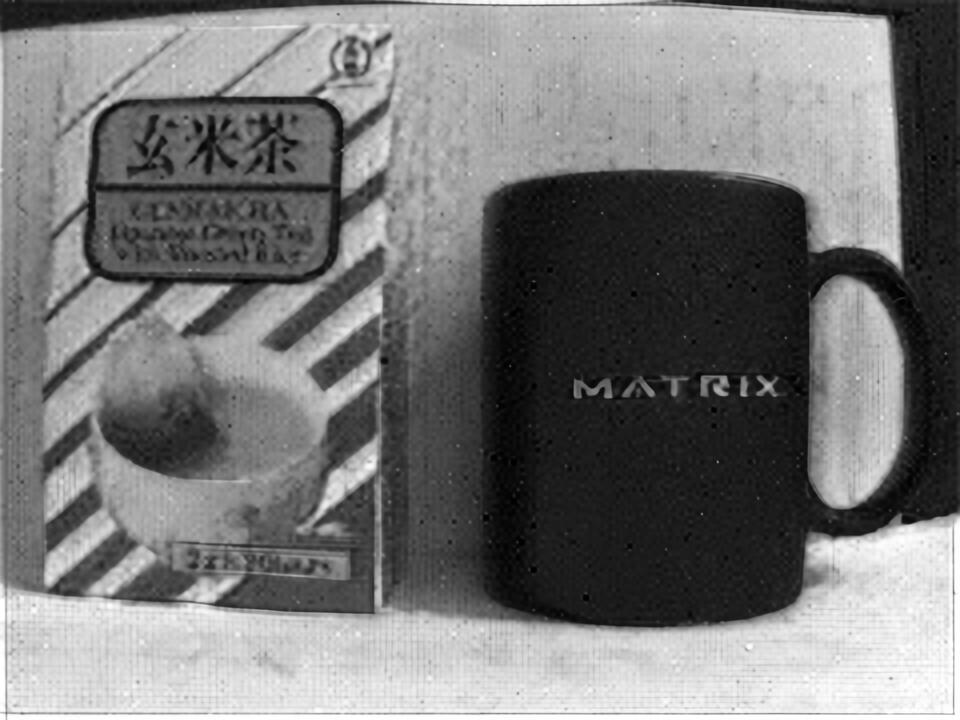}\hspace*{-0.5mm}
            \hspace*{\cimwid\linewidth}
			\hspace*{\removelag}
			\includegraphics[width=\cimwid\linewidth,trim={220px 230px 500px 100px},clip]{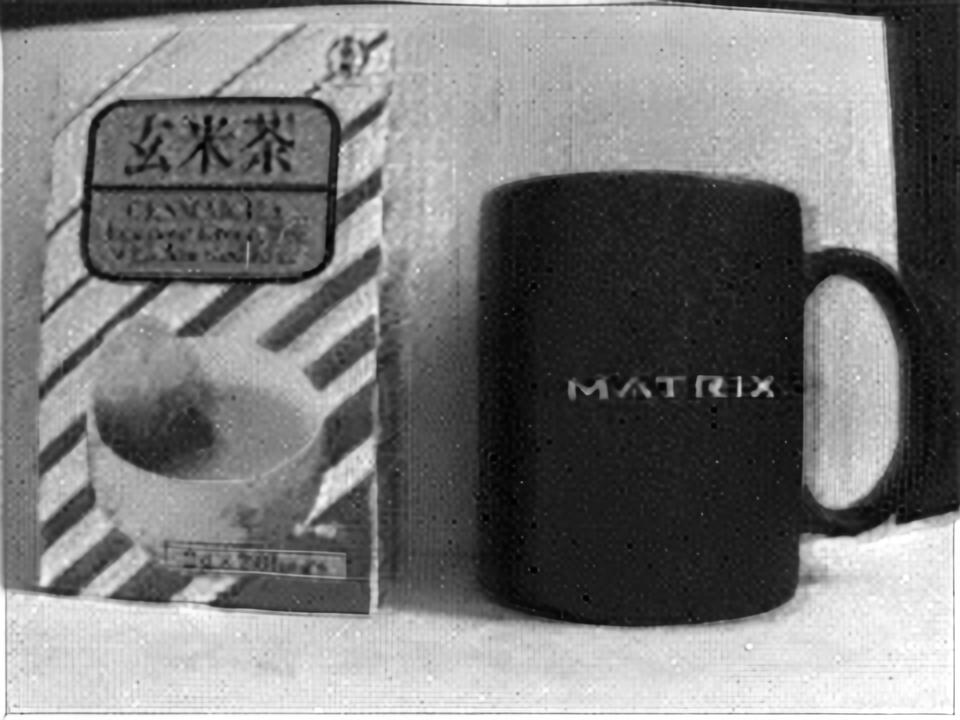}
% 			\hspace*{-0.5mm}
% 			\includegraphics[width=\cimwid\linewidth,trim={220px 230px 500px 100px},clip]{pic/mulframe/ledvdi/0019_11.jpg}\hspace*{-0.5mm}
% 			\includegraphics[width=\cimwid\linewidth,trim={220px 230px 500px 100px},clip]{pic/mulframe/ledvdi/0019_12.jpg}
	% }
  \vspace{.1em}
	\\
	% \subfigure{%left, bottom, right and top
	        \begin{tikzpicture}[inner sep=0]
            \node [label={[label distance=0.5cm,text depth=-1ex,rotate=90]right: {\scriptsize \textbf{eSL-Net}}}] at (0,8.7) {};
            \end{tikzpicture}
			\includegraphics[width=\cimwid\linewidth,trim={220px 230px 500px 100px},,clip=true]{pic/mulframe/eslnet/0019_0.png}\hspace*{-0.5mm}
			\includegraphics[width=\cimwid\linewidth,trim={220px 230px 500px 100px},clip]{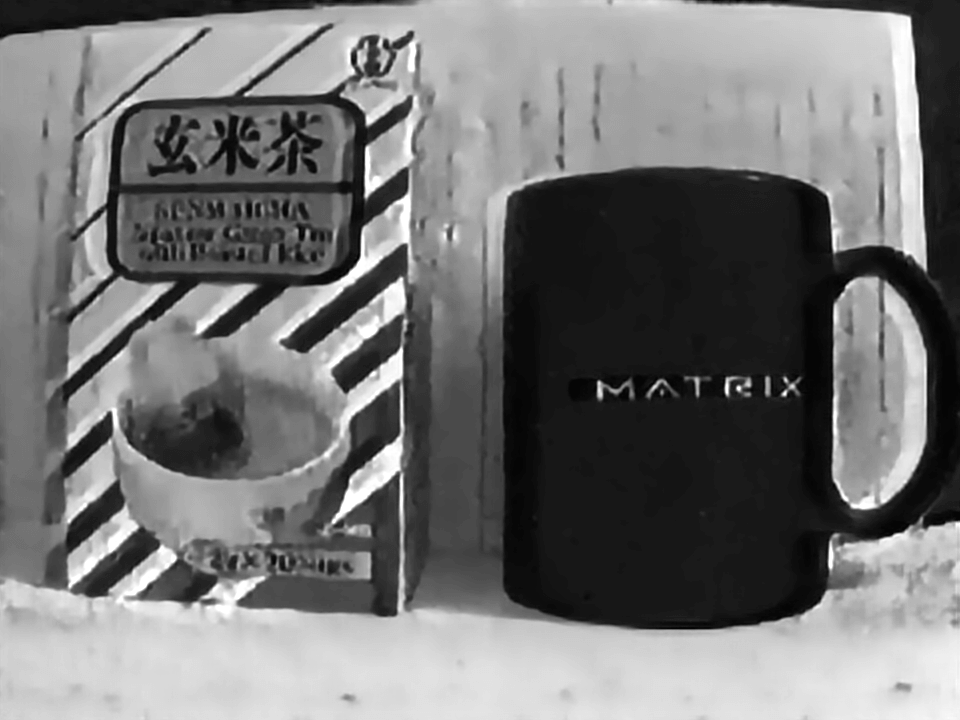}\hspace*{-0.5mm}
			\includegraphics[width=\cimwid\linewidth,trim={220px 230px 500px 100px},clip]{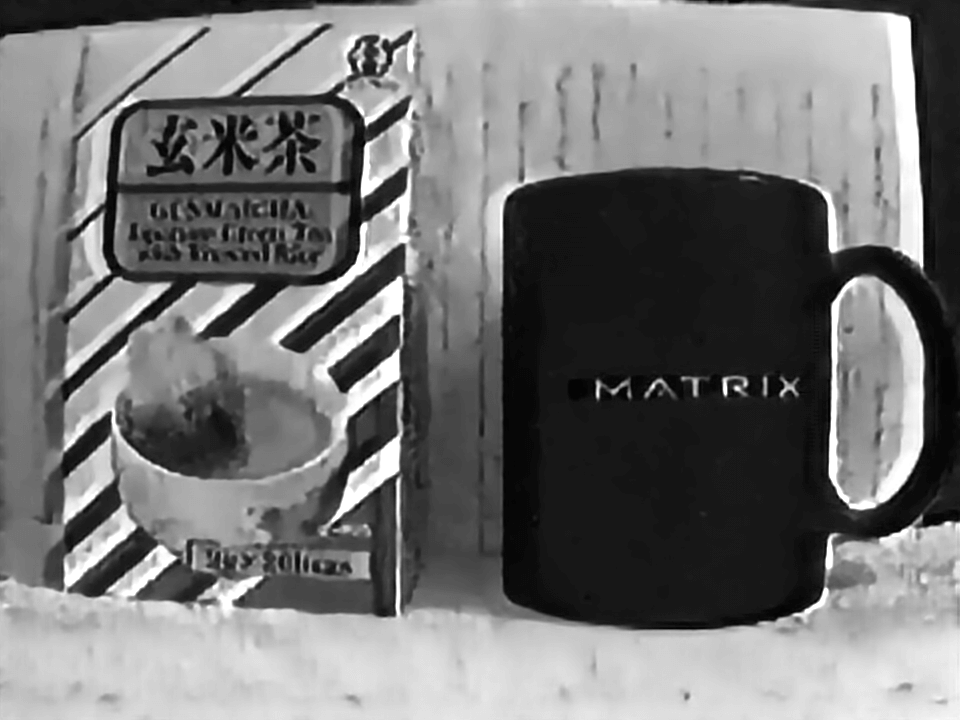}\hspace*{-0.5mm}
			\includegraphics[width=\cimwid\linewidth,trim={220px 230px 500px 100px},clip]{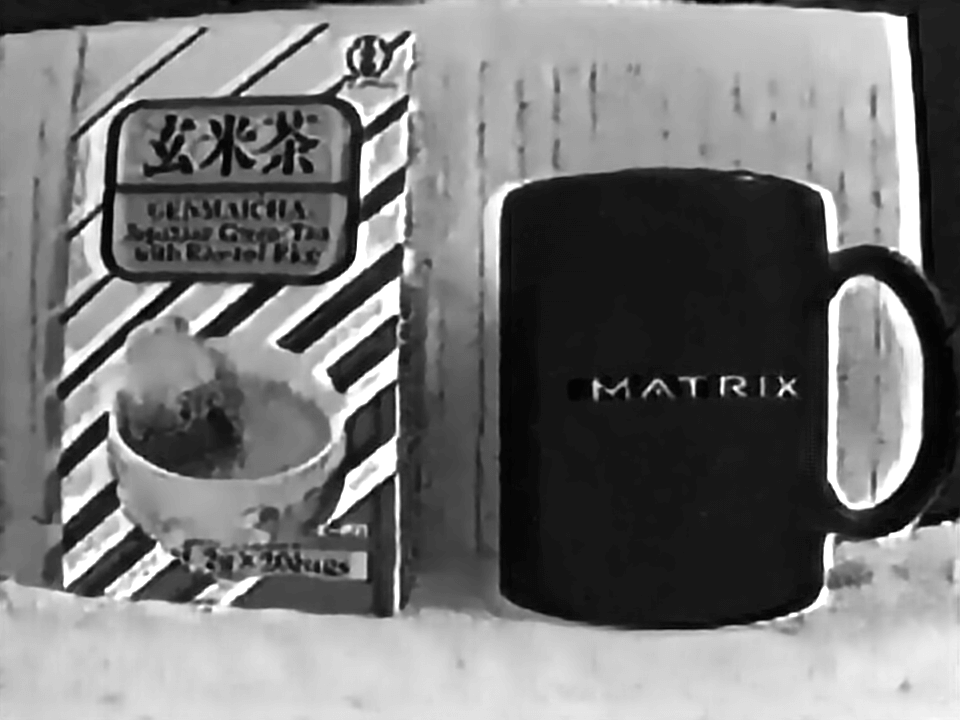}\hspace*{-0.5mm}
			\includegraphics[width=\cimwid\linewidth,trim={220px 230px 500px 100px},clip]{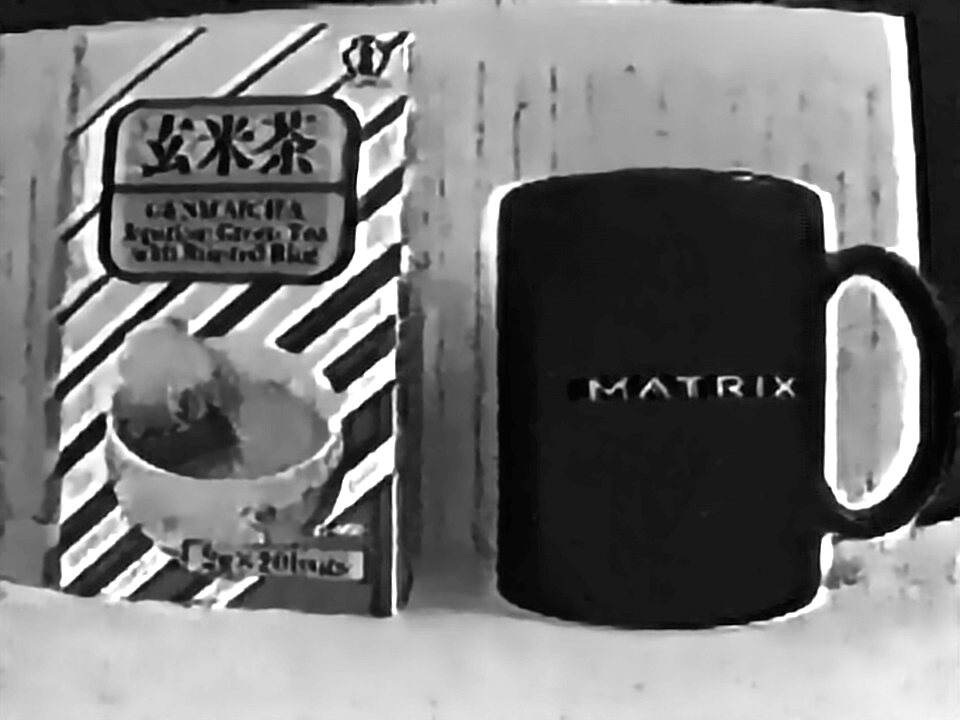}\hspace*{-0.5mm}
			\includegraphics[width=\cimwid\linewidth,trim={220px 230px 500px 100px},clip]{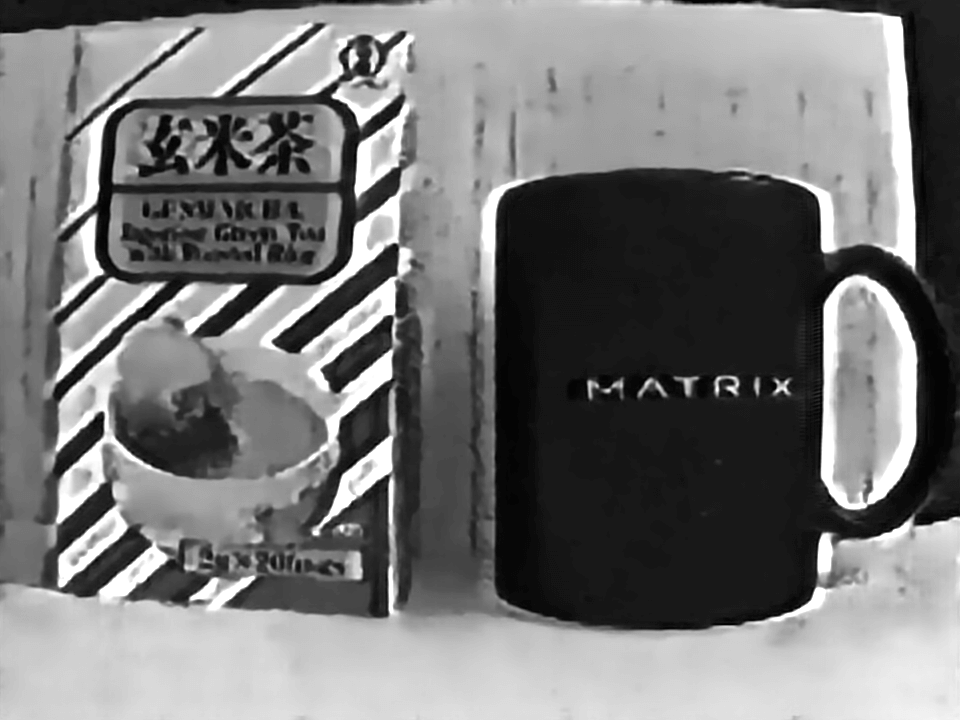}\hspace*{-0.5mm}
			\includegraphics[width=\cimwid\linewidth,trim={220px 230px 500px 100px},clip]{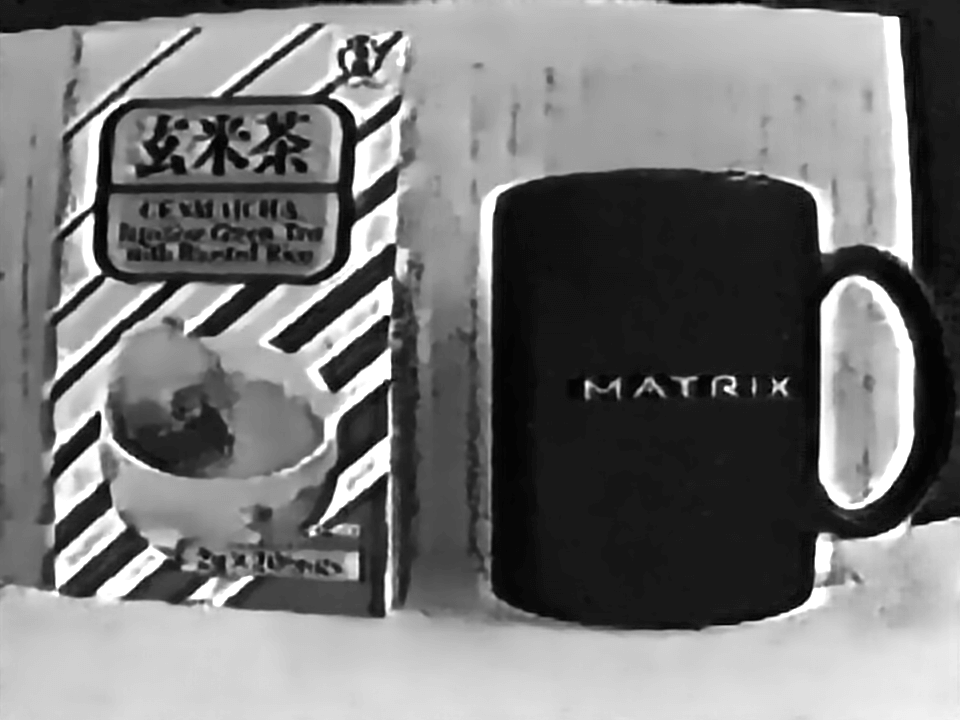}\hspace*{-0.5mm}
			\includegraphics[width=\cimwid\linewidth,trim={220px 230px 500px 100px},clip]{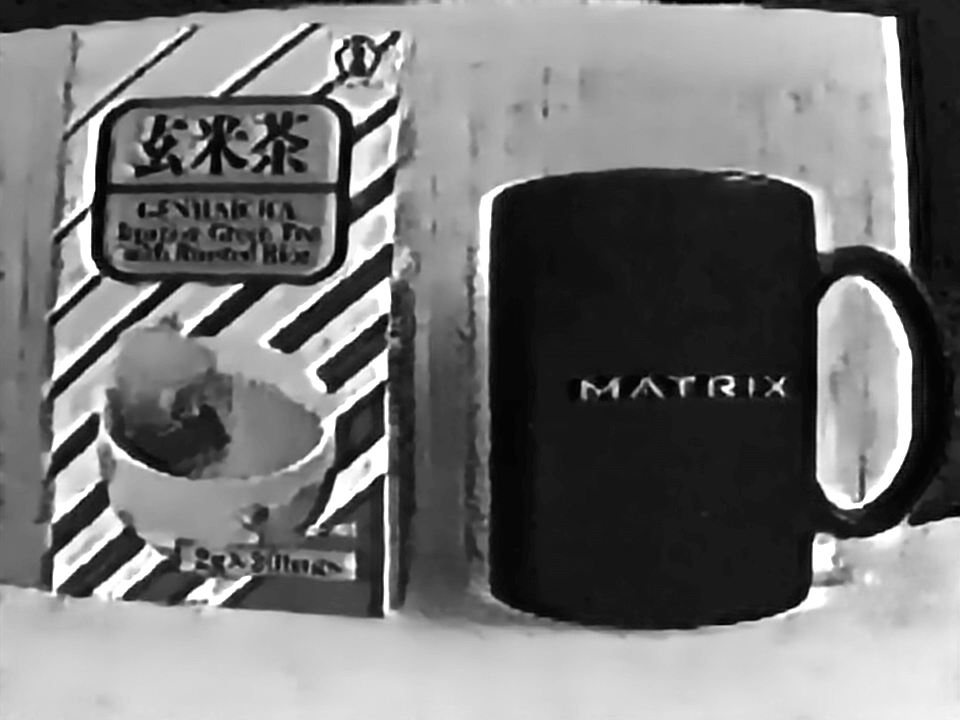}\hspace*{-0.5mm}
			\includegraphics[width=\cimwid\linewidth,trim={220px 230px 500px 100px},clip]{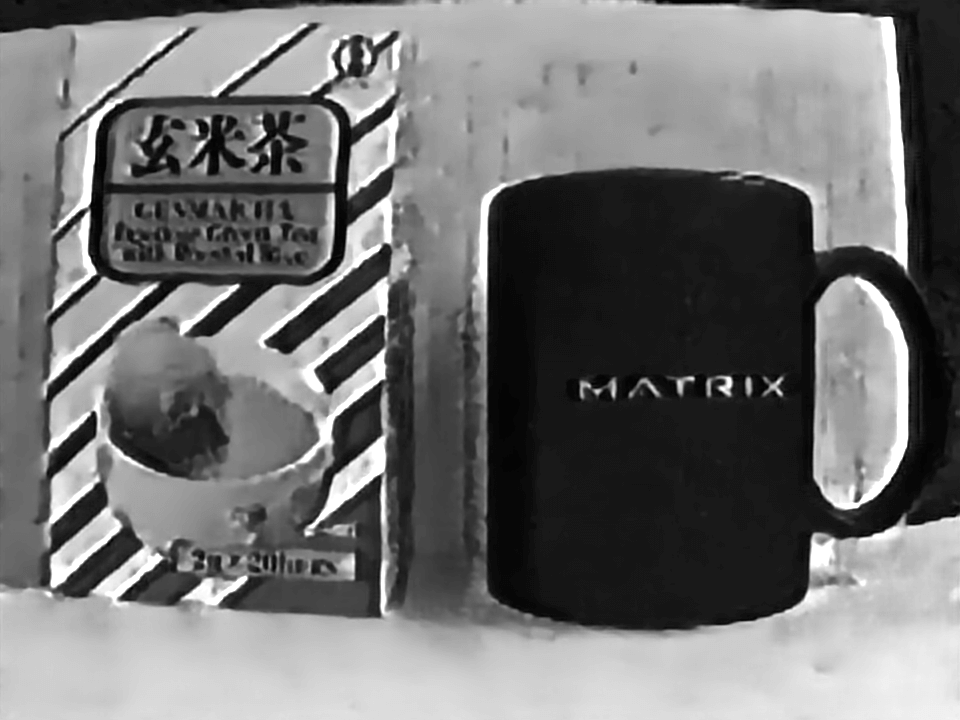}\hspace*{-0.5mm}
			\includegraphics[width=\cimwid\linewidth,trim={220px 230px 500px 100px},clip]{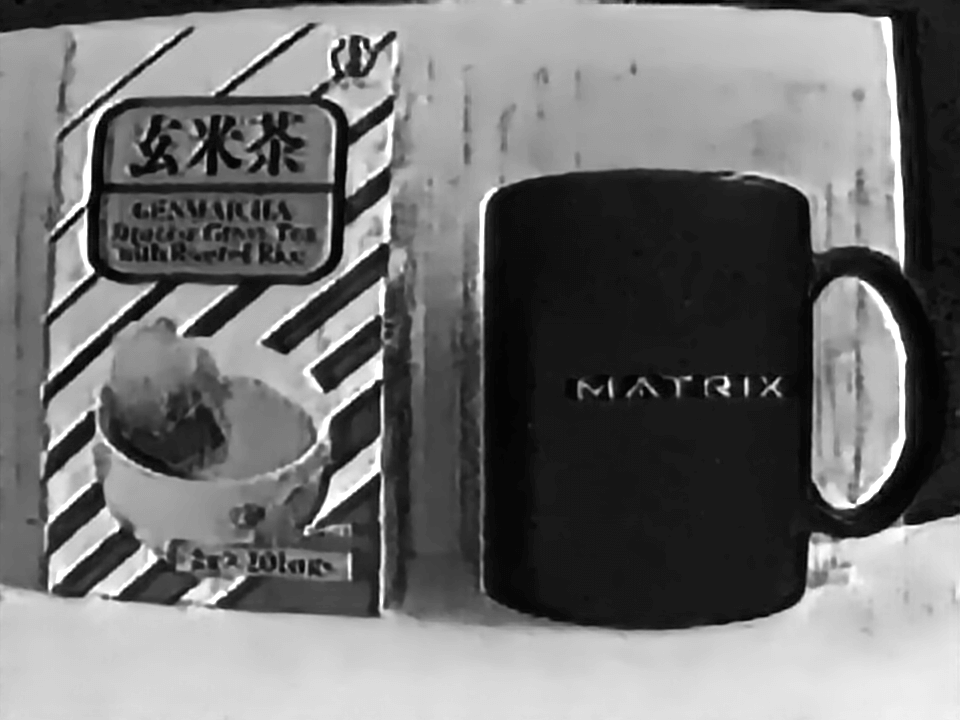}\hspace*{-0.5mm}
			\includegraphics[width=\cimwid\linewidth,trim={220px 230px 500px 100px},clip]{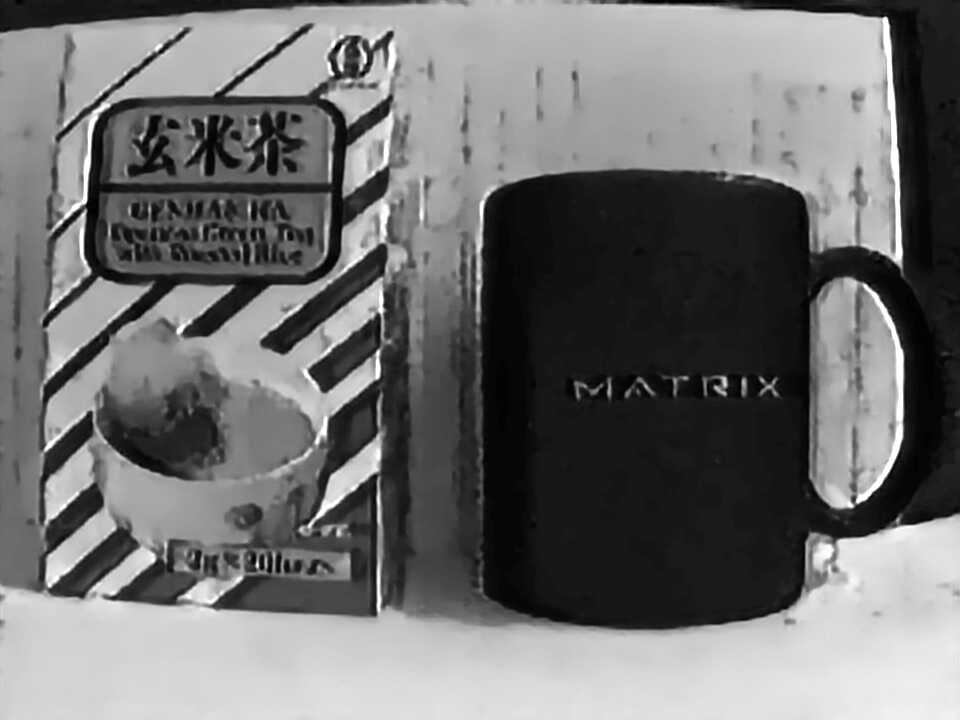}\hspace*{-0.5mm}
			\includegraphics[width=\cimwid\linewidth,trim={220px 230px 500px 100px},clip]{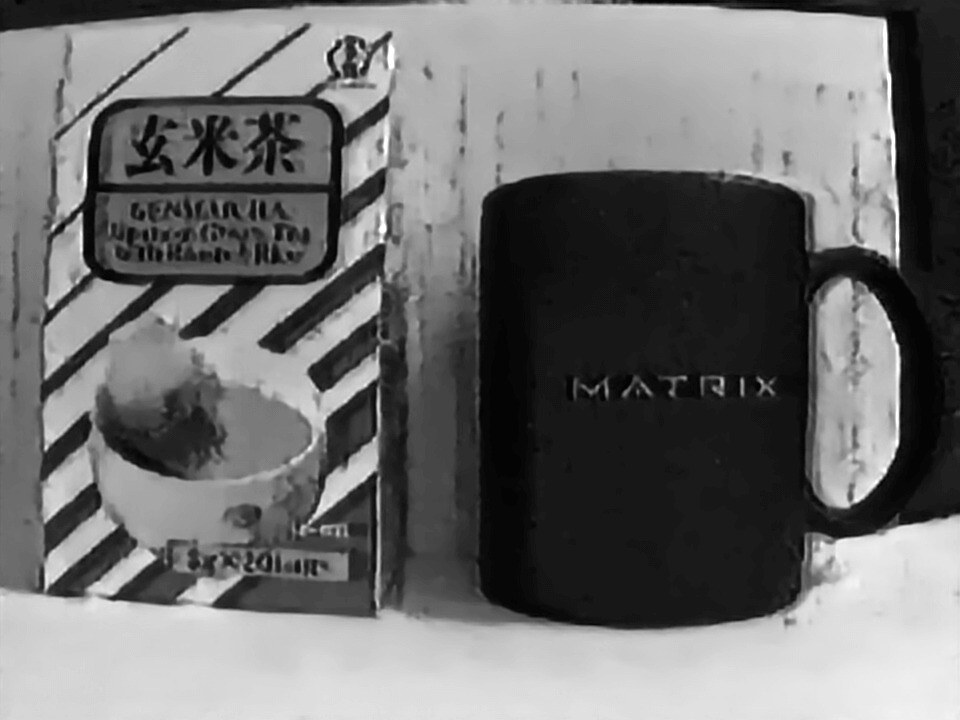}\hspace*{-0.5mm}
			\includegraphics[width=\cimwid\linewidth,trim={220px 230px 500px 100px},clip]{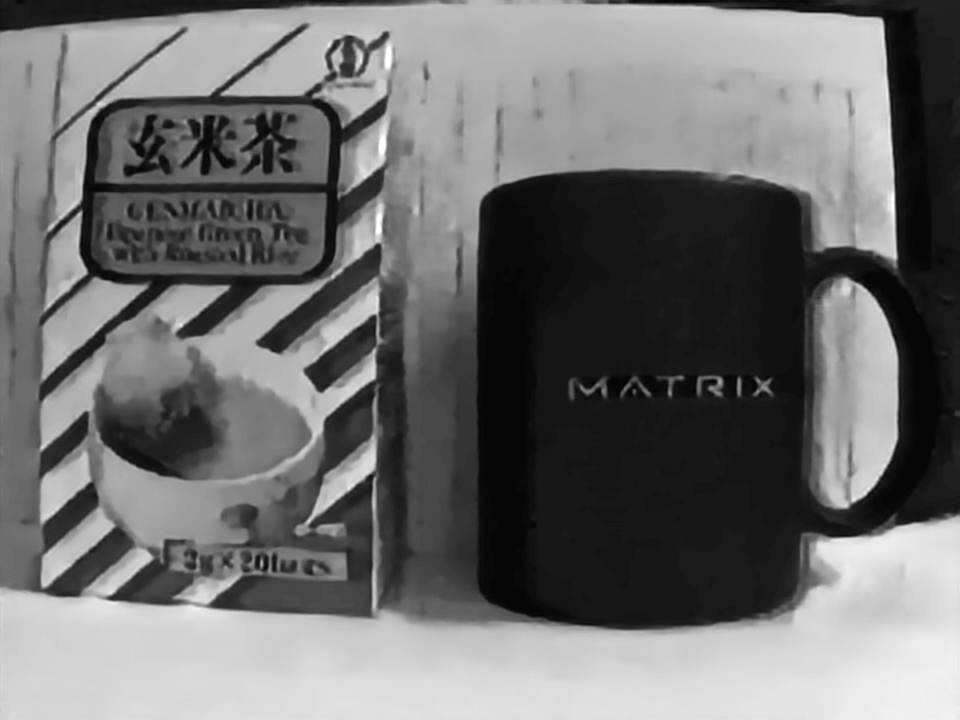}
	% }
 \vspace{.1em}
	\\
	% \subfigure{%left, bottom, right and top
	        \begin{tikzpicture}[inner sep=0]
            \node [label={[label distance=0.4cm,text depth=-1ex,rotate=90]right: {\scriptsize \textbf{eSL-Net++}}}] at (0,8.7) {};
            \end{tikzpicture}
			\includegraphics[width=\cimwid\linewidth,trim={220px 230px 500px 100px},,clip=true]{pic/mulframe/eslnet++/0019_0.png}\hspace*{-0.5mm}
			\includegraphics[width=\cimwid\linewidth,trim={220px 230px 500px 100px},clip]{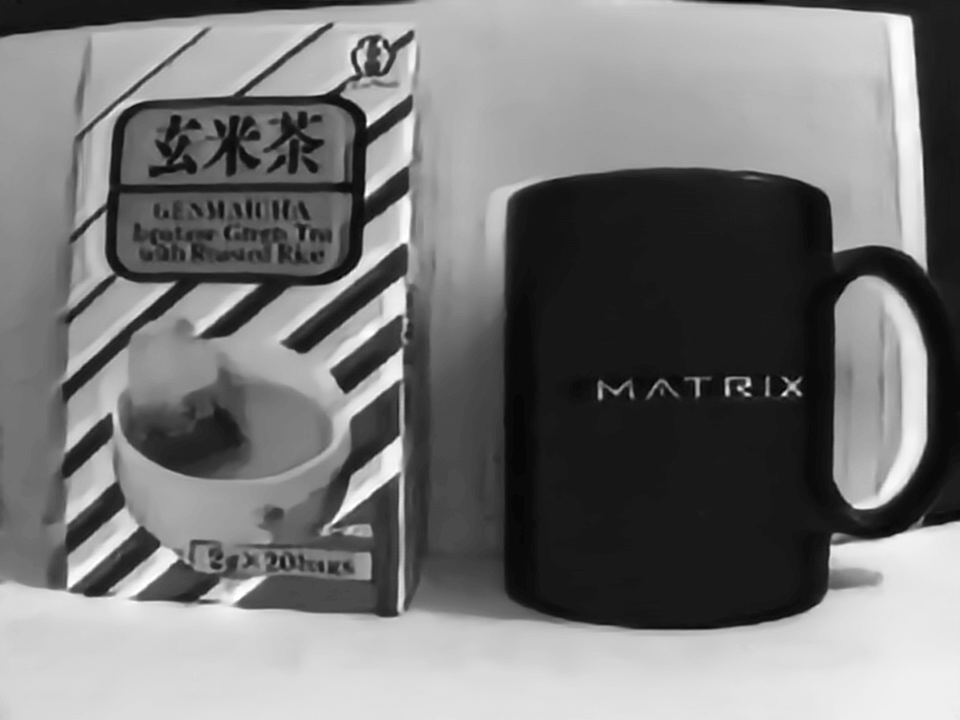}\hspace*{-0.5mm}
			\includegraphics[width=\cimwid\linewidth,trim={220px 230px 500px 100px},clip]{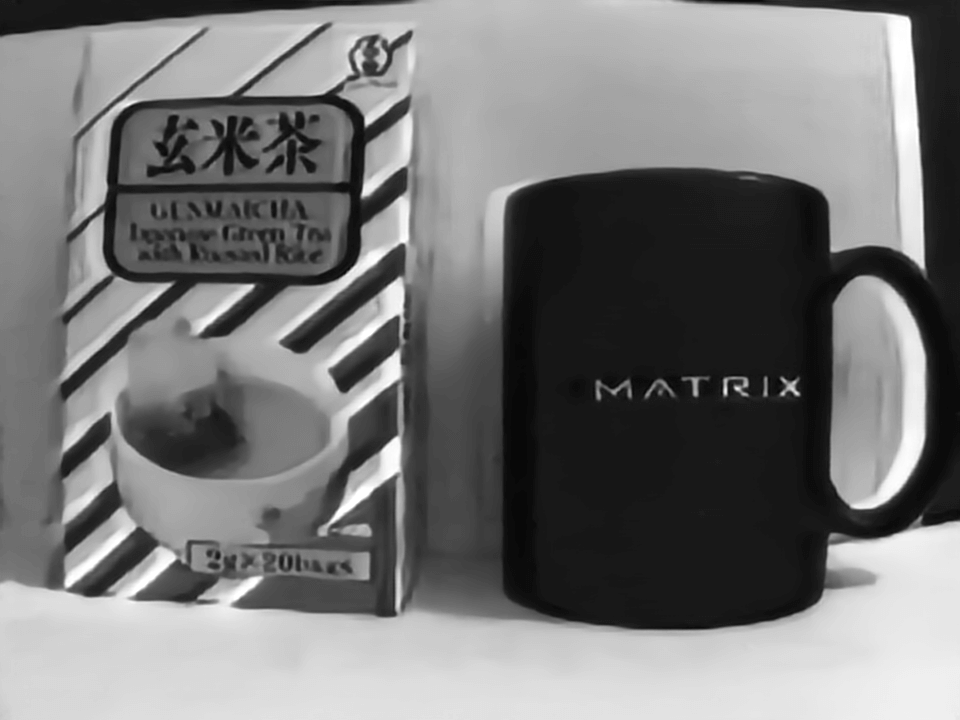}\hspace*{-0.5mm}
			\includegraphics[width=\cimwid\linewidth,trim={220px 230px 500px 100px},clip]{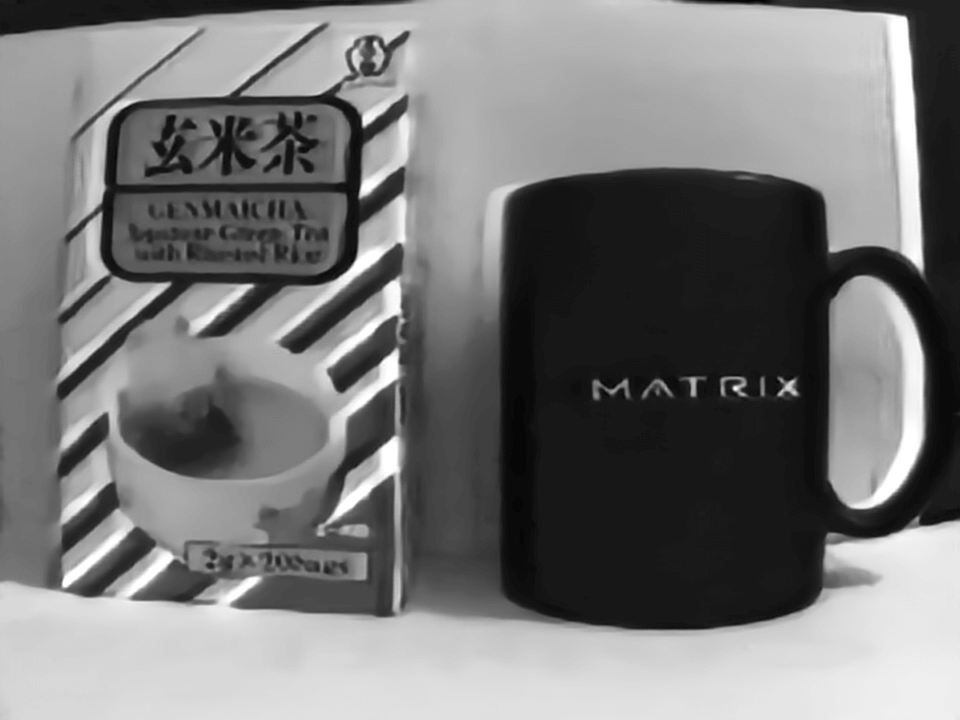}\hspace*{-0.5mm}
			\includegraphics[width=\cimwid\linewidth,trim={220px 230px 500px 100px},clip]{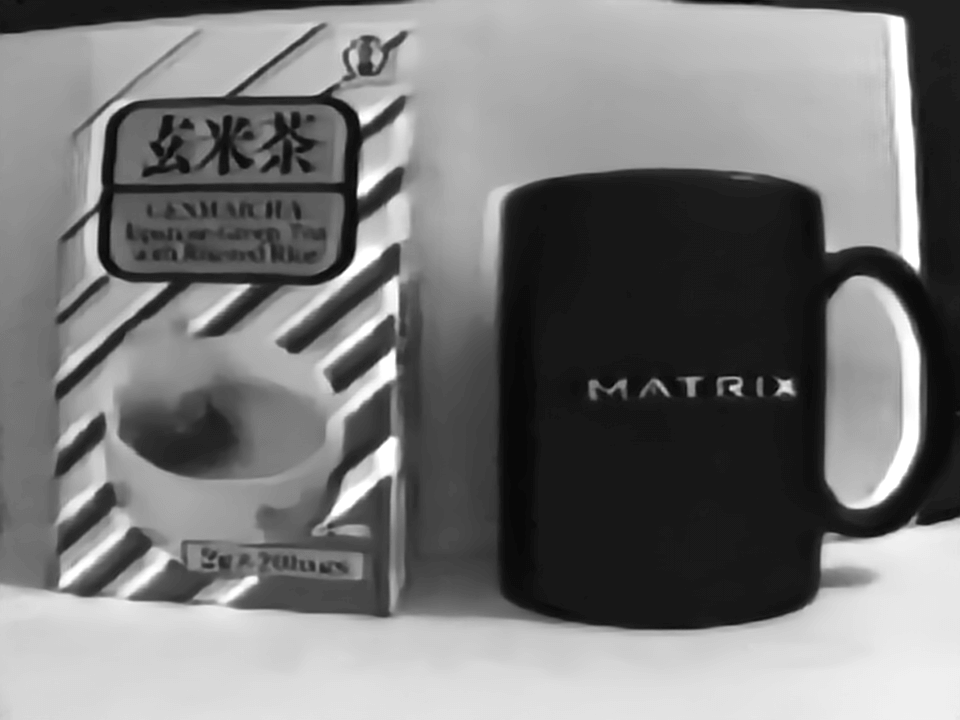}\hspace*{-0.5mm}
			\includegraphics[width=\cimwid\linewidth,trim={220px 230px 500px 100px},clip]{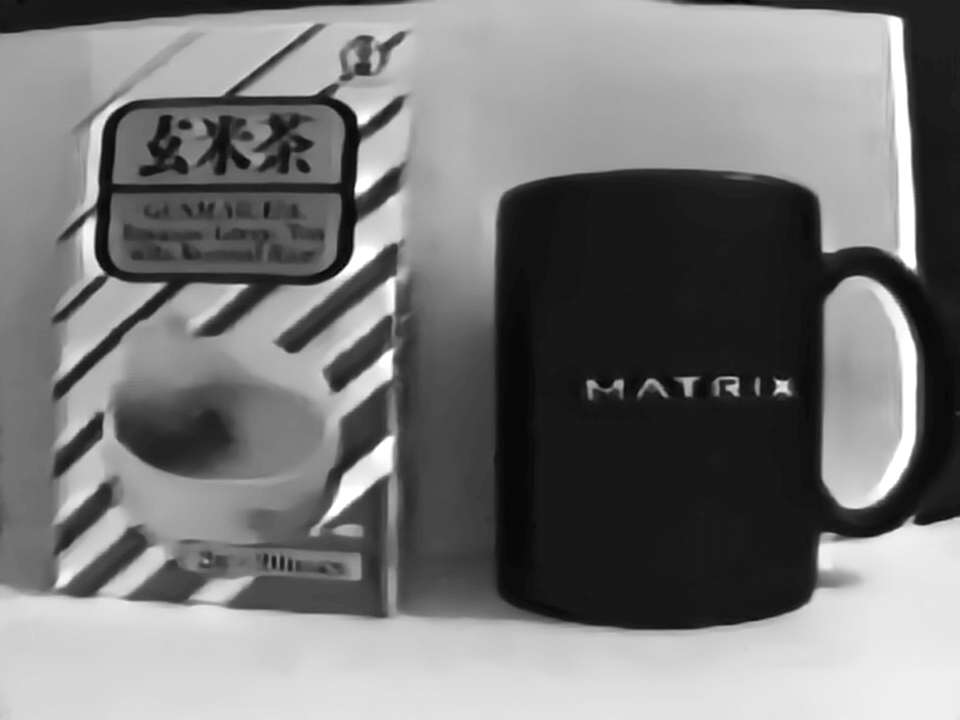}\hspace*{-0.5mm}
			\includegraphics[width=\cimwid\linewidth,trim={220px 230px 500px 100px},clip]{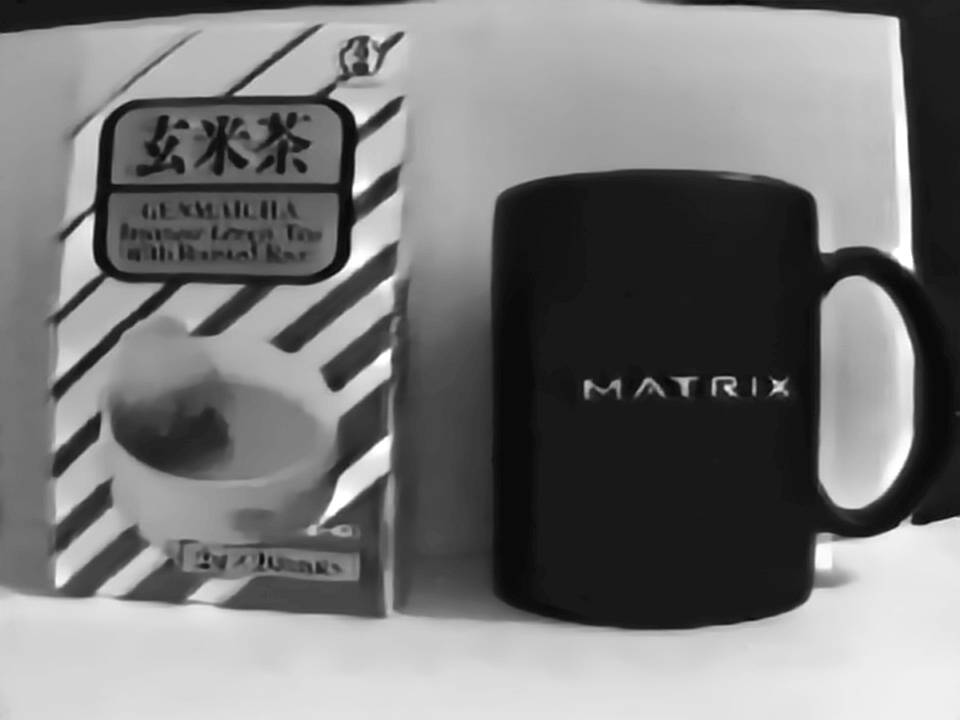}\hspace*{-0.5mm}
			\includegraphics[width=\cimwid\linewidth,trim={220px 230px 500px 100px},clip]{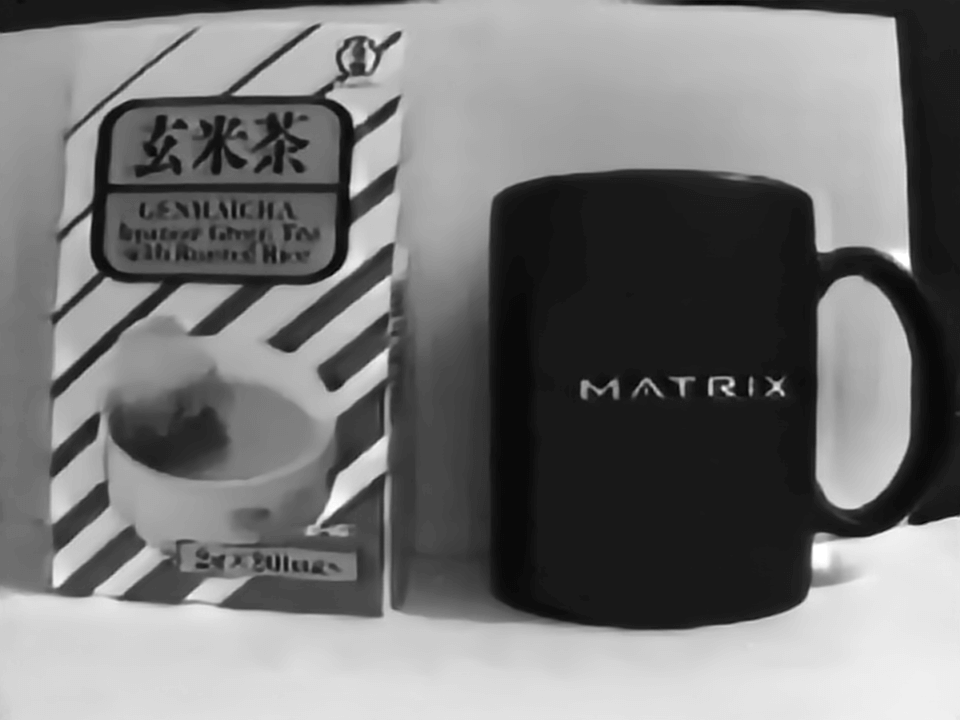}\hspace*{-0.5mm}
			\includegraphics[width=\cimwid\linewidth,trim={220px 230px 500px 100px},clip]{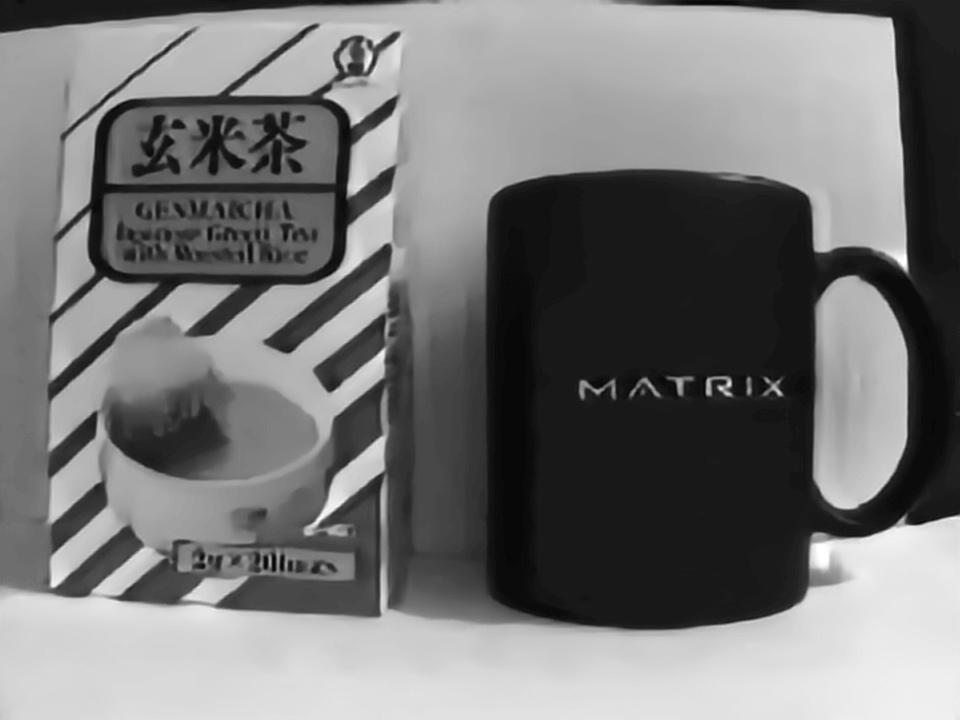}\hspace*{-0.5mm}
			\includegraphics[width=\cimwid\linewidth,trim={220px 230px 500px 100px},clip]{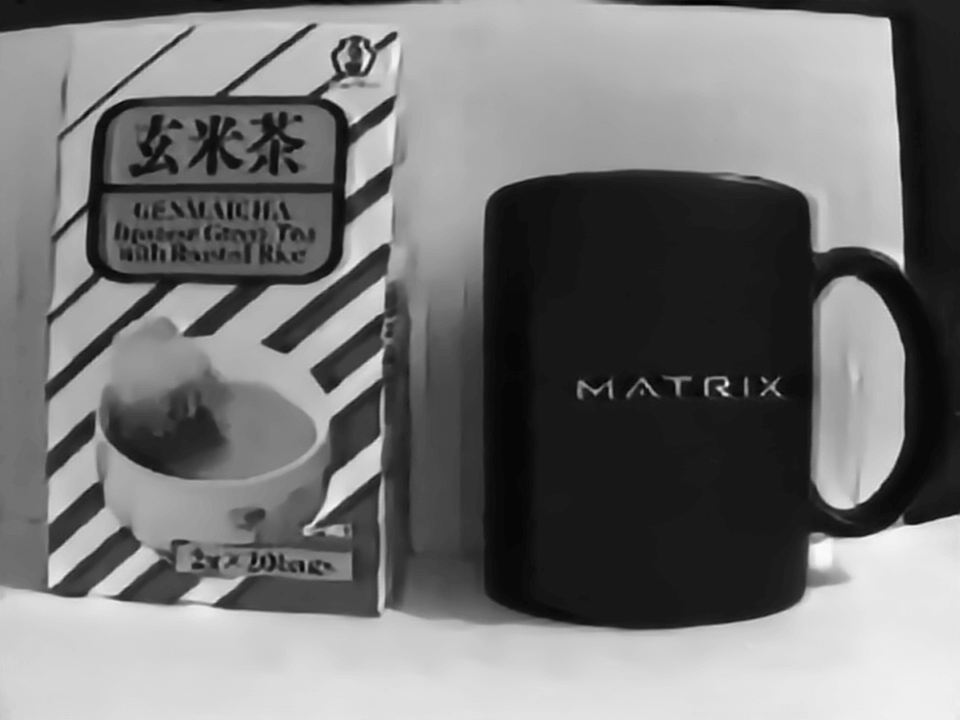}\hspace*{-0.5mm}
			\includegraphics[width=\cimwid\linewidth,trim={220px 230px 500px 100px},clip]{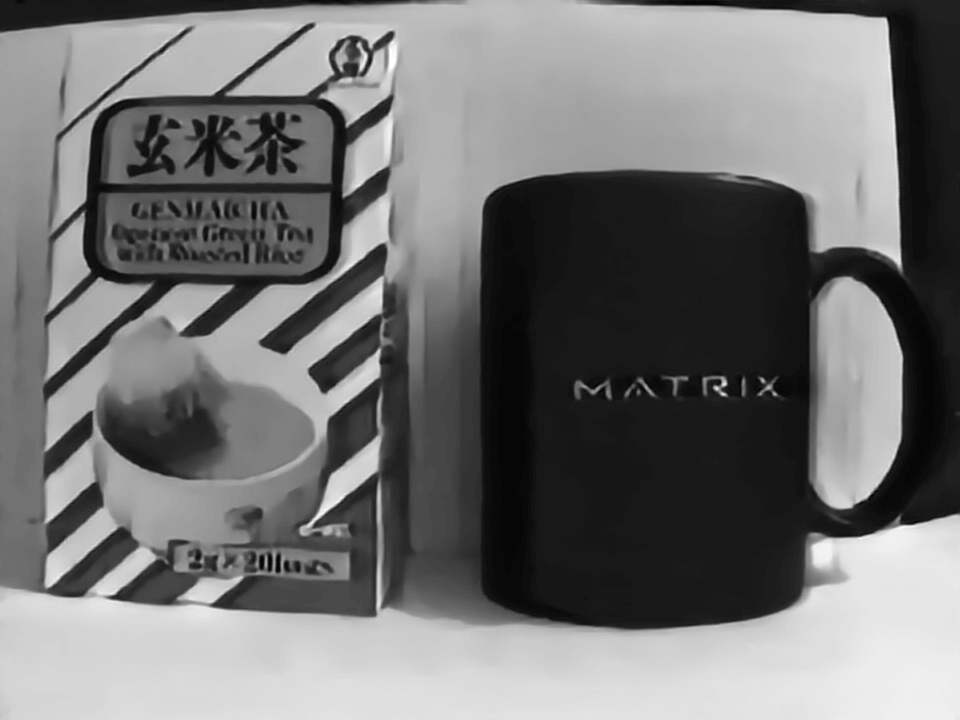}\hspace*{-0.5mm}
			\includegraphics[width=\cimwid\linewidth,trim={220px 230px 500px 100px},clip]{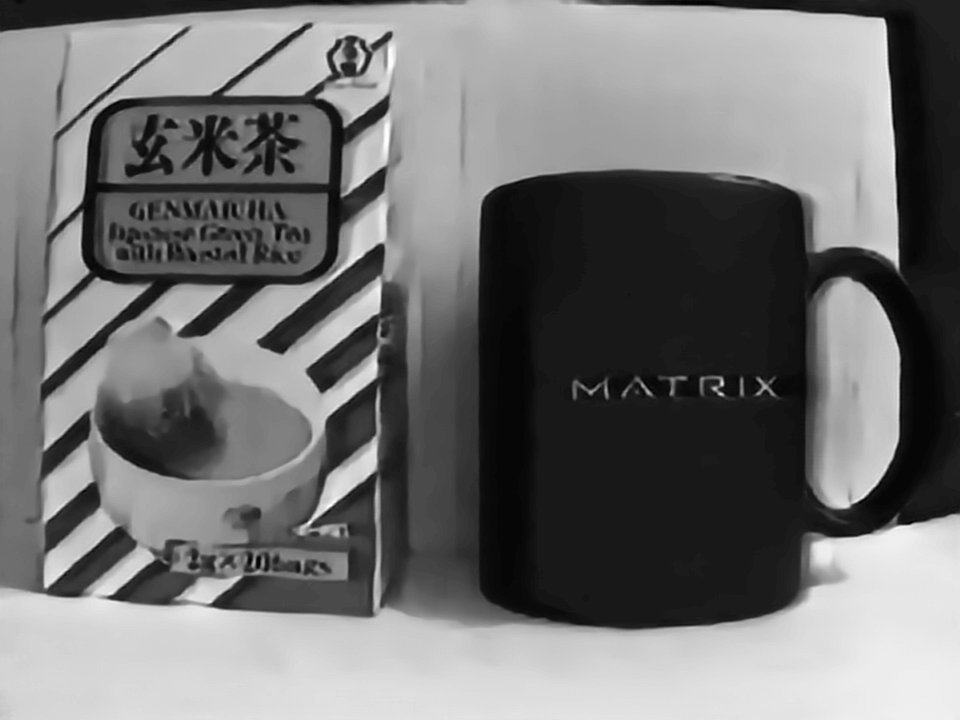}\hspace*{-0.5mm}
			\includegraphics[width=\cimwid\linewidth,trim={220px 230px 500px 100px},clip]{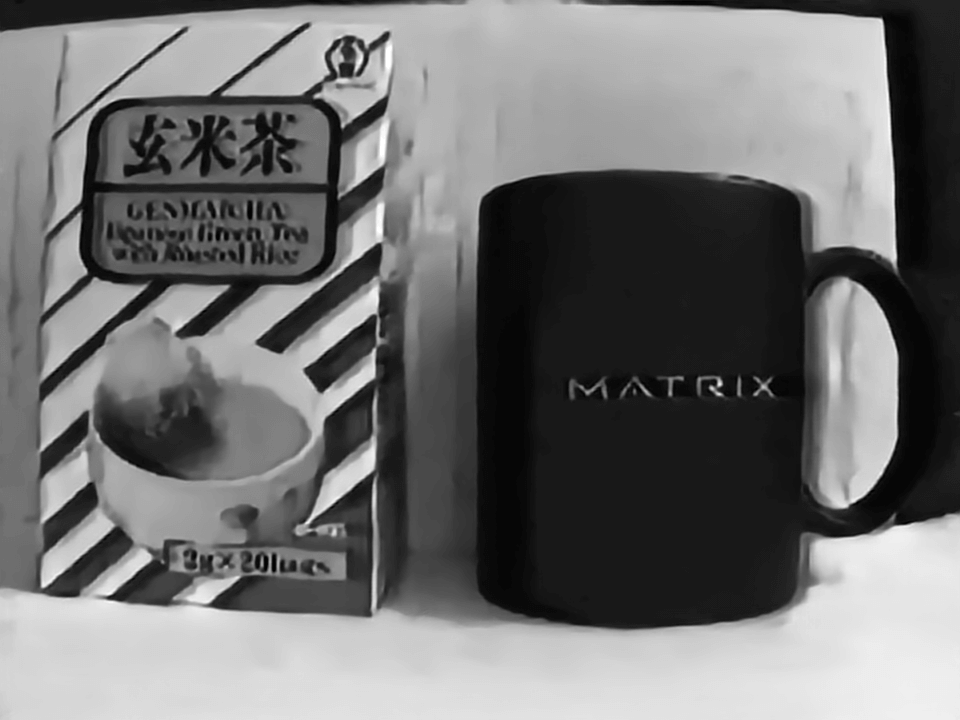}
    \vspace{.1em}\\
	% }%
	\caption{Qualitative results of single frame reconstructions (1st row) and their corresponding sequence reconstructions (2nd-6th rows) on the RWS dataset, where our proposed eSL-Net (5th row) and eSL-Net++ (6th row) are compared to EDI+RCAN (2nd row), LEDVDI+RCAN (3rd row) and RED-Net+RCAN (4th row).}
	\label{mul_image}
\end{figure*}

\begin{table*}[t]
\centering
\caption{\colored{Quantitative comparison of our proposed eSL-Net and eSL-Net++ with the state-of-the-art SR methods with inputs of pure images, pure events, and fusion of images and events. Evaluations are conducted for single frame and sequence reconstructions, respectively.}}
\resizebox{\linewidth}{!}{
\begin{tabular}{ccclcclcclcclccc}
\hline
\multirow{3}{*}{Method} & \multicolumn{2}{c}{\multirow{2}{*}{Inputs}}                                &  & \multicolumn{5}{c}{Single Frame Reconstruction}                                                                  & \multicolumn{1}{c}{} & \multicolumn{5}{c}{Sequence Reconstruction}                                                                      & \multirow{3}{*}{Params} \\ \cline{5-9} \cline{11-15}
                        & \multicolumn{2}{c}{}                                                       &  & \multicolumn{2}{c}{Synthetic events}          & \multicolumn{1}{c}{} & \multicolumn{2}{c}{Real events}               & \multicolumn{1}{c}{} & \multicolumn{2}{c}{Synthetic events}          & \multicolumn{1}{c}{} & \multicolumn{2}{c}{Real events}               &                                                                                        \\ \cline{2-3} \cline{5-6} \cline{8-9} \cline{11-12} \cline{14-15}
                        & \multicolumn{1}{c}{Events} & \multicolumn{1}{c}{Image}                     &  & PSNR                 & SSIM                 & \multicolumn{1}{c}{} & PSNR                 & SSIM                 & \multicolumn{1}{c}{} & PSNR                 & SSIM                 & \multicolumn{1}{c}{} & PSNR                 & SSIM                 &                                                                                        \\ \hline
%  \multicolumn{1}{l}{}    &                            &                                               &  & \multicolumn{1}{l}{} & \multicolumn{1}{l}{} &                      & \multicolumn{1}{l}{} & \multicolumn{1}{l}{} &                      & \multicolumn{1}{l}{} & \multicolumn{1}{l}{} &                      & \multicolumn{1}{l}{} & \multicolumn{1}{l}{} & \multicolumn{1}{l}{} 
 GFN  &  \XSolidBrush  & \Checkmark &  & 22.43  & 0.5295               &                      & 21.10      & 0.7507               &                      & /                    & /                    &                      & /                    & /                    & 12.2M\\
%\rowcolor{blue!30} 
DASR      &             \XSolidBrush                & \Checkmark                     &              & 20.87            & 0.5124            &  & 17.65          & 0.7128     &                      & /                    & /                    &                      & /                    & /                    & 5.97M    \\
%\rowcolor{blue!30} 
FKP              &             \XSolidBrush                & \Checkmark                     &          & 19.40            & 0.4717            &  & 17.25          & 0.6646       &                      & /                    & /                    &                      & /                    & /                    & \textbf{0.59M}     \\
%\rowcolor{blue!30} 
MANet       &             \XSolidBrush                & \Checkmark                     &           & 20.01            & 0.4993            &  & 17.48          & 0.6786    &                      & /                    & /                    &                      & /                    & /                    & 9.89M      \\ 
 SRN+$\text{RCAN}_{4\times}$          &          \XSolidBrush                   & \Checkmark                     &  & 22.86                & 0.5601               &                      & 20.91                    & 0.7562                    &                      & /                    & /                    &                      & /                    & /                    & \colored{34.9M}%19.4M
 \\
 CDVD+$\text{RCAN}_{4\times}$          &             \XSolidBrush                & \Checkmark                     &  & 22.20                & 0.5967               &                      & 20.95                    & 0.7577                    &                      & /                    & /                    &                      & /                    & /                    & \colored{47.3M} %31.8M
 \\ 
\hline
 E2VID+$\text{RCAN}_{4\times}$        & \Checkmark  &        \XSolidBrush                                       &  & 12.15                    & 0.3385                    &                      & 9.88                    &0.5392                    &                      & 12.14                & 0.3369               &                      & 9.86                    & 0.5382                    & \colored{41.8M} %26.3M
 \\
 E2SRI+$\text{RCAN}_{2\times}$      & \Checkmark  &              \XSolidBrush                                  &  & 11.90                    & 0.4183                    &                      & 11.28                    & 0.5832                    &                      & 11.85                & 0.4162               &                      & 11.34                    & 0.5853                    & \colored{40.9M} %25.5M
 \\ \hline
 EDI+$\text{RCAN}_{4\times}$           & \Checkmark  & \Checkmark                     &  & 22.48                & 0.6196               &                      & 21.34                    & 0.7689                    &                      & 21.77                & 0.6021               &                      & 19.79                    & 0.7473                    & \colored{31.1M} %15.6M
 \\
 LEDVDI+$\text{RCAN}_{4\times}$        & \Checkmark  & \Checkmark                     &  &  23.75              &  0.5676              &                      & 21.81                    & 0.7942                    &                      &       23.30               &    0.5670                &                      & 21.34                    & \textbf{0.7914}                    &  \colored{36.1M} %20.6M
 \\
 RED-Net+$\text{RCAN}_{4\times}$        & \Checkmark  & \Checkmark                     &  &  23.54              &  0.5608              &                      & 22.78                    & 0.7644                    &                      &       22.82               &    0.5443                &                      & 21.64                    & 0.7475                    &  \colored{25.3M}%9.76M
 \\ 
%\rowcolor{blue!30} 
 EFNet+$\text{DASR}_{4\times}$  & \Checkmark  & \Checkmark                     & & 23.85            & 0.5556            &  & 22.73          & 0.7996          &  & /            & /            &  & /          & /   &  \colored{14.4M}       \\
%\rowcolor{blue!30} 
LEDVDI+RealBasicVSR  & \Checkmark  & \Checkmark                     &   & 22.76            & 0.6286            &  & 18.11          & 0.6995          &  & 22.23            & 0.6150            &  & 16.97          & 0.6841      &  \colored{26.9M}    \\
%\rowcolor{blue!30} 
RED-Net+RealBasicVSR  & \Checkmark  & \Checkmark                     & & 23.41            & 0.5999            &  & 19.80          & 0.7273          &  & 20.49            & 0.5262            &  & 21.19          & 0.7450   &  \colored{16.0M}       \\
 \hline
 eSL-Net (Ours)             & \Checkmark  & \Checkmark                     &  & \underline{25.32}                & \underline{0.6705}               &                      & \underline{23.91}                    & \underline{0.8075}                    &                      & \underline{23.80}                & \underline{0.6455}               &                      & \underline{22.16}                    & {0.7790}                    & \underline{1.32M}                                                                                  \\
 eSL-Net++ (Ours)              & \Checkmark  & \Checkmark                     &  & \textbf{25.73}                & \textbf{0.6824}               &                      & \textbf{23.99}                   & \textbf{0.8087}                    &                      & \textbf{24.69}                & \textbf{0.6602}               &                      & \textbf{22.99}                   & \underline{0.7913}                   & {1.41M}                                                                                  \\ \hline
\label{table_dn} 
\end{tabular}
}
\end{table*}

\section{Experiments}\label{exper}
In this section, we evaluate the proposed E-SRB approaches, \ie, eSL-Nets (eSL-Net~\cite{wang2020event} and its extension eSL-Net++) and compare them with existing state-of-the-art SRB methods. The datasets, codes, and more results are available at \url{https://github.com/ShinyWang33/eSL-Net-Plusplus}. The performances of our methods are quantitatively evaluated by PSNR and SSIM~\cite{wang2003multiscale}. In addition, the qualitative evaluation is also given by visualizing reconstructed HR images.
\subsection{Datasets and Implementation Details}\label{dataset}
% Both synthetic and real-world datasets are built for training and evaluation. 
\colored{In this paper, both synthetic and real-world datasets are built.}
To train the proposed eSL-Net, we first synthesize the GoPro dataset consisting of LR blurry noisy images with labeled HR ground-truth images as well as the event streams, which are simulated from high frame-rate video sequences via ESIM~\cite{rebecq2018esim}. Meanwhile, the real-world dataset is also provided to validate the effectiveness of our proposed network in real-world scenarios.
\par
\noindent\textbf{{Dataset with Synthetic Events.}} 
We build the synthetic GoPro dataset containing {\it HR clear images}, {\it LR blurry images}, and {\it Event streams}.
% (1) \textbf{HR clear images dataset} consists of $25650$ HR sharp clear frames with various contents, locations, natural and handmade objects, from $270$ video, each of which contains 95 images. It is used as ground truth in training and testing of network. (2) \textbf{LR clear images dataset} consists of $25650$ LR sharp clear frames from $270$ video. It is used as ground truth in training and testing of network without SR. (3) \textbf{LR blurry images dataset} consists of $25650$ LR blurry noisy frames from $270$ video correspondingly, which simulates the APS frames of the event camera with motion blur and noises. (4) \textbf{Event sequences dataset} consists of event sequences with noises and without noises corresponding to LR blurry frames. 

{\it HR clear images.} We choose the continuous sharp clear images with resolution of $1280 \times 720$ from the GoPro dataset~\cite{nah2019ntire} as our ground truth. It consists of $25,650$ HR sharp clear frames with various natural and manmade objects captured at different locations. 

% \noindent{\bf LR clear images.} LR sharp clear images with resolution of $320 \times 180$ are obtained by down-sampling HR clear images with bicubic interpolation. Finally, it consists of $25650$ LR sharp clear frames from $270$ video. 

{\it LR blurry images.} Similar to~\cite{Nah2017Deep}, we first increase the frame-rate of LR sharp clear images to $ 960 $ fps using the method proposed in~\cite{Niklaus2017VideoFI}, and then generate motion blurred images by averaging $ 17 $ consecutive frames. %The LR blurry images are further contaminated by white noises to imitate real scenarios.
% with standard deviation $\sigma=4$ ($\sigma=4$ is the approximated mean of the standard deviations of many smooth patches in APS frames in the real dataset). Finally, it consists of $25650$ LR blurry noisy frames from $270$ video correspondingly, which simulates the APS frames of the event camera with motion blur and noises.
    
{\it Event streams.} For each blurry image frame, the ESIM~\cite{rebecq2018esim} is employed to synthesize concurrent events from the interpolated high frame-rate LR clear images. 
% We further add $30 \%$ noisy events with the uniform random distribution to the sequence to imitate real scenarios. 
{\colored{We also add noise to the synthesized events to imitate real scenarios. Specifically, we first calculate the number of the original events (denoted by $N_{o}$) in the exposure time of the corresponding blurry frames and then generate $N_{n}$ noise events with $N_{n}\triangleq \omega N_{o}$ with the event noise ratio $\omega$ ($\omega=0.3$ in our GoPro dataset). Afterward, we assign the noise events with pixel coordinates and polarities randomly sampled from the uniform distribution. Finally, we apply the rounding operation to the sampled coordinates and polarities to keep them in the integer format and add the noise events to the original event streams.}}

% In our experiments, 

% It consists of event sequences with noises and without noises corresponding to LR blurry frames. ($30 \%$ is artificially calculated approximate ratio of noise events to effective events in simple real scenes)

According to the partitions of the GoPro dataset~\cite{nah2019ntire}, images and event streams in the synthetic dataset from $ 240 $ video sequences are used for training and the rest from $30$ video sequences are for evaluation. 

% Although the eSL-Net is trained on the synthetic data, it is able to be generalized to real-world scenes~\cite{stoffregen2020reducing,pan2019bringing}.
\noindent\textbf{{Dataset with Real-world Events.}}  To validate the effectiveness of the proposed eSL-Net in real-world scenarios, the performance is further evaluated on two datasets with real-world events, \ie, HQF from~\cite{stoffregen2020reducing} and RWS partially built by ourselves.
% where the HQF is collected by the DAVIS240 camera with real-world events and well-posed ground-truth APS frames while the RWS is collected by the DAVIS346 camera with real-world events and real-world blurry images.

{\it HQF.} The HQF dataset~\cite{stoffregen2020reducing} contains real-world events and sharp LR clear ground-truth frames that are well-exposed to avoid motion blurs, while the motion blurs can be synthesized by averaging over $49$ consecutive sharp clear image frames. Finally, the HQF dataset contains real-world events and synthetic blurry images as well as the ground-truth frames, enabling quantitative evaluation with real-world events. Note that HQF only provides LR clear images and thus we first up-sample the LR clear images by RCAN~\cite{zhang2018image} and utilize the generated HR images as the ground truth.

{\it RWS.} The real-world scenes (RWS) dataset is built mainly to validate the effectiveness of our proposed method on different event cameras (a DAVIS346 camera and a DAVIS240 camera~\cite{pan2019bringing}) over different scenes. Since RWS only contains real-world events and real-world blurry APS frames while the HR clear images can not be provided as the ground truth, we only evaluate SRB methods qualitatively on the RWS dataset. 

% Furthermore, the RWS contains scenes with different types of motions including the foreground object moving and the camera moving, and different light conditions.

%\textcolor{red}{Wangbishan: add PSNR and SSIM here, and explain how to evaluate sequence. How to evaluate SR, deblur, denoise ....}

\noindent\textbf{Implementation Details.} We implement the proposed eSL-Nets using PyTorch on NVIDIA Titan-RTX 3090 GPUs. The step size $\eta$ in \eqref{ev-ista} is set to $0.01$ for stabilization of eSL-Nets. \colored{The parameter $\lambda_1$ in \eqref{ev-lasso} is a learnable parameter in our model, and $\lambda_2$ and $\lambda_3$ serve as the offsets in \eqref{ev-ista}, which are implicitly embedded in the convolutional layers. Thus, the parameters $\lambda_1$, $\lambda_2$, and $\lambda_3$ can be automatically learned during network training.}
Both eSL-Net~\cite{wang2020event} and eSL-Net++ are trained over the synthetic GoPro dataset. Adam
optimizer is used and the maximum epoch of training iterations
is set to $50$. The learning rate starts at $8 \times 10^{-4}$ and then decays by $50\%$ every $10$ epochs. To determine the number of recursions $n$, we train $5$ different models respectively with $5$, $10$, $15$, $20$, and $25$ recursion blocks and their SRB performance in terms of PSNR is respectively $25.13$ dB, $25.63$ dB, $25.73$ dB, $25.74$ dB, and $25.78$ dB. Apparently, increasing the recursion number $n$ can improve reconstruction performance, but at the sacrifice of the computation load. Thus, we choose $n=15$ to balance the computational burden and reconstruction performance.

\subsection{Results of Single Frame SRB}
% SRB is a joint task where both image SR and motion deblurring should be tackled. Therefore, we compare our proposed eSL-Net~\cite{wang2020event} and eSL-Net++ with the state-of-the-art methods including the end-to-end approaches, \ie, GFN~\cite{zhang2018e}, and approaches of cascading motion deblurring and image SR. State-of-the-art motion deblurring methods are employed including image-based methods, \ie, SRN~\cite{tao2018scale} and CDVD~\cite{pan2020cascaded}, event-based methods, \ie, E2VID~\cite{rebecq2019} and E2SRI~\cite{mostafavi2020e2sri}, and event-image-based methods, \ie, EDI~\cite{pan2019bringing}, RED-Net~\cite{xu2021motion} and LEDVDI~\cite{lin2020}. On the other hand, the task of the image SR is fulfilled by RCAN~\cite{zhang2018image}, the same method for generating the ground-truth HR images of the HQF dataset. Then, the performance of single frame SRB is evaluated quantitatively and qualitatively in the following of this subsection.

% New version
\colored{SRB is a joint task where both image SR and motion deblurring should be tackled. Thus we compare our eSL-Net and eSL-Net++ to state-of-the-art methods that can achieve SRB including the end-to-end methods, \ie, GFN~\cite{zhang2018e}, DASR~\cite{wang2021unsupervised}, FKP~\cite{liang2021}, and MANet~\cite{liang2021mutual}, and the two-stage Deblur-then-SR methods by cascading the motion deblurring with the image SR. Different motion deblurring methods are compared including the frame-based methods, \ie, SRN~\cite{tao2018scale} and CDVD~\cite{pan2020cascaded}, and the event-based methods, \ie, EDI~\cite{pan2019bringing}, RED-Net~\cite{xu2021motion}, and LEDVDI~\cite{lin2020ledvdi}. The task of the image SR is fulfilled by RCAN~\cite{zhang2018image}. We further make comparisons to the methods of intensity restoration from pure events, \ie, E2VID~\cite{rebecq2019} and E2SRI~\cite{mostafavi2020e2sri}. The performance of single-frame SRB is evaluated quantitatively and qualitatively in the following of this subsection.
}

\noindent{\bf Quantitative results.} The quantitative performances of our proposed eSL-Net and eSL-Net++ are evaluated over the GoPro dataset and the HQF dataset, where the ground-truth clear HR images are available. The PSNRs and SSIMs over different datasets are given in Tab.~\ref{table_dn}. Although our networks are trained on the synthetic GoPro dataset with synthesized motion blurs and events, they still perform well on datasets with real events as shown in Tab.~\ref{table_dn}, which validates the ability of our models to generalize from synthetic to real-world scenes. 

Both eSL-Net and eSL-Net++ outperform the state-of-the-art methods by a large margin. 
\colored{Specifically, the problem of motion blur can be well tackled with events and thus event-based SRB methods (RED-Net+RCAN, LEDVDI+RCAN, eSL-Net, and eSL-Net++) perform better than SRB methods without events, including joint SR-and-deblurring method (GFN), blind SR methods (DASR, FKP, and MANet), and deblurring-then-SR methods (SRN+RCAN and CDVD+RCAN).} 
Even though EDI is a traditional method, cascading EDI with RCAN still achieves comparable results to deep networks including SRN and CDVD. Among event-based SRB methods, our proposed eSL-Net and eSL-Net++ exhibit significant improvements over the two-stage event-based SRB methods cascaded with RCAN, \ie, EDI+RCAN, RED-Net+RCAN, and LEDVDI+RCAN, which validates the effectiveness of our unified framework with motion deblurring and super-resolution. 

\colored{Furthermore, we also compare our eSL-Net and eSL-Net++ with the blind SR method, \ie, DASR \cite{wang2021unsupervised}, with a motion-deblurring pre-processing, where the most recent state-of-the-art method for event-based motion deblurring, \ie, EFNet~\cite{sun2022event}, is applied to reduce the difficulty of SRB.
% , we have conducted one more comparison to the Deblur-then-SR approach cascading the most recent SOTA event-based motion deblurring method, \ie, EFNet~\cite{sun2022event}, with a blind SR method, \ie, 
% DASR \cite{wang2021unsupervised}.
The quantitative results are presented in Tab.~\ref{table_dn}. It is shown that the deblurring pre-processing by EFNet improves the performance of blind SR approaches, but our eSL-Net and eSL-Net++ still perform better than EFNet+DASR.}

On the other hand, eSL-Net++ outperforms its previous version eSL-Net on both datasets with synthetic and real events, which shows the effectiveness of the dual sparse learning scheme for event denoising.

\noindent{\bf Qualitative results.} We further evaluate the performances qualitatively on the GoPro, HQF, and RWS datasets. The results are visualized in Fig.~\ref{syn_sr} for the GoPro and HQF datasets and Fig.~\ref{real_sr_more} for the RWS dataset. 

% Since events only reflect brightness changes and are always contaminated, reconstructing SR images purely from events, \eg, E2VID+RCAN and E2SRI+RCAN, may suffer from inconsistency of brightness, leading to dramatic performance drop. Therefore, we only visualize the SR results via E2SRI+RCAN in Fig.~\ref{syn_sr}. 

Fig.~\ref{syn_sr} illustrates the qualitative results on the GoPro (first two rows) and HQF (second two rows) datasets with synthesized motion blurs. Both eSL-Net and eSL-Net++ can give precise HR reconstructions which exhibit the most similar appearances to the ground-truth HR images. Furthermore, eSL-Net++ still outperforms eSL-Net, producing less noises and artifacts, \eg, halo effects. 

Fig.~\ref{real_sr_more} visualizes the SR results from blurry images of the RWS dataset with real-world motion blurs and events, where the ground-truth HR images are not available. We only present the SR results of GFN, LEDVDI+RCAN, and our proposed eSL-Nets that have comparable quantitative results in Tab.~\ref{table_dn}. Compared to traditional SRB methods, \ie, GFN, E-SRB methods, \ie, LEDVDI+RCAN and eSL-Nets, can largely remove motion blurs benefiting from introducing events, which is consistent to the quantitative results. Furthermore, eSL-Nets produce better visualization results than cascading approaches, \eg, LEDVDI+RCAN, while eSL-Net++ achieves the best visual quality.

\def\imgWidth{0.192\textwidth} %子图大小
\def\sccone{(-1.7,-0.75)} % 文字位置
\def\ssxxsone{(0.15,0.15)} % 绿色小方框位置
\def\ssyysone{(1.7, 0.42)} % 绿色大方框位置

\def\scctwo{(-1.7,-1.1)} % 文字位置
\def\ssxxstwo{(.55,-0.7)} % 绿色小方框位置
\def\ssyystwo{(1.7, 0.75)} % 绿色大方框位置
\def\ssizz{1cm} %框大小
\def\ssmag{3}

\begin{figure*}[!htb] 
\centering
\tikzstyle{img} = [rectangle, minimum width=\imgWidth, draw=white]
    %% first row 
    \begin{tikzpicture}[spy using outlines={green,magnification=\ssmag,size=\ssizz},inner sep=0]
        \node [align=center, img] {\includegraphics[width=\imgWidth]{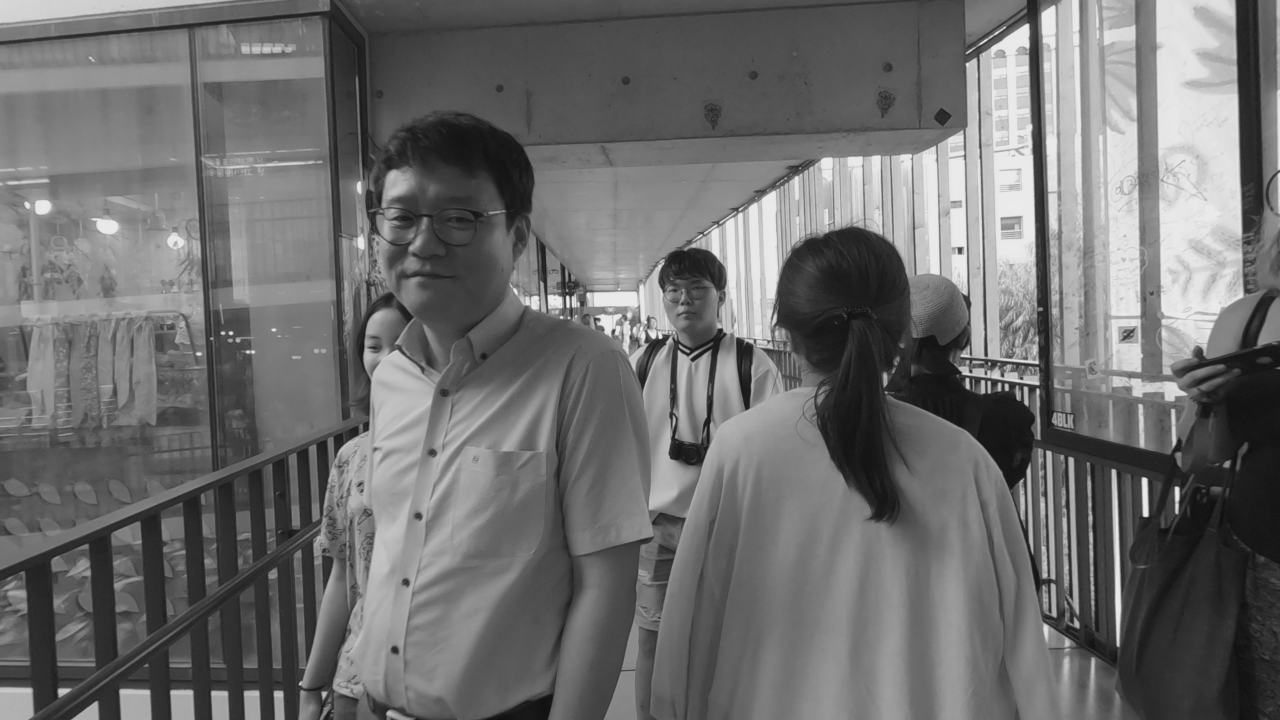}};
        \spy on \ssxxsone in node [left] at \ssyysone;
        \node [anchor=west] at \sccone {\textcolor{white}{\footnotesize \bf Ground Truth}};
    \end{tikzpicture}
    \begin{tikzpicture}[spy using outlines={green,magnification=\ssmag,size=\ssizz},inner sep=0]
        \node [align=center, img] {\includegraphics[width=\imgWidth]{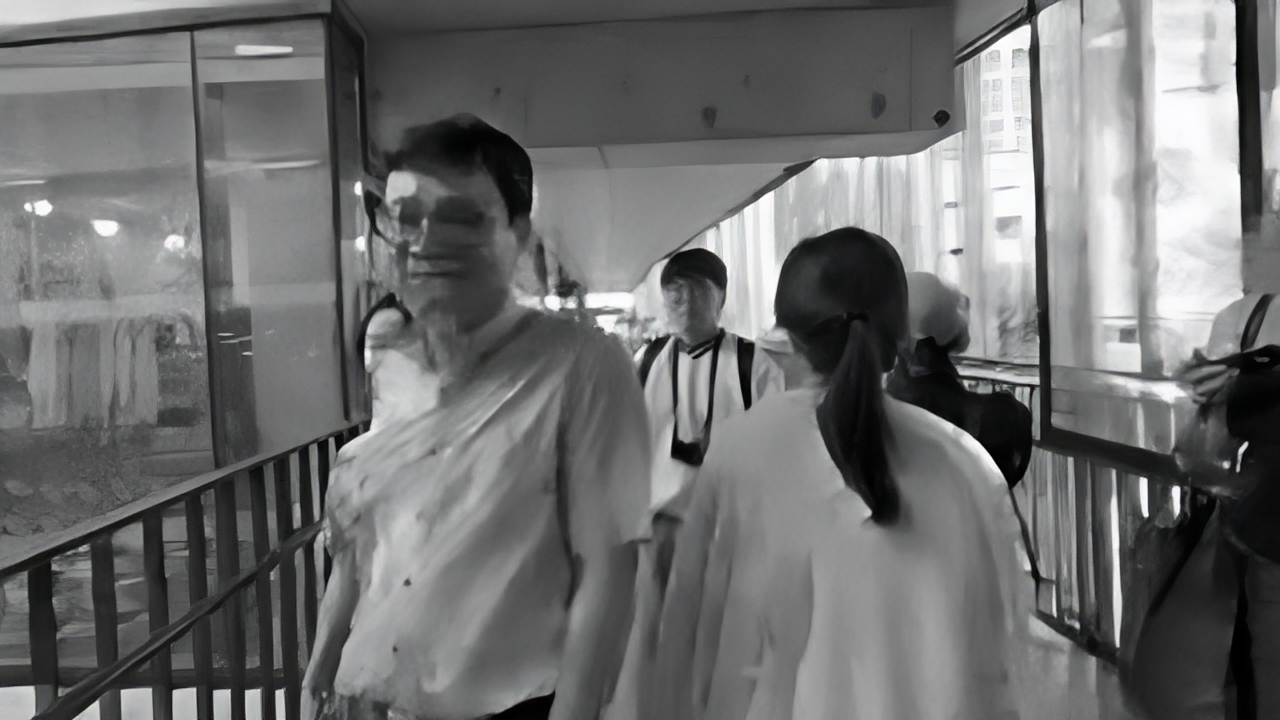}};
        \spy on \ssxxsone in node [left] at \ssyysone;
        \node [anchor=west] at \sccone {\textcolor{white}{\footnotesize \bf LEDVDI+VSR}};
    \end{tikzpicture}
    \begin{tikzpicture}[spy using outlines={green,magnification=\ssmag,size=\ssizz},inner sep=0]
        \node [align=center, img] {\includegraphics[width=\imgWidth]{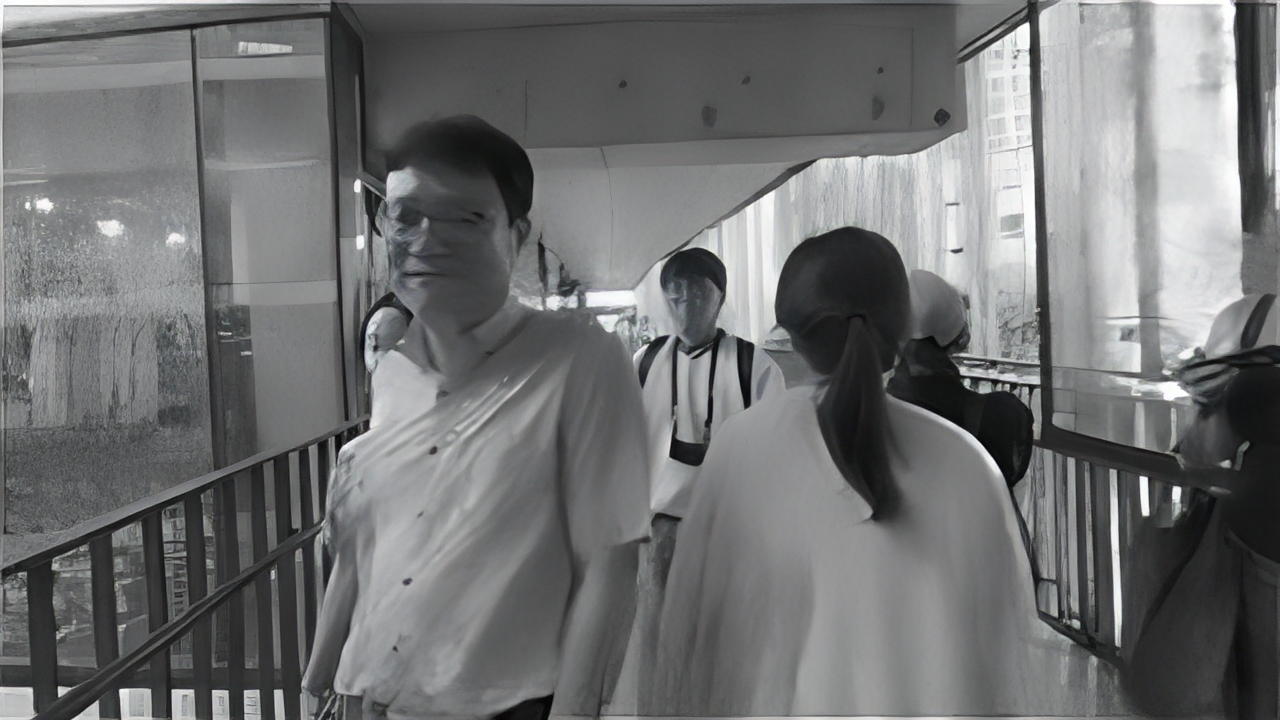}};
        \spy on \ssxxsone in node [left] at \ssyysone;
        \node [anchor=west] at \sccone {\textcolor{white}{\footnotesize \bf RED-Net+VSR}};
    \end{tikzpicture}
    \begin{tikzpicture}[spy using outlines={green,magnification=\ssmag,size=\ssizz},inner sep=0]
        \node [align=center, img] {\includegraphics[width=\imgWidth]{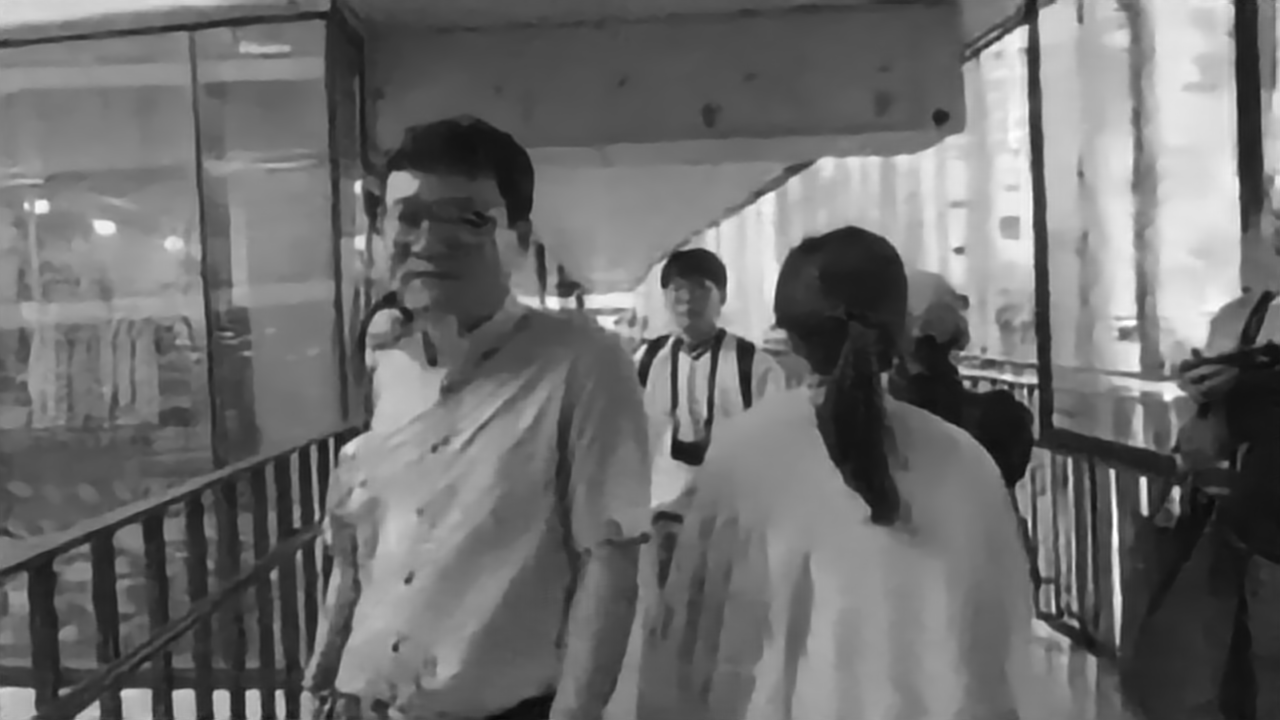}};
        \spy on \ssxxsone in node [left] at \ssyysone;
        \node [anchor=west] at \sccone {\textcolor{white}{\footnotesize \bf eSL-Net}};
    \end{tikzpicture}
    \begin{tikzpicture}[spy using outlines={green,magnification=\ssmag,size=\ssizz},inner sep=0]
        \node [align=center, img] {\includegraphics[width=\imgWidth]{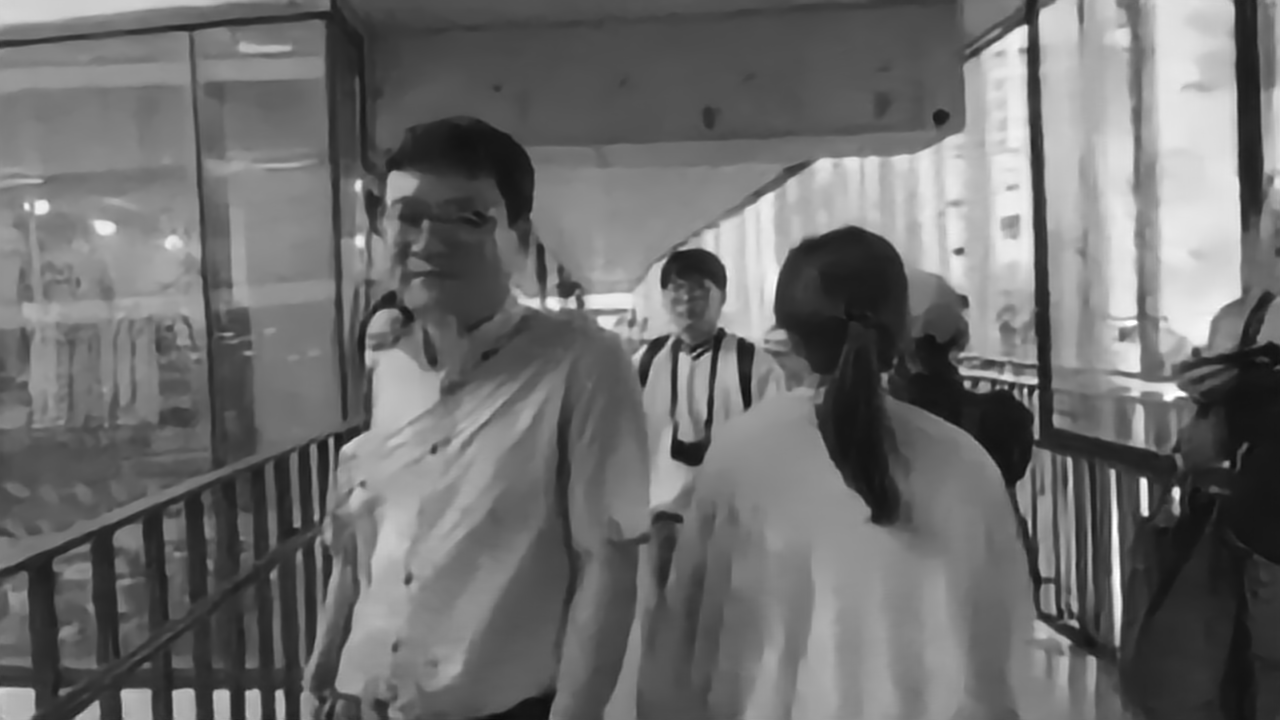}};
        \spy on \ssxxsone in node [left] at \ssyysone;
        \node [anchor=west] at \sccone {\textcolor{white}{\footnotesize \bf eSL-Net++}};
    \end{tikzpicture}
    \\
    %% first row 
    \begin{tikzpicture}[spy using outlines={green,magnification=\ssmag,size=\ssizz},inner sep=0]
        \node [align=center, img] {\includegraphics[width=\imgWidth]{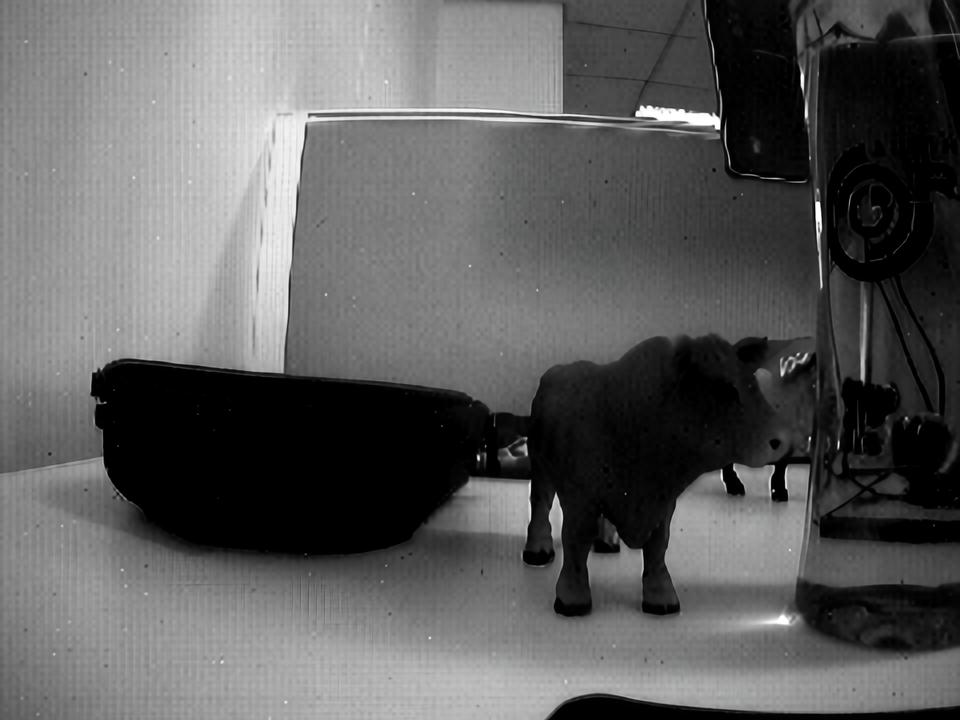}};
        \spy on \ssxxstwo in node [left] at \ssyystwo;
        \node [anchor=west] at \scctwo {\textcolor{white}{\footnotesize \bf Ground Truth}};
    \end{tikzpicture}
    \begin{tikzpicture}[spy using outlines={green,magnification=\ssmag,size=\ssizz},inner sep=0]
        \node [align=center, img] {\includegraphics[width=\imgWidth]{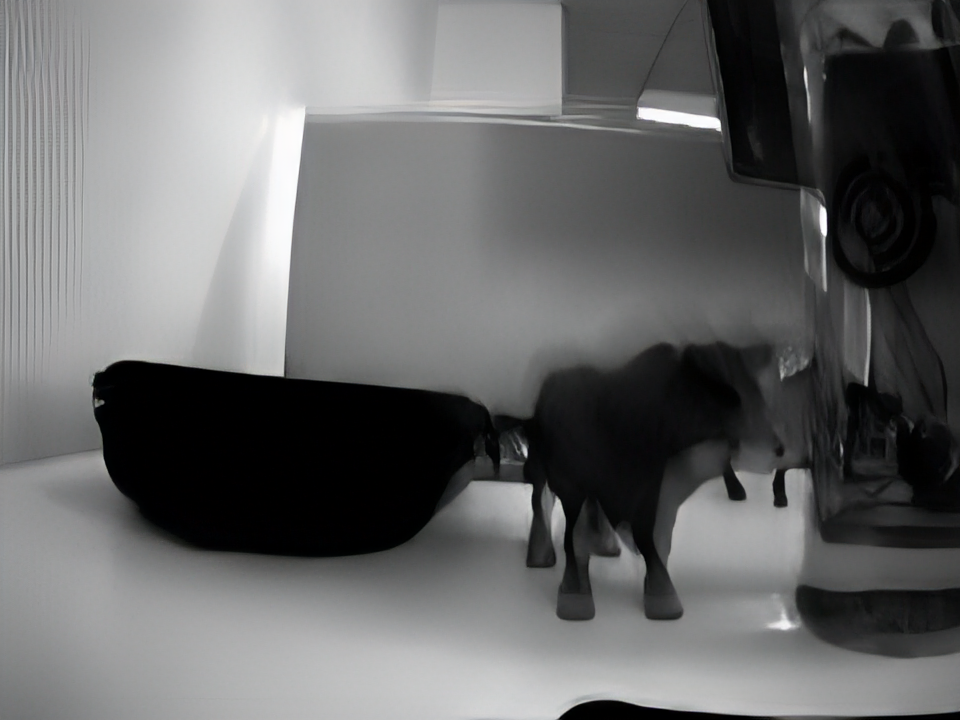}};
        \spy on \ssxxstwo in node [left] at \ssyystwo;
        \node [anchor=west] at \scctwo {\textcolor{white}{\footnotesize \bf LEDVDI+VSR}};
    \end{tikzpicture}
    \begin{tikzpicture}[spy using outlines={green,magnification=\ssmag,size=\ssizz},inner sep=0]
        \node [align=center, img] {\includegraphics[width=\imgWidth]{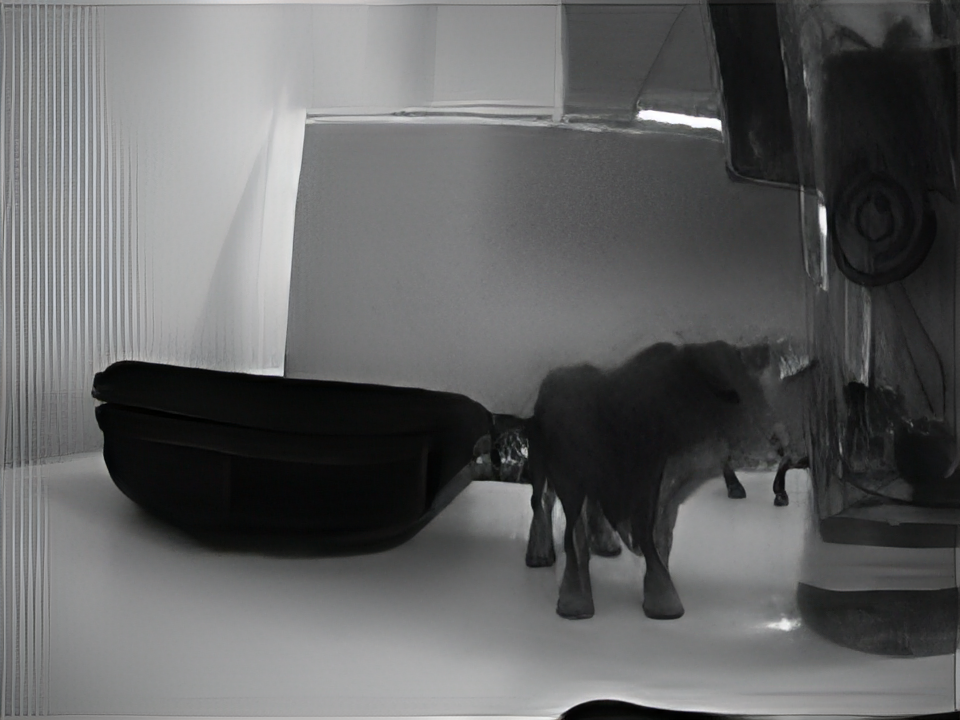}};
        \spy on \ssxxstwo in node [left] at \ssyystwo;
        \node [anchor=west] at \scctwo {\textcolor{white}{\footnotesize \bf RED-Net+VSR}};
    \end{tikzpicture}
    \begin{tikzpicture}[spy using outlines={green,magnification=\ssmag,size=\ssizz},inner sep=0]
        \node [align=center, img] {\includegraphics[width=\imgWidth]{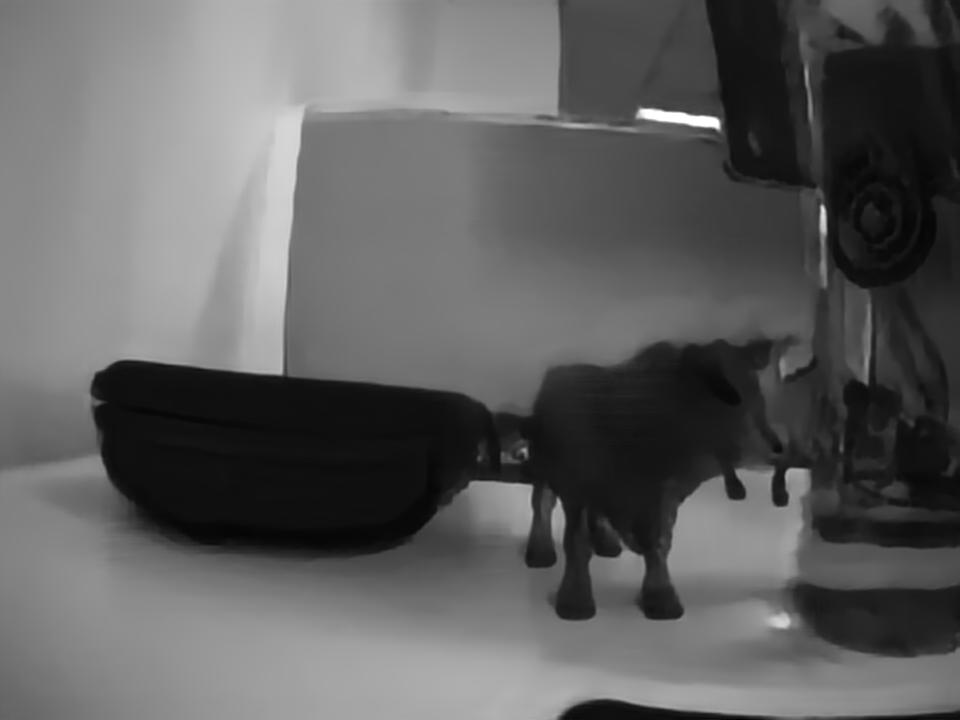}};
        \spy on \ssxxstwo in node [left] at \ssyystwo;
        \node [anchor=west] at \scctwo {\textcolor{white}{\footnotesize \bf eSL-Net}};
    \end{tikzpicture}
    \begin{tikzpicture}[spy using outlines={green,magnification=\ssmag,size=\ssizz},inner sep=0]
        \node [align=center, img] {\includegraphics[width=\imgWidth]{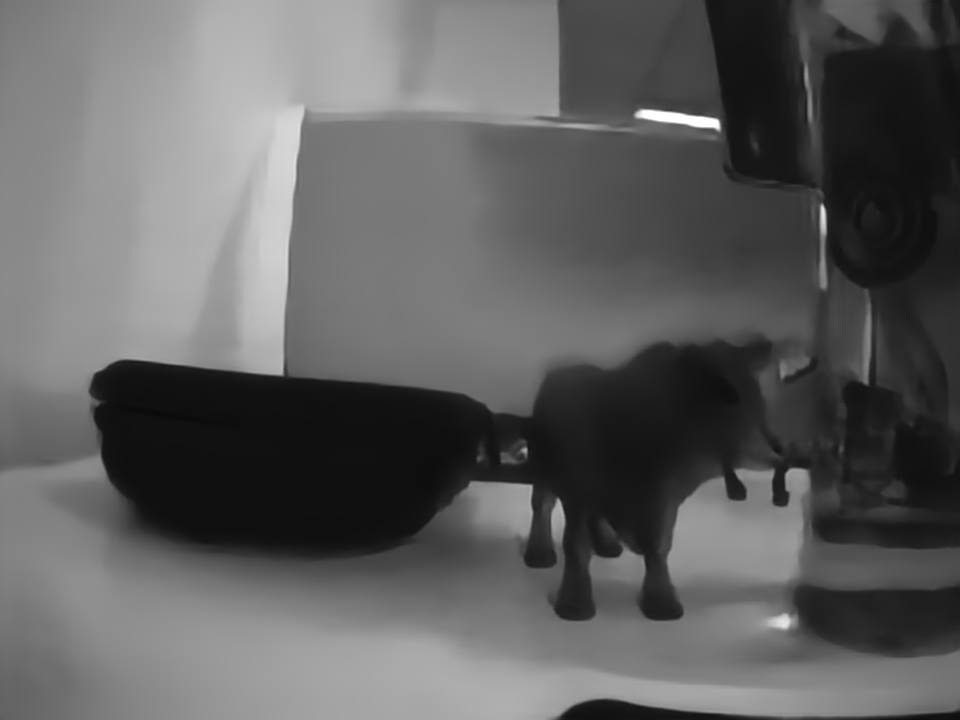}};
        \spy on \ssxxstwo in node [left] at \ssyystwo;
        \node [anchor=west] at \scctwo {\textcolor{white}{\footnotesize \bf eSL-Net++}};
    \end{tikzpicture}
	\caption{\colored{Qualitative comparisons with Deblur-then-VSR methods on the GoPro (top row) and HQF (bottom row) datasets, where the VSR algorithm employed in this experiment is RealBasicVSR \cite{chan2022investigating}. }}
	\label{fig:vsr}
\end{figure*}
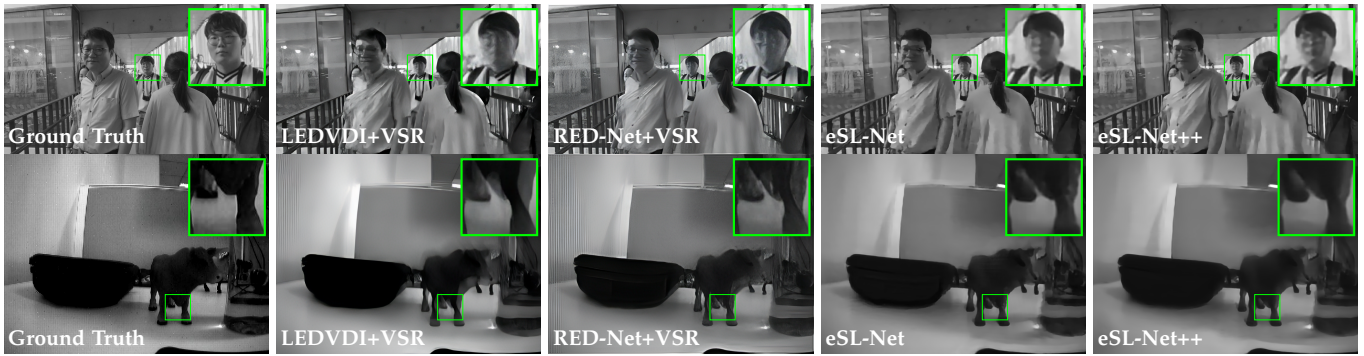

\subsection{Results of Sequence Frame SRB}
The task of sequence frame SRB is more challenging than single frame SRB. To the best of our knowledge, existing approaches for SRB can only tackle single frame SRB, \eg, GFN, DASR, FKP, and MANet. Thus, we only compare our proposed eSL-Nets with the event-based SRB methods, \ie, EDI+RCAN, LEDVDI+RCAN, and RED-Net+RCAN. Quantitative and qualitative results are respectively given in Tab.~\ref{table_dn} and Fig.~\ref{mul_image}. 

Regarding the quantitative results, we evaluate the performance of sequence frame SRB methods with respect to three reconstructed latent sharp HR images from a single blurry image, respectively corresponding to the start, middle, and end of the exposure time. Accordingly, EDI and LEDVDI are utilized to reconstruct latent HR images of the same timestamps. For the datasets with synthetic or real events, our eSL-Nets are with significantly better performances compared to the other E-SRB methods in terms of both PSNR and SSIM.

% Theoretically, our method can generate sequence frames with frame-rate as high as the DVS's eps according to \mysecref{ev-represent}.

% To verify the ability of sequence frame reconstruction, we show the quantitative comparisons tabulated in \mytabref{table_dn} and \mytabref{table_sr}. Three images at the start, middle and end of the exposure time for a input blur image are reconstructed on evaluation of sequence frame reconstruction. No matter in the synthetic testing dataset with synthetic events and HQF dataset with real events, or in the case of Recon-Sequence and Recon-SR-Sequence, our method is significantly higher than other methods in terms of metrics, \ie, PSNR and SSIM. In addition, eSL-Net-M++ trained for sequence reconstruction is better than eSL-Net++, and eSL-Net++ is superior to the remaining methods, which verifies effectiveness of the event representation proposed for sequence reconstruction.

For qualitative comparisons on sequence frame SRB, we show the sequence frame reconstructions from a single blurry image of the RWS dataset by four methods, \ie, EDI+RCAN, LEDVDI+RCAN, eSL-Net and eSL-Net++, as shown in \myfigref{mul_image}. As EDI and eSL-Nets can output arbitrary number of frames, while LEDVDI only outputs 6 frames, we reconstruct 13 frames for EDI and eSL-Nets to facilitate the frame alignment. Obviously, the reconstructions of EDI+RCAN still suffer from blurry effects and noises as illustrated in the 2nd row of \myfigref{mul_image}. LEDVDI+RCAN also produces noisy outputs as illustrated in the third row of \myfigref{mul_image} and reconstruction of higher frame-rate HR videos requires additional training phase. Our eSL-Net is likely to generate halo artifacts around black edges and the reconstructed images are still noisy as illustrated in the 4th row of \myfigref{mul_image}. Benefiting from the dual sparse learning module and the rigorous event shuffle-and-merge module, eSL-Net++ can alleviate the effects of event noises and halo artifacts, leading to the best visualization performance on sequence reconstruction.

\subsection{\colored{Comparisons to Video SR}}
\colored{The video SR (VSR) can boost SR performance by leveraging the temporal inter-frame correlations~\cite{chan2022investigating}. Thus we also make comparisons of our eSL-Net and eSL-Net++ to Deblur-then-VSR approaches, where we first exploit two event-based motion deblurring methods, \ie, LEDVDI and RED-Net, to reconstruct sharp LR video sequences from a single blurry image, and then apply the VSR approach, \ie, RealBasicVSR \cite{chan2022investigating} as the consecutive SR procedure. The quantitative results are given in Tab.~\ref{table_dn} and our eSL-Net and eSL-Net++ perform much better than the two Deblur-then-VSR approaches. According to the qualitative results in Fig.~\ref{fig:vsr}, we can observe that eSL-Net and eSL-Net++ give clearer upscaled faces in the top row and ox feet in the bottom row than the Deblur-then-VSR methods. The performance of two-stage Deblur-then-VSR methods is confined and the deblurring errors may even be magnified by the SR procedure, \eg, halo effects on the right shoulder of the girl in the top row and the vertical stripes on the wall in the bottom row. 
}

\def\ssxxx{(1.4,0.6)} % 噪声
\def\ssyy{(2.2,-0.65)} % 左下角
\def\ssmag{3}
\def\ssizz{1cm}
\begin{figure*}[!htb]
	\centering
	% \subfigure[Noisy events]{
		\begin{minipage}[t]{0.25\linewidth}
			\centering
			\begin{tikzpicture}[spy using outlines={green,magnification=\ssmag,size=\ssizz},inner sep=0]
				\node {\includegraphics[width=0.98\linewidth]{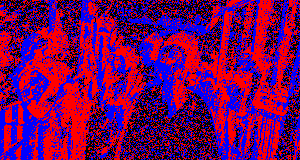}};
				\spy on \ssxxx in node [left] at \ssyy;
				\end{tikzpicture}
    \footnotesize Noisy events
		\end{minipage}%
	% }\hspace*{-1.5mm}
	% \subfigure[$\boldsymbol{E}$ of eSL-Net w/ event noise]{
		\begin{minipage}[t]{0.25\linewidth}
			\centering
			\begin{tikzpicture}[spy using outlines={green,magnification=\ssmag,size=\ssizz},inner sep=0]
				\node {\includegraphics[width=0.98\linewidth]{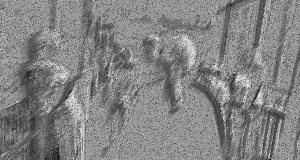}};
				\spy on \ssxxx in node [left] at \ssyy;
				\end{tikzpicture}
    \footnotesize $\boldsymbol{E}$ of eSL-Net w/ event noise
		\end{minipage}%
	% }\hspace*{-1.5mm}
	% \subfigure[$\boldsymbol{E}$ of eSL-Net w/o event noise]{
		\begin{minipage}[t]{0.25\linewidth}
			\centering
			\begin{tikzpicture}[spy using outlines={green,magnification=\ssmag,size=\ssizz},inner sep=0]
				\node {\includegraphics[width=0.98\linewidth]{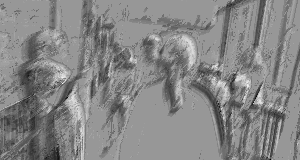}};
				\spy on \ssxxx in node [left] at \ssyy;
				\end{tikzpicture}
    \footnotesize $\boldsymbol{E}$ of eSL-Net w/o event noise
		\end{minipage}%
	% }\hspace*{-1.5mm}
	% \subfigure[$\bar{\boldsymbol{E}}$ of eSL-Net++ w/ event noise]{
		\begin{minipage}[t]{0.25\linewidth}
			\centering
			\begin{tikzpicture}[spy using outlines={green,magnification=\ssmag,size=\ssizz},inner sep=0]
				\node {\includegraphics[width=0.98\linewidth]{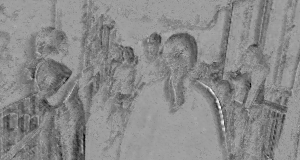}};
				\spy on \ssxxx in node [left] at \ssyy;
				\end{tikzpicture}
    \footnotesize $\bar{\boldsymbol{E}}$ of eSL-Net++ w/ event noise
		\end{minipage}%
	% }
	
	% \subfigure[Blurry image]{
		\begin{minipage}[t]{0.25\linewidth}
			\centering
			\begin{tikzpicture}[spy using outlines={green,magnification=\ssmag,size=\ssizz},inner sep=0]
				\node {\includegraphics[width=0.98\linewidth]{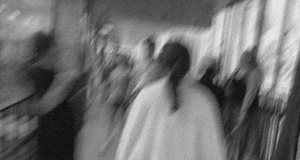}};
				\spy on \ssxxx in node [left] at \ssyy;
				\end{tikzpicture}
        \footnotesize Blurry image\\
		    \footnotesize (PSNR, SSIM)\vspace*{1.5mm}
		\end{minipage}%
	% }\hspace*{-1.5mm}
	% \subfigure[$\boldsymbol{X}$ of eSL-Net w/ event noise]{
		\begin{minipage}[t]{0.25\linewidth}
			\centering
			\begin{tikzpicture}[spy using outlines={green,magnification=\ssmag,size=\ssizz},inner sep=0]
				\node {\includegraphics[width=0.98\linewidth]{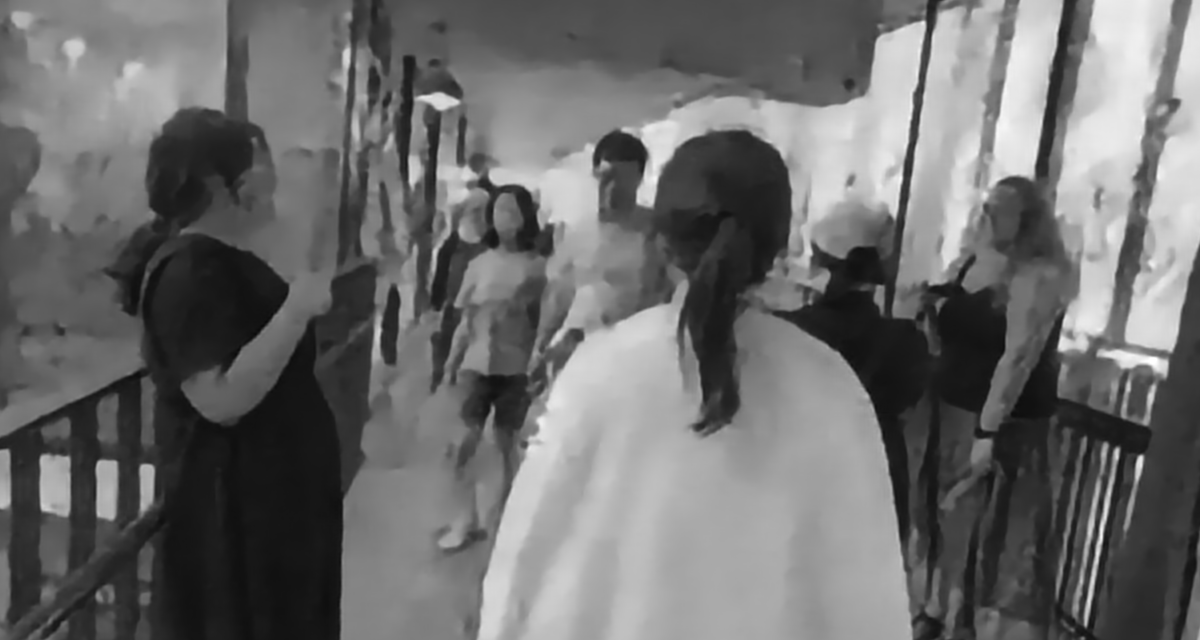}};
				\spy on \ssxxx in node [left] at \ssyy;
				\end{tikzpicture}
\footnotesize $\boldsymbol{X}$ of eSL-Net w/ event noise \\
			\footnotesize (24.54, 0.7642)\vspace*{1.5mm}
		\end{minipage}%
	% }\hspace*{-1.5mm}
	% \subfigure[$\boldsymbol{X}$ of eSL-Net w/o event noise]{
		\begin{minipage}[t]{0.25\linewidth}
			\centering
			\begin{tikzpicture}[spy using outlines={green,magnification=\ssmag,size=\ssizz},inner sep=0]
				\node {\includegraphics[width=0.98\linewidth]{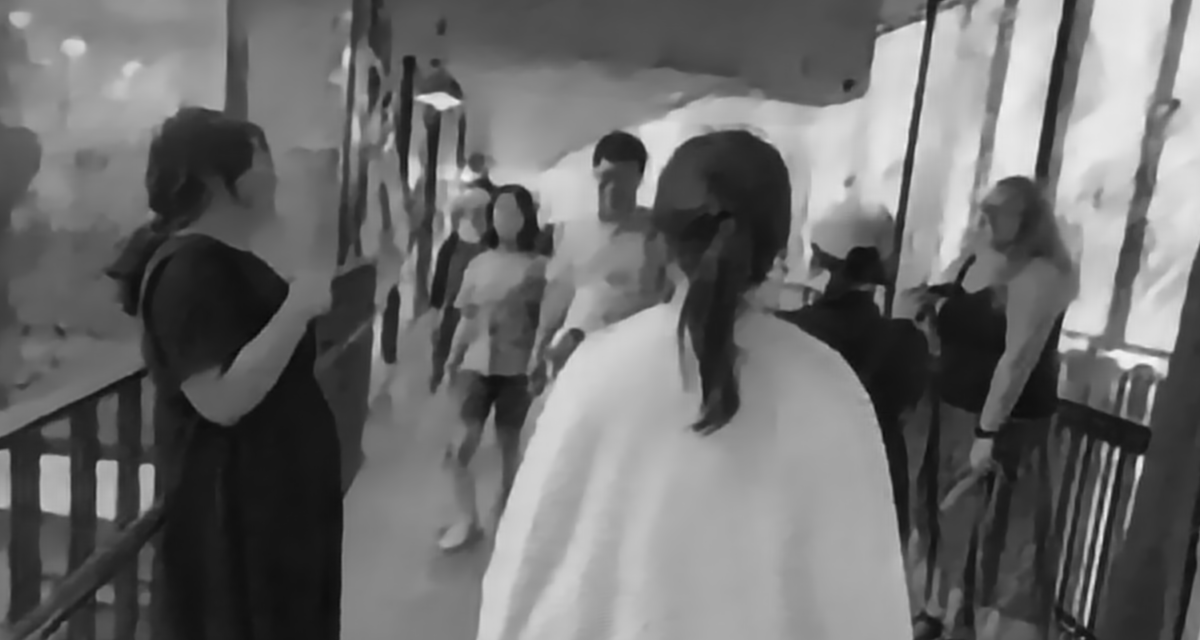}};
				\spy on \ssxxx in node [left] at \ssyy;
				\end{tikzpicture}
    \footnotesize $\boldsymbol{X}$ of eSL-Net w/o event noise\\
			\footnotesize (\underline{24.97}, \underline{0.7768})\vspace*{1.5mm}
		\end{minipage}%
	% }\hspace*{-1.5mm}
	% \subfigure[$\boldsymbol{X}$ of eSL-Net++ w/ event noise]{
		\begin{minipage}[t]{0.25\linewidth}
			\centering
			\begin{tikzpicture}[spy using outlines={green,magnification=\ssmag,size=\ssizz},inner sep=0]
				\node {\includegraphics[width=0.98\linewidth]{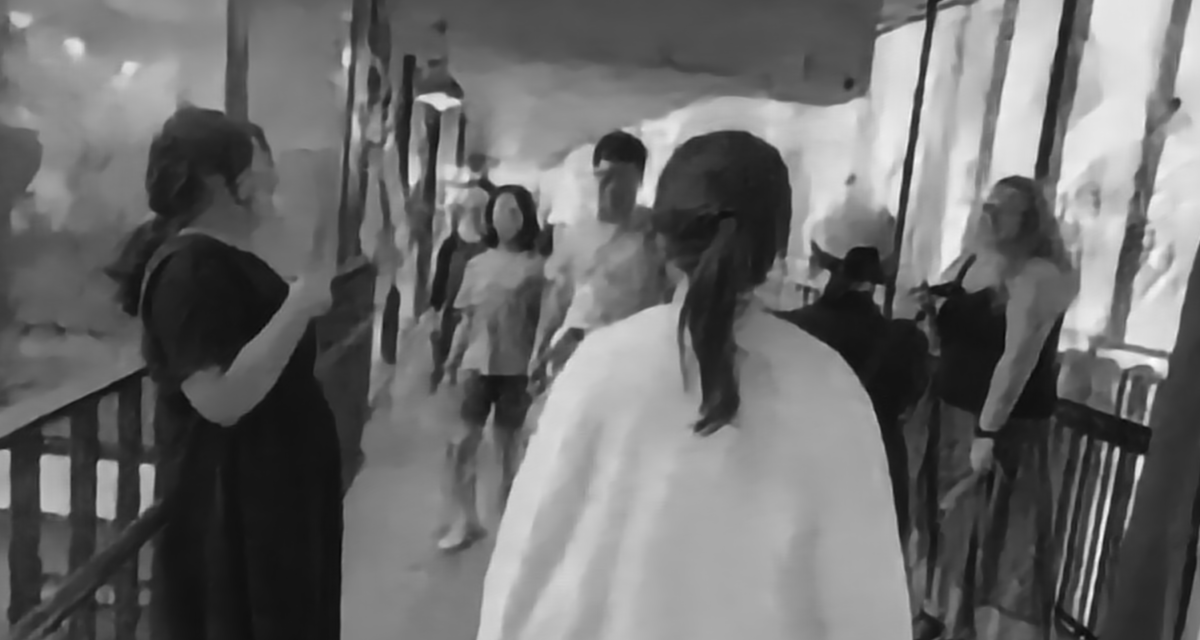}};
				\spy on \ssxxx in node [left] at \ssyy;
				\end{tikzpicture}
    \footnotesize $\boldsymbol{X}$ of eSL-Net++ w/ event noise \\
		\footnotesize ({\bf 25.17, 0.7786})\vspace*{1.5mm}
		\end{minipage}%
	\caption{Qualitative ablation study on the DSL module. (b) and (f) are respectively the output of LDI, \ie, $\boldsymbol{E}$ and the final SR result of eSL-Net \cite{wang2020event} when inputting noisy synthesized events (a) and the corresponding blurry image (e). Accordingly, (c) and (g) are the results of eSL-Net when inputting noiseless events, while (d) and (h) are the results of eSL-Net++ when inputting noisy events.}
% 	(a) (c) the output of LDI in eSL-Net while the input of event stream without noises; (d) the output of LDI in eSL-Net++, \ie, $\bar{\boldsymbol{E}}$; (e) the input of a blur image; (f) the results of eSL-Net; (g) the results of eSL-Net while the input of event stream without noises; (h) the results of eSL-Net++.}
	\label{input-event-frame}
\end{figure*}

% \input{sparse_code_pic}
% \usepackage{multirow}
\begin{table}[htb]
\caption{Ablation studies on {\it DSL} and {\it RESM} over the GoPro dataset with synthetic events and the HQF dataset with real events.}
\label{ablation}
\centering
\begin{tabular}{cclcclcc}
\hline
\multirow{2}{*}{\begin{tabular}[c]{@{}c@{}} DSL \end{tabular}} & \multirow{2}{*}{\begin{tabular}[c]{@{}c@{}} RESM \end{tabular}} &  & \multicolumn{2}{c}{Synthetic events} &  & \multicolumn{2}{c}{Real events} \\ \cline{4-5} \cline{7-8} 
                                                                         &                                                                          &  & PSNR              & SSIM             &  & PSNR            & SSIM          \\ \hline
\XSolidBrush                                                             & \XSolidBrush                                                             &  & 23.80           & 0.6455           &  & 22.16         & 0.7790         \\
\XSolidBrush                                                             & \Checkmark                                                               &  & 24.42           & 0.6530            &  & 22.86         & 0.7890         \\
\Checkmark                                                               & \XSolidBrush                                                                  &  & 24.12           & 0.6579           &  & 22.43         & 0.7898        \\
\Checkmark                                                               & \Checkmark                                                               &  & \textbf{24.69}  & \textbf{0.6602}  &  & \textbf{22.99}& \textbf{0.7913} \\ \hline
\end{tabular}
\end{table}

\def\imwidth{0.195}
\def\cimwid{0.072}

\def\zuoxia{(-1,-0.5)}
\def\youshang{(-0.1,1)}

\def\ssyy{(-0.8,-0.85)}
\def\ssizz{0.5cm}
\def\sswidth{0.245\textwidth}
\def\ssmag{3}
\def\scc{(2.12,1.4)}

\begin{figure*}[htb]
\footnotesize
	\centering
    % \subfigure{
    	\begin{minipage}[t]{\imwidth\linewidth}
    		\centering
    		%ground truth
			\begin{tikzpicture}[spy using outlines={rectangle,green,magnification=\ssmag,size=\ssizz},inner sep=0]
				\node {\includegraphics[width=\linewidth]{pic/mulframe/0019.jpg}};
				% \spy on \ssxxs in node [left] at \ssyys;
				\draw[green] \zuoxia rectangle \youshang;
				\end{tikzpicture}
            Blurry Image\vspace{0.3em}
    	\end{minipage}%
    % }
    \hspace*{-1mm}
    % \subfigure{
    	\begin{minipage}[t]{\imwidth\linewidth}
    		\centering
    		\begin{tikzpicture}[spy using outlines={rectangle,green,magnification=\ssmag,size=\ssizz},inner sep=0]
				\node {\includegraphics[width=\linewidth]{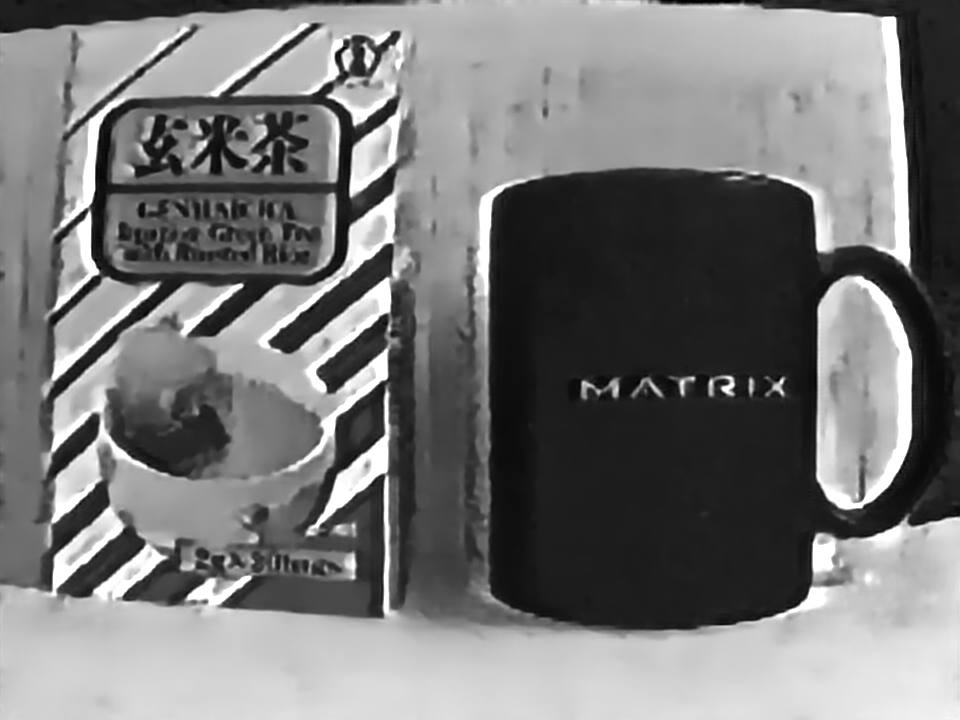}};
				% \spy on \ssxxs in node [left] at \ssyys;
				\draw[green] \zuoxia rectangle \youshang;
				\end{tikzpicture}
			
			Baseline\vspace{0.5em}
    	\end{minipage}%
    % }
    \hspace*{-1mm}
    % \subfigure{
    	\begin{minipage}[t]{\imwidth\linewidth}
    		\centering
    	    \begin{tikzpicture}[spy using outlines={green,magnification=\ssmag,size=\ssizz},inner sep=0]
				\node {\includegraphics[width=\linewidth]{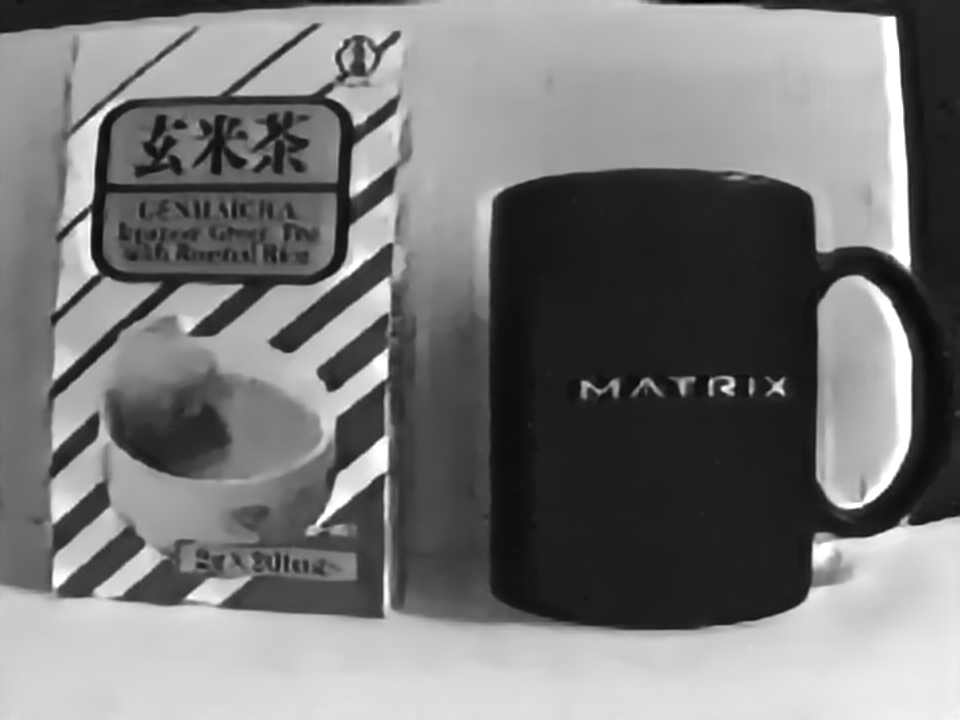}};
				% \spy on \ssxxs in node [left] at \ssyys;
				\draw[green] \zuoxia rectangle \youshang;
				\end{tikzpicture}
			RESM\vspace{0.5em}
    	\end{minipage}%
    % }
    \hspace*{-1mm}
    % \subfigure{
    	\begin{minipage}[t]{\imwidth\linewidth}
    		\centering
    	    \begin{tikzpicture}[spy using outlines={green,magnification=\ssmag,size=\ssizz},inner sep=0]
				\node {\includegraphics[width=\linewidth]{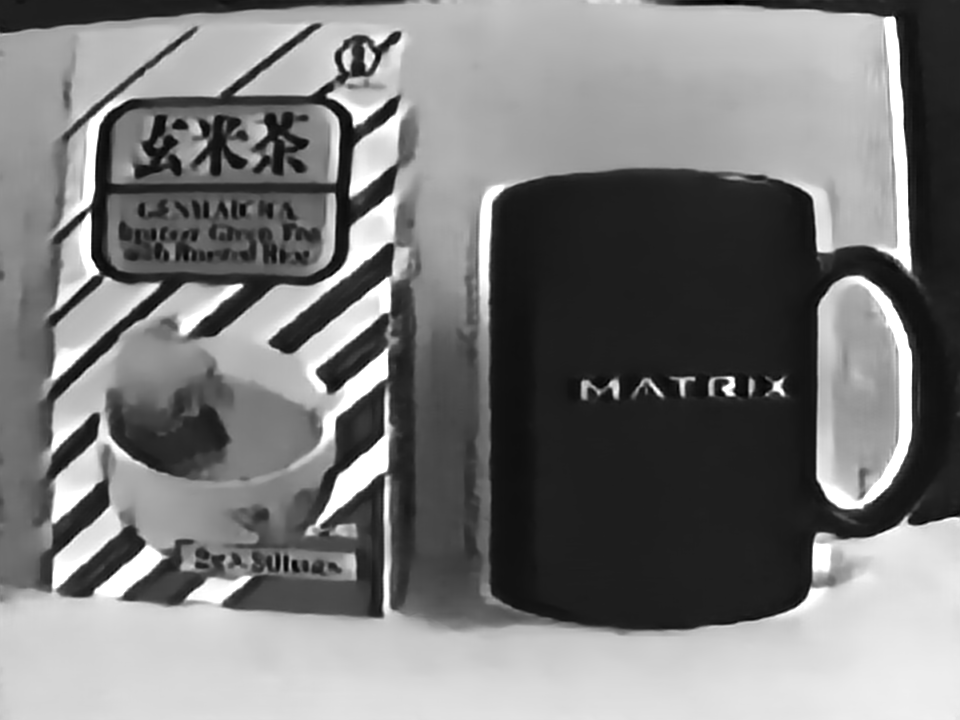}};
				% \spy on \ssxxs in node [left] at \ssyys;
				\draw[green] \zuoxia rectangle \youshang;
				\end{tikzpicture}
			DSL\vspace{0.5em}
    	\end{minipage}%
    % }
    \hspace*{-1mm}
	% \subfigure{
		\begin{minipage}[t]{\imwidth\linewidth}
			\centering
			\begin{tikzpicture}[spy using outlines={green,magnification=\ssmag,size=\ssizz},inner sep=0]
				\node {\includegraphics[width=\linewidth]{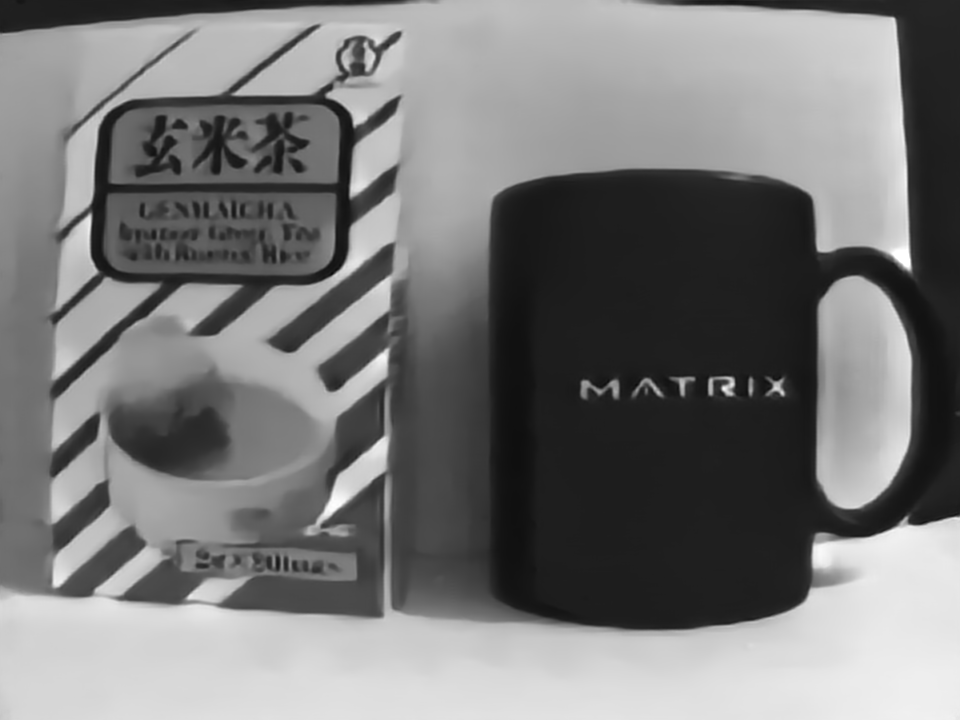}};
				% \spy on \ssxxs in node [left] at \ssyys;
				\draw[green] \zuoxia rectangle \youshang;
				\end{tikzpicture}
			DSL+RESM \vspace{.5em}
			
% 			(27.60/26.49, 0.8318/0.7907)\\
			
		\end{minipage}%
	% }
 
 % \hspace*{-1mm}
 %    % \subfigure{
 %    	\begin{minipage}[t]{\imwidth\linewidth}
 %    		\centering
 %    		\begin{tikzpicture}[spy using outlines={green,magnification=\ssmag,size=\ssizz},inner sep=0]
	% 			\node {\includegraphics[width=\linewidth]{pic/mulframe/eslnet++/0019_0.png}};
	% 			% \spy on \ssxxs in node [left] at \ssyys;
	% 			\draw[green] \zuoxia rectangle \youshang;
	% 			\end{tikzpicture}
 %    		{\bf eSL-Net++} \vspace{.5em}
 %    % 		(28.50/26.98, 0.8548/0.8060)\\
 %    	\end{minipage}%
 
	% \subfigure{%left, bottom, right and top
	        \begin{tikzpicture}[inner sep=0]
            \node [label={[label distance=0.5cm,text depth=-1ex,rotate=90]right: {\scriptsize \textbf{Baseline}}}] at (0,8.7) {};
            \end{tikzpicture}
			\includegraphics[width=\cimwid\linewidth,trim={220px 230px 500px 100px},,clip=true]{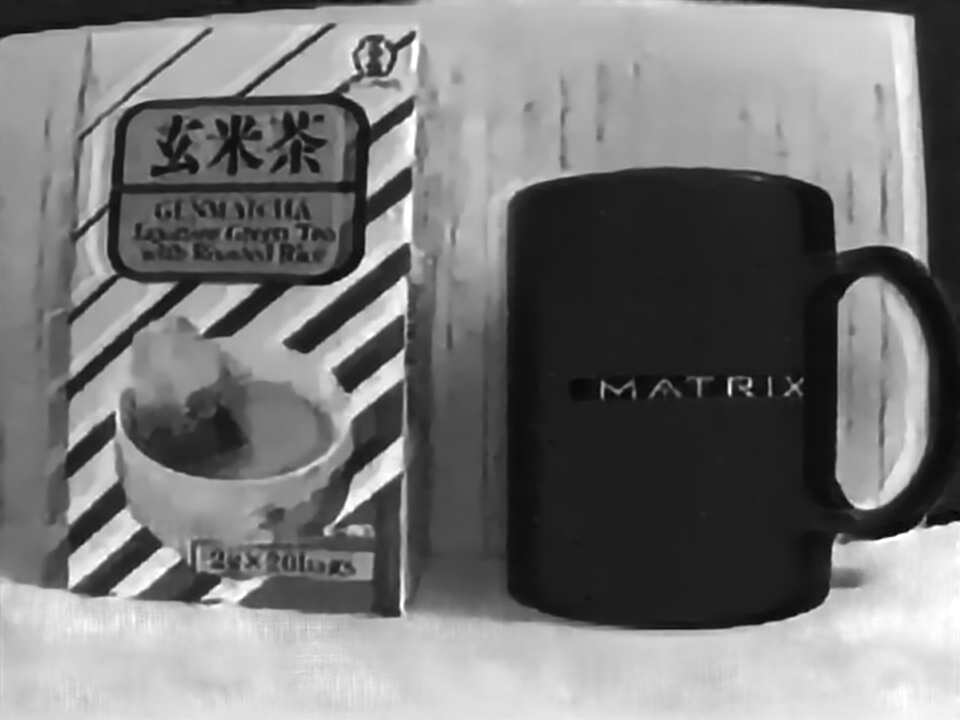}\hspace*{-0.5mm}
			\includegraphics[width=\cimwid\linewidth,trim={220px 230px 500px 100px},clip]{pic/Ablation/eSL-wo-ESM/0001_00.png}\hspace*{-0.5mm}
			\includegraphics[width=\cimwid\linewidth,trim={220px 230px 500px 100px},clip]{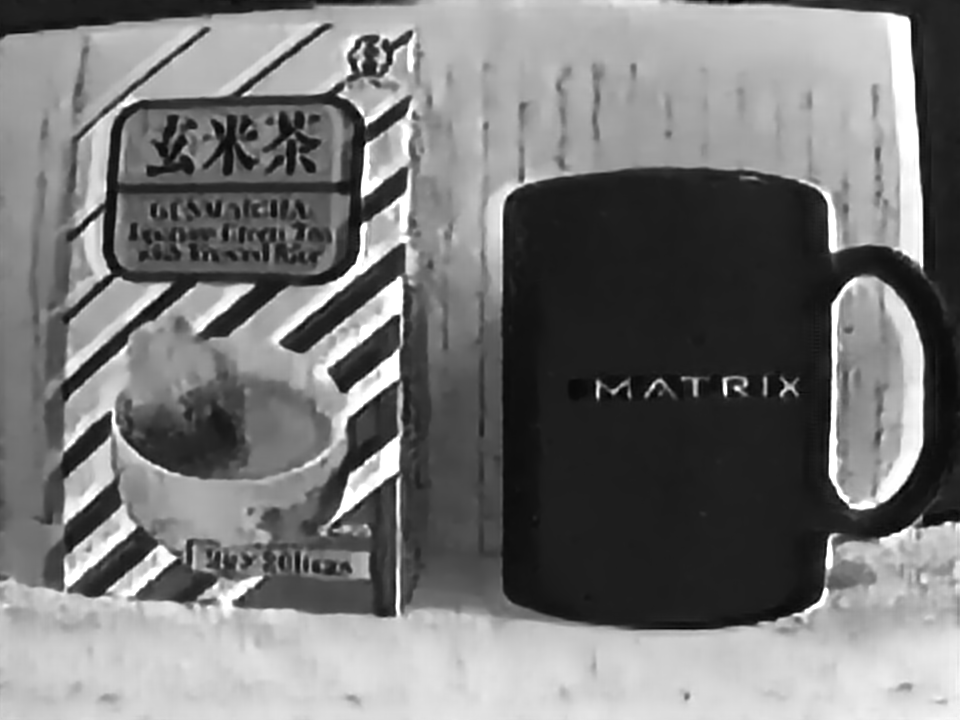}\hspace*{-0.5mm}
			\includegraphics[width=\cimwid\linewidth,trim={220px 230px 500px 100px},clip]{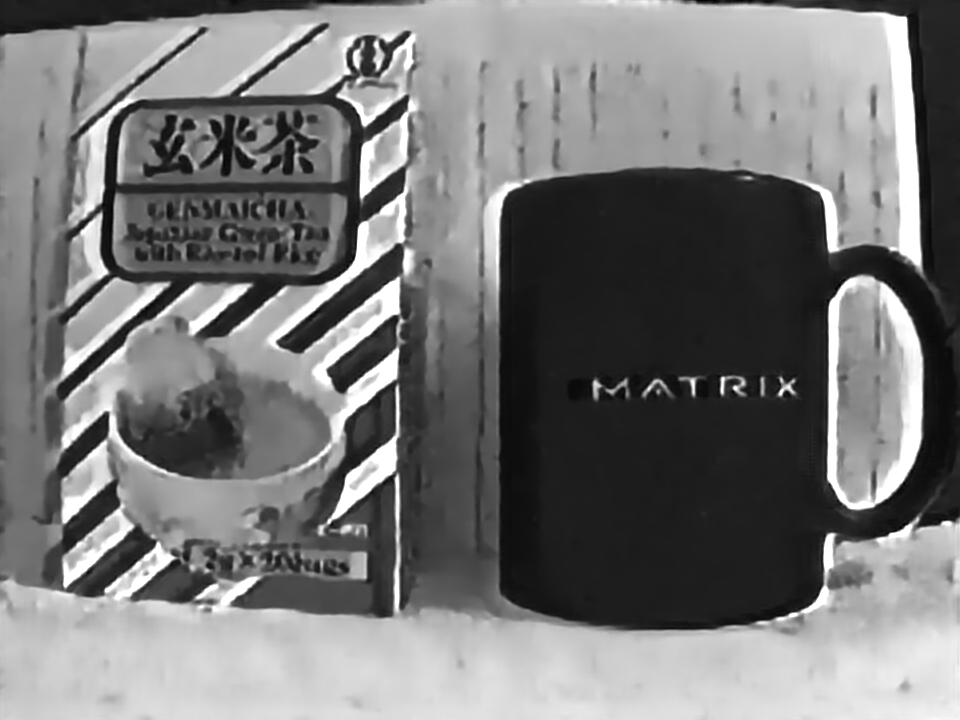}\hspace*{-0.5mm}
			\includegraphics[width=\cimwid\linewidth,trim={220px 230px 500px 100px},clip]{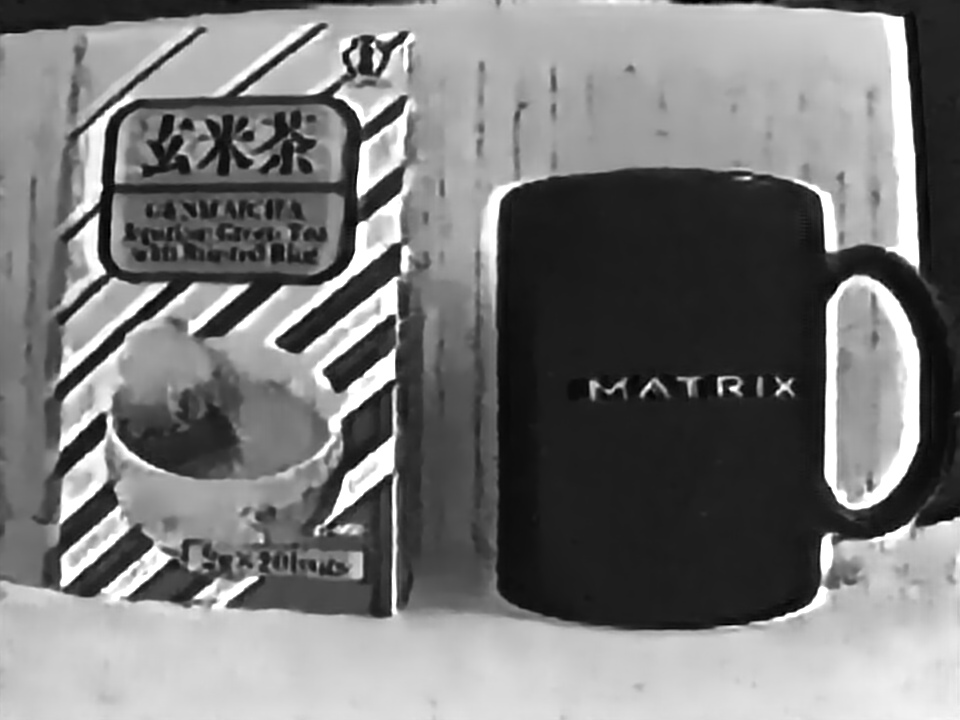}\hspace*{-0.5mm}
			\includegraphics[width=\cimwid\linewidth,trim={220px 230px 500px 100px},clip]{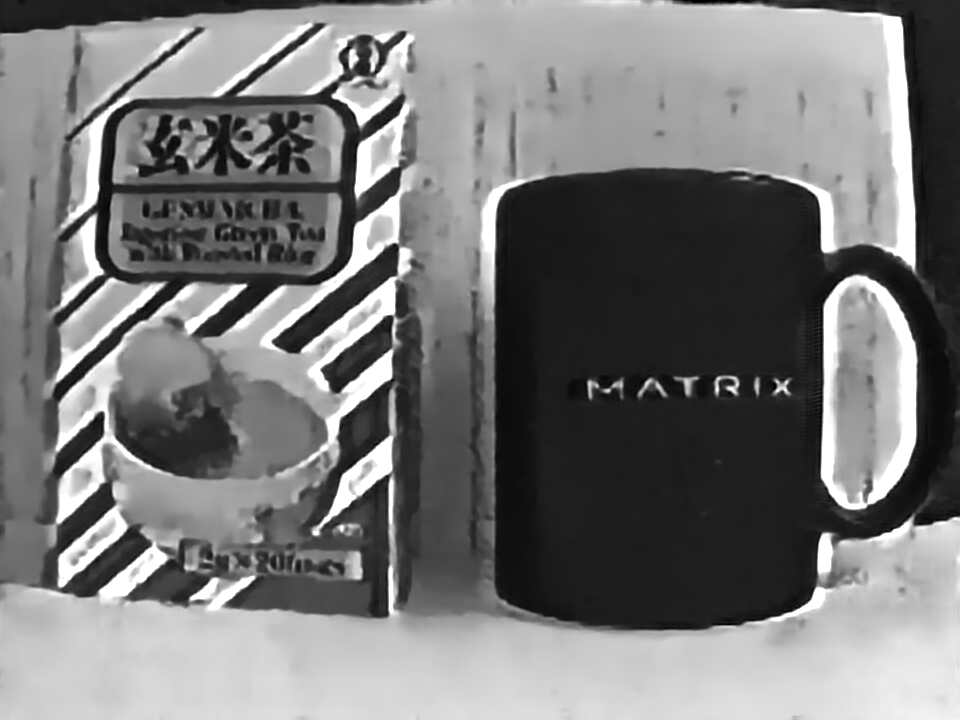}\hspace*{-0.5mm}
			\includegraphics[width=\cimwid\linewidth,trim={220px 230px 500px 100px},clip]{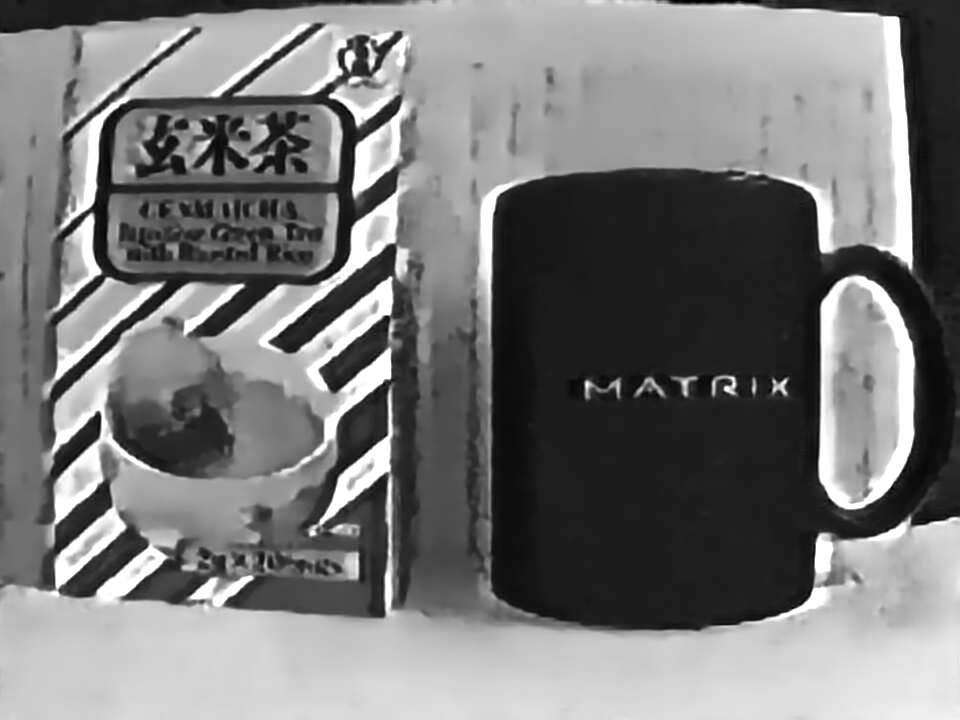}\hspace*{-0.5mm}
			\includegraphics[width=\cimwid\linewidth,trim={220px 230px 500px 100px},clip]{pic/Ablation/eSL-wo-ESM/0001_07.png}\hspace*{-0.5mm}
			\includegraphics[width=\cimwid\linewidth,trim={220px 230px 500px 100px},clip]{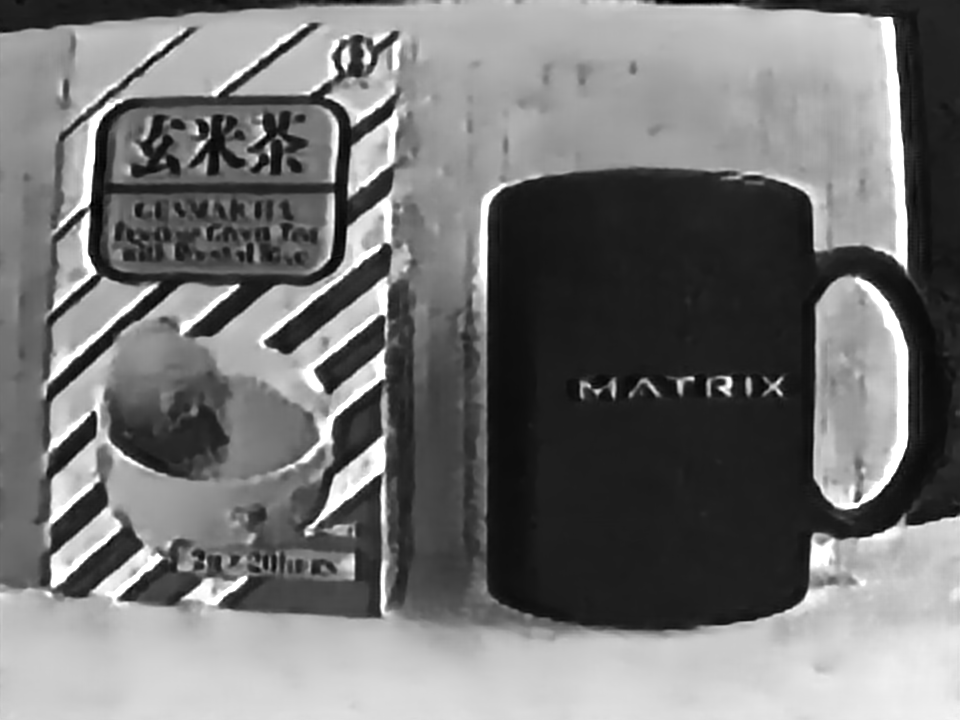}\hspace*{-0.5mm}
			\includegraphics[width=\cimwid\linewidth,trim={220px 230px 500px 100px},clip]{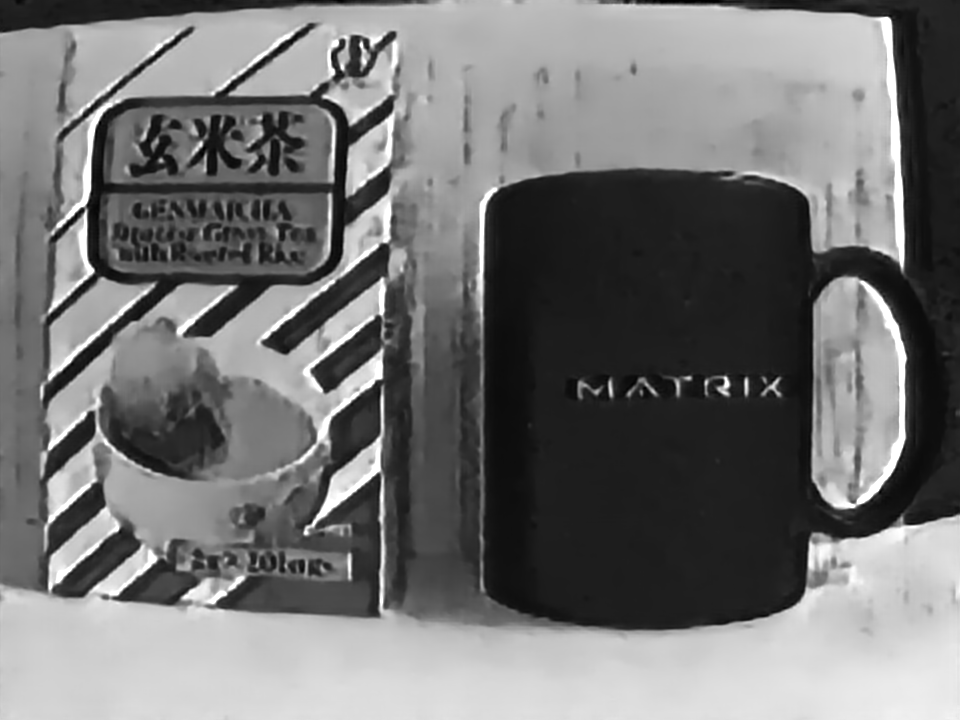}\hspace*{-0.5mm}
			\includegraphics[width=\cimwid\linewidth,trim={220px 230px 500px 100px},clip]{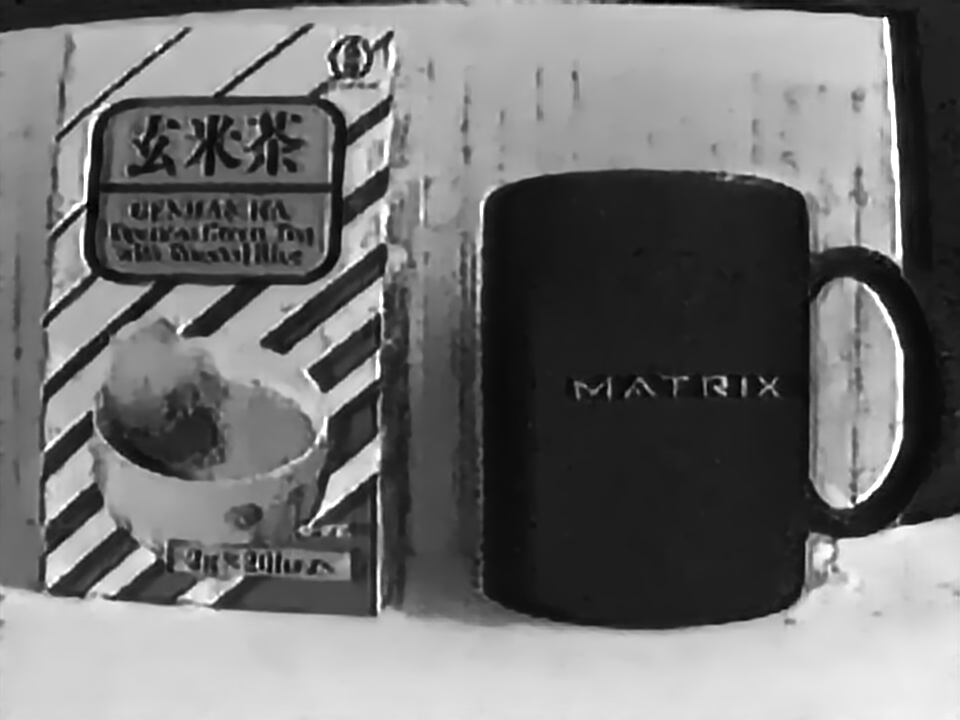}\hspace*{-0.5mm}
			\includegraphics[width=\cimwid\linewidth,trim={220px 230px 500px 100px},clip]{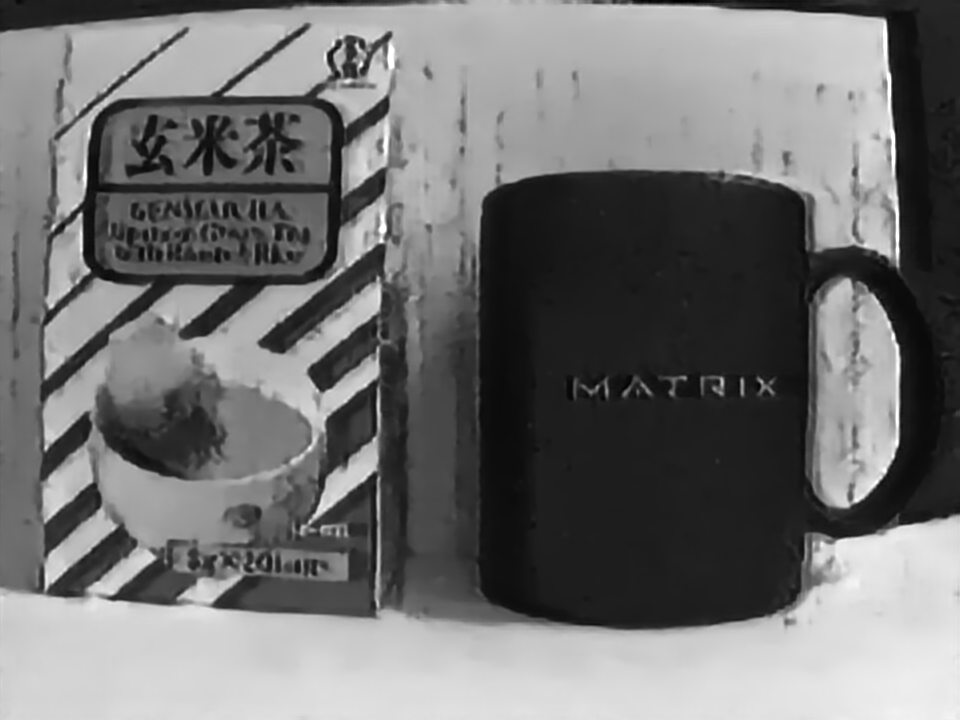}\hspace*{-0.5mm}
			\includegraphics[width=\cimwid\linewidth,trim={220px 230px 500px 100px},clip]{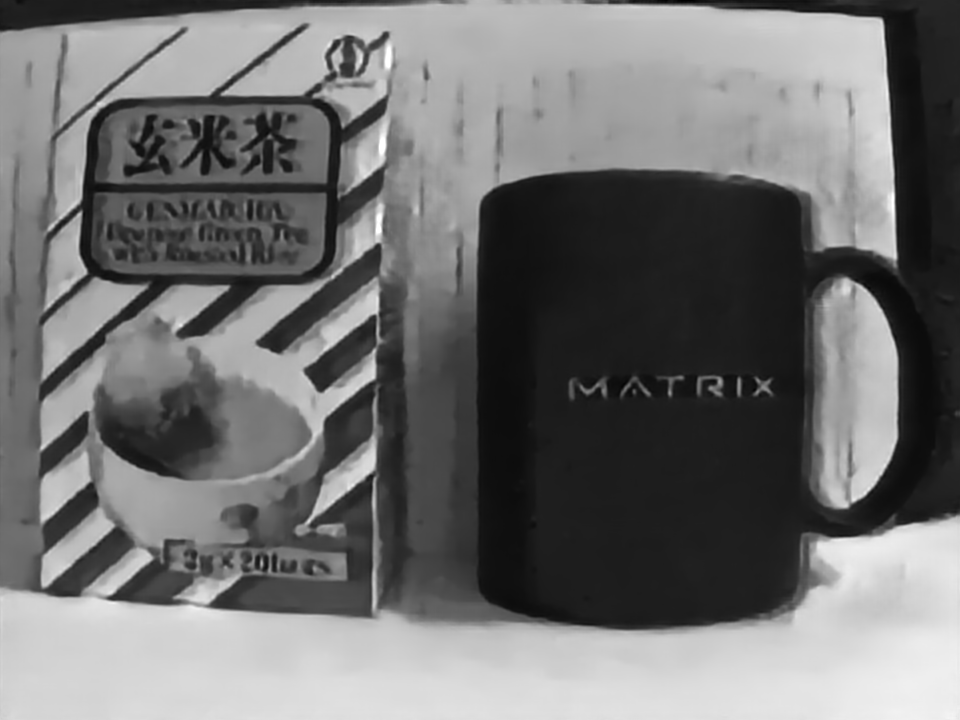}
	% }
 \vspace{.1em}
	\\
	% \subfigure{%left, bottom, right and top
	        \begin{tikzpicture}[inner sep=0]
            \node [label={[label distance=0.7cm,text depth=-1ex,rotate=90]right: {\scriptsize \textbf{RESM}}}] at (0,8.7) {};
            \end{tikzpicture}
			\includegraphics[width=\cimwid\linewidth,trim={220px 230px 500px 100px},,clip=true]{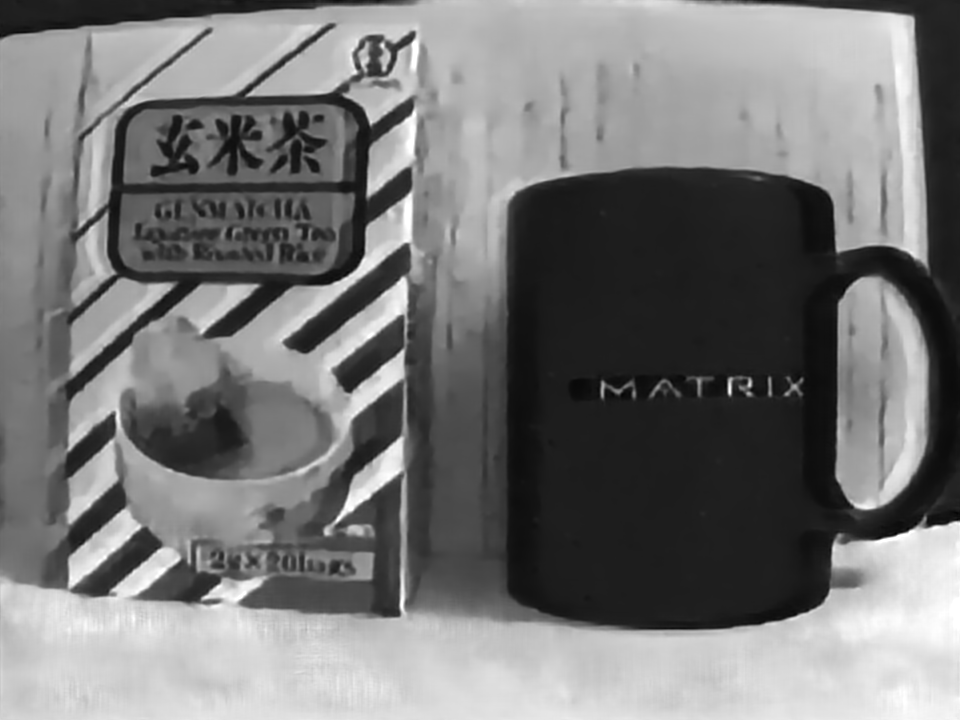}\hspace*{-0.5mm}
			\includegraphics[width=\cimwid\linewidth,trim={220px 230px 500px 100px},clip]{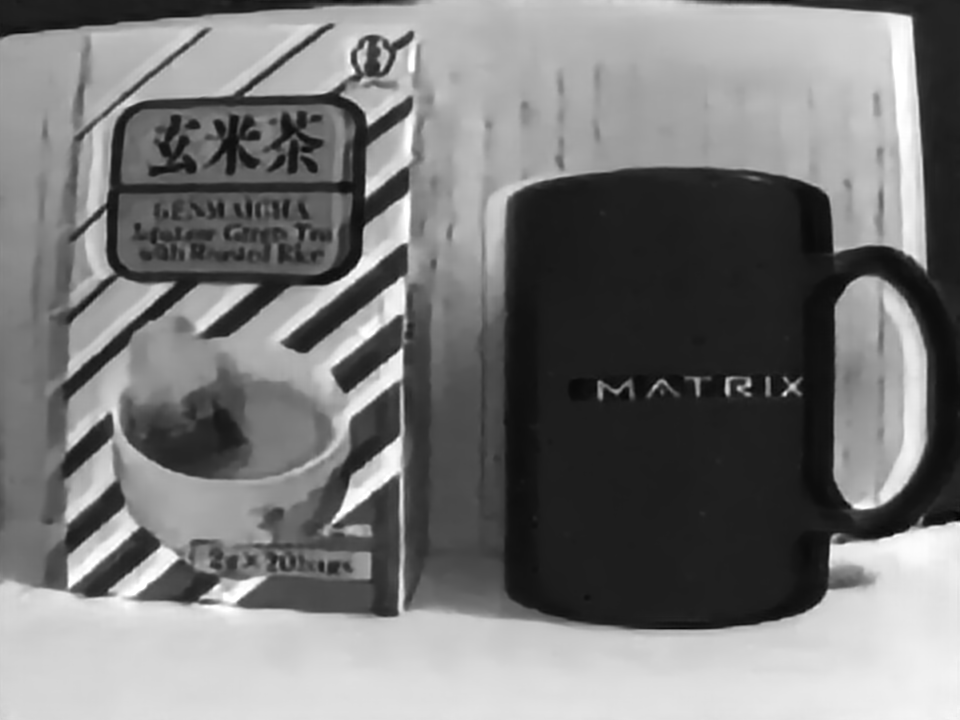}\hspace*{-0.5mm}
			\includegraphics[width=\cimwid\linewidth,trim={220px 230px 500px 100px},clip]{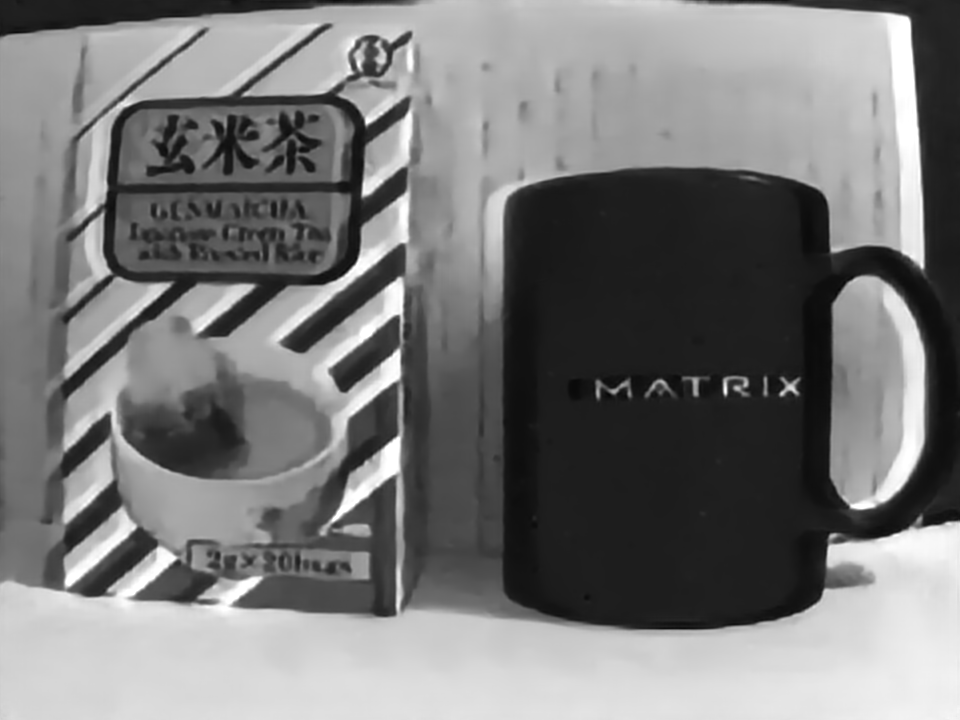}\hspace*{-0.5mm}
			\includegraphics[width=\cimwid\linewidth,trim={220px 230px 500px 100px},clip]{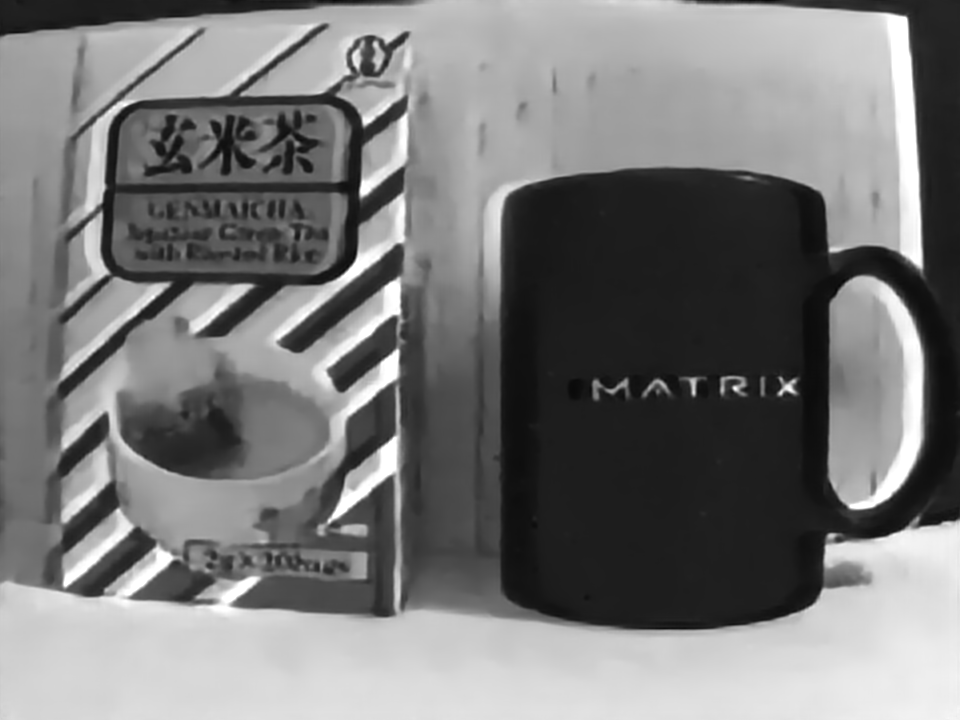}\hspace*{-0.5mm}
			\includegraphics[width=\cimwid\linewidth,trim={220px 230px 500px 100px},clip]{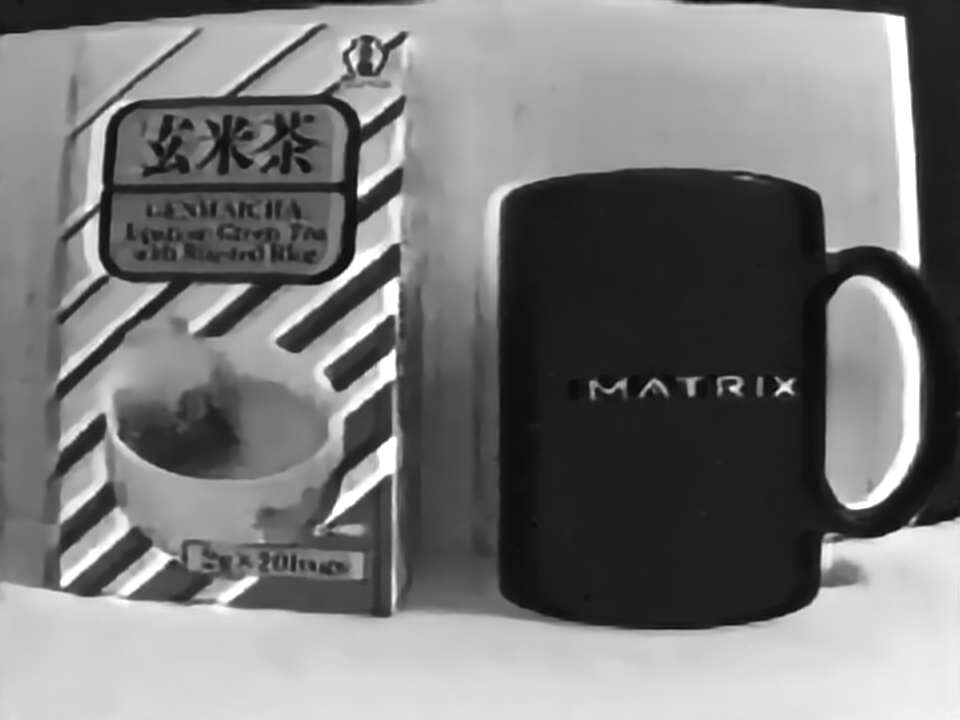}\hspace*{-0.5mm}
			\includegraphics[width=\cimwid\linewidth,trim={220px 230px 500px 100px},clip]{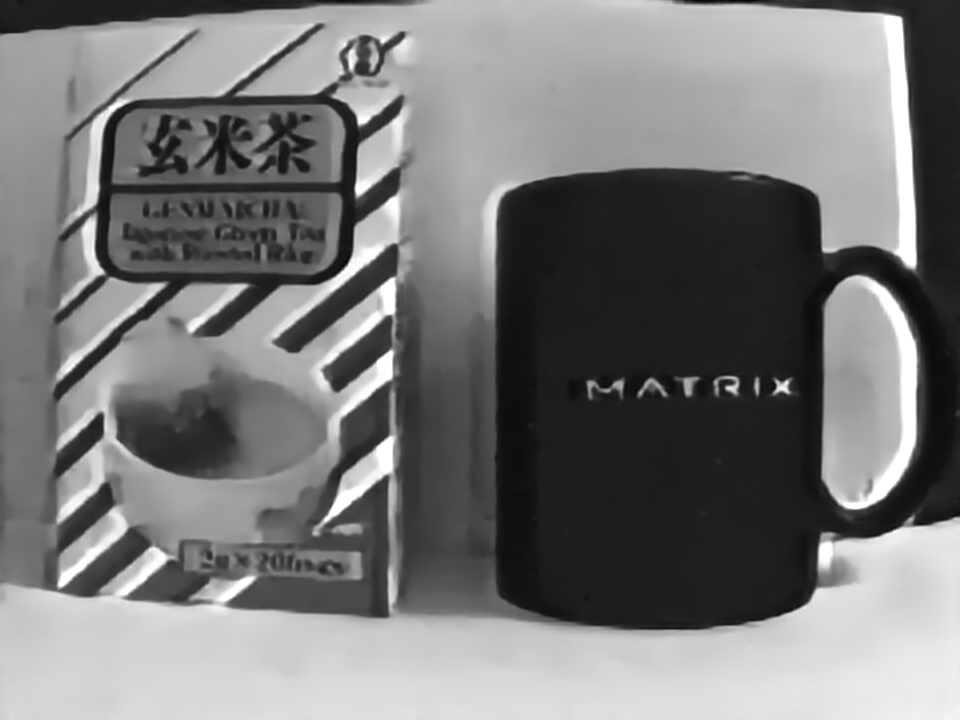}\hspace*{-0.5mm}
			\includegraphics[width=\cimwid\linewidth,trim={220px 230px 500px 100px},clip]{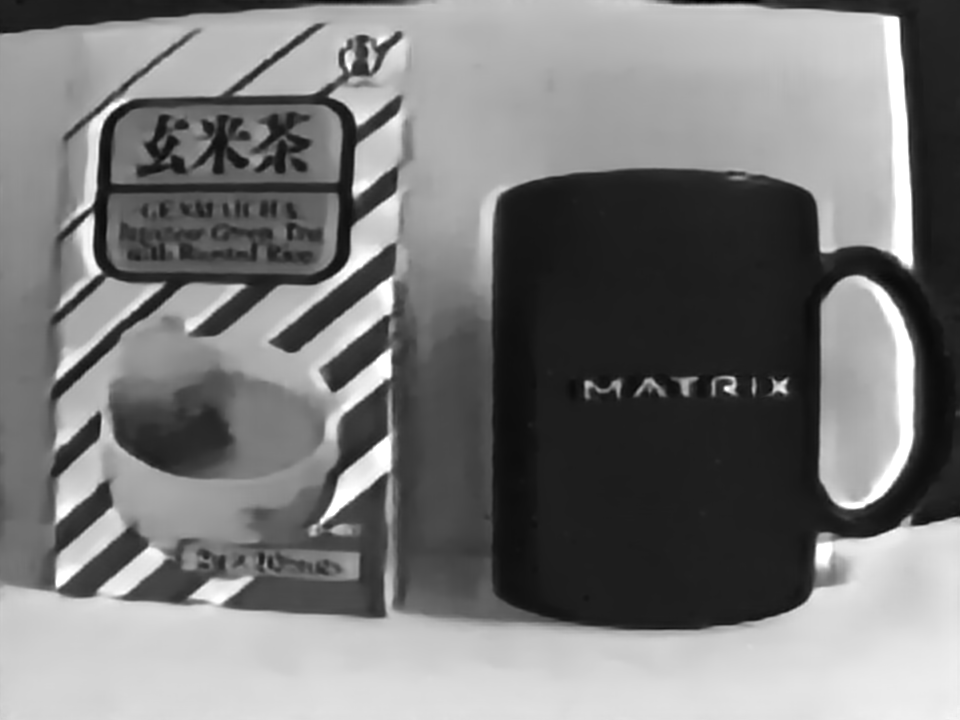}\hspace*{-0.5mm}
			\includegraphics[width=\cimwid\linewidth,trim={220px 230px 500px 100px},clip]{pic/Ablation/eSL-w-ESM/0001_07.png}\hspace*{-0.5mm}
			\includegraphics[width=\cimwid\linewidth,trim={220px 230px 500px 100px},clip]{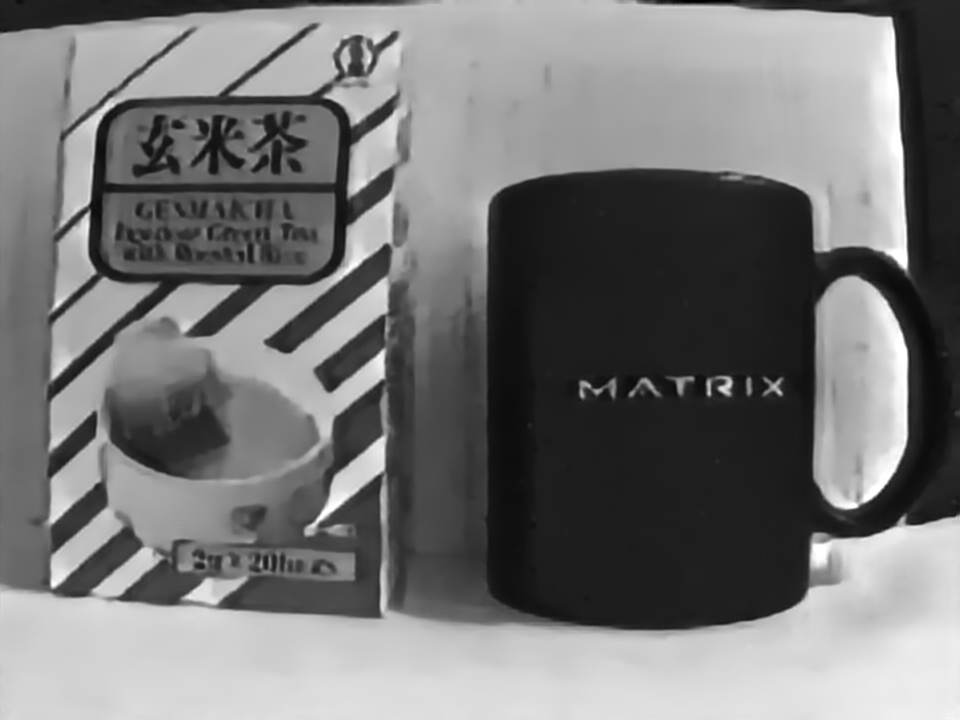}\hspace*{-0.5mm}
			\includegraphics[width=\cimwid\linewidth,trim={220px 230px 500px 100px},clip]{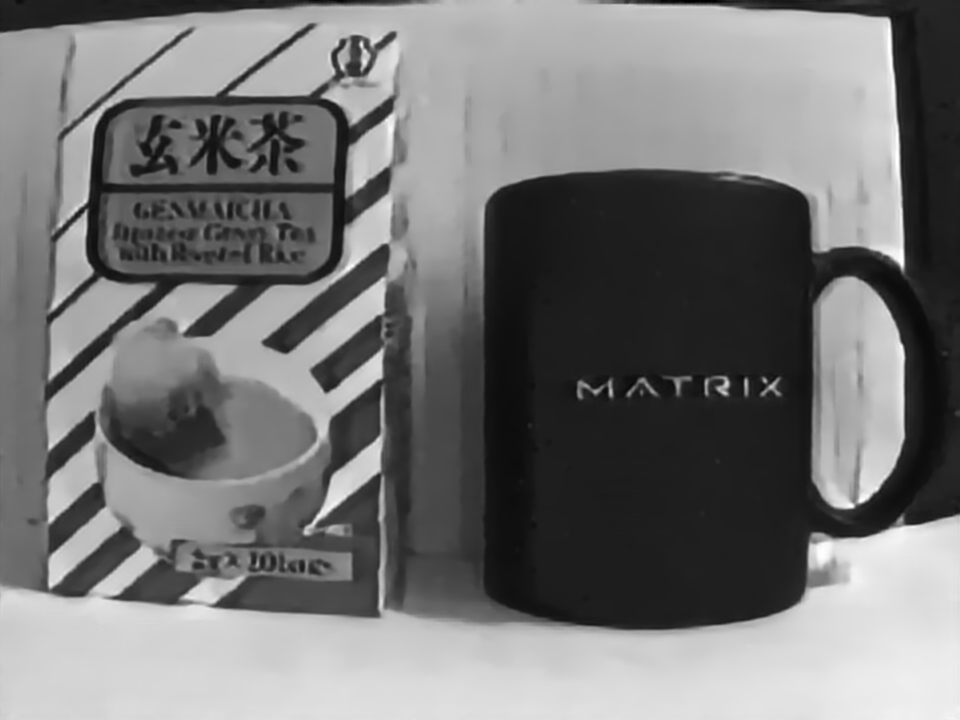}\hspace*{-0.5mm}
			\includegraphics[width=\cimwid\linewidth,trim={220px 230px 500px 100px},clip]{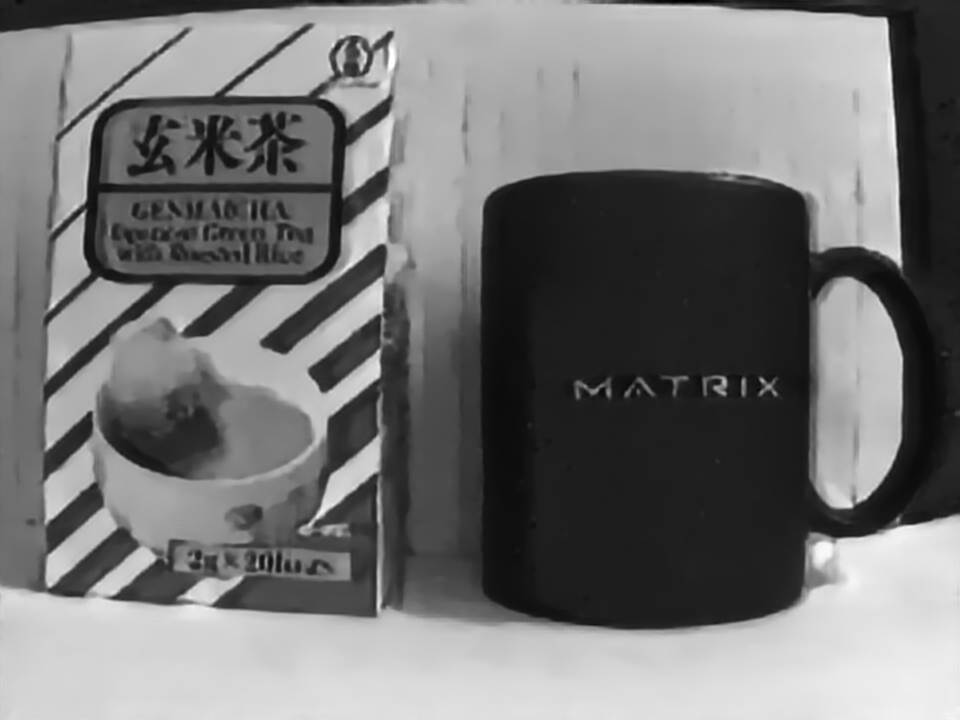}\hspace*{-0.5mm}
			\includegraphics[width=\cimwid\linewidth,trim={220px 230px 500px 100px},clip]{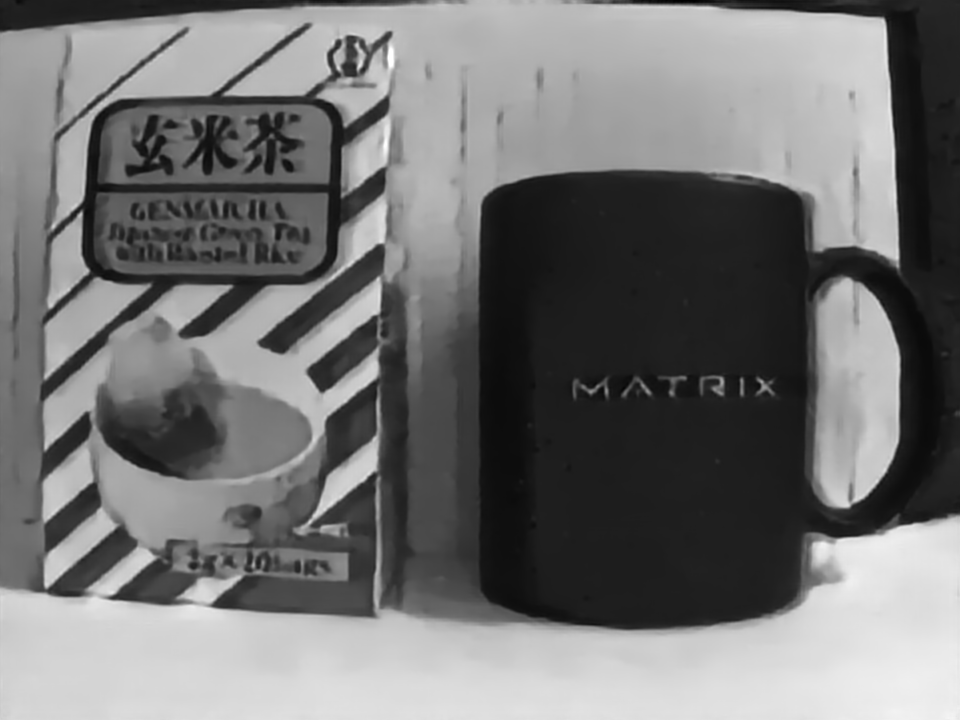}\hspace*{-0.5mm}
			\includegraphics[width=\cimwid\linewidth,trim={220px 230px 500px 100px},clip]{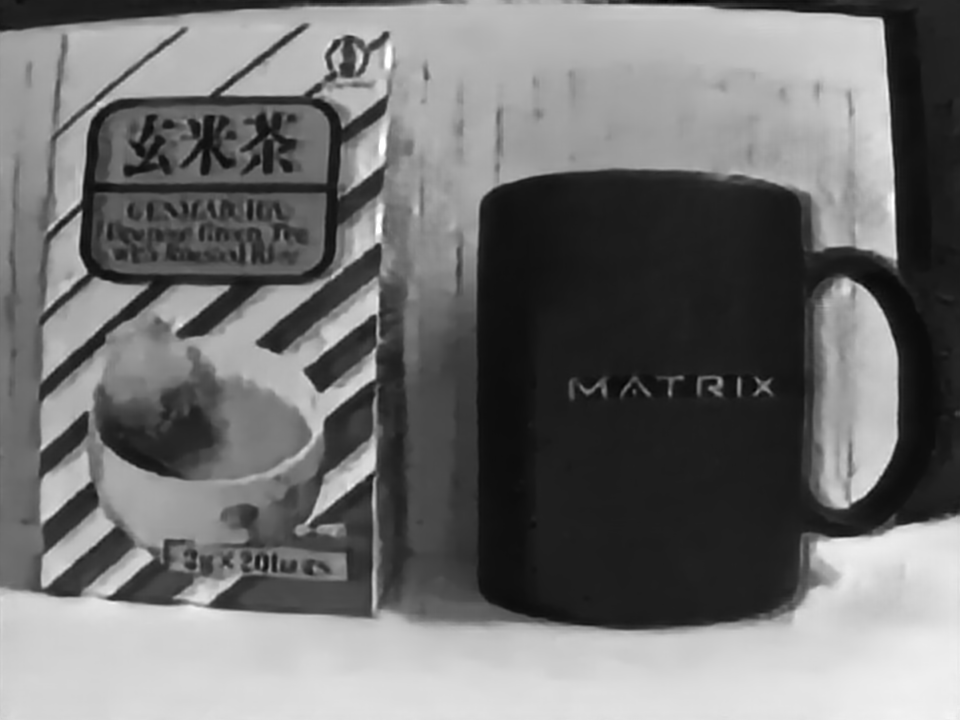}
    \vspace{.1em}\\
    % third row
    \begin{tikzpicture}[inner sep=0]
            \node [label={[label distance=0.8cm,text depth=-1ex,rotate=90]right: {\scriptsize \textbf{DSL}}}] at (0,8.7) {};
            \end{tikzpicture}
			\includegraphics[width=\cimwid\linewidth,trim={220px 230px 500px 100px},,clip=true]{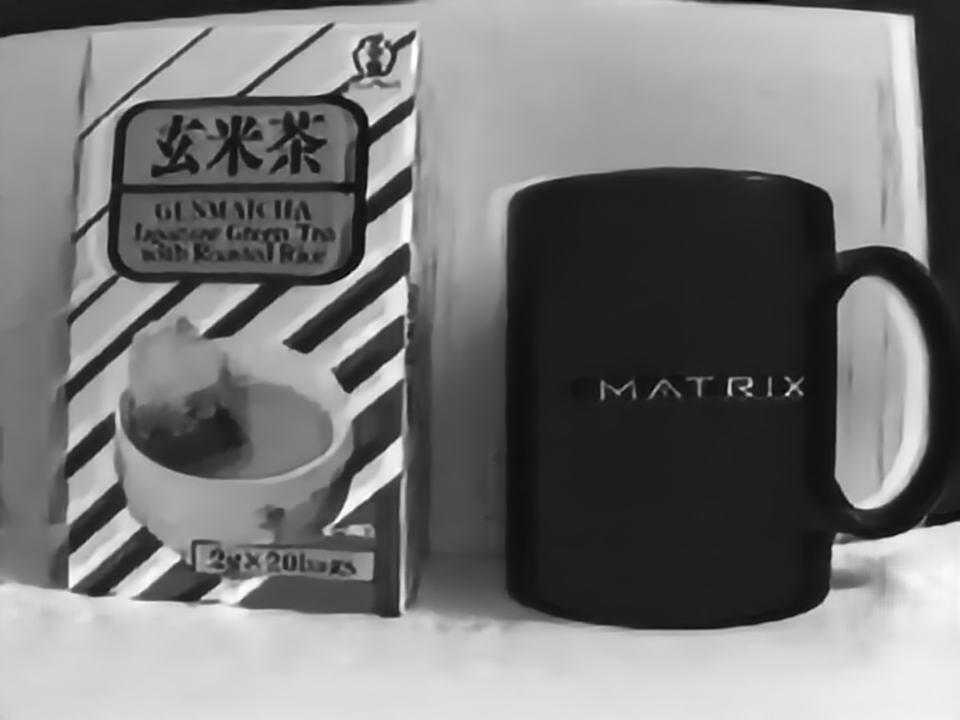}\hspace*{-0.5mm}
			\includegraphics[width=\cimwid\linewidth,trim={220px 230px 500px 100px},clip]{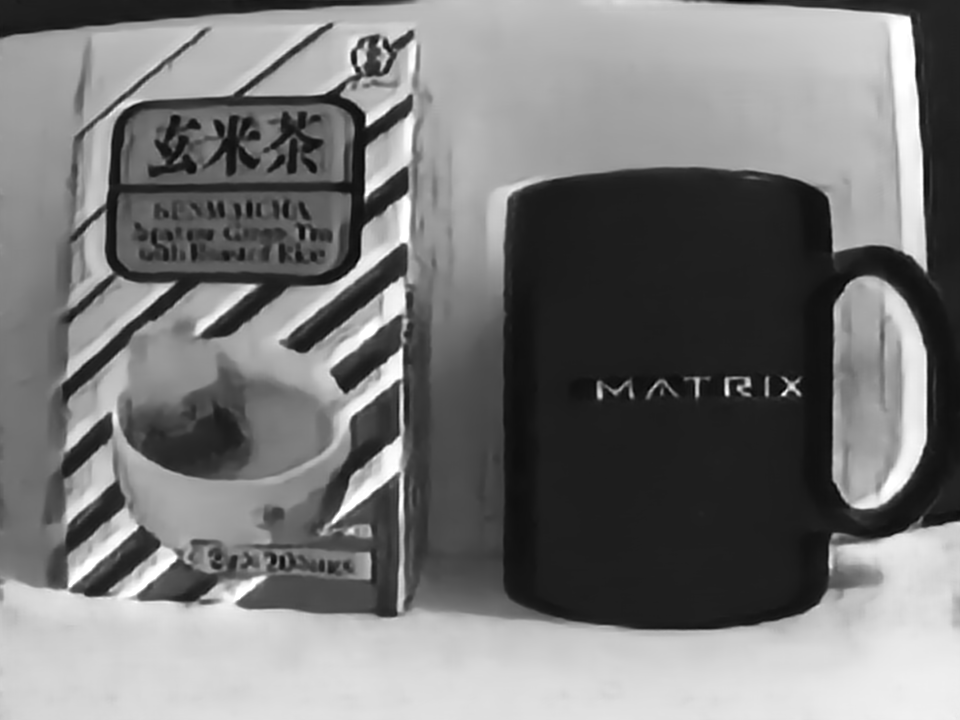}\hspace*{-0.5mm}
			\includegraphics[width=\cimwid\linewidth,trim={220px 230px 500px 100px},clip]{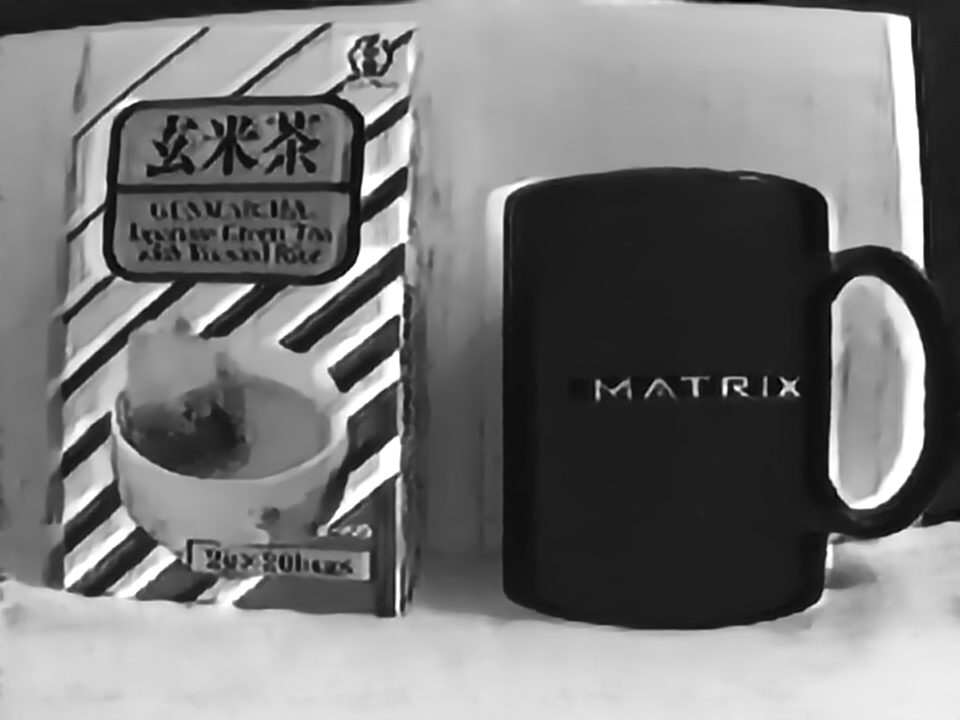}\hspace*{-0.5mm}
			\includegraphics[width=\cimwid\linewidth,trim={220px 230px 500px 100px},clip]{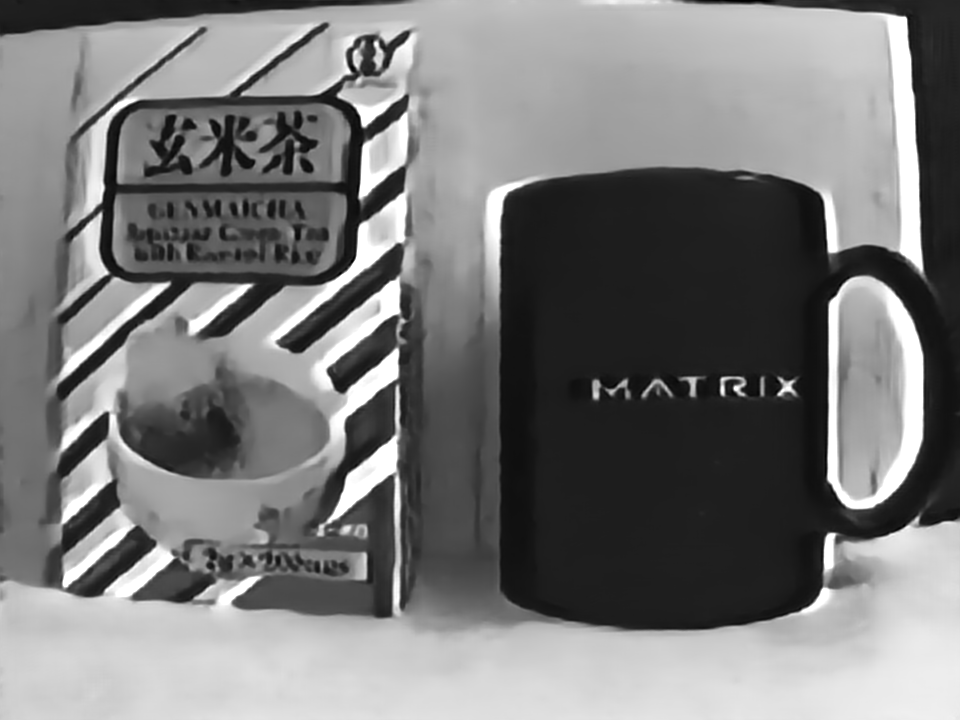}\hspace*{-0.5mm}
			\includegraphics[width=\cimwid\linewidth,trim={220px 230px 500px 100px},clip]{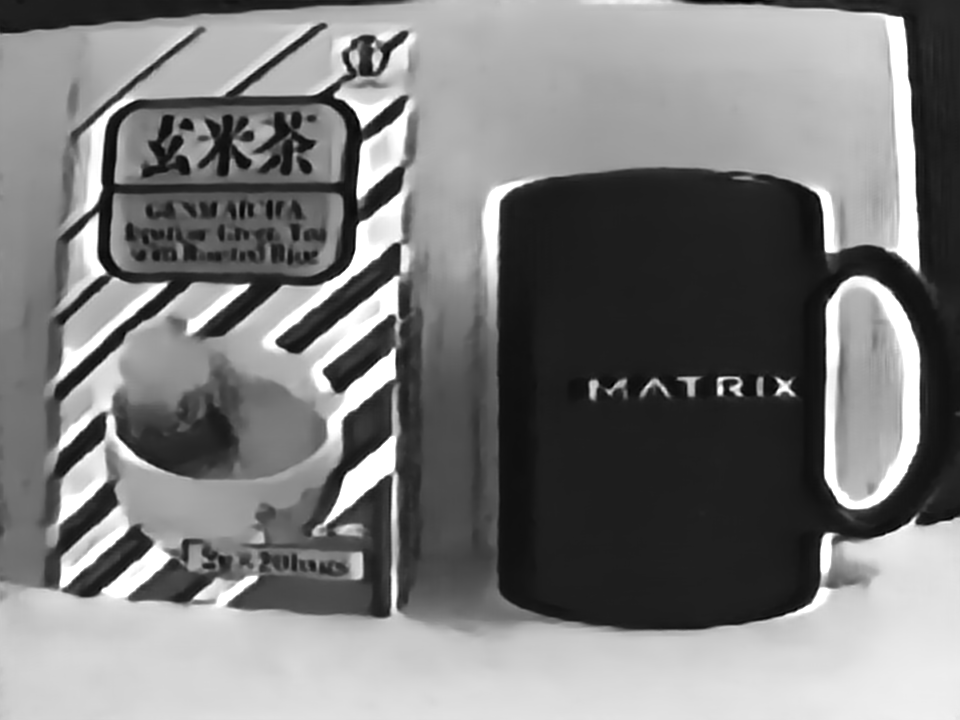}\hspace*{-0.5mm}
			\includegraphics[width=\cimwid\linewidth,trim={220px 230px 500px 100px},clip]{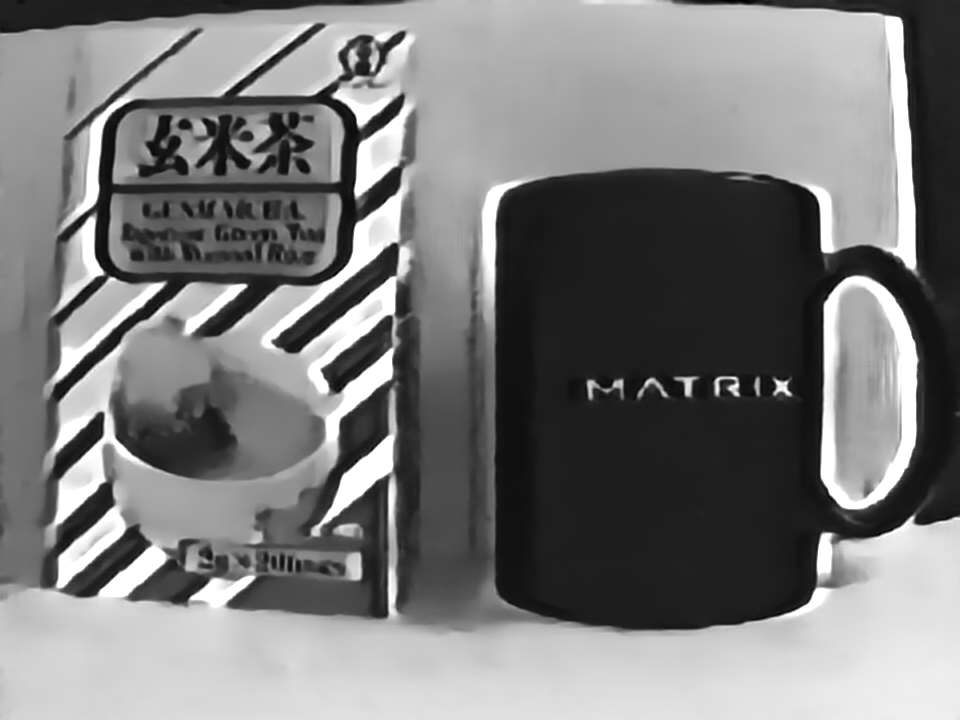}\hspace*{-0.5mm}
			\includegraphics[width=\cimwid\linewidth,trim={220px 230px 500px 100px},clip]{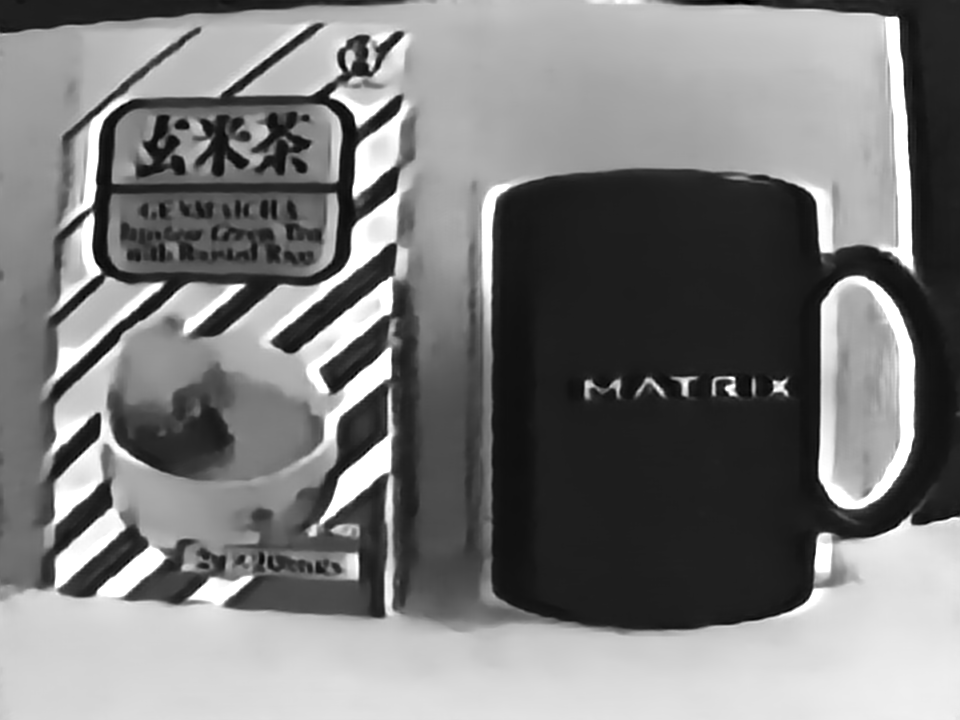}\hspace*{-0.5mm}
			\includegraphics[width=\cimwid\linewidth,trim={220px 230px 500px 100px},clip]{pic/Ablation/eSL++-wo-ESM/0001_07.png}\hspace*{-0.5mm}
			\includegraphics[width=\cimwid\linewidth,trim={220px 230px 500px 100px},clip]{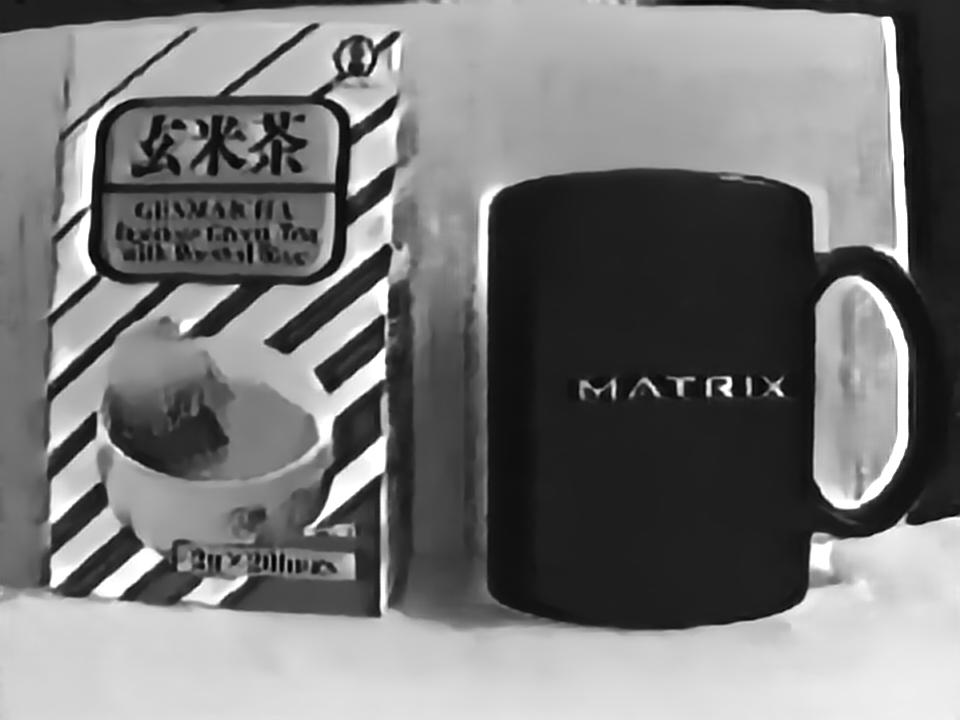}\hspace*{-0.5mm}
			\includegraphics[width=\cimwid\linewidth,trim={220px 230px 500px 100px},clip]{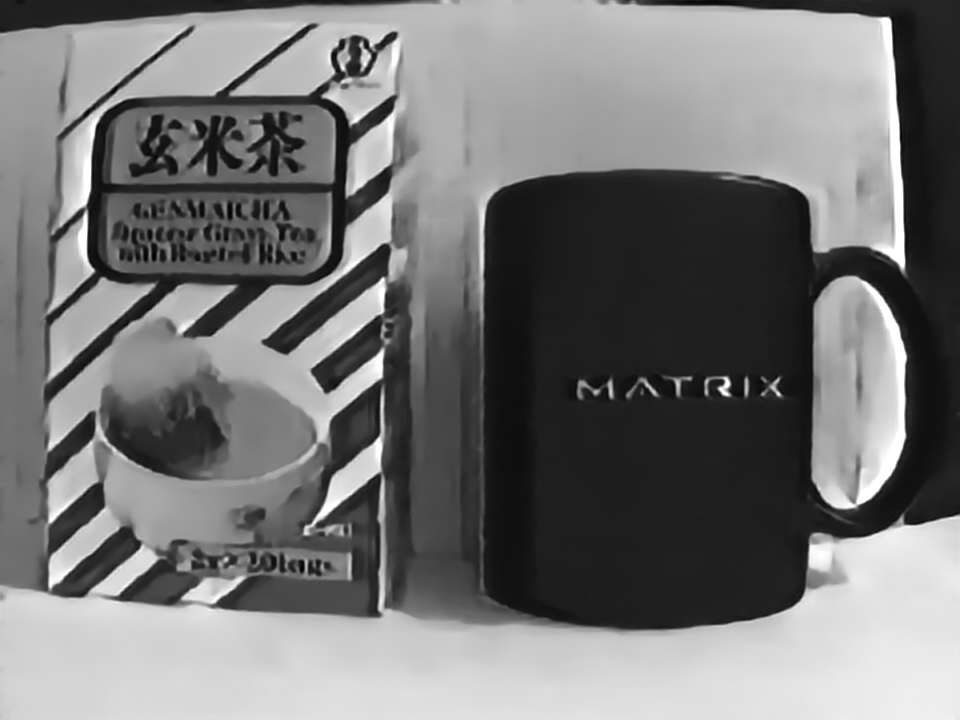}\hspace*{-0.5mm}
			\includegraphics[width=\cimwid\linewidth,trim={220px 230px 500px 100px},clip]{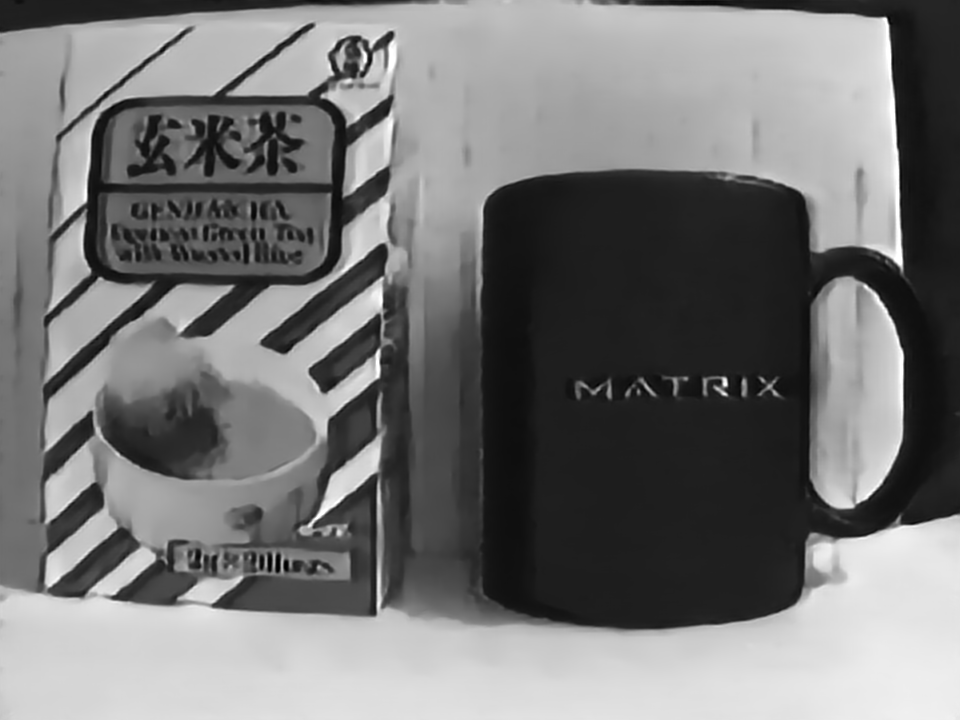}\hspace*{-0.5mm}
			\includegraphics[width=\cimwid\linewidth,trim={220px 230px 500px 100px},clip]{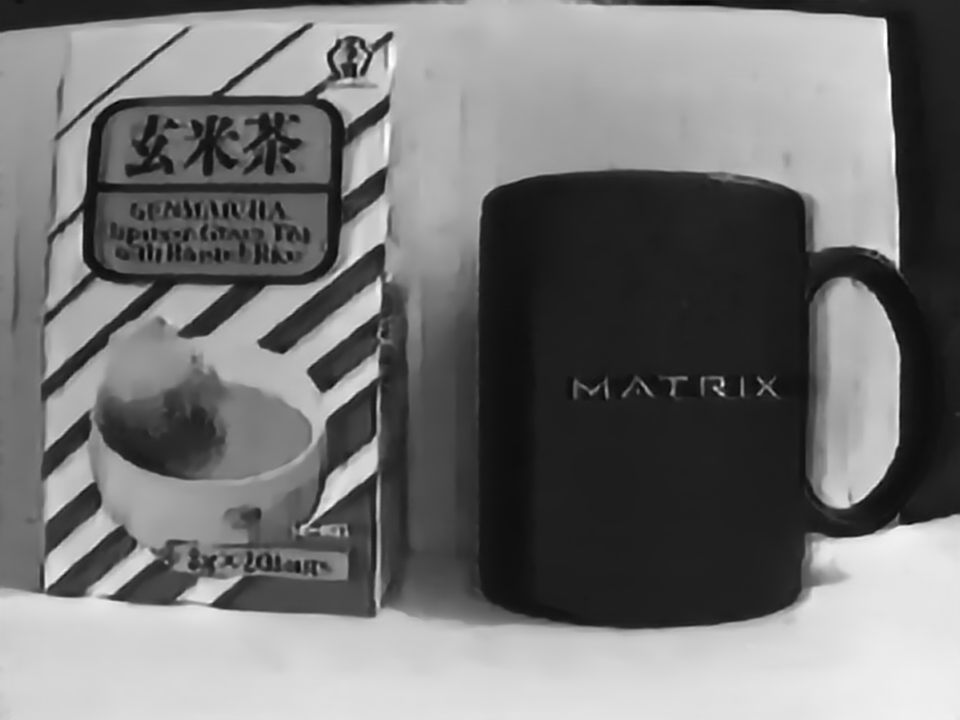}\hspace*{-0.5mm}
			\includegraphics[width=\cimwid\linewidth,trim={220px 230px 500px 100px},clip]{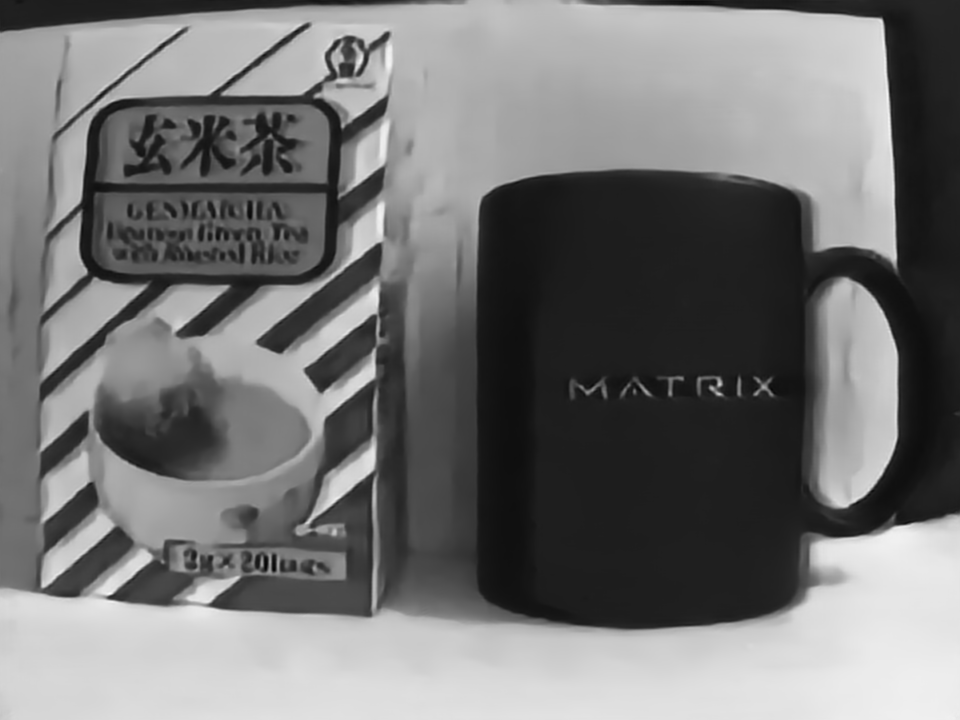}
    \vspace{.1em}\\
    % forth row
    \begin{tikzpicture}[inner sep=0]
            \node [label={[label distance=0.3cm,text depth=-1ex,rotate=90]right: {\scriptsize \textbf{DSL+RESM}}}] at (0,8.7) {};
            \end{tikzpicture}
			\includegraphics[width=\cimwid\linewidth,trim={220px 230px 500px 100px},,clip=true]{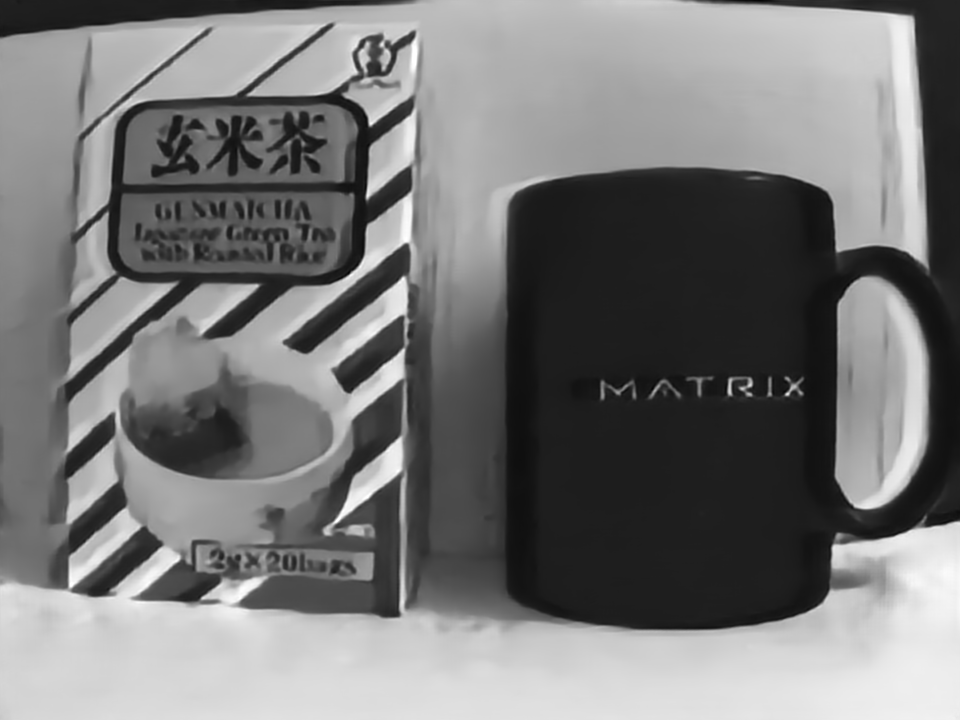}\hspace*{-0.5mm}
			\includegraphics[width=\cimwid\linewidth,trim={220px 230px 500px 100px},clip]{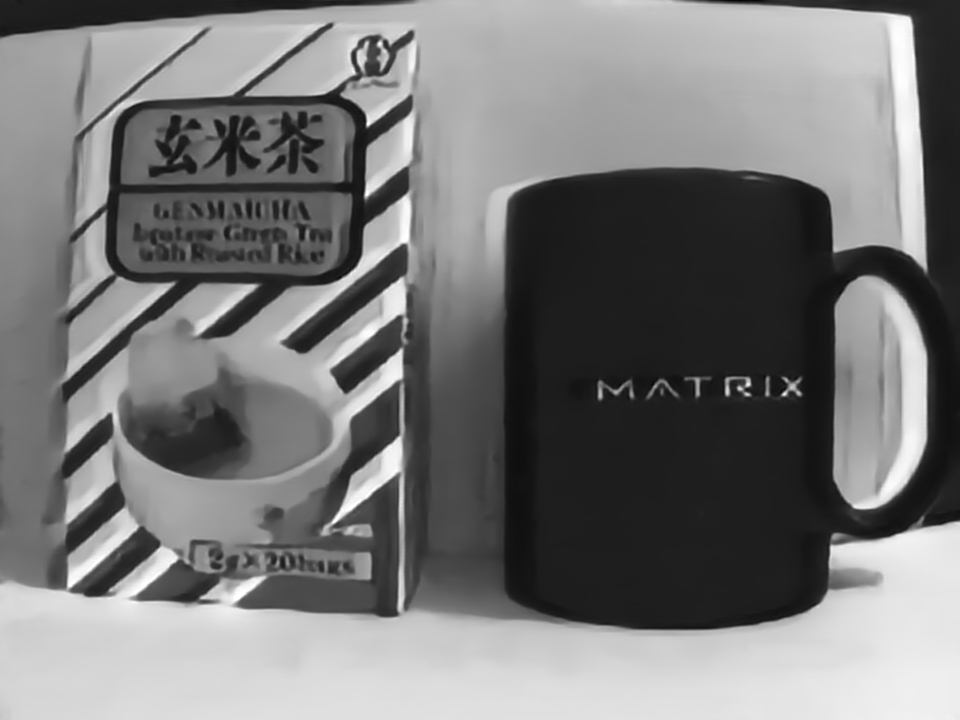}\hspace*{-0.5mm}
			\includegraphics[width=\cimwid\linewidth,trim={220px 230px 500px 100px},clip]{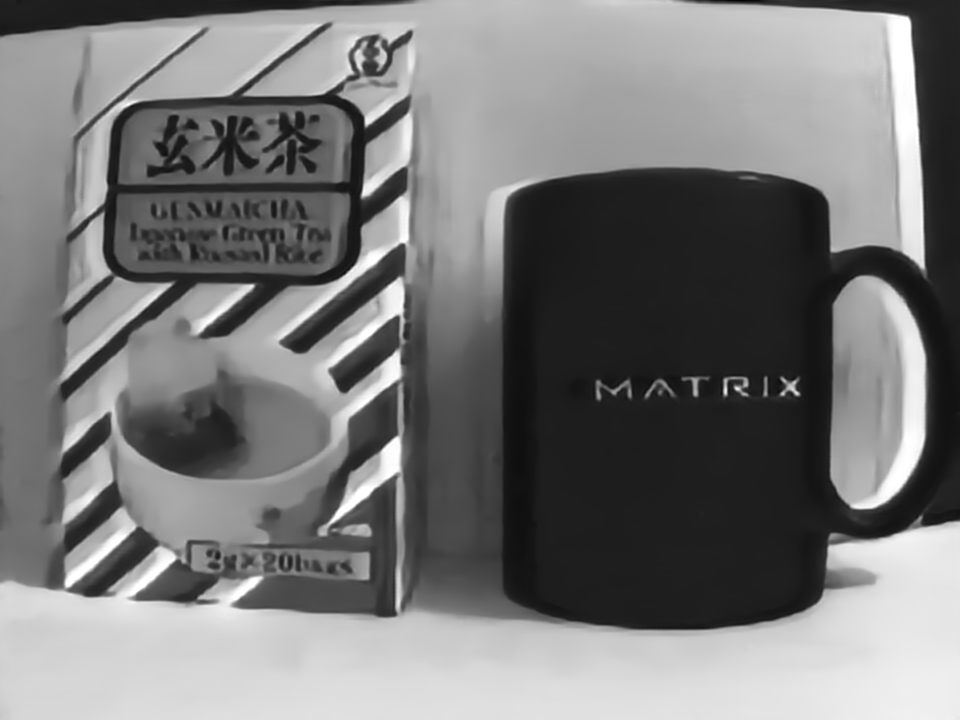}\hspace*{-0.5mm}
			\includegraphics[width=\cimwid\linewidth,trim={220px 230px 500px 100px},clip]{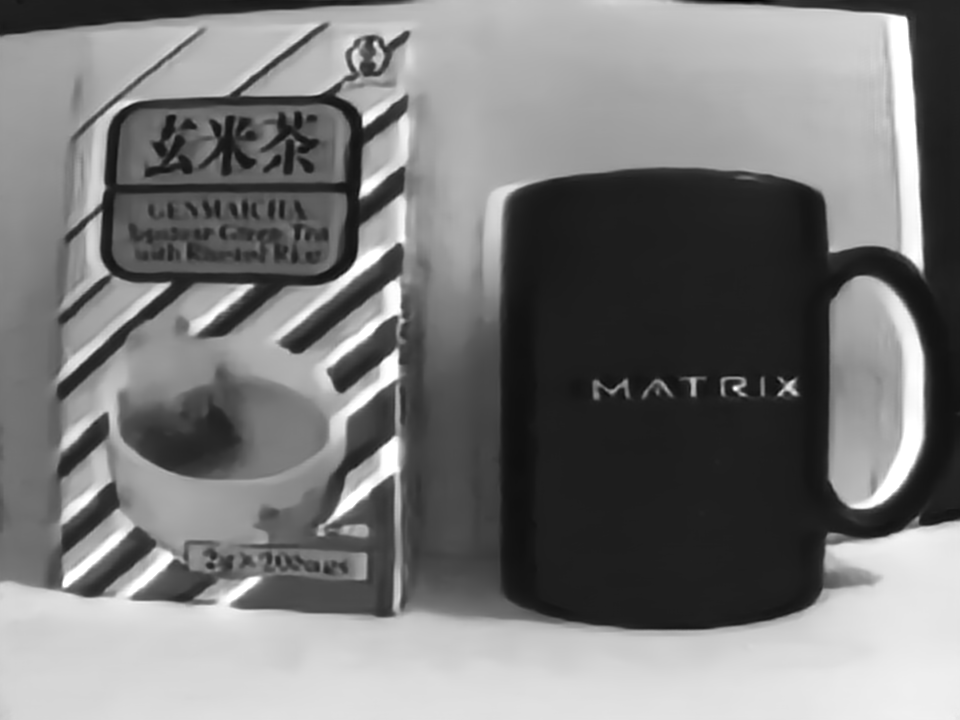}\hspace*{-0.5mm}
			\includegraphics[width=\cimwid\linewidth,trim={220px 230px 500px 100px},clip]{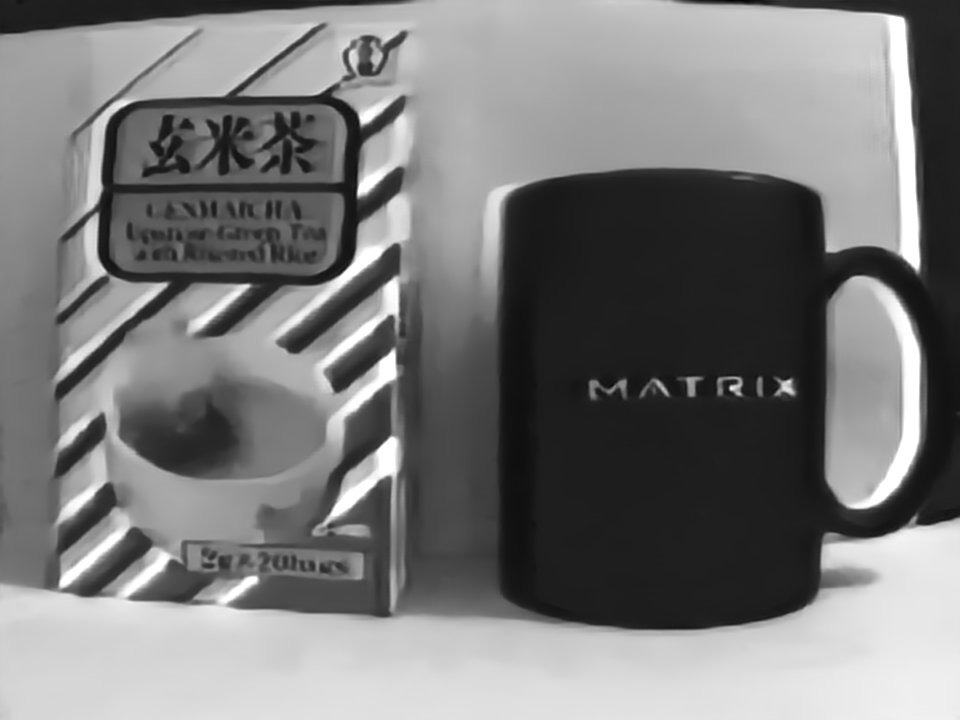}\hspace*{-0.5mm}
			\includegraphics[width=\cimwid\linewidth,trim={220px 230px 500px 100px},clip]{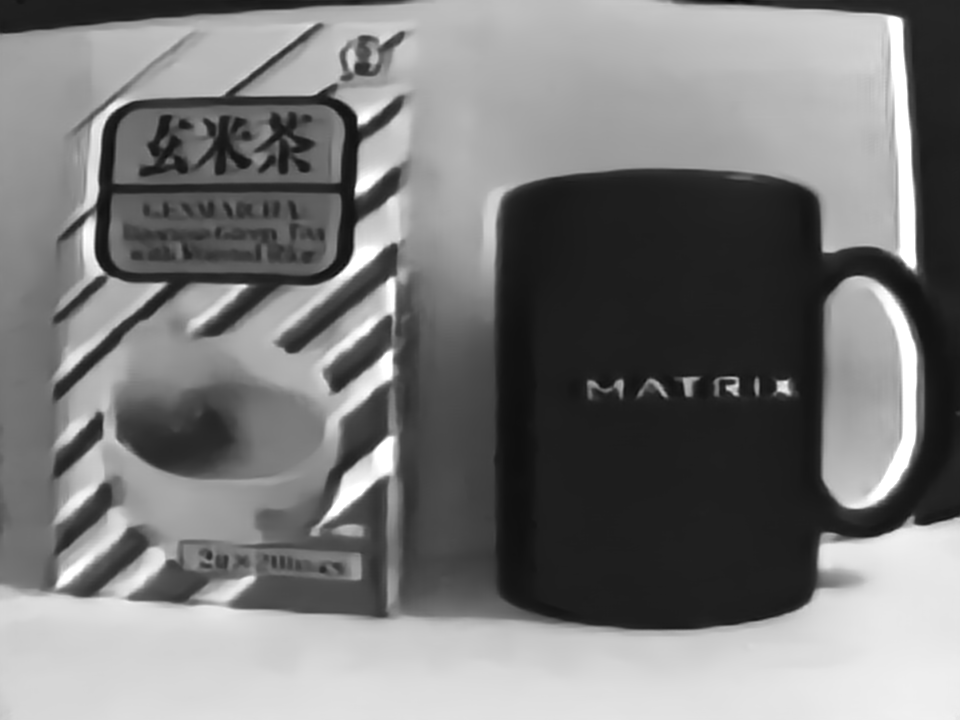}\hspace*{-0.5mm}
			\includegraphics[width=\cimwid\linewidth,trim={220px 230px 500px 100px},clip]{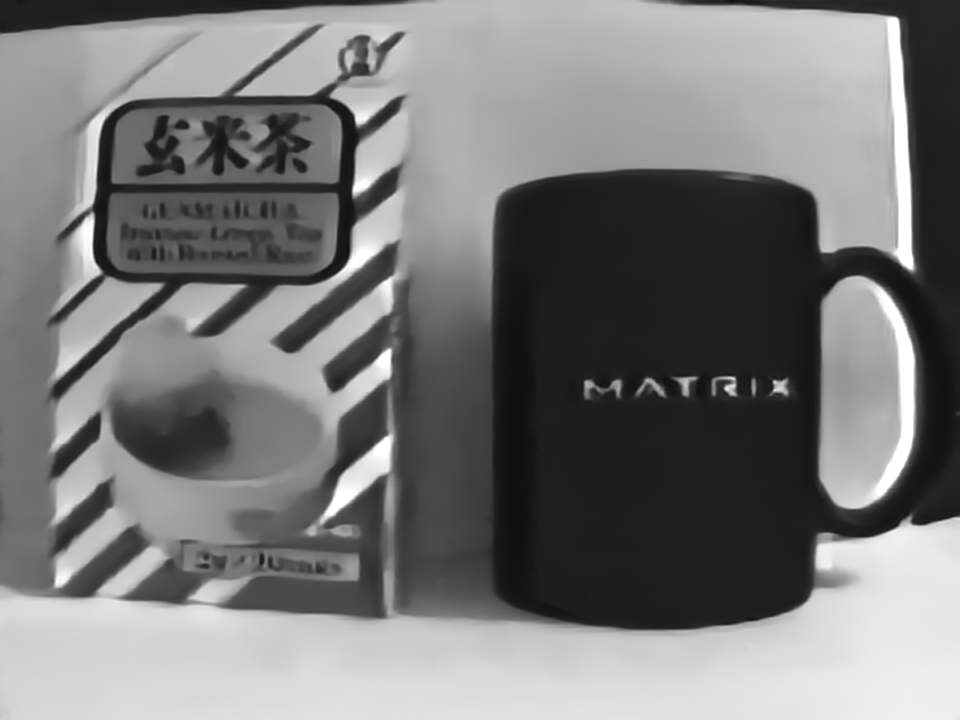}\hspace*{-0.5mm}
			\includegraphics[width=\cimwid\linewidth,trim={220px 230px 500px 100px},clip]{pic/Ablation/eSL++-w-ESM/0001_07.png}\hspace*{-0.5mm}
			\includegraphics[width=\cimwid\linewidth,trim={220px 230px 500px 100px},clip]{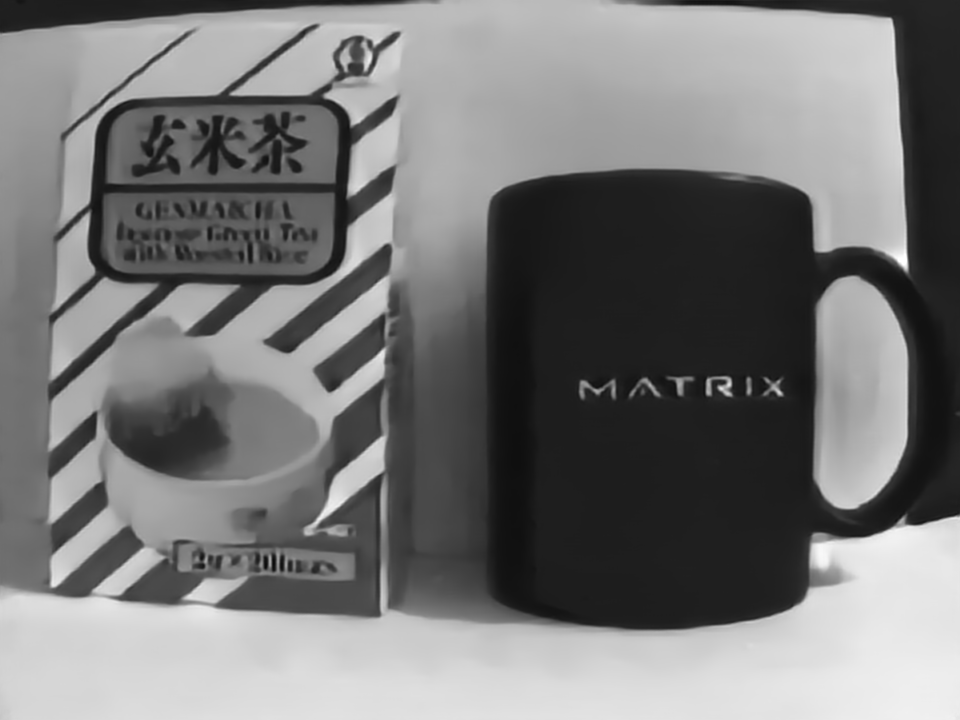}\hspace*{-0.5mm}
			\includegraphics[width=\cimwid\linewidth,trim={220px 230px 500px 100px},clip]{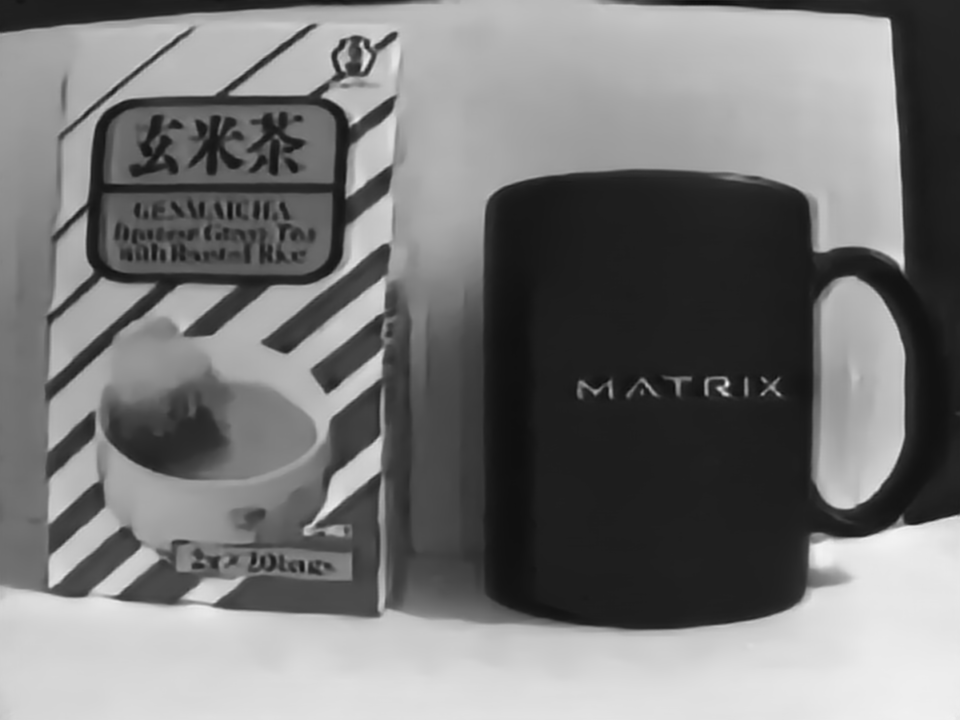}\hspace*{-0.5mm}
			\includegraphics[width=\cimwid\linewidth,trim={220px 230px 500px 100px},clip]{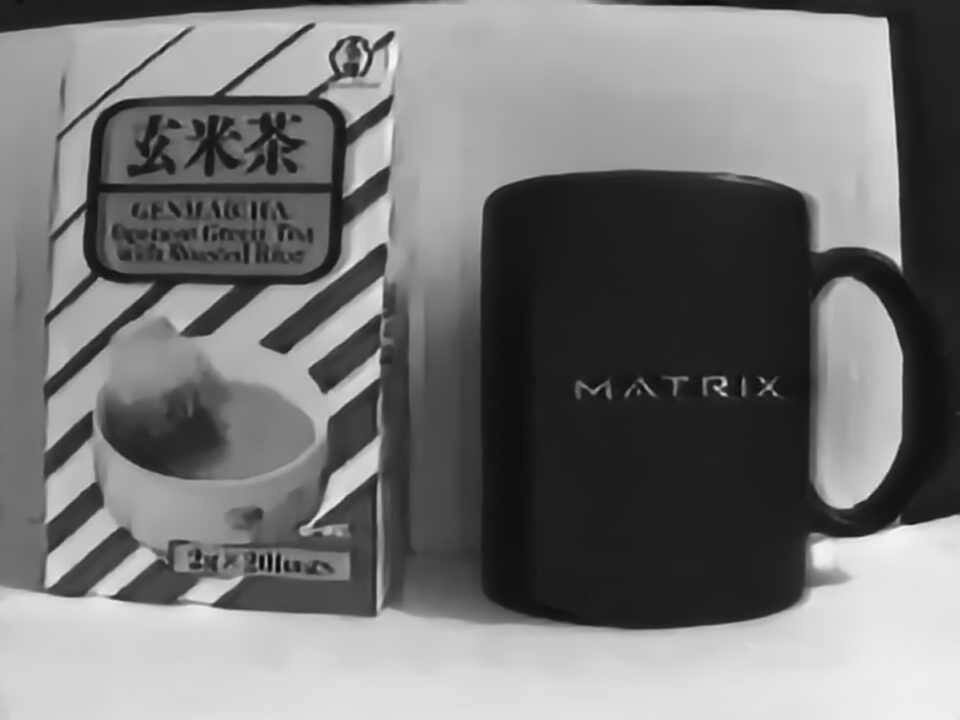}\hspace*{-0.5mm}
			\includegraphics[width=\cimwid\linewidth,trim={220px 230px 500px 100px},clip]{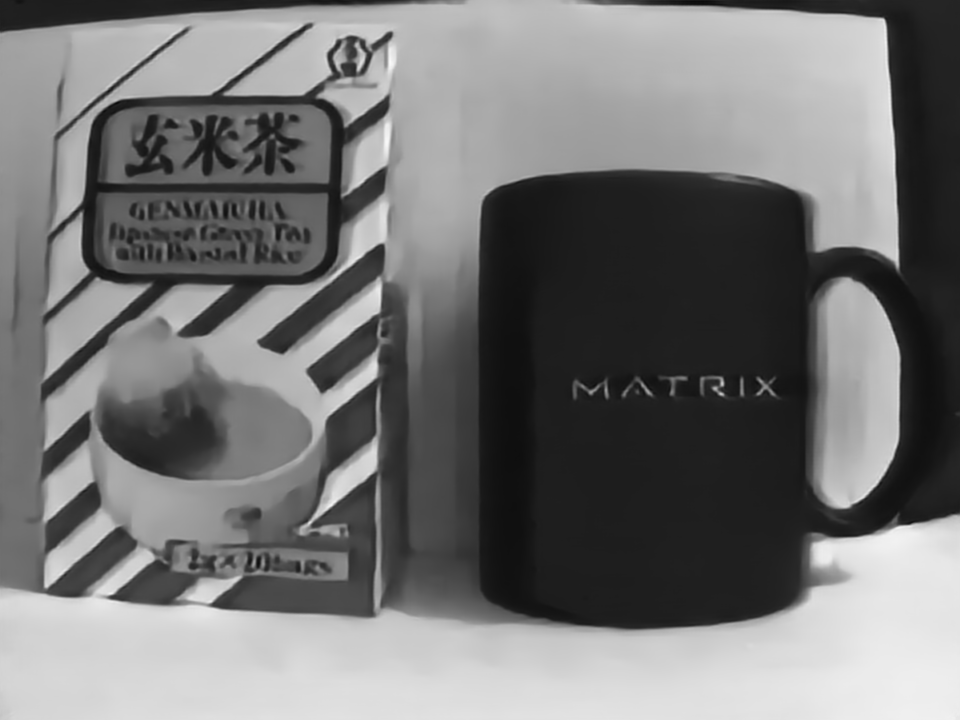}\hspace*{-0.5mm}
			\includegraphics[width=\cimwid\linewidth,trim={220px 230px 500px 100px},clip]{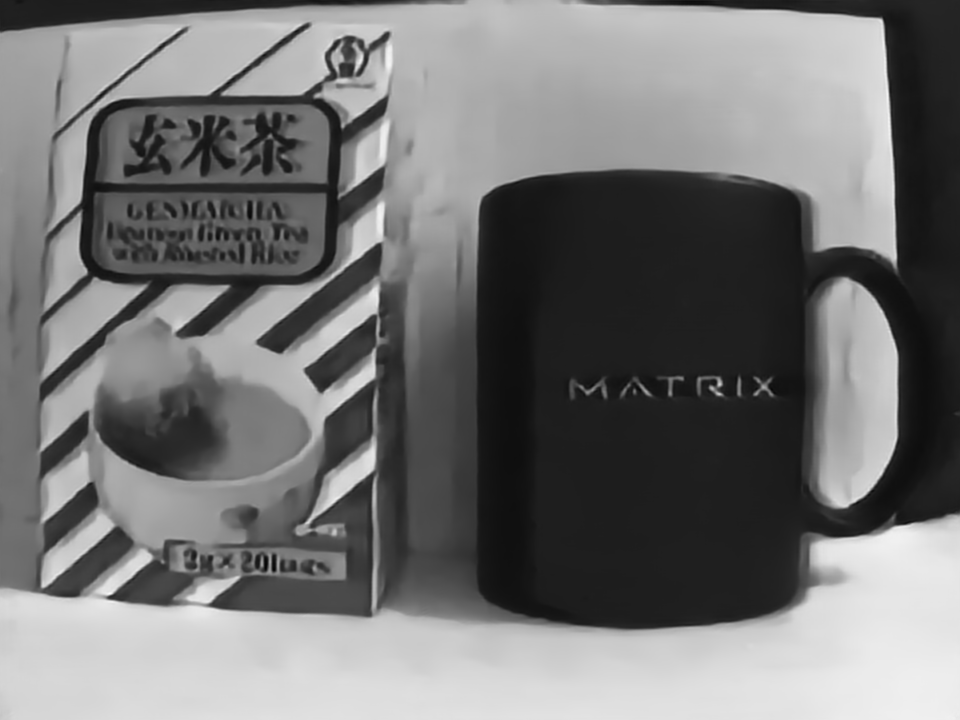}
    \vspace{.1em}\\
	\caption{\colored{Qualitative results of single frame reconstructions (1st row) and their corresponding sequence reconstructions (2nd-5th rows) on the RWS dataset, where Baseline indicates eSL-Net without DSL and RESM.}}
	\label{fig:exp-ablation}
\end{figure*}
\subsection{Ablation Study}

Different from the preliminary version~\cite{wang2020event}, \ie, eSL-Net, eSL-Net++ further contains the {\it D}ual {\it S}parse {\it L}earning (DSL) scheme to take into account event noises and a {\it R}igorous {\it E}vent {\it S}huffle-and-{\it M}erge (RESM) scheme for sequence reconstruction. In this section, ablation studies on these two schemes, \ie, DSL and RESM, are analyzed with the sequence reconstruction on the GoPro dataset with synthetic events and the HQF dataset with real events. Four different experiments are implemented to analyze the effectiveness of DSL and RESM, as shown in Tab.~\ref{ablation}. 

\noindent\textbf{Importance of DSL.} 
Since events $\boldsymbol{\mathcal{E}}_{\mathcal{T}}$ and image frames $\boldsymbol{I}$ are coupled together in a multiplicative manner as \eqref{general_model}, noises in events $\boldsymbol{\mathcal{E}}_{\mathcal{T}}$ would inevitably affect the final SRB results, as shown in Fig.~\ref{input-event-frame}. Without DSL, eSL-Net directly utilizes the output of the LDI module which suffers from serious noises, as shown in Fig.~\ref{input-event-frame} (b). And accordingly, these noises would be aggregated and degrade the final SRB performance, as shown in Fig.~\ref{input-event-frame} (f). On the other hand, the DSL scheme utilized in eSL-Net++ alleviates the burden of event noises by leveraging the smoothness of images (Fig.~\ref{input-event-frame} (h)), finally leading to better SRB reconstructions than eSL-Net (Fig.~\ref{input-event-frame} (f)). We further compare eSL-Net++ to eSL-Net without event noise (\ie, Fig.~\ref{input-event-frame} (c)(g)) which outputs less noisy $\boldsymbol{E}$ and $\boldsymbol{X}$ than eSL-Net with event noise (\ie, Fig.~\ref{input-event-frame} (b)(f)). Nevertheless, eSL-Net++ with noisy events still achieves the best performance, which validates the effectiveness of the DSL module.

Quantitative ablations of the DSL scheme are shown in Tab.~\ref{ablation}. Specifically, we respectively implement two types of experiments, \ie, with or without the RESM scheme. For methods without the RESM scheme (1st and 3rd rows in Tab.~\ref{ablation}), the DSL scheme brings remarkable improvements in terms of PSNR and SSIM on both GoPro and HQF datasets, \ie, $0.32/0.0122$ on the GoPro dataset and $0.27/0.0108$ on the HQF dataset. When with the RESM scheme (2nd and 4th rows in Tab.~\ref{ablation}), the DSL scheme can further boost the SRB accuracy with improvements of $0.27/0.0072$ on the GoPro dataset and $0.13/0.0023$ on the HQF dataset.

\noindent\textbf{\colored{Importance of Rigorous ESM (RESM).}} The eSL-Nets are originally trained for the task of single-frame SRB. To extend eSL-Nets to sequence SRB without additional training process, an event shuffle scheme has been proposed according to the chronological order of events and image frames in \cite{wang2020event}. Instead, we further  analyze the theoretical relationship between single-frame E-SRB and sequence-frame E-SRB and accordingly propose a rigorous ESM method as \eqref{eq:ldief}, extending eSL-Net++ to the task of sequence SRB. To validate its effectiveness, two groups of experiments are done with different settings, \ie, with or without DSL, and the quantitative results are given in Tab.~\ref{ablation}. It shows that the RESM scheme brings notable improvements on both experimental settings, \ie, up to $0.7$ dB in PSNR and $0.01$ in SSIM without the DSL scheme, and $0.57$ dB in PSNR and $0.0023$ in SSIM with the DSL scheme. 

\colored{To further illustrate the effectiveness of DSL and RESM, we conduct qualitative comparisons of the sequence restorations respectively by eSL-Net (Baseline), eSL-Net+RESM (RESM), eSL-Net+DSL (DSL), and eSL-Net++ (DSL+RESM). The results on the RWS dataset are shown in Fig.~\ref{fig:exp-ablation}.}
    \begin{itemize}
        \item \colored{The improvement from DSL can be found by comparing DSL to Baseline. Clearly, the restorations of DSL are less noisy than those of Baseline. Similarly, DSL+RESM gives less noisy restorations than RESM due to the usage of the DSL module.}
        \item \colored{The inaccurate ESM scheme used in eSL-Net often leads to halo effects on high-contrast edges for sequence restoration, as shown in the intermediate frames restored by Baseline and DSL, while these halo effects can be effectively suppressed by the RESM scheme used in eSL-Net++, as shown in the intermediate frames restored by RESM and DSL+RESM.}
    \end{itemize}
 \colored{Either the DSL scheme or the RESM scheme can improve the SRB performance. Specifically, DSL can effectively reduce performance degradation caused by noises, while RESM can avoid halo effects brought by the inaccurate ESM in \cite{wang2020event} and thus improves the accuracy of sequence restoration without additional network training process. Thus, eSL-Net++ achieves the best performance benefiting from these two schemes.}
% The qualitative ablation of DSL and RESM on sequence-frame restoration can be found in Fig.~\ref{fig:exp-ablation}. 

% In summary, either the DSL scheme or the ESM scheme can improve the SRB accuracy. Thus, eSL-Net++ achieves the best performance benefiting from these two schemes.

\begin{figure*}[!htb]
    \centering
    \begin{minipage}[b]{0.48\textwidth}
	\includegraphics[width=0.48\linewidth]{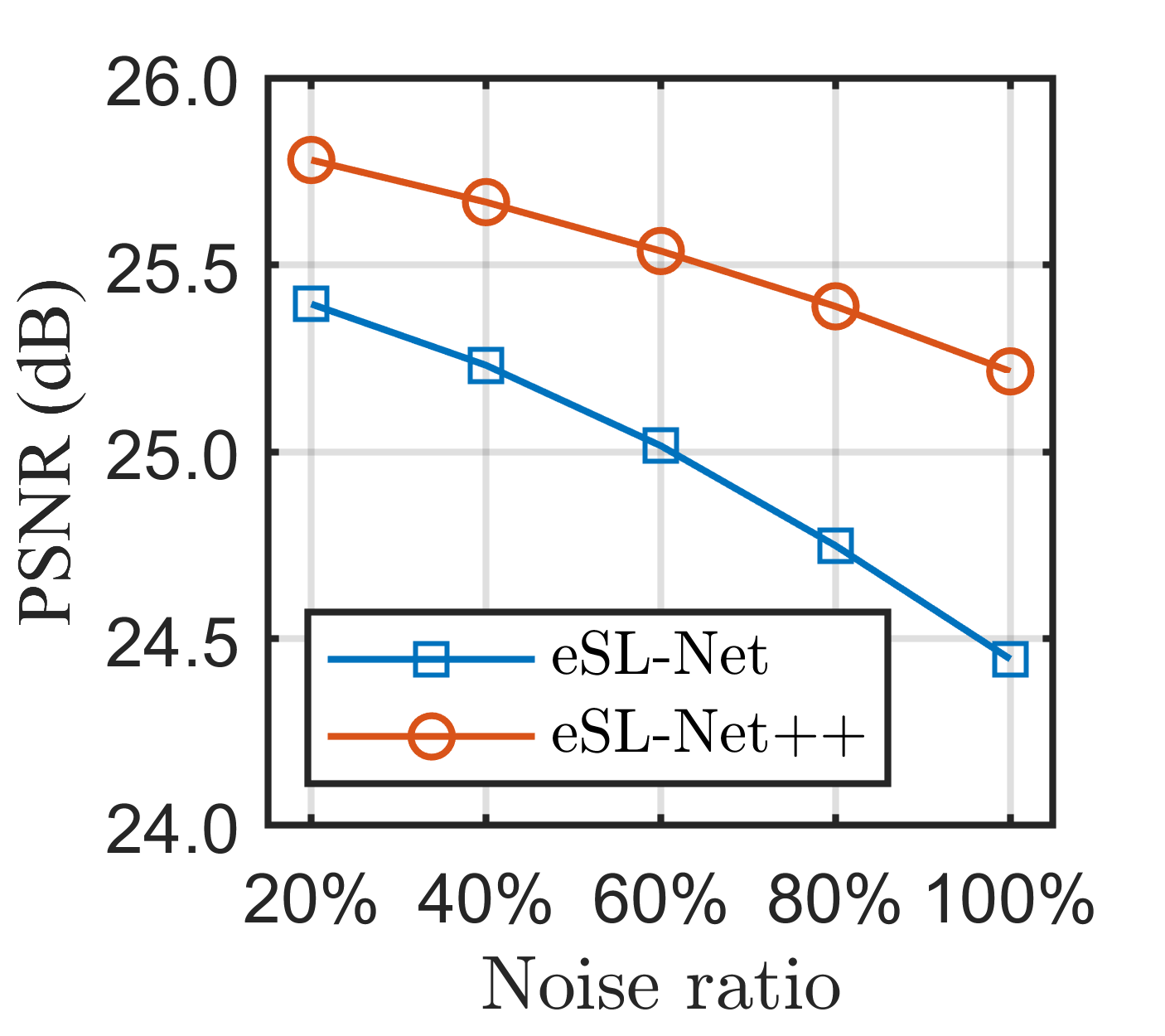}
    \includegraphics[width=0.48\linewidth]{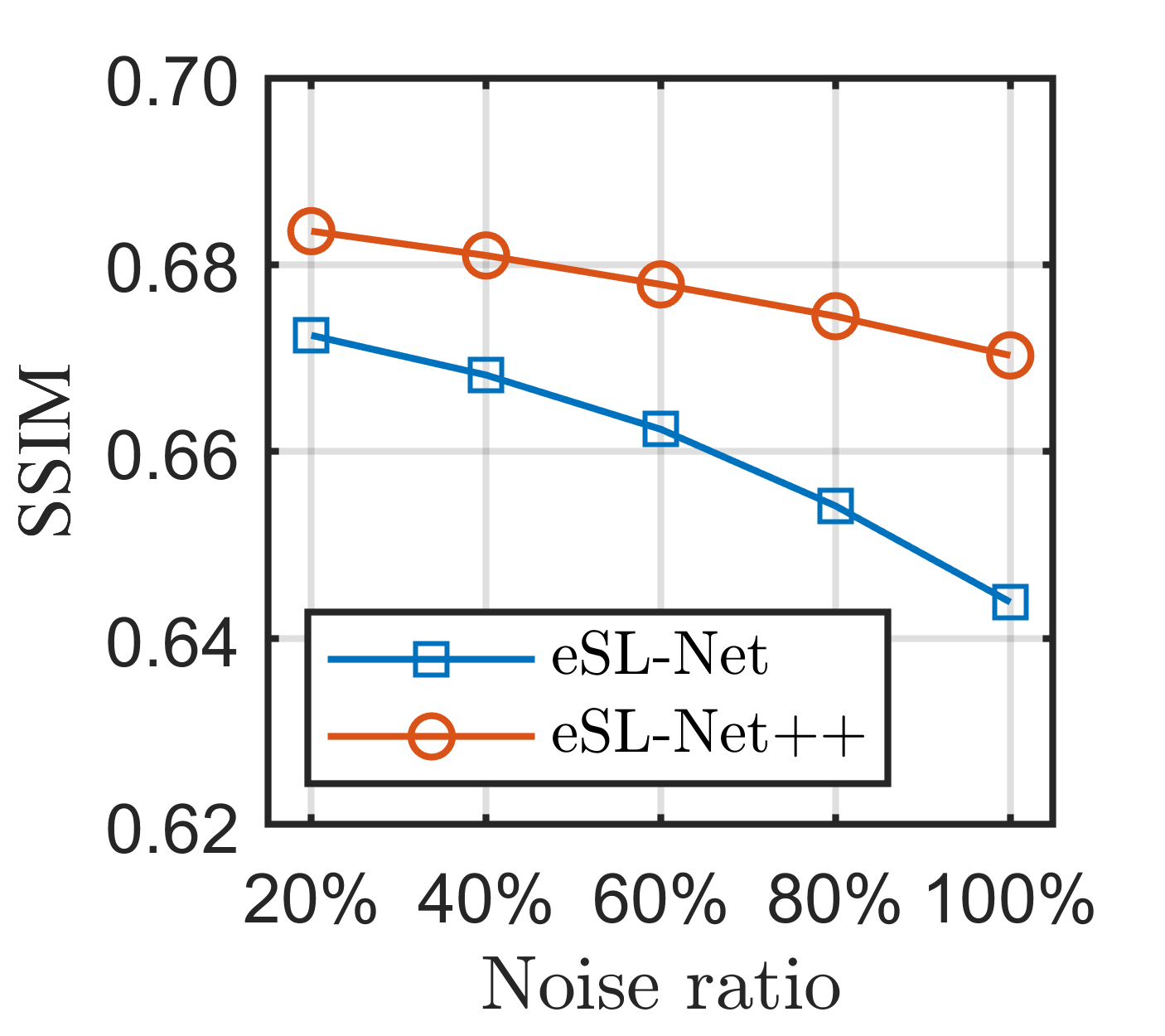}
	    \subcaption{\rmfamily \fontsize{8pt}{0} Results under different event noise}
	    \label{fig:event-noise}
    \end{minipage}
    \begin{minipage}[b]{0.48\textwidth}
	\includegraphics[width=0.48\linewidth]{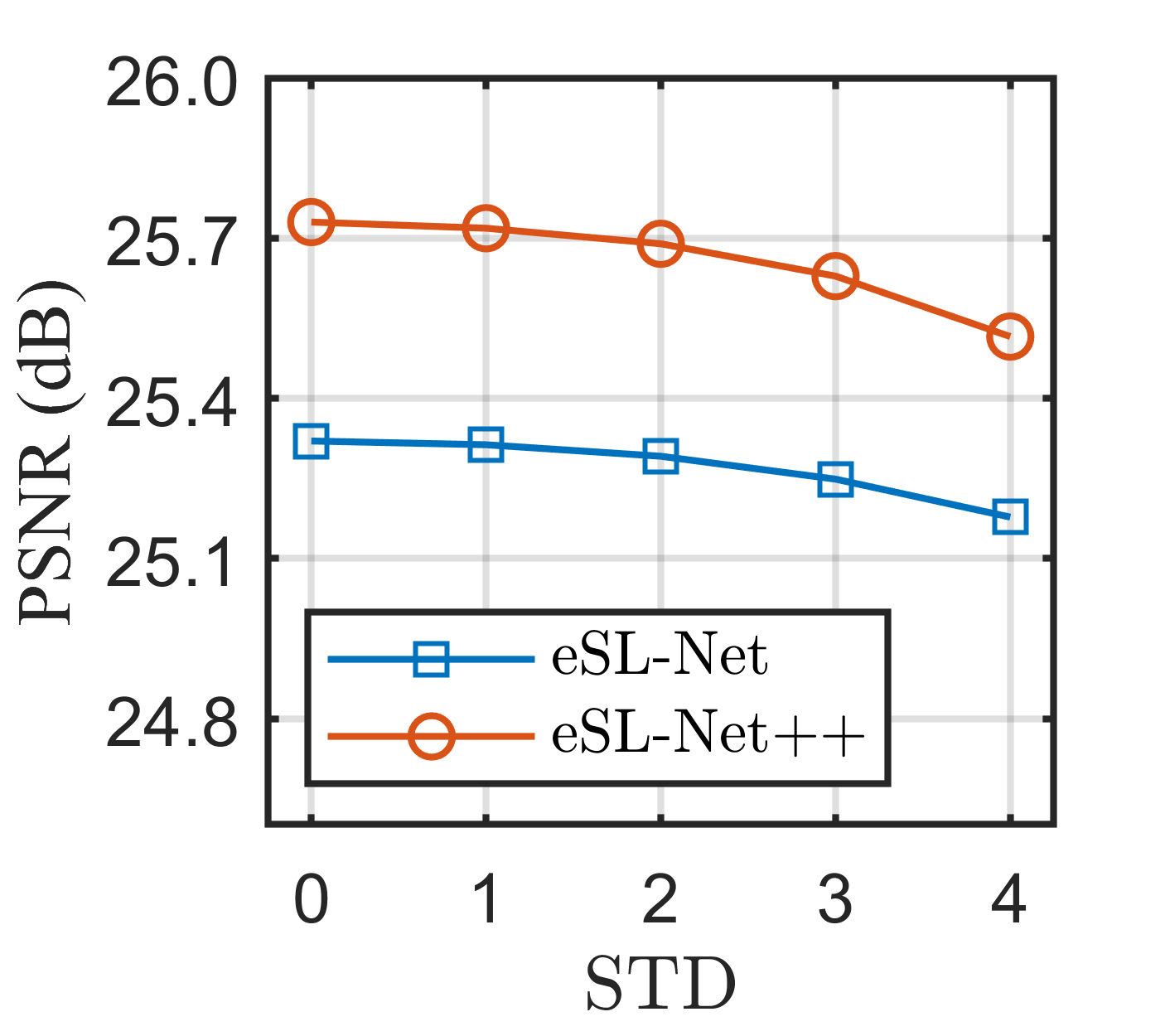}
    \includegraphics[width=0.48\linewidth]{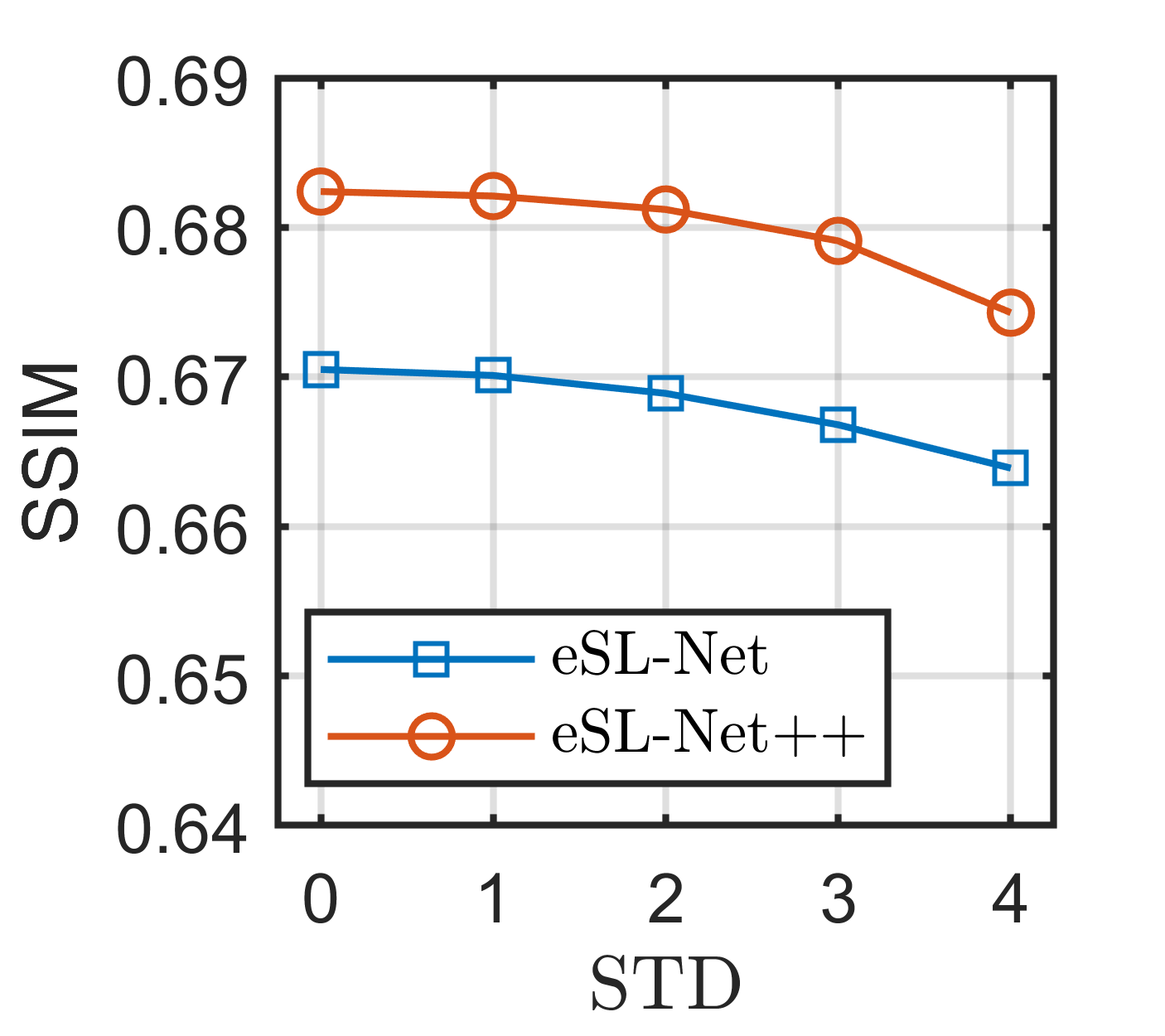}
	    \subcaption{\rmfamily \fontsize{8pt}{0} Results under different image noise}
	    \label{fig:image-noise}
    \end{minipage}
    \caption{\colored{Quantitative comparisons of single frame reconstruction  under different levels of event and image noise on the GoPro dataset. (a) Results under different event noise, where the noise ratio indicates the ratio of noise events to the original events. (b) Results under different image noise, where STD represents the standard deviation of the white Gaussian noise added to images.}}
    \label{fig:noise-exp}
\end{figure*}

\def\imgWidth{0.235\textwidth} %子图大小
\def\sccone{(-2.1,-1.)} % 文字位置
\def\ssxxsone{(0,0)} % 绿色小方框位置
\def\ssyysone{(2.1, 0.5)} % 绿色大方框位置

\def\scctwo{(-2,-1.35)} % 文字位置
\def\ssxxstwo{(1.9,0.1)} % 绿色小方框位置
\def\ssyystwo{(-0.88, 0.5)} % 绿色大方框位置
\def\ssizz{1.2cm} %框大小
\def\ssmag{3}

\begin{figure*}[!htb] 
\centering
\tikzstyle{img} = [rectangle, minimum width=\imgWidth, draw=white]
    %% first row 
    \begin{minipage}[b]{\textwidth}
        \begin{tikzpicture}[spy using outlines={green,magnification=\ssmag,size=\ssizz},inner sep=0]
            \node [align=center, img] {\includegraphics[width=\imgWidth]{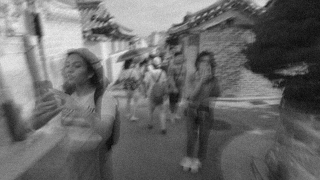}};
            \spy on \ssxxsone in node [left] at \ssyysone;
            \node [anchor=west] at \sccone {\textcolor{white}{\footnotesize \bf Blurry Image}};
    	\end{tikzpicture}
		\begin{tikzpicture}[spy using outlines={green,magnification=\ssmag,size=\ssizz},inner sep=0]
            \node [align=center, img] {\includegraphics[width=\imgWidth]{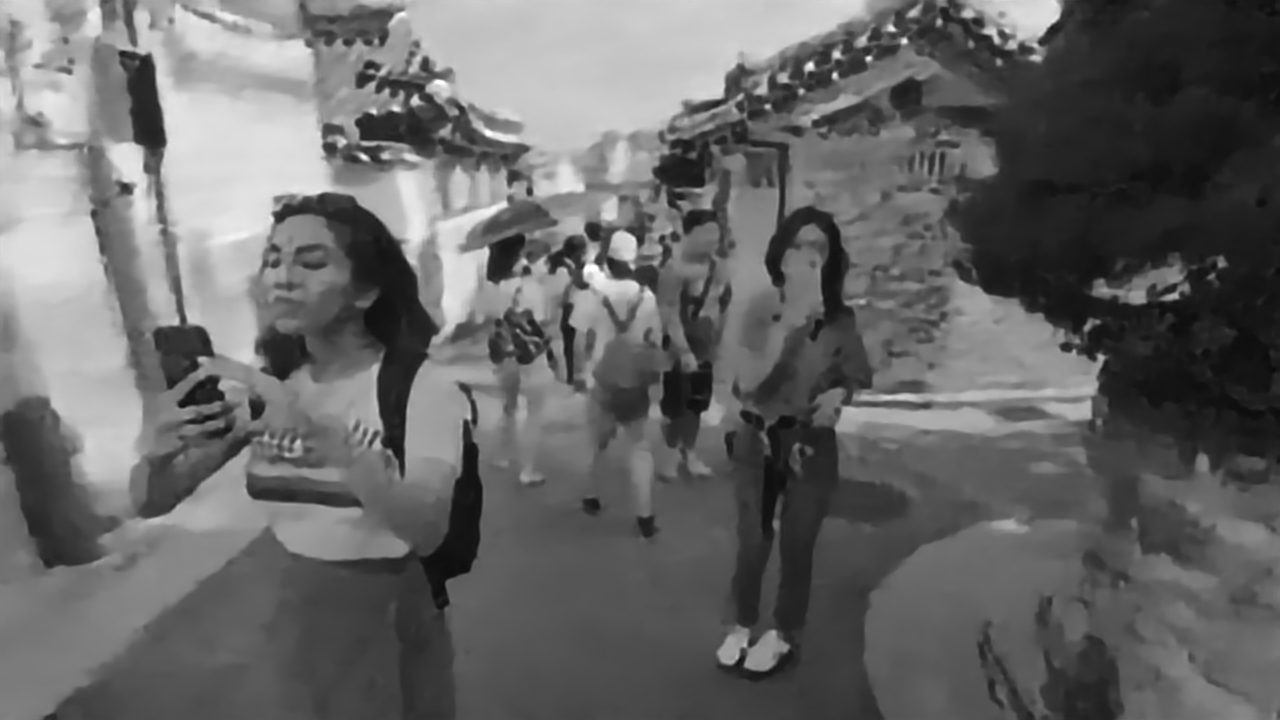}};
            \spy on \ssxxsone in node [left] at \ssyysone;
            \node [anchor=west] at \sccone {\textcolor{white}{\footnotesize \bf eSL-Net / 20\%}};
    	\end{tikzpicture}
        \begin{tikzpicture}[spy using outlines={green,magnification=\ssmag,size=\ssizz},inner sep=0]
            \node [align=center, img] {\includegraphics[width=\imgWidth]{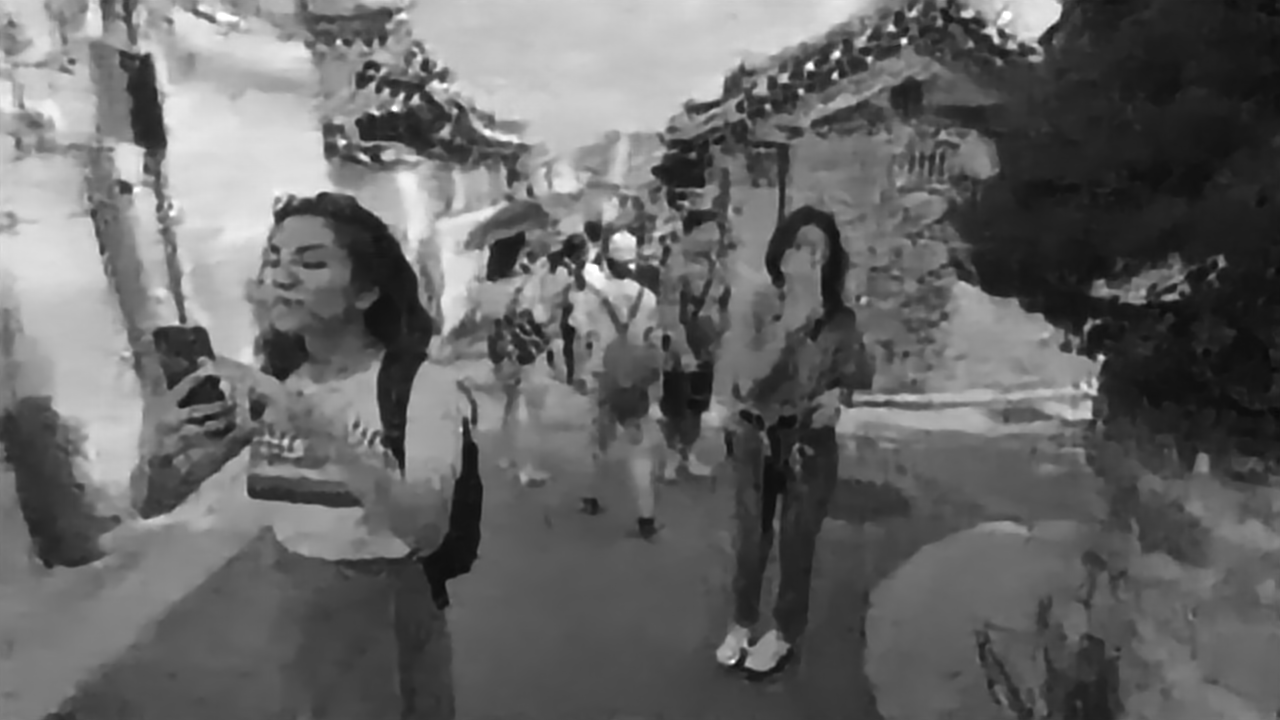}};
            \spy on \ssxxsone in node [left] at \ssyysone;
            \node [anchor=west] at \sccone {\textcolor{white}{\footnotesize \bf eSL-Net / 60\%}};
    	\end{tikzpicture}
		\begin{tikzpicture}[spy using outlines={green,magnification=\ssmag,size=\ssizz},inner sep=0]
            \node [align=center, img] {\includegraphics[width=\imgWidth]{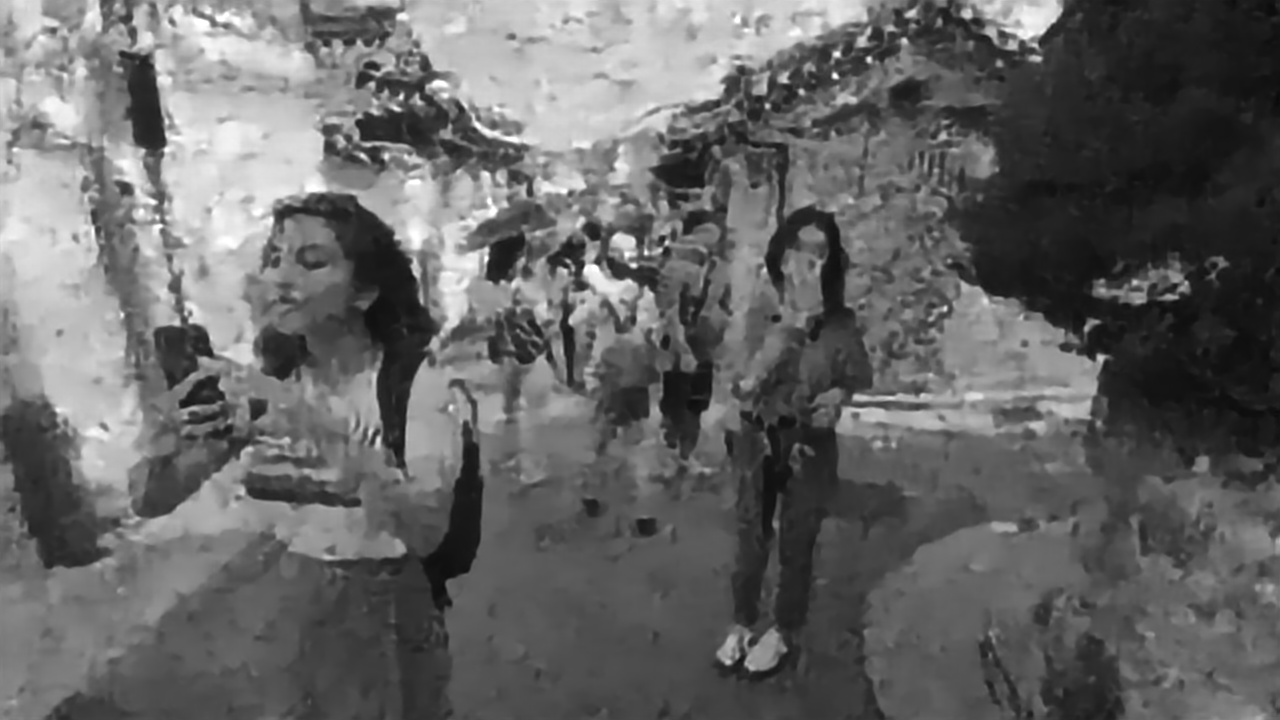}};
            \spy on \ssxxsone in node [left] at \ssyysone;
            \node [anchor=west] at \sccone {\textcolor{white}{\footnotesize \bf eSL-Net / 100\%}};
    	\end{tikzpicture}
\\
    %% second row
		\begin{tikzpicture}[spy using outlines={green,magnification=\ssmag,size=\ssizz},inner sep=0]
            \node [align=center, img] {\includegraphics[width=\imgWidth]{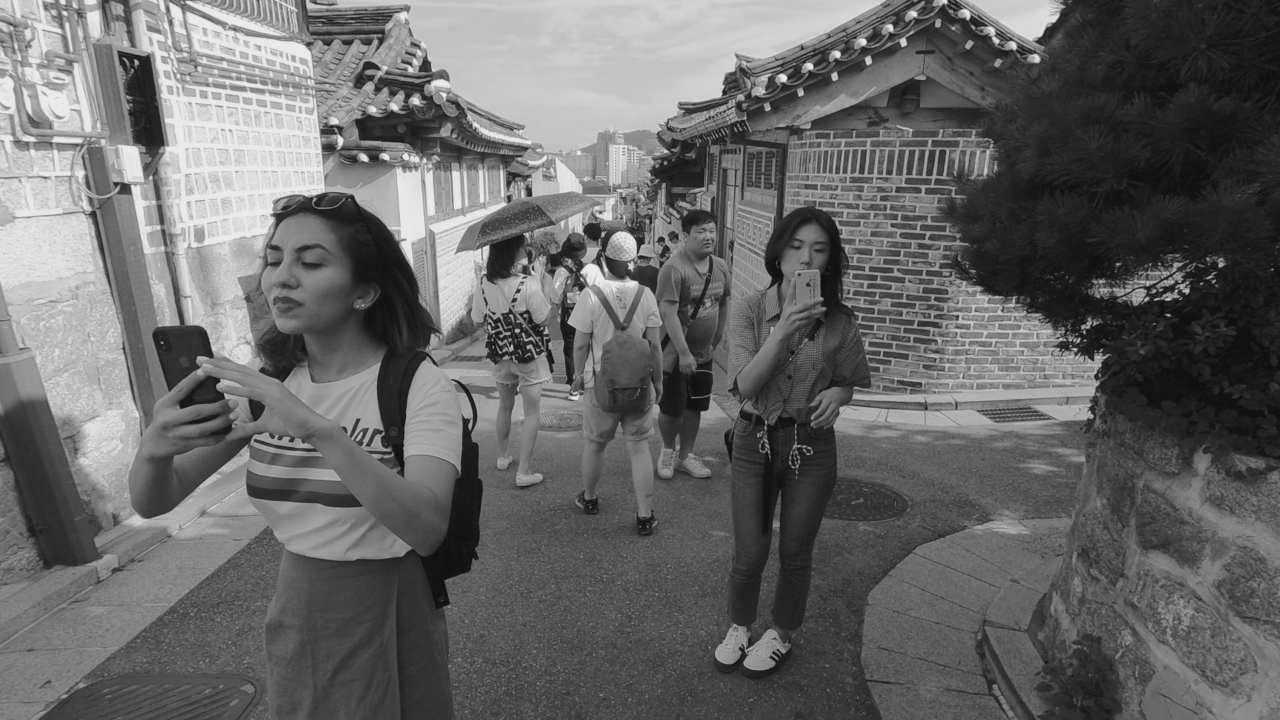}};
            \spy on \ssxxsone in node [left] at \ssyysone;
            \node [anchor=west] at \sccone {\textcolor{white}{\footnotesize \bf Ground Truth}};
    	\end{tikzpicture}
		\begin{tikzpicture}[spy using outlines={green,magnification=\ssmag,size=\ssizz},inner sep=0]
            \node [align=center, img] {\includegraphics[width=\imgWidth]{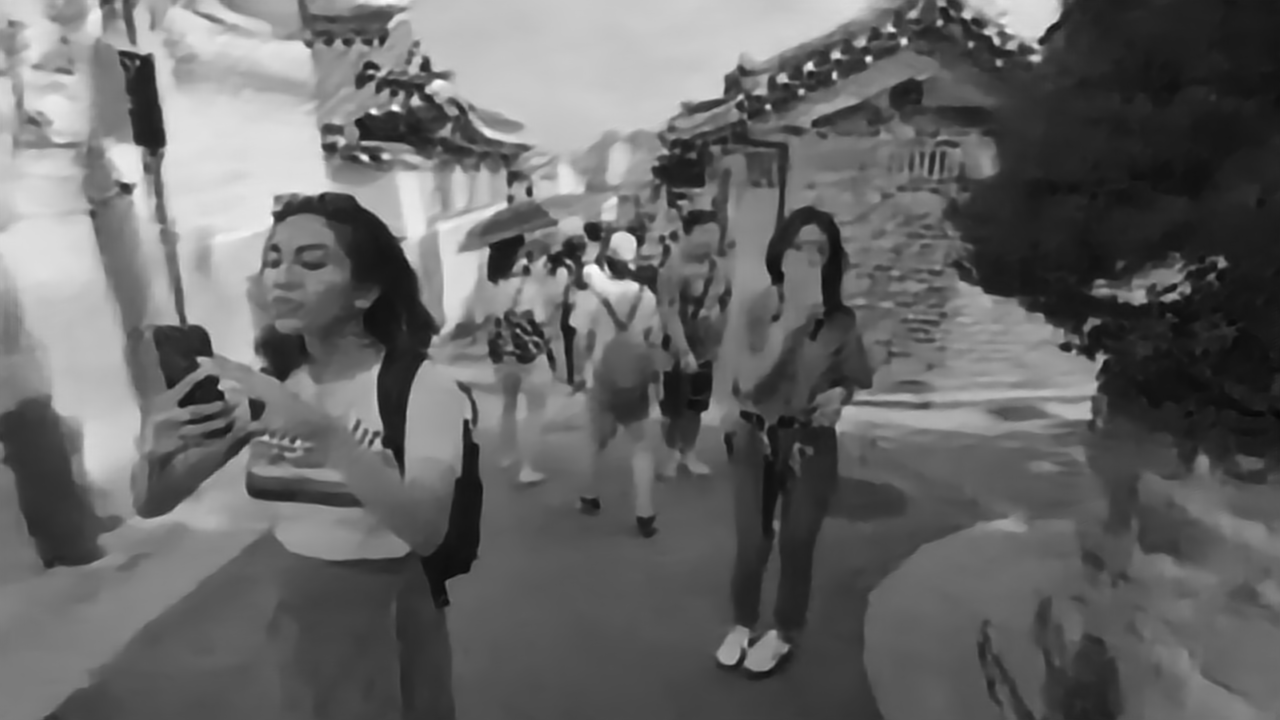}};
            \spy on \ssxxsone in node [left] at \ssyysone;
            \node [anchor=west] at \sccone {\textcolor{white}{\footnotesize \bf eSL-Net++ / 20\%}};
    	\end{tikzpicture}
		\begin{tikzpicture}[spy using outlines={green,magnification=\ssmag,size=\ssizz},inner sep=0]
           \node [align=center, img] {\includegraphics[width=\imgWidth]{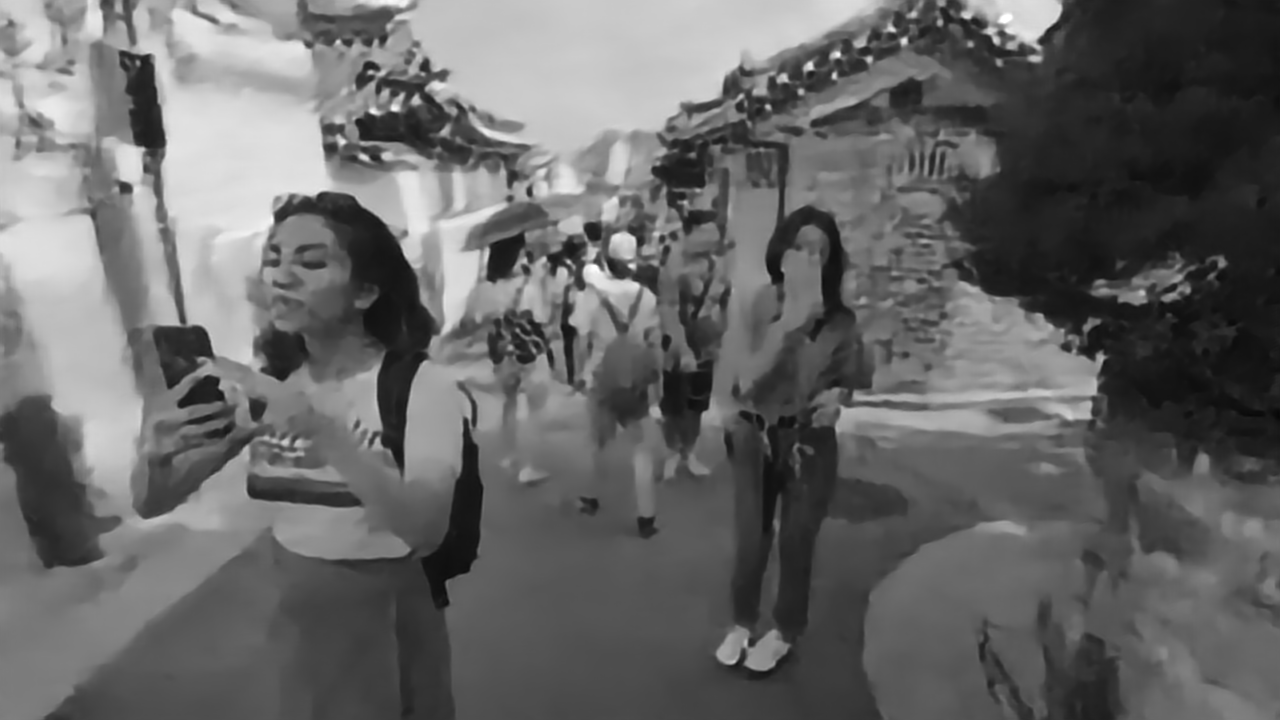}};
           \spy on \ssxxsone in node [left] at \ssyysone;
            \node [anchor=west] at \sccone {\textcolor{white}{\footnotesize \bf eSL-Net++ / 60\%}};
    	\end{tikzpicture}
		\begin{tikzpicture}[spy using outlines={green,magnification=\ssmag,size=\ssizz},inner sep=0]
            \node [align=center, img] {\includegraphics[width=\imgWidth]{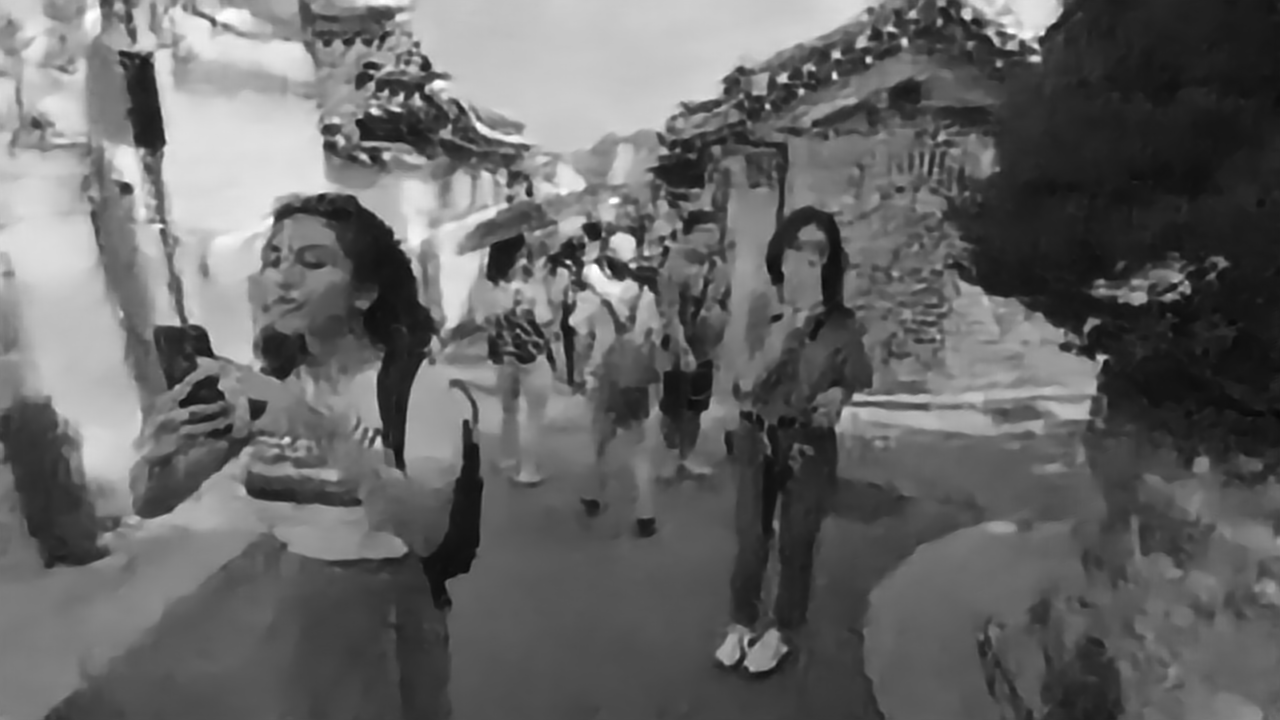}};
            \spy on \ssxxsone in node [left] at \ssyysone;
            \node [anchor=west] at \sccone {\textcolor{white}{\footnotesize \bf eSL-Net++ / 100\%}};
    	\end{tikzpicture}
    	\subcaption{Results under different event noise}
    % third row
    \end{minipage}
    \begin{minipage}[b]{\textwidth}
        \begin{tikzpicture}[spy using outlines={green,magnification=\ssmag,size=\ssizz},inner sep=0]
            \node [align=center, img] {\includegraphics[width=\imgWidth]{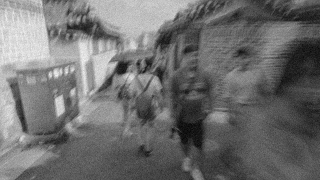}};
            \spy on \ssxxstwo in node [left] at \ssyystwo;
            \node [anchor=west] at \sccone {\textcolor{white}{\footnotesize \bf Blurry Image}};
    	\end{tikzpicture}
		\begin{tikzpicture}[spy using outlines={green,magnification=\ssmag,size=\ssizz},inner sep=0]
            \node [align=center, img] {\includegraphics[width=\imgWidth]{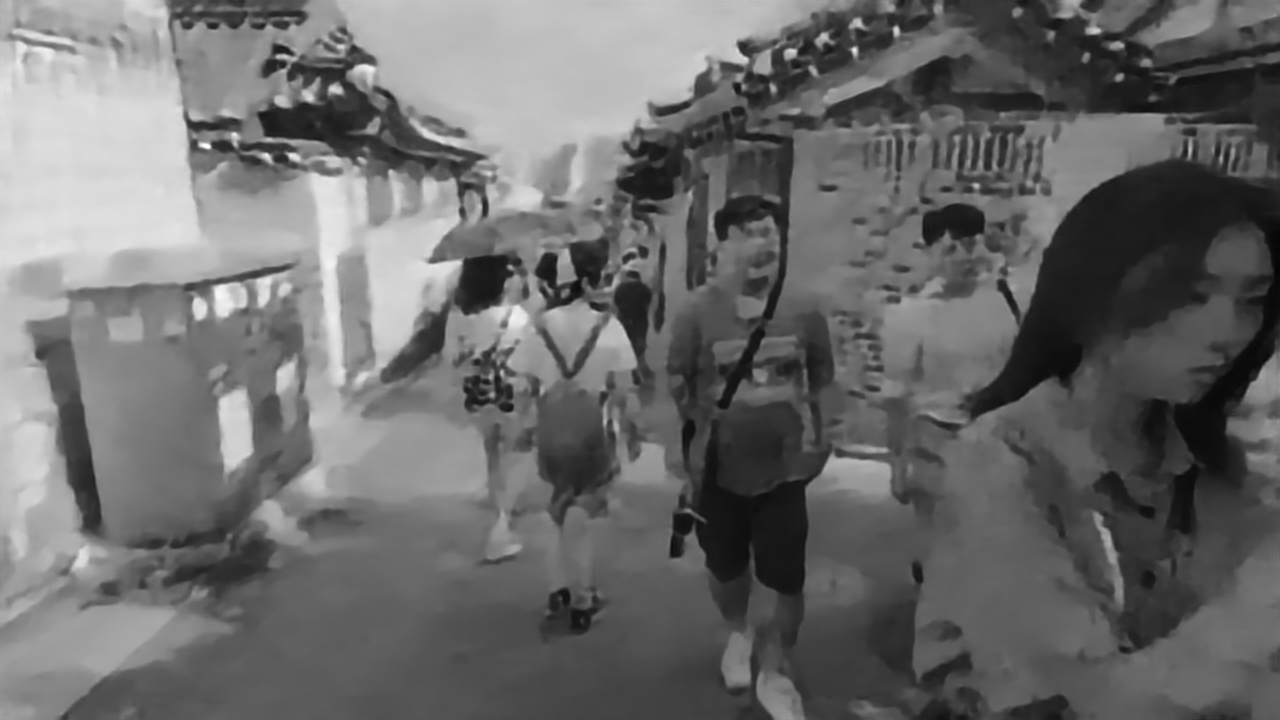}};
            \spy on \ssxxstwo in node [left] at \ssyystwo;
            \node [anchor=west] at \sccone {\textcolor{white}{\footnotesize \bf eSL-Net / 0}};
    	\end{tikzpicture}
        \begin{tikzpicture}[spy using outlines={green,magnification=\ssmag,size=\ssizz},inner sep=0]
            \node [align=center, img] {\includegraphics[width=\imgWidth]{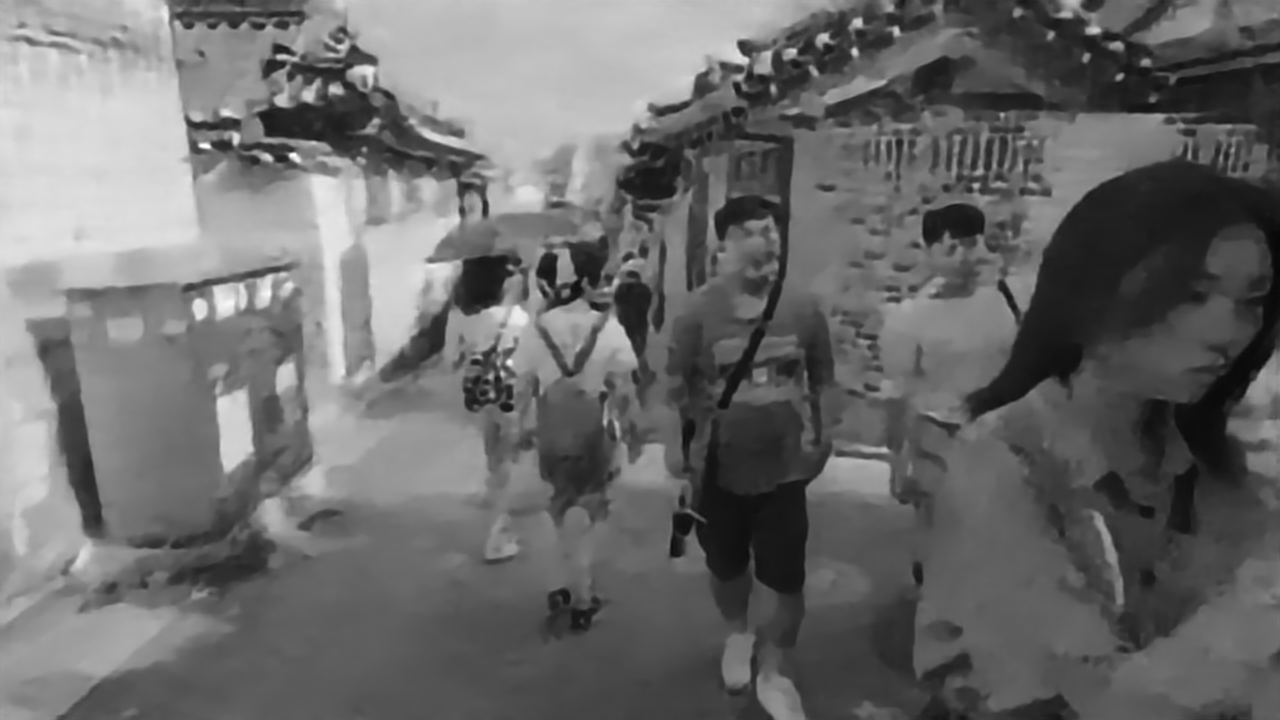}};
            \spy on \ssxxstwo in node [left] at \ssyystwo;
            \node [anchor=west] at \sccone {\textcolor{white}{\footnotesize \bf eSL-Net / 2}};
    	\end{tikzpicture}
		\begin{tikzpicture}[spy using outlines={green,magnification=\ssmag,size=\ssizz},inner sep=0]
            \node [align=center, img] {\includegraphics[width=\imgWidth]{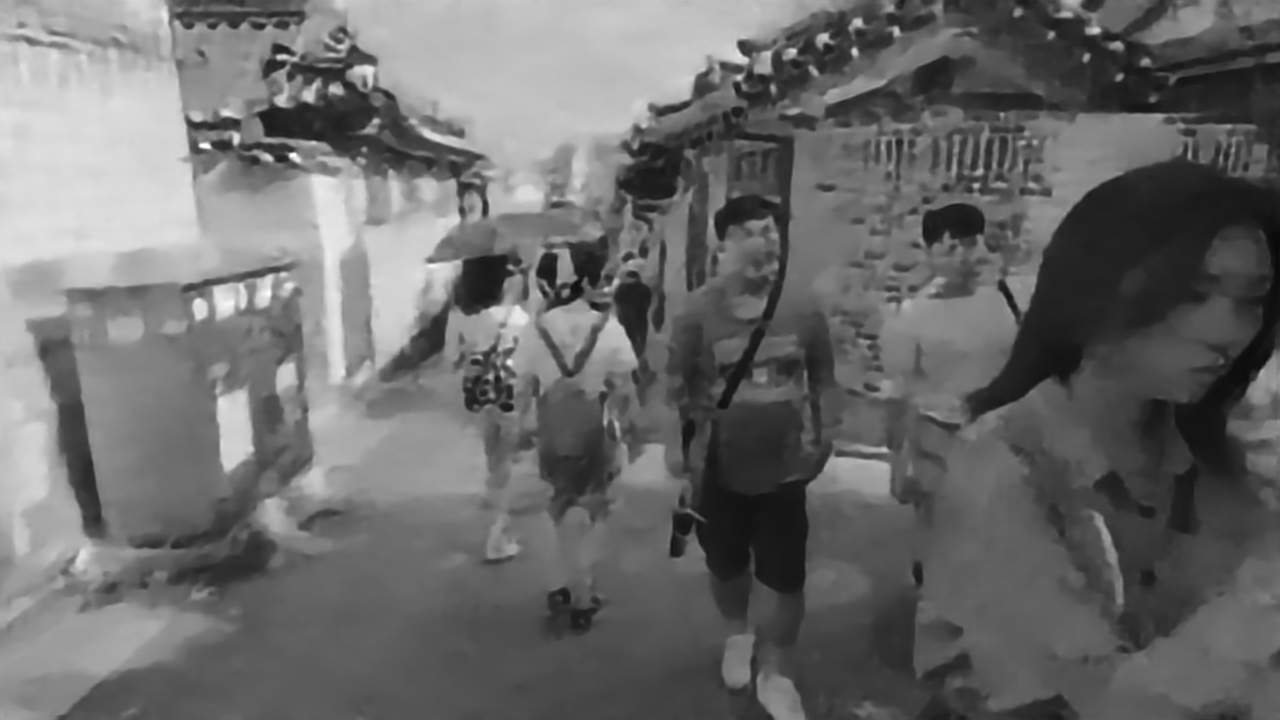}};
            \spy on \ssxxstwo in node [left] at \ssyystwo;
            \node [anchor=west] at \sccone {\textcolor{white}{\footnotesize \bf eSL-Net / 4}};
    	\end{tikzpicture}
\\
    %% second row
		\begin{tikzpicture}[spy using outlines={green,magnification=\ssmag,size=\ssizz},inner sep=0]
            \node [align=center, img] {\includegraphics[width=\imgWidth]{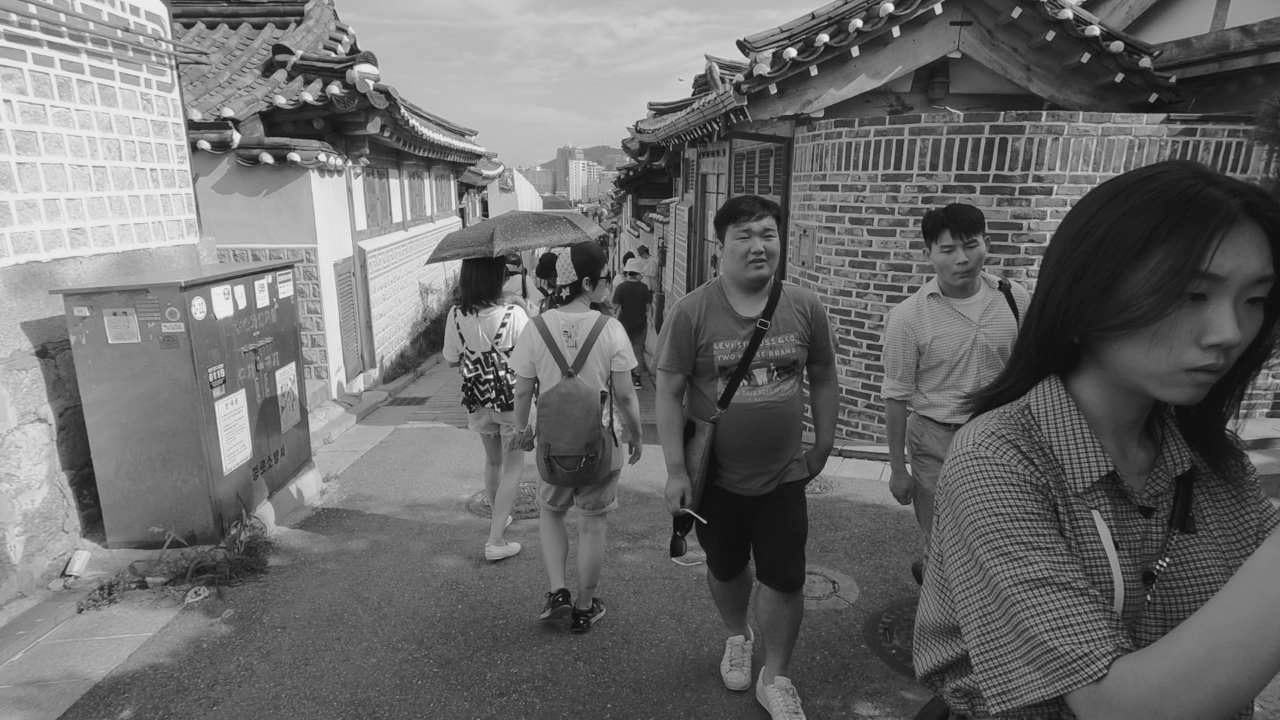}};
            \spy on \ssxxstwo in node [left] at \ssyystwo;
            \node [anchor=west] at \sccone {\textcolor{white}{\footnotesize \bf Ground Truth}};
    	\end{tikzpicture}
		\begin{tikzpicture}[spy using outlines={green,magnification=\ssmag,size=\ssizz},inner sep=0]
            \node [align=center, img] {\includegraphics[width=\imgWidth]{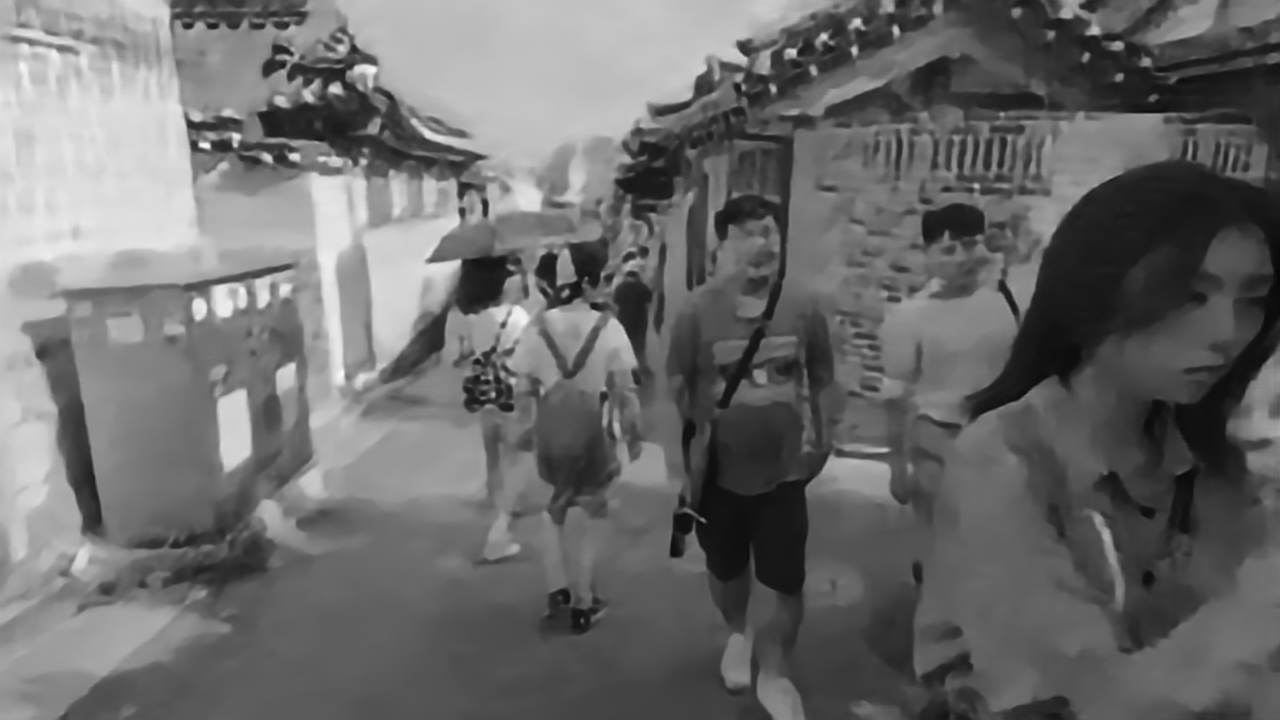}};
            \spy on \ssxxstwo in node [left] at \ssyystwo;
            \node [anchor=west] at \sccone {\textcolor{white}{\footnotesize \bf eSL-Net++ / 0}};
    	\end{tikzpicture}
		\begin{tikzpicture}[spy using outlines={green,magnification=\ssmag,size=\ssizz},inner sep=0]
           \node [align=center, img] {\includegraphics[width=\imgWidth]{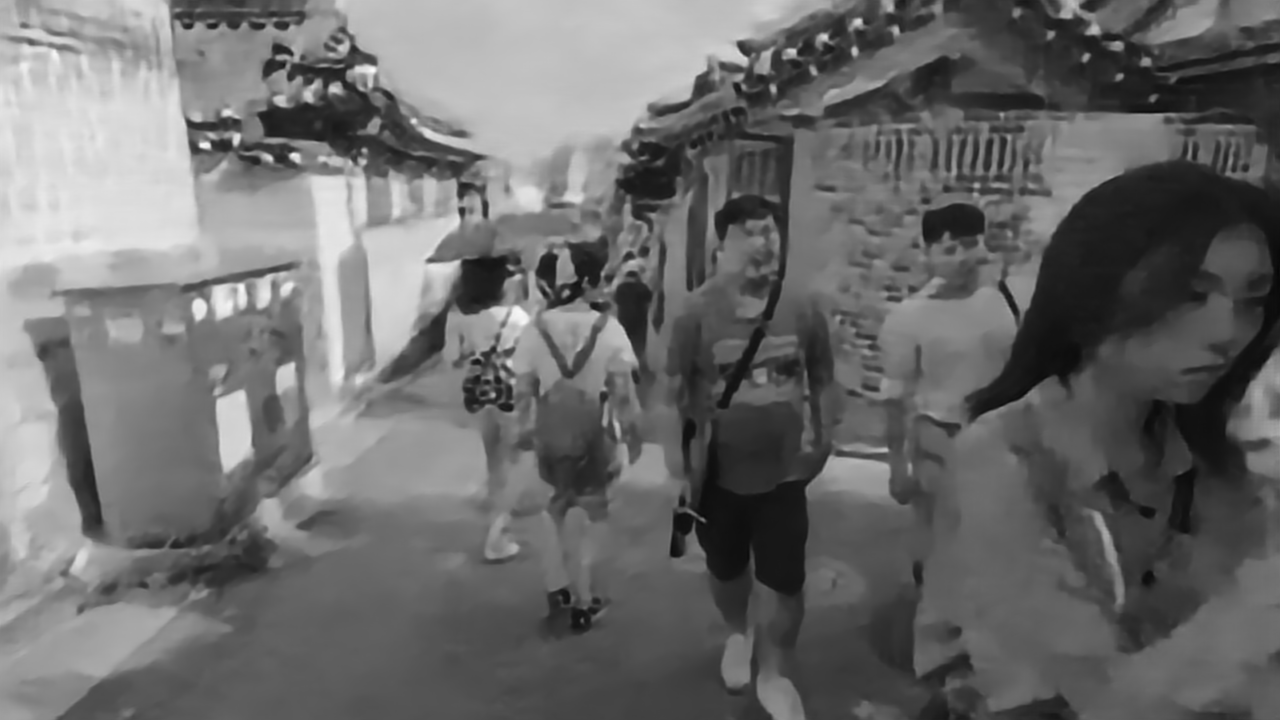}};
           \spy on \ssxxstwo in node [left] at \ssyystwo;
            \node [anchor=west] at \sccone {\textcolor{white}{\footnotesize \bf eSL-Net++ / 2}};
    	\end{tikzpicture}
		\begin{tikzpicture}[spy using outlines={green,magnification=\ssmag,size=\ssizz},inner sep=0]
            \node [align=center, img] {\includegraphics[width=\imgWidth]{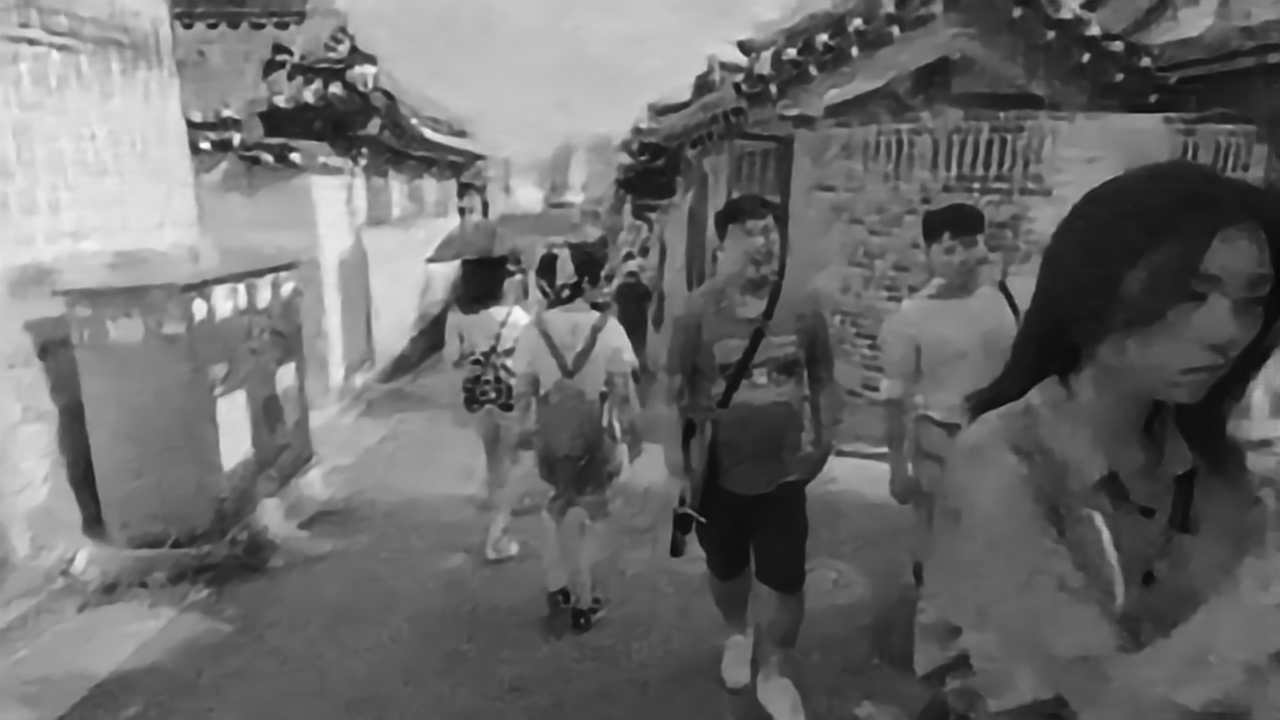}};
            \spy on \ssxxstwo in node [left] at \ssyystwo;
            \node [anchor=west] at \sccone {\textcolor{white}{\footnotesize \bf eSL-Net++ / 4}};
    	\end{tikzpicture}
    	\subcaption{Results under different image noise}
    \end{minipage}
	\caption{\colored{
	Qualitative comparisons under different levels of event noise and image noise. (a) Results under different event noise, where 20\%, 60\%, and 100\% indicate that the noise events are 20\%, 60\%, and 100\% of the original events, respectively. (b) Results under different image noise, where 0, 2, and 4 indicate the standard deviations of the white Gaussian noise added to images are 0, 2, and 4, respectively.  }}
	\label{fig:noise-exp-visual}
\end{figure*}
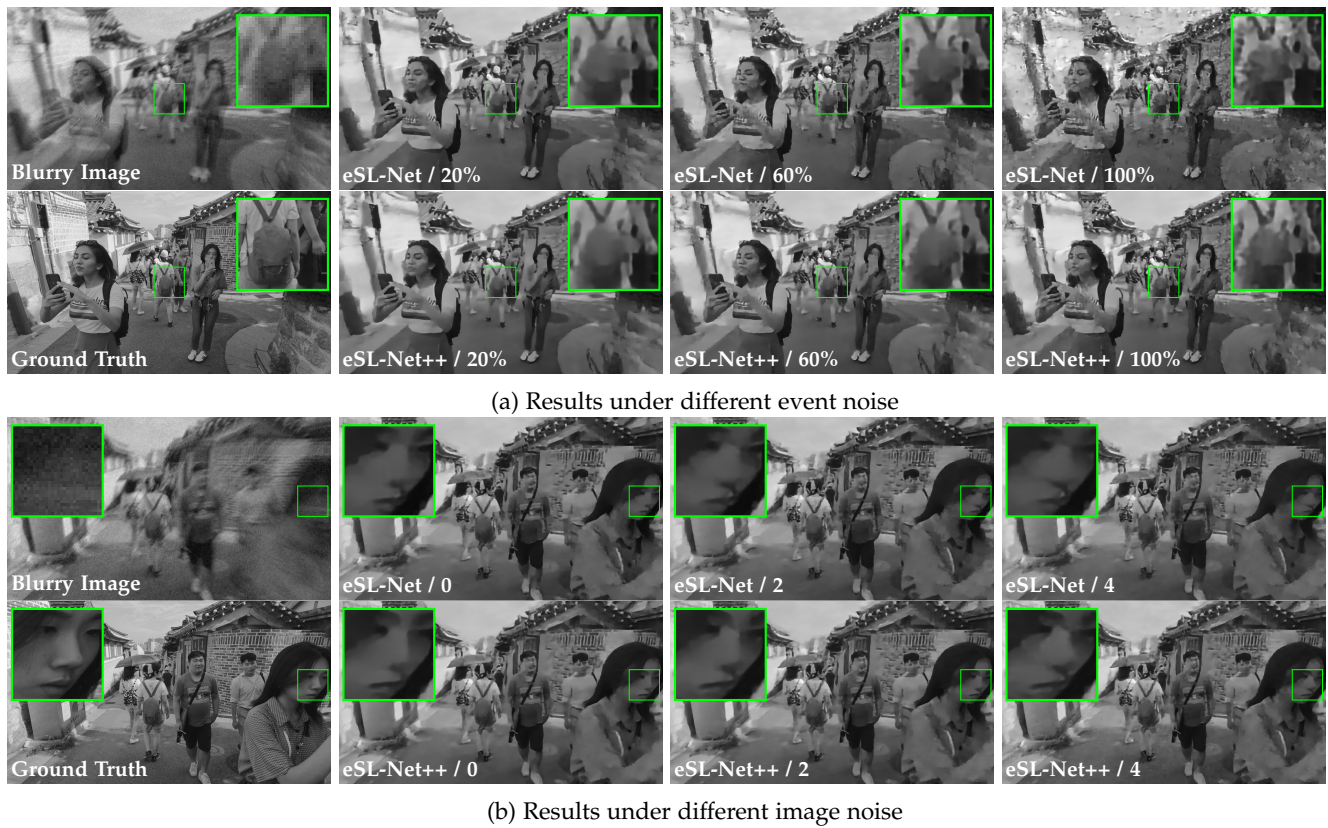

\subsection{\colored{Robustness to Event and Image Noise}}

\colored{To validate the robustness of our proposed algorithms, we conduct quantitative and qualitative comparisons of single frame reconstruction under different levels of event and image noise on the GoPro dataset. Regarding the event noise, we add event noise with varying ratios of noise ranging from $20$\% to $100$\% and evaluate the performance of eSL-Net and eSL-Net++. The quantitative results are plotted in Fig.~\ref{fig:event-noise}. Regarding the image noise, we add Gaussian noise to the blurry image with different standard deviations. The corresponding performance of eSL-Net and eSL-Net++ can be referred from Fig.~\ref{fig:image-noise}. The quantitative results validate that eSL-Net++ outperforms eSL-Net at different levels of noise no matter on events or images, which validates that eSL-Net++ is more robust than eSL-Net.
% we conduct quantitative and qualitative comparisons of single frame reconstruction under different levels of event and image noise on the GoPro dataset. Regarding the event noise, we add event noise with different noise ratios ranging from $20$\% to $100$\% and evaluate the performance of eSL-Net and eSL-Net++. The quantitative results are plotted in Fig.~\ref{fig:event-noise}. Regarding the image noise, we add Gaussian noise to the blurry image with different standard deviations and the corresponding performance of eSL-Net and eSL-Net++ can be referred from Fig.~\ref{fig:image-noise}. Quantitative results validate that eSL-Net++ outperforms eSL-Net at the different levels of noise no matter on events or images, which validates the robustness of eSL-Net++.
}

\colored{Correspondingly, we also provide qualitative comparisons under different levels of event noise and image noise as shown in Fig.~\ref{fig:noise-exp-visual}, where we can draw conclusions consistent with the quantitative comparisons.}

\section{Conclusion}
In this paper, we have proposed a novel network named \textbf{eSL-Net++} for E-SRB. We start from formulating an {\it E}vent-enhanced {\it D}generation {\it M}odel (EDM) to simultaneously take into account the image degradation caused by noises, motion blurs, and down-sampling. We present a {\it D}ual {\it S}parse {\it L}earning scheme (DSL) by assuming the sparsity over latent images and events, and then built a deep neural network, \ie, eSL-Net++, by unfolding DSL iterations. Compared to its previous version, \ie, eSL-Net, eSL-Net++ further takes into account event noises and extends to sequence frame SRB by a rigorous event shuffle-and-merge scheme, leading to superior performance to eSL-Net. Extensive experiments on synthetic and real-world data have demonstrated the effectiveness and superiority of our eSL-Net++.
\bibliographystyle{IEEEtran}
% \bibliography{refpaper}
\bibliography{lib}
\vskip -2.3\baselineskip plus -1fil

\begin{IEEEbiography}[{\includegraphics[width=1in,height=1.25in,clip, keepaspectratio, trim={0 50 0 90}]{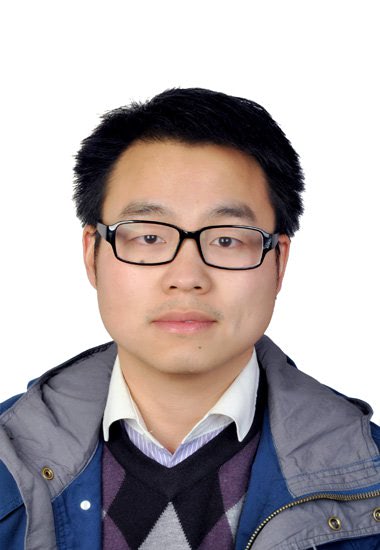}}]{Lei Yu}
received his B.S. and Ph.D. degrees in signal processing from Wuhan University, Wuhan, China, in 2006 and 2012, respectively. From 2013 to 2014, he has been a Postdoc Researcher with the VisAGeS Group at the Institut National de Recherche en Informatique et en Automatique (INRIA) for one and half years. He is currently working as an associate professor at the School of Electronics and Information, Wuhan University, Wuhan, China. From 2016 to 2017, he has also been a Visiting Professor at Duke University for one year. He has been working as a guest professor in the École Nationale Supérieure de l'Électronique et de ses Applications (ENSEA), Cergy, France, for one month in 2018. His research interests include neuromorphic vision and computation.
\end{IEEEbiography}

\vskip -2.\baselineskip plus -1fil

\begin{IEEEbiography}[{\includegraphics[width=1in,height=1.25in,clip,keepaspectratio, trim={0 35 0 10}]{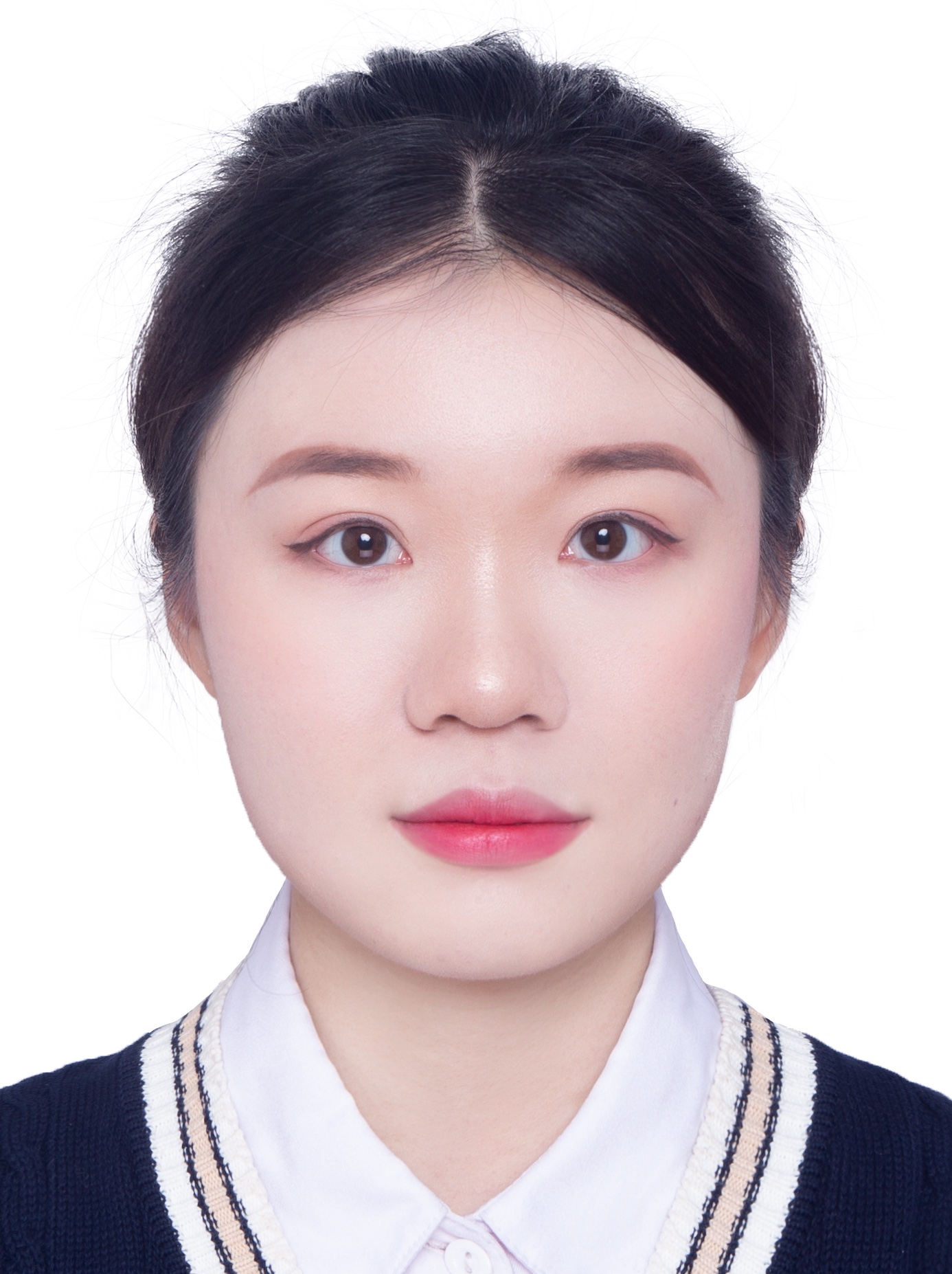}}]{Bishan Wang}
received her B.E. degree in communication engineering from Wuhan University, Wuhan, China, in 2019. She is currently working toward an M.S. degree in information and communication engineering with the electronic information school, Wuhan University, Wuhan, China. Her research interests include image processing and computer vision.
\end{IEEEbiography}

% \vskip -2.3\baselineskip plus -1fil
\vskip -2.\baselineskip plus -1fil

\begin{IEEEbiography}[{\includegraphics[width=1in,height=1.25in,clip,keepaspectratio, trim={20 40 20 40}]{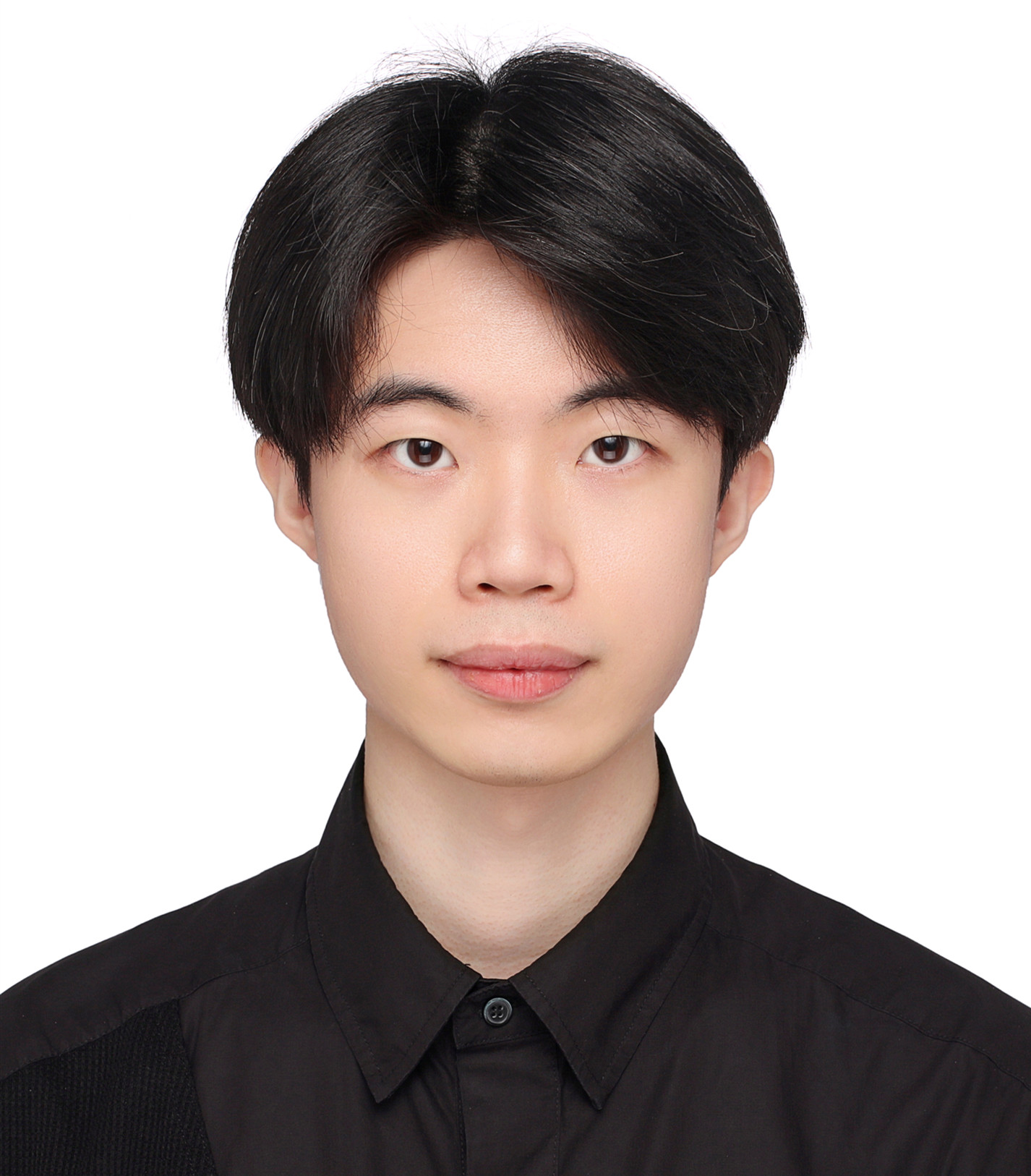}}]{Xiang Zhang}
received his B.E. degree in communication engineering from Wuhan University, Wuhan, China, in 2020. He is currently working toward an M.S. degree in information and communication engineering with the electronic information school, Wuhan University, Wuhan, China. His research interests include computer vision and neuromorphic computation.
\end{IEEEbiography}

% % \vskip -2.3\baselineskip plus -1fil

% \vskip -2.3\baselineskip plus -1fil
\vskip -2.\baselineskip plus -1fil

\begin{IEEEbiography}[{\includegraphics[width=1in,height=1.25in,clip,keepaspectratio]{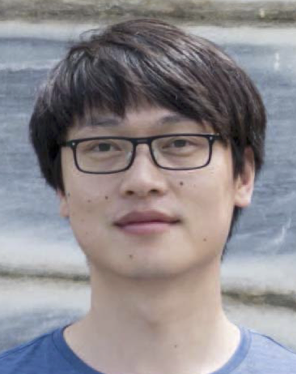}}]{Haijian Zhang}
received the B.Eng. degree in electronic information engineering from Wuhan University, Wuhan, China, in 2006, and the joint Ph.D. degree from the Conservatoire National des Arts et Metiers, Paris, France, and Wuhan University, in 2011. From 2011 to 2014, he was a Research Fellow with the School of Electrical and Electronic Engineering, Nanyang Technological University, Singapore. He is currently an Associate Professor with the School of Electronic Information, Wuhan University. His main research interests include time–frequency analysis, array signal processing, and multimedia forensics and security.

\end{IEEEbiography}

% \vskip -2.3\baselineskip plus -1fil
\vskip -2.\baselineskip plus -1fil

\begin{IEEEbiography}[{\includegraphics[width=1in,height=1.25in,clip,keepaspectratio, trim={0 35 0 10}]{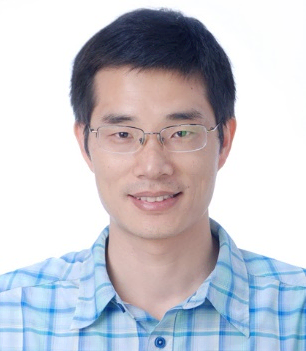}}]{Wen Yang} received the B.S. degree in electronic apparatus and surveying technology, and the M.S. degree in computer application technology and the Ph.D. degree in communication and information system from Wuhan University, Wuhan, China, in 1998, 2001, and 2004, respectively. From 2008 to 2009, he worked as a Visiting Scholar with the Apprentissage et Interfaces (AI) Team, Laboratoire Jean Kuntzmann, Grenoble, France. From 2010 to 2013, he worked as a Post-Doctoral Researcher with the State Key Laboratory of Information Engineering, Surveying, Mapping and Remote Sensing, Wuhan University. Since then, he has been a Full Professor with the School of Electronic Information, Wuhan University. He is also a guest professor of the Future Lab AI4EO in Technical University of Munich. His research interests include object detection and recognition, multisensor information fusion, and remote sensing image processing.
\end{IEEEbiography}

% \vskip -2.3\baselineskip plus -1fil
\vskip -2.\baselineskip plus -1fil

\begin{IEEEbiography}[{\includegraphics[width=1in,height=1.25in,clip,keepaspectratio]{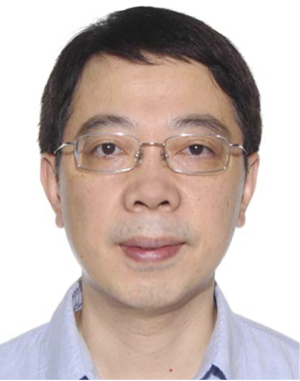}}]{Jianzhuang Liu}
received the Ph.D. degree in computer vision from The Chinese University of Hong Kong, Hong Kong, in 1997. From 1998 to 2000, he was a Research Fellow with Nanyang Technological University, Singapore. From 2000 to 2012, he was a Postdoctoral Fellow, an Assistant Professor, and an Adjunct Associate Professor with The Chinese University of Hong Kong. In 2011, he joined the Shenzhen Institute of Advanced Technology, University of Chinese Academy of Sciences, Shenzhen, China,
as a Professor. He is currently a Principal Researcher with Huawei Technolo- gies Company Ltd., Shenzhen. He has authored more than 150 papers. His research interests include computer vision, image processing, deep learning, and graphics.
\end{IEEEbiography}

% \vskip -2.3\baselineskip plus -1fil
\vskip -2.\baselineskip plus -1fil

\begin{IEEEbiography}[{\includegraphics[width=1in,height=1.25in,clip,keepaspectratio]{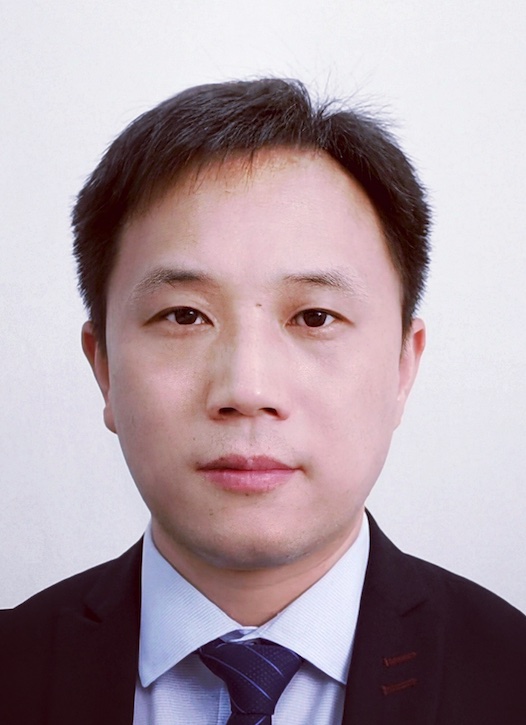}}]{Gui-Song Xia}
received his Ph.D. degree in image processing and computer vision from CNRS LTCI, T{\'e}l{\'e}com ParisTech, Paris, France, in 2011. From 2011 to 2012, he has been a Post-Doctoral Researcher with the Centre de Recherche en Math{\'e}matiques de la Decision, CNRS, Paris-Dauphine University, Paris, for one and a half years.
He is currently working as a full professor in computer vision and photogrammetry at Wuhan University. He has also been working as Visiting Scholar at DMA, {\'E}cole Normale Sup{\'e}rieure (ENS-Paris) for two months in 2018. He is also a guest professor of the Future Lab AI4EO in Technical University of Munich (TUM). His current research interests include mathematical modeling of images and videos, structure from motion, perceptual grouping, and remote sensing image understanding. He serves on the Editorial Boards of several journals, including {\em ISPRS Journal of Photogrammetry and Remote Sensing, Pattern Recognition, Signal Processing: Image Communications, EURASIP Journal on Image \& Video Processing, Journal of Remote Sensing, and Frontiers in Computer Science: Computer Vision}.
\end{IEEEbiography}

\end{document}